%% file: main.tex
\pdfoutput=1
\documentclass{article}

\input{header.tex}

\begin{document}

\maketitle

\begin{abstract}
  Policy-gradient methods in Reinforcement Learning(RL) are very universal and widely applied in practice but their performance suffers from the high variance of the gradient estimate. Several procedures were proposed to reduce it including actor-critic(AC) and advantage actor-critic(A2C) methods. Recently the approaches have got new perspective due to the introduction of Deep RL: both new control variates(CV) and new sub-sampling procedures became available in the setting of complex models like neural networks. The vital part of CV-based methods is the goal functional for the training  of the CV, the most popular one is the least-squares criterion of A2C. Despite its practical success, the criterion is not the only one possible. In this paper we for the first time investigate the performance of the one called Empirical Variance(EV). We observe in the experiments that not only EV-criterion performs not worse than A2C but sometimes can be considerably better. Apart from that, we also prove some theoretical guarantees of the actual variance reduction under very general assumptions and show that A2C least-squares goal functional is an upper bound for EV goal. Our experiments indicate that in terms of variance reduction EV-based methods are much better than A2C and allow stronger variance reduction.
\end{abstract}

\section{Introduction}
\label{intro}

\input{intro.tex}

\section{EV-Algorithms}

\input{section2.tex}

\section{Theoretical Guarantees}

\input{section3.tex}

\section{Experimental Results}

\input{experiments.tex}

\section{Conclusion}

\input{conclusion.tex}

\section{Acknowledgements}

\input{acknowledgements.tex}

\clearpage
\bibliography{main.bib}

\clearpage

\input{supplement}

\end{document}

%% file: header.tex
\usepackage{graphicx}
\usepackage{subfigure}

\usepackage{dsfont}
\usepackage{amsfonts,amsmath,amsthm,amssymb}
\usepackage{thmtools,thm-restate}
\usepackage{xargs}

\usepackage{multirow}

\newcounter{asscnt}
\newtheorem{assum}[asscnt]{Assumption}
\newcounter{thcnt}
\newtheorem{theorem}[thcnt]{Theorem}
\newtheorem{proposition}[thcnt]{Proposition}
\newtheorem{lemma}[thcnt]{Lemma}

\DeclareMathOperator*{\argmin}{arg\,min}
\newcommandx{\norm}[2]{\left\Vert #1 \right\Vert_#2}
\newcommandx{\E}[1]{\mathds{E} \left[ #1 \right] }

\newcommandx{\Q}[1]{Q^{#1}}
\newcommandx{\V}[1]{V^{#1}}

\usepackage{natbib}
\setcitestyle{numbers,square}

\usepackage[utf8]{inputenc} 
\usepackage[T1]{fontenc}    
\usepackage{hyperref}       
\usepackage{url}            
\usepackage{booktabs}       
\usepackage{amsfonts}       
\usepackage{nicefrac}       
\usepackage{microtype}      
\usepackage{xcolor}         

\usepackage[preprint]{neurips_2021}

\input{authorTitle.tex}

%% file: authorTitle.tex
\title{Variance Reduction for Policy-Gradient Methods via Empirical Variance Minimization}

\author{%
  Maxim Kaledin \\
  CMAP, École Polytechnique, Palaiseau, France;\\
  HDI Lab, HSE University, Moscow, Russia;\\
  \texttt{mkaledin@hse.ru} \\
  \And
  Alexander Golubev\\
  HDI Lab, HSE University, Moscow, Russia;\\
  \texttt{agolubev@hse.ru}
  \And
  Denis Belomestny\\
  University of Duisburg-Essen, Essen, Germany;\\
  HDI Lab, HSE University, Moscow, Russia;\\
  \texttt{denis.belomestny@gmail.com}
}

%% file: intro.tex
Reinforcement learning (RL) is a framework of stochastic control problems where one needs to derive a decision rule (policy) for an agent performing in some stochastic environment giving him rewards or penalties for taking actions. The decision rule is naturally desired to be optimal with respect to some criterion; commonly, expected sum of discounted rewards is used as such criterion\citep{sutton:rlbook:2018}. Reinforcement learning is currently a fast-developing area with promising and existing applications in numerous innovative areas of the society: starting from AI for games \citep{VinyalsSCII2019,bernerDota22019,silverGo2017} and going to energy management systems \citep{levent:energyManagementRL2019,ernstDeepRLEMM}, manufacturing and robotics \citep{akkayaRubik2019} to name a few. Naturally, RL gives the practitioners new sets of control tools for any kind of automatization \citep{bellemareIntroDeepRL2018}.

Policy-gradient methods constitute the family of gradient algorithms which directly model the policy and exploit various formulas to approximate the gradient of expected reward with respect to the policy parameters \citep{williamsPG1992,suttonA2C2000}. One of the main drawbacks of these approaches is the variance emerging from the estimation of the gradient \citep{WeaverORBaseline2001}, which typically is high-dimensional. Apart from that, the total sum of rewards is itself a random variable with high variance. Both facts imply that the problem of gradient estimation might be quite challenging. The straightforward way to tackle gradient estimation is Monte Carlo scheme resulting in the algorithm called REINFORCE \citep{williamsPG1992}. In REINFORCE increasing the number of trajectories for gradient estimation naturally reduces the variance but costs a lot of time spent on simulation. Therefore,  variance reduction is necessarily required to construct procedures with gradient estimates of lower variance and lower computational cost than increasing the sample size.

The main developments in this direction include actor-critic by \cite{kondaAC2000} and advantage actor-critic: A2C \citep{suttonA2C2000} and asynchronous version of it, A3C \citep{mnihA3C2016}. Recently a new interest in such methods has emerged due to the introduction of deep reinforcement learning \citep{mnihHuman2015} and the frameworks for training nonlinear models like a neural network in RL setting, a very comprehensive review of this area is done by  \cite{bellemareIntroDeepRL2018}. During several decades a large number of new variance reduction methods were proposed, including sub-sampling methods like SVRPG \citep{papiniSVRPG2018,XuSVRPGConv2019} and various control variate approaches of \cite{schulmanGAE2015}, \cite{guQProp2017}, \cite{liuSEPolicyOpt2017}, \cite{tuckerMirage2018}, \cite{wuFactorizedBaselines2018}. There are also approaches of a bit different nature: trajectory-wise control variates \citep{ChengTrajwise2020} using the control variate based on future rewards and variance reduction in input-driven environments \citep{maoVar2018}. Apart from that, in ergodic case there were both theoretic \citep{greensmithVRGE2004} and also some practical advancements \citep{ciosekExpectedPG2020}. The importance of the criteria for variance reduction is well-known in Monte-Carlo and MCMC \citep{siScalableCV2021} and recently was also addressed in RL by  \cite{fletResidVariance2021}, where the Actor with Variance Estimated Critic (AVEC) was proposed.

Being successful in practice, A2C method is difficult to analyze theoretically. In particular, it remains unclear how the goal functional used in A2C is related to the variance of the gradient estimator. Moreover, the empirical studies of the variance of the gradient estimator are still very rare and available mostly for artificial problems. In the community there is still an ongoing discussion, whether the variance of the gradient really plays main role in the performance of the developed algorithms, according to \cite{tuckerMirage2018}. In our paper we try to answer some of these questions and suggest a more direct approach inspired by the Empirical Variance(EV) Minimization recently studied by \cite{belomEVM2017}. We show that the proposed EV-algorithm  is not only theoretically justifiable but can also perform better than the classic A2C algorithm. It should be noted that the idea of using some kind of empirical variance functional is not new: some hints appeared, for instance, in \citep{liuSEPolicyOpt2017}. Despite that, the implementation and theoretical studies of this approach are still missing in the literature.

\subsection{Main Contributions}

\begin{itemize}
    \item We provide two new policy-gradient methods (EV-methods) based on EV-criterion and show that they perform well in several practical problems in comparison to A2C-criterion. We have deliberately chosen A2C and Reinforce as baseline algorithms to be less design-specific and have fair comparison of the two criteria. We show that in terms of training and mean rewards EV-methods perform at least as good as A2C but are considerably better in cases with complex policies.
    \item Theoretical variance bounds are proven for EV-methods. Also we show that EV-criterion addresses the stability of the gradient scheme directly while A2C-criterion is in fact an upper bound for EV. As far as we know, we are the first in the setting of RL who formulates the variance bounds with high probability with the help of the tools of statistical learning.
    \item We also provide the measurements of the variance of the gradient estimates which present several somewhat surprising observations. Firstly, EV-methods are able to solve this task much better allowing for reduction ratios of $10^{3}$ times . Secondly, in general we see another confirmation the hypothesis of \cite{tuckerMirage2018}: variance reduction has its effect but some environments are not so responsive to this.
    \item To our knowledge, we are the first who provide an experimental investigation of EV-criterion of policy-gradient methods in classic benchmark problems and the first implementation of it in the framework of PyTorch. Despite the idea is not new (it is mentioned, for example, by \cite{liuSEPolicyOpt2017}), so far EV-criterion was out of sight mainly because of A2C-criterion is computationally cheaper and is simpler to implement in the current deep learning frameworks since it does not need any complex operations with the gradient.
\end{itemize}

%% file: section2.tex
\subsection{Preliminaries}

Let us assume a Markov Decision Problem (MDP) $(\mathcal{S},\mathcal{A},R,\mathrm{P},\Pi,\mu_0,\gamma)$ with a finite horizon $T$, an arbitrary state space $\mathcal{S},$  an action space $\mathcal{A}$, a reward function $R: \mathcal{S} \times \mathcal{A} \to \mathds{R}$, Markov transition kernel $\mathrm{P}$. We are also given a class of policies $\Pi=\left\lbrace \pi_\theta: \mathcal{S} \to \mathcal{P}(\mathcal{A}) ~\vert~ \theta \in \Theta\right\rbrace$ parametrized by $\theta \in \Theta \subset \mathds{R}^D$ where $\mathcal{P}(\mathcal{A})$ is the set of probability distributions over the action set $\mathcal{A}$. We will omit the  subscript in $\pi_\theta$ wherever possible for shorter notation, in all occurrences $\pi \in \Pi$. Additionally we are provided with an initial distribution $\mu_0$, so that  $S_0 \sim \mu_0$, and a discounting factor $\gamma \in (0,1)$. The optimization problem reads as
\[
\text{ maximize } J(\theta)=\mathds{E}\left[ \sum_{t=0}^{T-1} \gamma^t R(S_t,A_t)\right] \quad \text{ w.r.t. }  ~ \theta \in \Theta,
\]
where we have assumed that the horizon $T$ is fixed. Let us note that any sequence of states, actions, and rewards can be represented as an element \(X\) of the product space
\[
 (\mathcal{S} \times \mathcal{A} \times \mathds{R})^T.
\]
 A generalization to the cases of infinite horizon and episodes is straightforward: we need to consider the space of sequences
\[
 (\mathcal{S} \times \mathcal{A} \times \mathds{R})^\infty
\]
 for infinite horizon or
\[
 \bigcup_{L=1}^\infty (\mathcal{S} \times \mathcal{A} \times \mathds{R})^L
\]
for the episodes, where the union is the set of all finite sequences. It turns out that the gradient scheme described below still works for these two cases, so we will focus on the finite horizon case only to simplify  the exposition.

\subsection{General Policy Gradient Scheme and REINFORCE}

Let $\widetilde{\nabla} J\vert_{\theta'}: (\mathcal{S} \times \mathcal{A} \times \mathds{R})^T \to \mathds{R}^D$ be an unbiased estimator of the gradient $\nabla_\theta J$ at point $\theta=\theta'$. 
With this notation the gradient descent algorithm for  minimization of $J(\theta)$ using  the estimate  $\widetilde{\nabla} J$ reads as follows:
\begin{equation}
\theta_{n+1} = \theta_n + \eta_n \frac{1}{K} \sum_{k=1}^K \widetilde{\nabla} J\vert_{\theta_n}(X_n^{(k)}),\quad n=1,2,\ldots ~
\label{eq:gradScheme}
\end{equation}
with $\eta_n$ being a positive sequence of step sizes. We will omit the subscript $\theta_n$ in the gradient estimate if it is clear from the context at which point  the gradient is computed. A simple example of the estimator $\widetilde{\nabla} J$ is the one called REINFORCE \citep{williamsPG1992}:
\[
\widetilde{\nabla}^{\rm{reinf}}J:\, X\mapsto \sum_{t=0}^{T-1} \gamma^t G_t(X) \nabla_\theta \log \pi(A_t \vert S_t)
\]
with 
\[
G_t(X):=\sum_{t'=t}^{T-1}\gamma^{t'-t} R_t,
\]
where $R_t=R(S_t,A_t)$ and 
\[
X=\left[ (S_0,A_0,R_0),..,(S_{T-1},A_{T-1},R_{T-1})\right]^\top.
\]
This form is obtained with the help of the following policy gradient theorem. 

\begin{proposition}(Policy gradient theorem \citep{williamsPG1992})
If  $X\in (\mathcal{S} \times \mathcal{A} \times \mathds{R})^\infty$ is sampled from the MDP, then under mild regularity conditions on $\pi, \mathrm{P},$
\[
\nabla_\theta J = \mathds{E}\left[ \sum_{t=0}^\infty \gamma^t G_t(X) \nabla_{\theta} \log \pi (A_t \vert S_t)\right].
\]
\end{proposition}
Note that the above proposition is formulated for the infinite horizon case, but similar statement also holds  for the finite-horizon and episodic cases. To see that, one can rewrite the problem as the one with infinite horizon giving zero reward after the end of the trajectory and almost sure transition from the end state to itself. 
\par
The baseline approach modifies the above Monte Carlo estimate by incorporating a family of state-action-dependent baselines $b_\phi: \mathcal{S} \times \mathcal{A} \to \mathds{R}$ (SA-baselines) or state-dependent baselines $b_\phi:\mathcal{S} \to \mathds{R}$ (S-baselines) parametrized by  $\phi$. The resulting gradient estimate reads as
\begin{equation}
\widetilde{\nabla}^{b_\phi} J: X \mapsto \sum_{t=0}^{T-1} \gamma^t (G_t(X)-b_\phi(S_t,A_t)) \nabla_\theta \log \pi(A_t \vert S_t).
\label{eq:baselinedGradient}
\end{equation}
In order to keep this estimate  unbiased, we need to additionally require that for all $\theta\in \Theta,$
\begin{equation*}
    \mathds{E}\left[ \sum_{t=0}^{T-1} \gamma^t b_\phi(S_t,A_t) \nabla_\theta \log \pi(A_t \vert S_t) \right]=0.
\end{equation*}
It is known that  every S-baseline $b_\phi: \mathcal{S} \to \mathds{R}$ will satisfy this requirement (we have placed the proof in Supplementary Materials, Prop. \ref{prop:sbaseline}). 
Such baselines, in particular, are often used in A2C algorithms \citep{bellemareIntroDeepRL2018}. When action dependence is presented, special care is required. In fact, the SA-baselines  keeping the gradient estimate unbiased  are known only in the case of continuous action spaces, see \citep{guQProp2017,liuSEPolicyOpt2017,tuckerMirage2018,wuFactorizedBaselines2018}. The main drawback of these methods is that they often are problem-specific. For example, QProp and SteinCV algorithms require the actions to be from continuous set so that one could differentiate the policy with respect to them. QProp additionally needs a notion of mean action.  In the case of factorized baselines we need to require the policy to be factorized in coordinates and to construct a vector representation for each action which is  not trivial in practice. In this paper we experiment with S-baselines since these allow us fair comparison of the variance reduction procedures with the same models for baseline and policy but the algorithm is applicable generally: it could be used in the gradient routines in place of A2C least-squares criterion.

\subsection{Two-Timescale Gradient Algorithm with Variance Reduction}

If we consider A2C algorithm, we might notice that it can be written as a two-timescale scheme with two step sizes $\alpha_n, \beta_n$
\begin{align}
& \theta_{n+1} = \theta_n + \alpha_n \frac{1}{K} \sum_{k=1}^K \widetilde{\nabla}^{b_\phi}J(X^{(k)}_n),\\
& \phi_{n+1} = \phi_n - \beta_n \nabla_\phi V^{A2C}_{K,n}(\phi)\vert_{\phi_n}
\end{align}
where
\begin{equation}
 V^{A2C}_{K,n}(\phi):= \frac{1}{K} \sum_{k=1}^K \sum_{t=0}^{T-1} (G_t(X_n^{(k)})-b_{\phi}(S_t^{(k)}))^2
\end{equation}
is A2C goal reflecting our desire to approximate the corresponding value function from its noisy estimates $(G_t(X_n^{(k)}))$  via least squares. The motivation behind it is that if one chooses the value function as baseline, the variance will be minimized. This strategy works well in practical problems \citep{mnihA3C2016}.
\par
If one would like to improve the baseline method there are two ways. One can either construct better baseline families (in which much effort was already invested) or  change variance functional in the second timescale. In this work we address the variance of the gradient estimate directly via empirical variance (EV). Since a gradient estimate at iteration $n$, $\widetilde{\nabla}^{b_\phi} J(X_n)\vert_{\theta_n}$, is a random vector, we could define its variance as
\begin{align}
&V_n(\theta,\phi):= \mathrm{Tr}\left(\E{ \left(\widetilde{\nabla}^{b_\phi}J(X_n)\vert_\theta - \E{\widetilde{\nabla}^{b_\phi}J(X_n)\vert_\theta}\right) \cdot  \left(\widetilde{\nabla}^{b_\phi}J(X_n)\vert_\theta - \E{\widetilde{\nabla}^{b_\phi}J(X_n)\vert_\theta}\right)^* } \right),
\label{eq:variancedef1}
\end{align}
or, what is the same, as
\begin{align}
&V_n(\theta,\phi) =\E{\norm{\widetilde{\nabla}^{b_\phi}J(X_n)\vert_{\theta} }{2}^2} -\norm{  \E{ \widetilde{\nabla}^{b_\phi}J(X_n)\vert_{\theta}}}{2}^2 .
\label{eq:variancedef2}
\end{align}
Therefore, its empirical analogue is
\begin{align}
&V^{EVv}_{n,K}(\theta,\phi) :=\\
&=\frac{1}{K}\sum_{k=1}^K \norm{\widetilde{\nabla}^{b_\phi}J(X_n^{(k)})\vert_{\theta}}{2}^2 - \frac{1}{K^2} \norm{ \sum_{k=1}^K \widetilde{\nabla} ^{b_\phi}J(X_n^{(k)})\vert_{\theta}}{2}^2.
\label{eq:empVariance}
\end{align}
It can be noticed also, that the second term in the variance \eqref{eq:variancedef2} does not depend on $\phi$ if the baseline does not add any bias. In this case we could safely discard it before going to sample estimates and use instead
\[
V^{EVm}_{K}(\theta,\phi) := \frac{1}{K}\sum_{k=1}^K \norm{\widetilde{\nabla}^{b_\phi}J(X_n^{(k)})\vert_{\theta} }{2}^2.
\]
The corresponding gradient descent algorithms can be described as
\begin{align}
& \theta_{n+1} = \theta_n + \alpha_n \frac{1}{K} \sum_{k=1}^K \widetilde{\nabla}^{b_\phi}J(X^{(k)}_n),\\
& \phi_{n+1} = \phi_n - \beta_n \nabla_\phi V^{EV}_K(\phi,\theta)\vert_{\phi_n, \theta_n}.
\label{eq:policyGradEV}
\end{align}
So we have constructed two methods. The first one uses the full variance $V_K^{EVv}$ and is called EVv, the second one is titled EVm and exploits $V_K^{EVm}$, the same variance functional but without the second term. The important fact to note is that EVv routine would work only if $K\geq 2$, otherwise we try to estimate the variance with one observation.\\
\par
As was pointed out in \citep{liuSEPolicyOpt2017}, the methods addressing the minimization of empirical variance would be computationally very demanding. This though strongly depends on the implementation. EV-methods are indeed more time-consuming than A2C, partially because of PyTorch which is not made for parallel computing of the gradients: the larger $K$ we want, the more time is needed. We are inclined to think that our implementation can be significantly optimized. The main complexity discussion with charts is placed in Supplementary.

%% file: section3.tex
The main advantage of using empirical variance is that we have the machinery of statistical learning to prove the upper bounds for the variance of the gradient estimator. Our main theoretical result is concerned about one step of the update of $\theta_n$ with the best possible baseline chosen from the class of control variates. In this section we give some background and problem formulation and in the end discuss how the results are applied to our initial problem and give theoretical guarantees for EV-methods.

\subsection{Variance Reduction}

Classic problem for variance reduction is formulated for Monte Carlo estimation of some expectation. Let $X$ be random variable and $X_1,..,X_K$ be a sample from the same distribution. Given a function $h:\mathds{R} \to \mathds{R}$ we want to evaluate $\mathds{E}[h(X)]=\mathcal{E}$ using $\frac{1}{K}\sum_{k=1}^K h(X_k)$. This estimate, however, may possess large variance $V(h):=\mathds{E}\left[(h(X)-\mathcal{E})^2\right]$, one could avoid that using other function $h'$ such that $\mathds{E}[h'(X)]=\mathcal{E}$ but $V(h')<V(h)$. Such estimate would be more reliable since there is less uncertainty.
\par
This leads us to the following formulation. Given a class $\mathcal{H}$ of functions $h: \mathds{R}^d \to \mathds{R}$ find  function $h_* \in \mathcal{H}$ with the least possible variance $V(h_*)\leq V(h)~~ \forall h \in \mathcal{H}$. Such problem is well-investigated in the literature and many methods have been suggested for variance reduction both for Monte Carlo and MCMC settings, for examples see \cite{girolamiMC2017,southZV2020}.

\subsection{Variance Reduction in Multivariate Case}

Let us consider scheme \eqref{eq:gradScheme}, the variance of the gradient estimate affects the convergence properties of the scheme so one is interested in reducing it, as can be seen in \citep{koppelGlobalConv2020, panxuConvAnalSVRPG2020} . We also provide a discussion about it in Supplementary. Note that now we are in setting different from the one above: it is needed to construct a vector estimate.\\

Let   $X_1,...,X_K \sim P$ be a sample of random vectors taking values in $\mathcal{X} \subset \mathds{R}^d$ and let  $\mathcal{H}$ be a class of functions $h: \mathds{R}^d \to \mathds{R}^D$ such that $\mathds{E}[h(X)]=\mathcal{E}$. Later we will also need the corresponding empirical measure $P_K$ based on $X_1,...,X_K$. Define the variance 
\[
V(h):=\mathds{E}[\Vert h(X) - \mathcal{E} \Vert^2]
\]
with $\Vert \cdot \Vert$ being Euclidean 2-norm. Our goal  is to find  a function $h_*\in \mathcal{H}$ such that $V(h_*)\leq V(h)$ for all $h \in \mathcal{H}$. Then we have a variance reduced Monte Carlo estimate  $\frac{1}{K}\sum_{k=1}^K h_*(X_k)$.

\subsection{Variance Representation in Terms of Excess Risk}

It is obvious that the exact solution $h_*$ cannot be computed meaning that we are left always with some suboptimal solution $\hat{h} \in \mathcal{H}$ given by a particular method of ours. The quantity $V(\hat{h})-V(h_*)$ where $h_*$ is defined with
\[
V(h_*):=\inf_{h \in \mathcal{H}} V(h)
\]
is usually called \textit{excess risk} in statistics and represents optimality gap, i.e. it shows how far the current solution $h$ is from the optimal one. We can always write the variance of $\hat{h}$ as
\begin{equation}
V(\hat{h})= \left[ V(\hat{h}) - \inf_{h \in \mathcal{H}} V(h) \right] + \inf_{h \in \mathcal{H}} V(h)
\label{eq:exriskdecomposition}
\end{equation}
from which we can clearly see the
excess risk (the first term) and the second term representing approximation richness of the class $\mathcal{H}$: generally speaking, the better this class is, the lower the infimum can be. As more concrete example, consider variance reduction using the method of control variates. In this setting the goal is to estimate $\mathds{E}[f(X)]$ for a fixed $f: \mathds{R}^d \to \mathds{R}^D$. To reduce the variance of the Monte Carlo sample mean one adds some \textit{control variate} $g \in \mathcal{G}$ with zero expectation giving us the class of unbiased estimates
\[
\mathcal{H}=\left\lbrace f-g: ~~ g\in \mathcal{G},~~ \mathds{E}[g(X)]=0 \right\rbrace.
\]
The excess risk is now of the form
\[
\left[ V(f-\hat{g}) - \inf_{g\in \mathcal{G}} V(f-g) \right],
\]
sometimes called \textit{stochastic error} and the second term
\[
\inf_{g \in \mathcal{H}} V(f-g)
\]
is known as \textit{approximation error}. So, the way to analyze variance reduction is to estimate the excess risk and the approximation error.\\

In our analysis we consider a class of estimators with control variates implemented as baselines. Specifically, the class of estimators is
\begin{equation}
\mathcal{H}:= \left\lbrace \widetilde{\nabla}^{b_\phi} J ~\vert ~ b_\phi \in \mathcal{B}_\Phi \right\rbrace,
\label{eqs:polgradestimators}
\end{equation}
where 
\[
\widetilde{\nabla}^{b_\phi}_{\theta} J: X \mapsto \sum_{t=0}^{T-1} \gamma^t (G_t-b_\phi(S_t,A_t)) \nabla_\theta \log \pi(A_t \vert S_t)
\]
and $b_\phi \in \mathcal{B}_\Phi$ is a map $\mathcal{S} \times \mathcal{A} \to \mathds{R}$. The set of baselines $\mathcal{B}_\Phi$ is a parametric class parametrized by $\phi \in \Phi$.  We require that for each $b_\phi \in \mathcal{B}_\Phi$ and for all policies $\pi \in \Pi$
\begin{equation*}
    \mathds{E}\left[ \sum_{t=0}^{T-1} \gamma^t b_\phi(S_t,A_t) \nabla_\theta \log \pi(A_t \vert S_t)\right]=0,
\end{equation*}
requiring therefore that the estimator $\widetilde{\nabla}^{b_\phi} J$ is unbiased for all $b_\phi \in \mathcal{B}_\Phi$. For example, any set of maps $b_\phi: \mathcal{S} \to \mathds{R}$ will satisfy the above condition leading to S-baselines.

We start with one-step analysis, showing how well the variance behaves when variance reduction with EV is applied at $n$th iteration. Let us further notate the estimator as $h: \mathds{R}^d \to \mathds{R}^D$ and note that $\mathds{E}[h(X)]=\mathcal{E}$ with constant $\mathcal{E}=\nabla_\theta J$ since the estimate is assumed to be unbiased. In order to reduce the variance in the gradient estimator we would like to pick on each epoch $n$ the best possible estimator
\[
h^* = \argmin_{h \in \mathcal{H}} V(h)
\]
where variance functional $V$ is defined for any $h \in \mathcal{H}$ via
\[
V(h) := \mathds{E}\left[ \Vert h(X) - \mathcal{E} \Vert^2\right]
\]
where $X$ is random vector of concatenated states, actions and rewards described before. To solve the above optimization problem, we use empirical analogue of the variance and define
\[
\hat{h} := \text{arg}\min_{h \in \mathcal{H}} V_K(h)
\]
with the empirical variance functional of the form:
\[
V_K(h):=\frac{1}{K-1} \sum_{k=1}^K \Vert h(X^{(k)}) - P_K h\Vert^2
\]
with $P_K$ being the empirical measure, so with the given sample we could notate sample mean as
\[
P_K h := \frac{1}{K}\sum_{k=1}^K h(X^{(k)}).
\]
Let us pose several key assumptions.

\begin{assum}
Class $\mathcal{H}$ consists of bounded functions: 
\[
 \sup_{x\in \mathcal{X}}\Vert h(x) \Vert \leq b, \quad \forall h \in \mathcal{H}.
\] 
\label{ass:bounded}
\end{assum}
\begin{assum}
The solution $h_*$ is unique and $\mathcal{H}$ is star-shaped around $h_*$: 
\[
\alpha h + (1-\alpha)h_* \in \mathcal{H},\quad \forall h \in \mathcal{H},~\alpha \in [0,1].
\]
\label{ass:starshape}
\end{assum}
\begin{assum}
The class $\mathcal{H}$ has covering of polynomial size: there are $\alpha \geq 2$ and $c>0$ such that for all $u \in (0,b],$
\[
\mathcal{N}( \mathcal{H}, \Vert \cdot \Vert_{L^2(P_K)},u) \leq \left( \frac{c}{u}\right)^\alpha ~a.s.
\]
where 
\[
\Vert h \Vert_{L^2(P_K)}= \sqrt{ P_K \Vert h\Vert_2^2}
\]
\label{ass:polycover}
\end{assum}
The following result holds.
\begin{theorem}
Under Assumptions \ref{ass:bounded}-\ref{ass:polycover} it holds with probability at least $1-4e^{-t},$
\[
V(h_K)-V(h_*) \leq \max_{j=1,\ldots,4} \beta_j(t)
\]
with
\begin{align*}
& \beta_{1}  \leq C_1 \frac{ \log K}{K}, ~\beta_{2}  \leq C_2 \frac{ \log K}{K}, \\
&\beta^{3}(t) =  \frac{ 8 (40b^2t +72b^2)}{3K}, \beta^{4}(t) =  \frac{ 9216 b^2t}{K},
\end{align*}
where $C_1,C_2$ are constants not depending on the dimension $D$ or the sample size $K$.
\label{th:exrisk}
\end{theorem}

This allows to conclude from the variance decomposition \eqref{eq:exriskdecomposition} that as sample size $K$ grows, the variance reduces to that of $h_*$. From practical perspective, Theorem \ref{th:exrisk} firstly gives some reliability guarantee. Secondly, it also shows that if we have $K$ large enough, we can reduce the variance even more. 

\subsection{Verifying the Assumptions in Policy-Gradient Setting}

Let us now discuss how we can satisfy the assumptions in our policy-gradient scheme \eqref{eq:policyGradEV}.

As to Assumption \ref{ass:bounded}, we can prove 
\begin{proposition}
(see Supplementary) If there exist constants $C_L>0$ and $C_R>0$ such that 
\begin{align*}
\forall \theta \in \Theta, ~a\in \mathcal{A}, s\in \mathcal{S} \quad &\Vert \nabla_\theta \log \pi(a \vert s)\Vert \leq C_L, \\
&\vert R(s,a)\vert \leq C_R,
\end{align*}
then Assumption 1 is satisfied.
\label{prop:boundedrewgrad}
\end{proposition}

In order to satisfy Assumption \ref{ass:starshape} in the context of policy gradient estimators $\mathcal{H}$ defined in \eqref{eqs:polgradestimators}, one might notice that
\[
V(h_K)-\argmin_{h \in \mathcal{H}}V(h)\leq V(h_K)-\argmin_{h \in \text{conv}(\mathcal{H})} V(h).
\]
Indeed, Assumption 2  is a weaker notion than convexity.\\

\begin{proposition}
(see Supplementary) If Assumption 3 holds for $\mathcal{B}_\phi$, under the conditions of Proposition \ref{prop:boundedrewgrad} Assumption 3 holds also for $\mathcal{H}$ with other constants $c,\alpha$.
\end{proposition}


Let us also note that we could use the more realistic As. \ref{ass:polycover} stating the same for $\log \mathcal{N}$ (therefore considering more complex classes of baselines) and get weaker bounds with rates $1/\sqrt{K}$ and $\log K/ K$, see \citep{belomEVM2017}.

\subsection{Asymptotic Equivalence of EVv and EVm}

Let us have a closer look on the variance functional with fixed baseline $b_\phi$,
\[
V(\tilde{\nabla}^{b_\phi} J(X_n))= \mathds{E}[\Vert\tilde{\nabla}^{b_\phi} J(X_n) \Vert^2] - \Vert\mathds{E}[\tilde{\nabla}^{b_\phi} J(X_n)]\Vert^2.
\]
Note that the right term equals $\Vert \nabla_\theta J\Vert^2$ since the estimate is unbiased. Therefore, if the gradient scheme converges to local optimum, i.e. $\theta_n \to \theta_*$  with $\nabla_{\theta_*} J=0$ and the baseline parameters $\phi_n \to \phi_*$ as $n \to \infty$ a.s., then we can define the limiting variance as
\[
V_\infty(\tilde{\nabla}^{b_{\phi_*}} J)= \mathds{E}[\Vert\tilde{\nabla}^{b_{\phi_*}} J\vert_{\theta=\theta_*} \Vert^2]
\]
which will strongly depend on the baseline we have chosen. This fact, firstly, implies that EVm and EVv algorithms are asymptotically equivalent because they differ in the second term converging to $0$ and the first term is dominating by Jensen's inequality. Indeed, in our experiments we see that EVm and EVv behave similarly, so one would accept EVm as computationally cheaper version which works with $K\geq 1$. Secondly, EV-methods give additional stability guarantees for large $n$  because they are directly related to the asymptotic gradient variance. It is an open question though to characterize the convergence of the presented two-timescale scheme to $(\theta_*,\phi_*)$ more precisely.

\subsection{Relation to A2C}

\begin{proposition}
(see Supplementary) If the conditions of Proposition \ref{prop:boundedrewgrad} are satisfied, then for all $K\geq 2$ A2C goal function $V^{A2C}_K(\phi)$ is an upper bound (up to a constant) for EV goal functions:
\[
V^{EVm}_K(\phi) \leq 2 C_L^2 V^{A2C}_K(\phi), \quad V^{EVv}_K(\phi) \leq 2 C_L^2 V^{A2C}_K(\phi).
\]

\end{proposition}

So, A2C is more computationally friendly method which exploits the upper bound on empirical variance for baseline training. This, in a sense, explains the success of A2C and different performance of A2C and EV-methods.

%% file: experiments.tex
We empirically investigate the behavior of EV-algorithms on several benchmark problems:
\begin{itemize}
    \item Gym Minigrid \citep{gym_minigrid} (\texttt{Unlock-v0, GoToDoor-5x5-v0});\\
    \item Gym Classic Control \citep{brockmanOpenAI2016} (\texttt{CartPole-v1, LunarLander-v2, Acrobot-v1}).
\end{itemize}
For each of these we provide charts with mean rewards illustrating the training process, the study of gradient variance and reward variance and time complexity discussions. Here because of small amount of space we present the most important results but the reader is welcome in the Supplementary materials where more experiments and investigations are presented together with all the implementation details. The code and config-files can be found on GitHub page \cite{github}. \\

\subsection{Overview}

Below we show the discussions about several key indicators of the algorithms.

\begin{enumerate}
    \item \textbf{Mean rewards.} They are computed at each epoch based on the rewards obtained during the training in 40 runs and characterize how good is the algorithm in interaction with the environment.
    \item \textbf{Standard deviation of the rewards.} These are computed in the same way but standard deviation is computed instead of mean. This values show how stable the training goes: high values indicate that there are frequent drops or increases in rewards.
    \item \textbf{Gradient variance.} It is measured every 200 epochs using  \eqref{eq:empVariance} with separate set of 50 sampled trajectories with relevant policy. This is the key indicator in the discussion of variance reduction. Surprisingly, as far as we know, we are the first in the RL community presenting such results for classic benchmarks. The resulting curves are averaged over 40 runs.
    \item \textbf{Variance Reduction Ratio.} Together with Gradient Variance itself we also measure reduction ratio computed as sample variance of the estimator with baseline divided by the sample variance  without baseline (assuming $b_\phi=0$) in the computations of Gradient Variance. The reduction ratio is the main value of interest in variance reduction research in Monte Carlo and MCMC.
\end{enumerate}


\begin{figure}[h!]
    \centering
    \begin{tabular}{lcr}
    (a) \includegraphics[scale=.2]{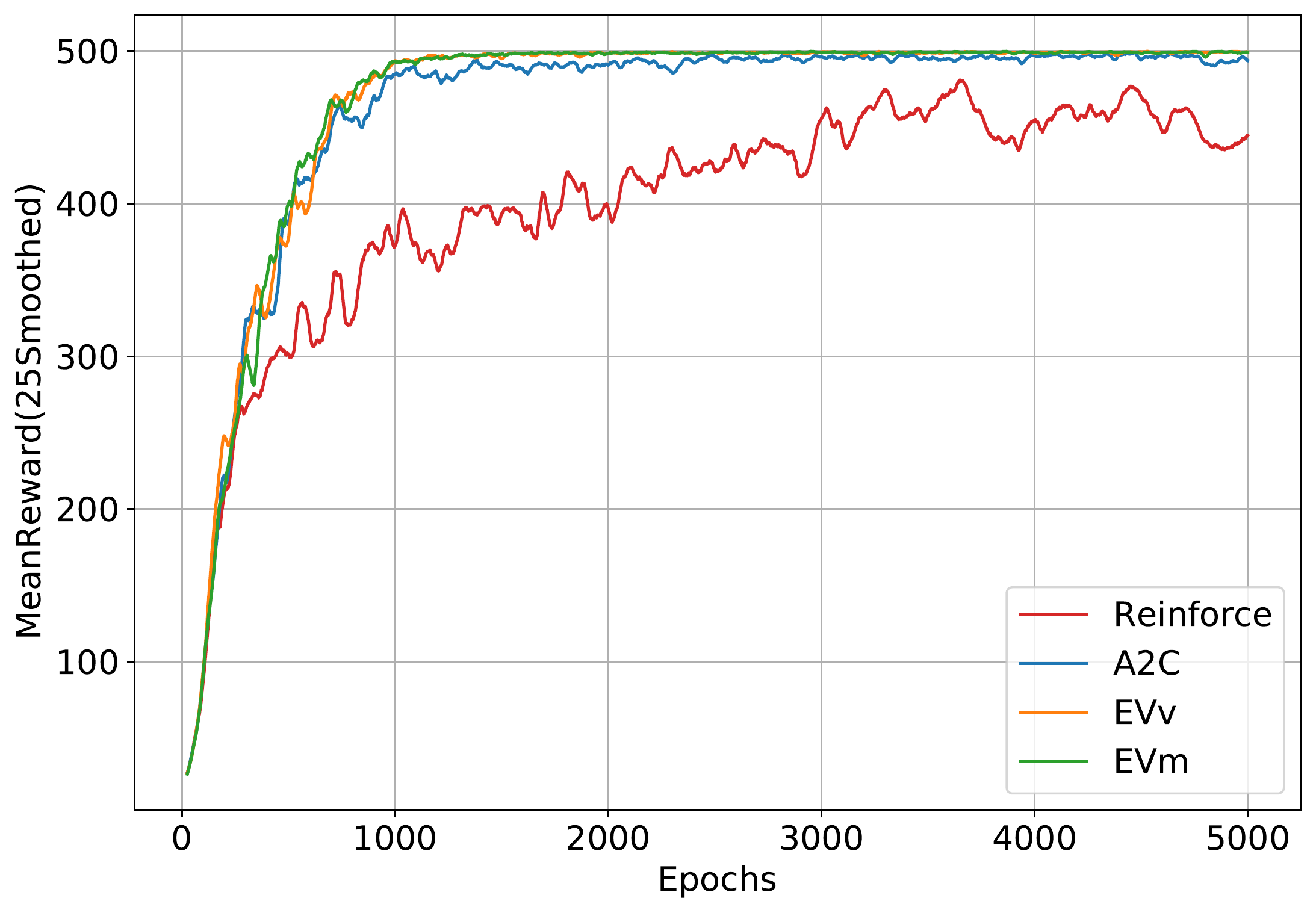} & (b) \includegraphics[scale=.2]{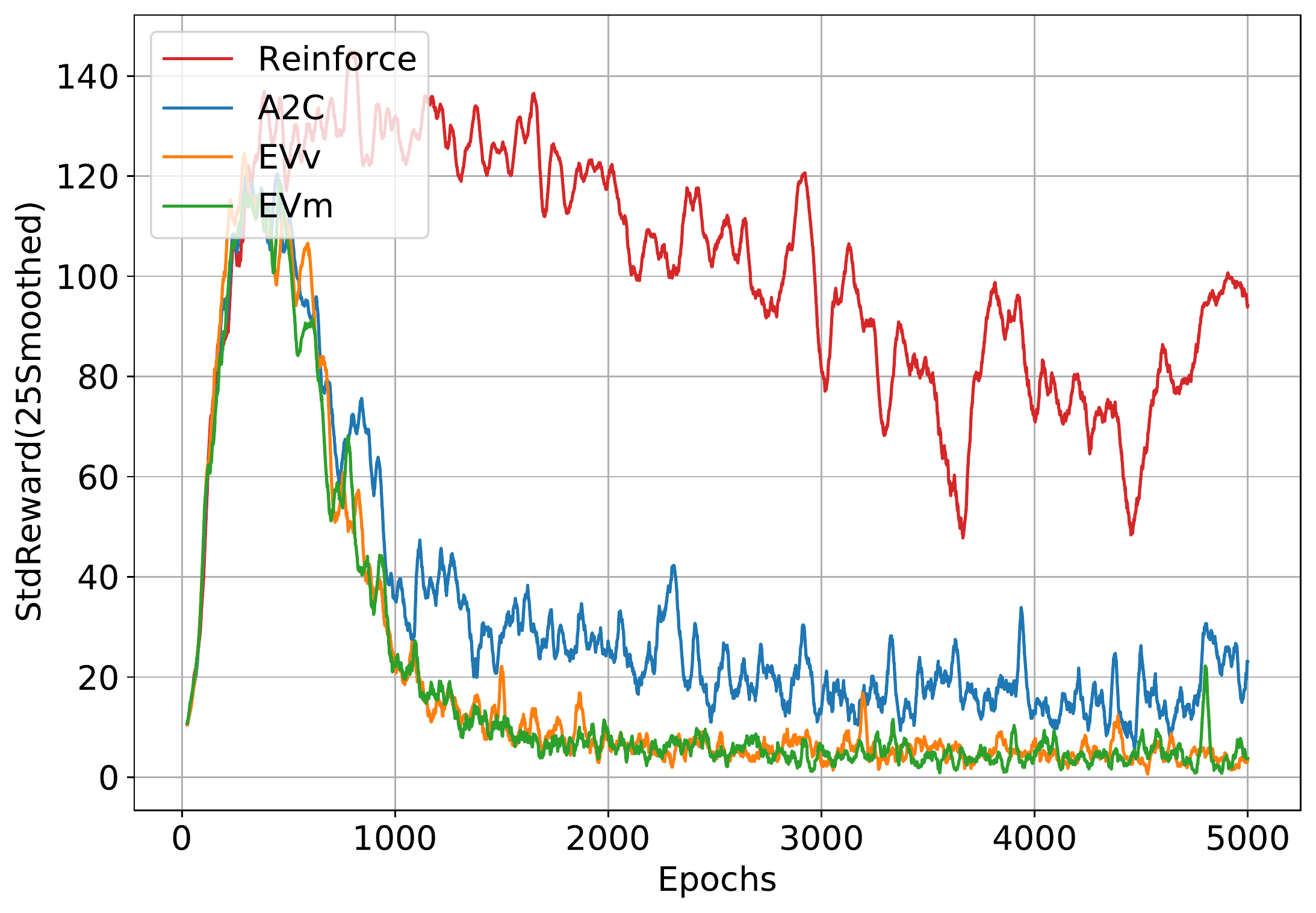} & (c)\includegraphics[scale=.2]{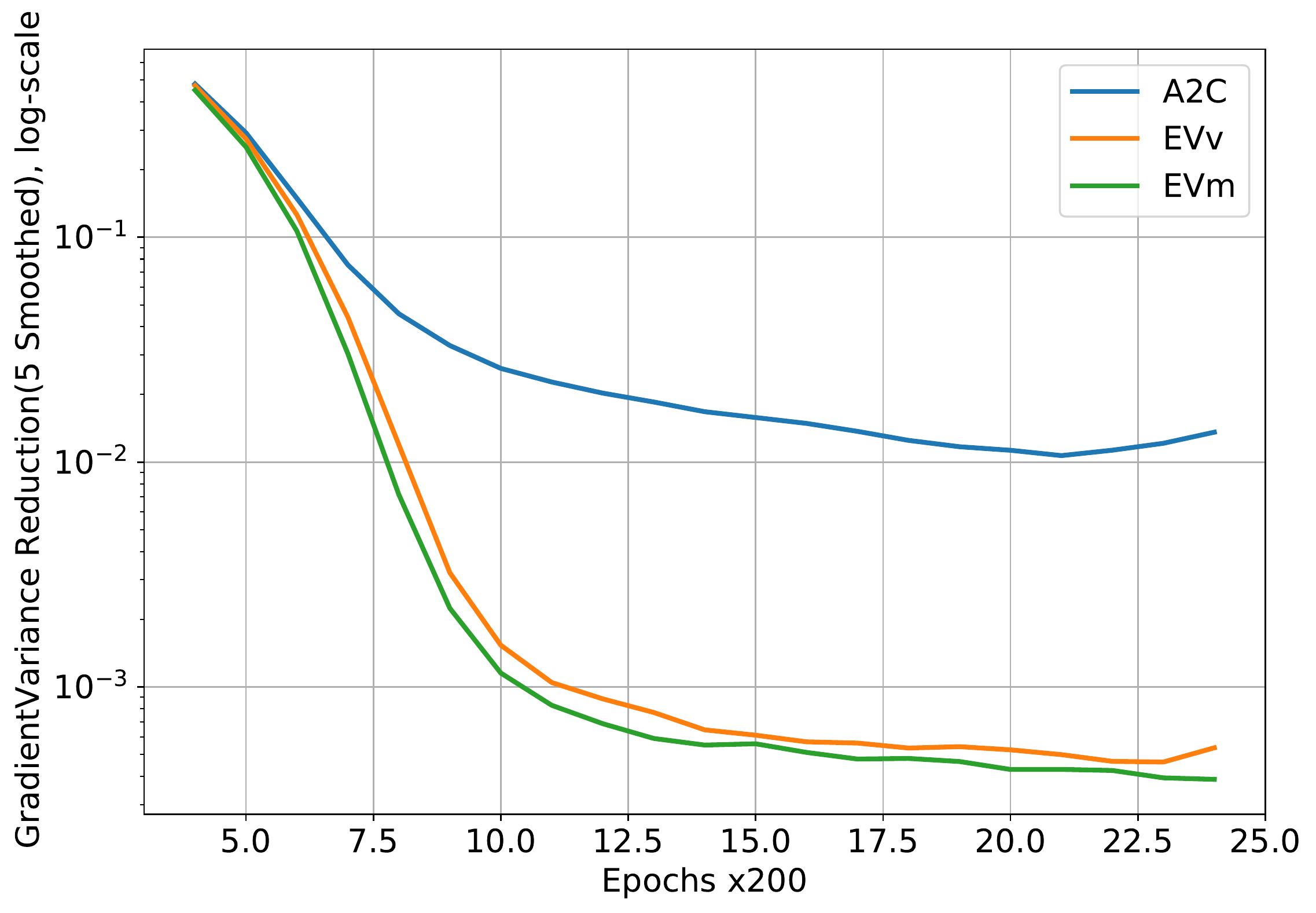} \\
    (d) \includegraphics[scale=.2]{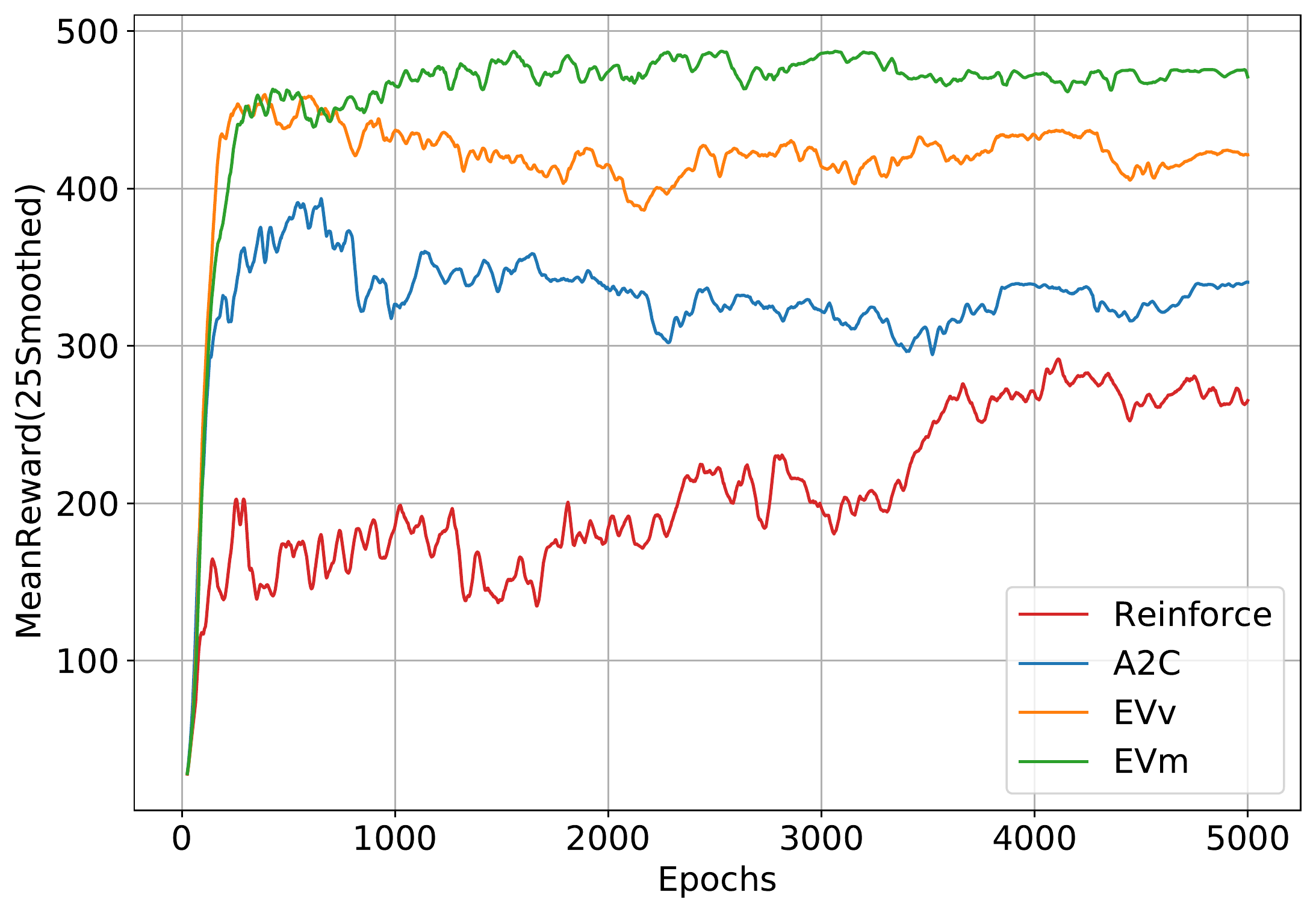} & (e) \includegraphics[scale=.2]{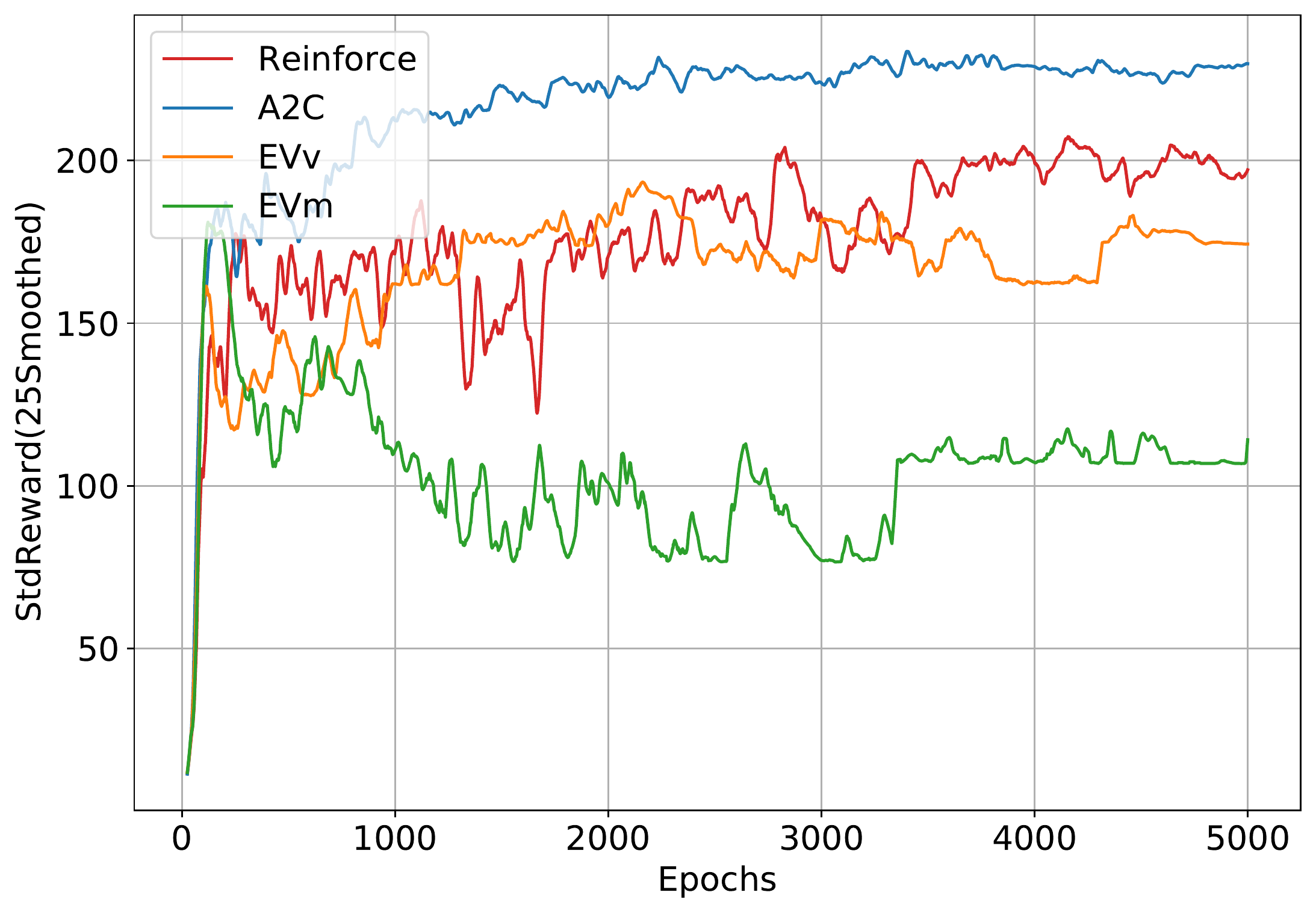} & (f)\includegraphics[scale=.2]{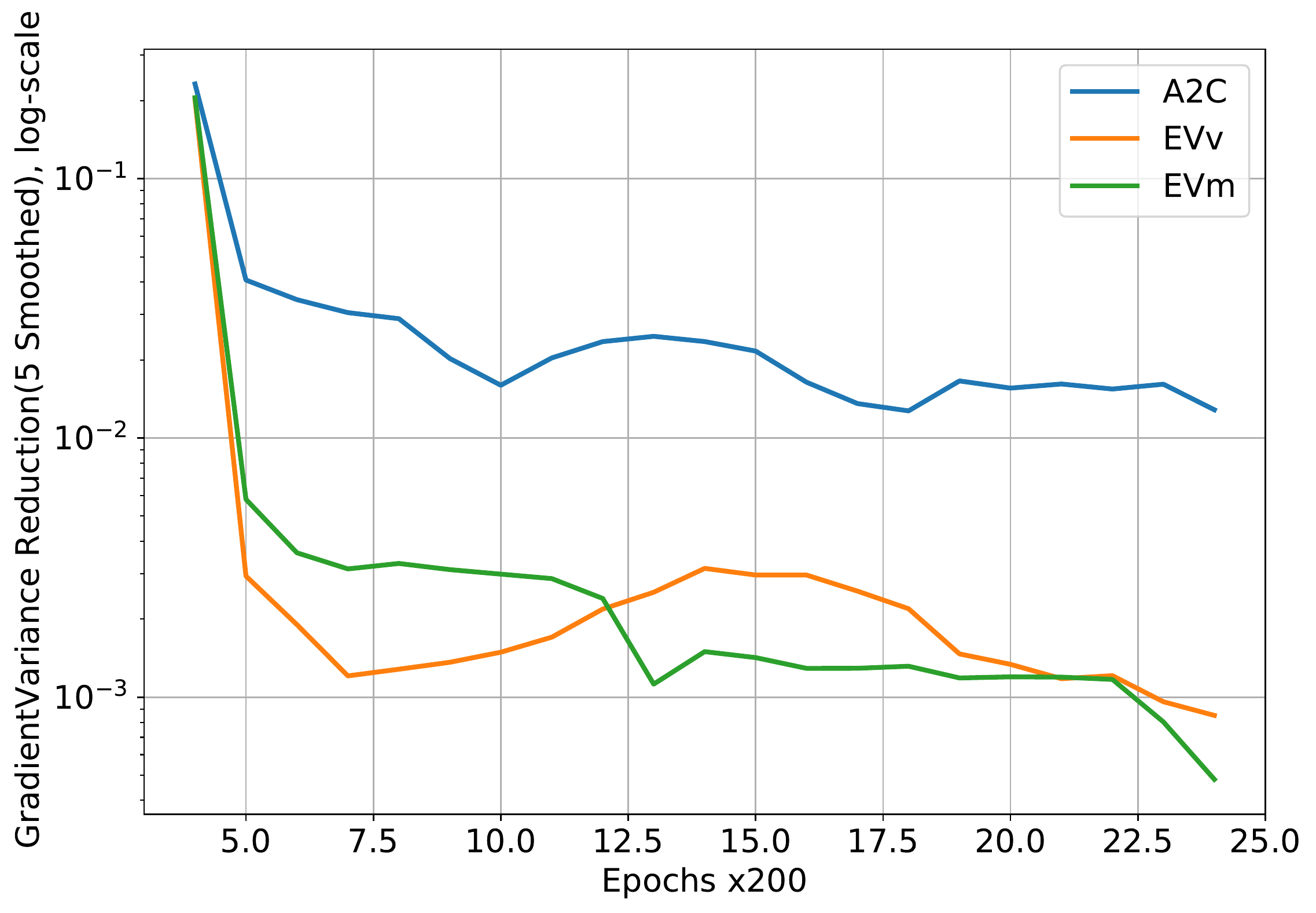} \\
    \end{tabular}
    \caption{The charts representing the results for \texttt{CartPole} environment: (a,b,c) represent mean rewards, standard deviation of the rewards and gradient variance reduction ratio for config5 and (d,e,f) show the same information about config8.}
    \label{fig:plotsCartpole}
\end{figure}

\subsection{Algorithm Performance}

While observing mean rewards during the training we may notice immediately that EV-algorithms are at least as good as A2C.\\

In \texttt{CartPole} environment (Fig. \ref{fig:plotsCartpole} ) we conducted several experiments and present here two policy configurations: one with simpler neural network (config5, see Fig. \ref{fig:plotsCartpole}(a,b,c) ) and one with more complex network (config8, see Fig. \ref{fig:plotsCartpole}(d,e,f) ). In the first case both A2C and EV have very similar performance but in the second case the agent learns considerably faster with EV-based variance reduction  and we get approximately 50\% improvement over A2C agent and 75\% over Reinforce agent in the end and even more during the training. The phenomenon of better performance of EV in CartPole with more complex policies is observed often, more detailed discussion is placed in Supplementary.\\

\begin{figure}[h!]
    \centering
    \begin{tabular}{lcr}
    (a) \includegraphics[scale=.2]{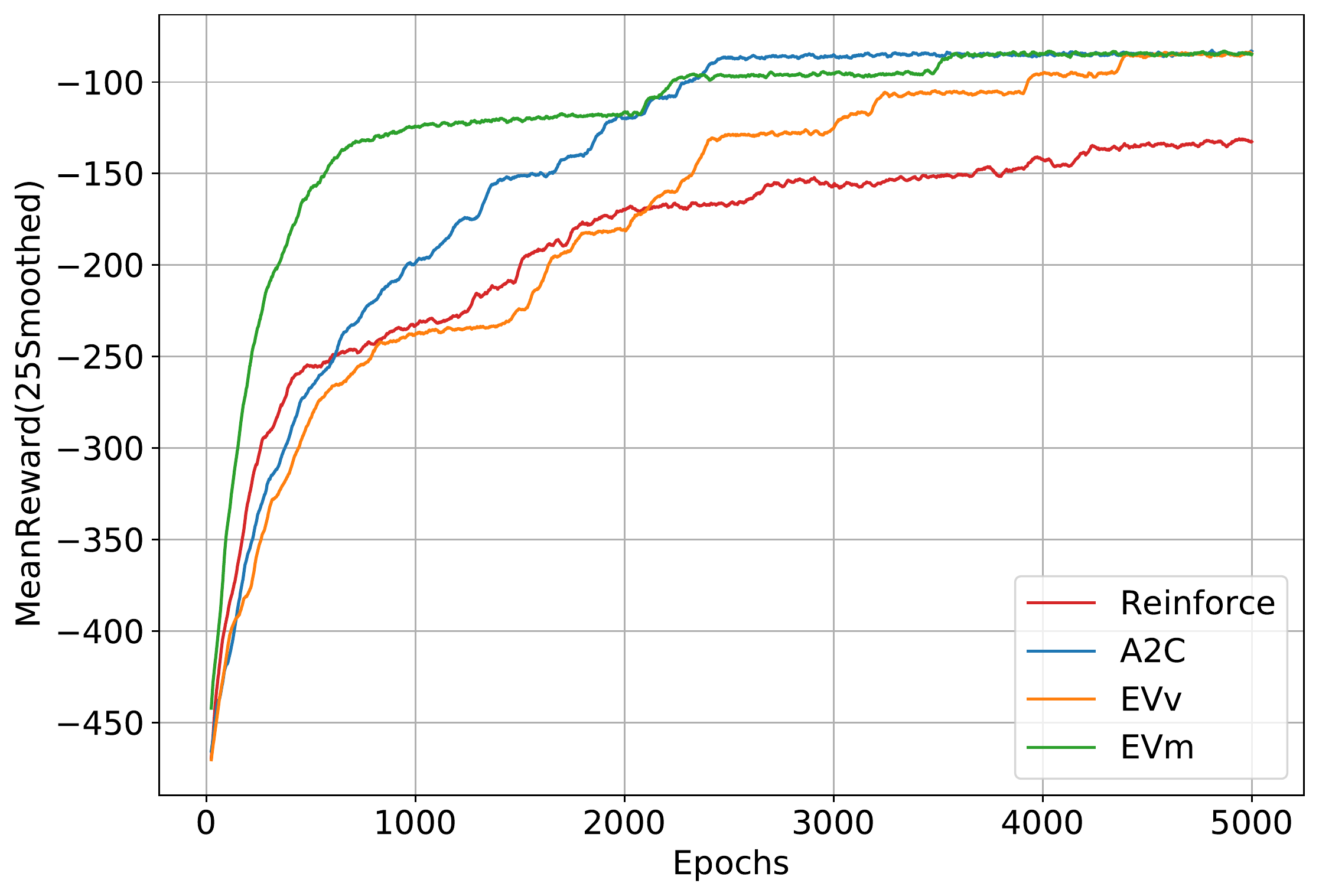} & (b) \includegraphics[scale=.2]{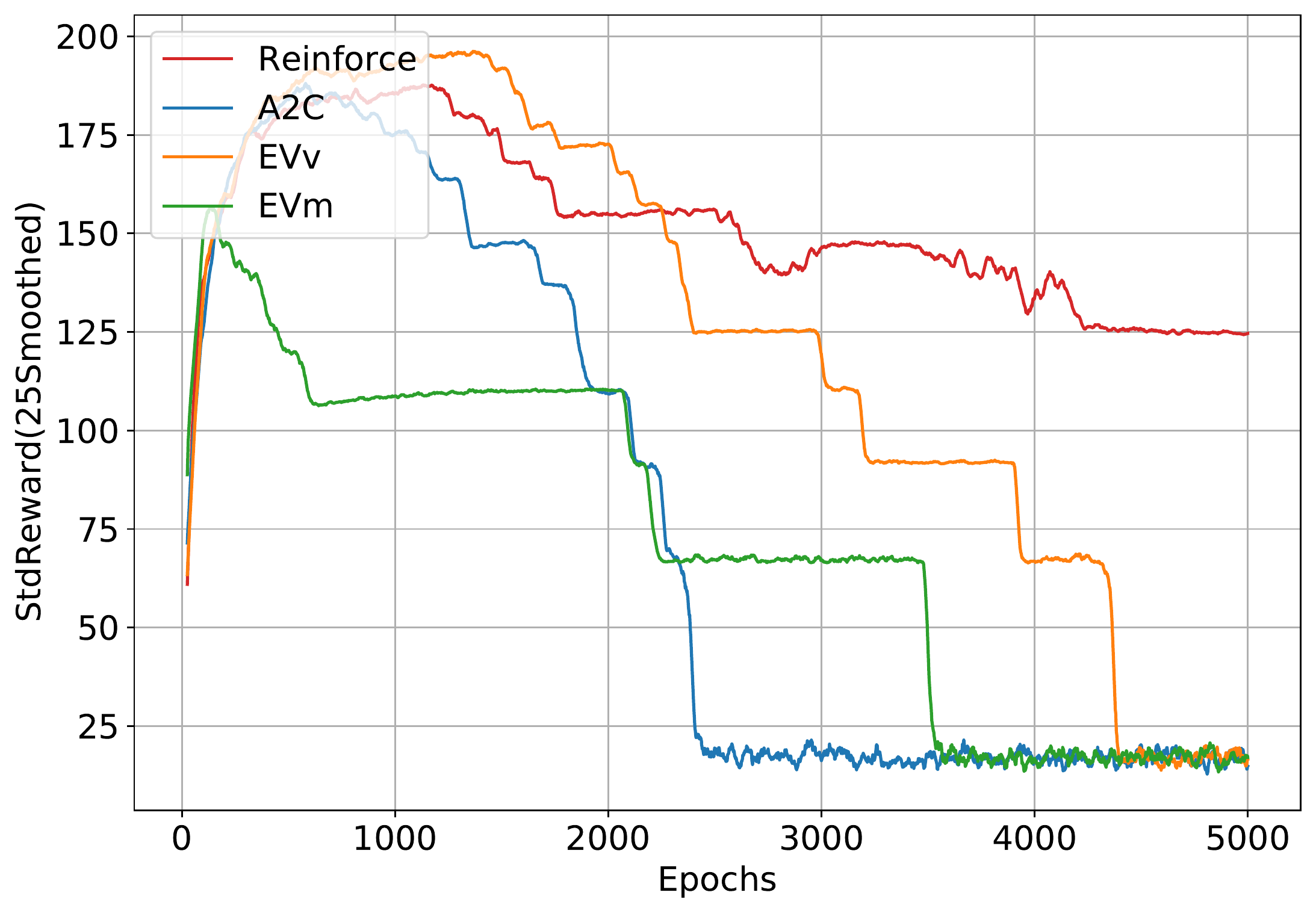} & (c) \includegraphics[scale=.2]{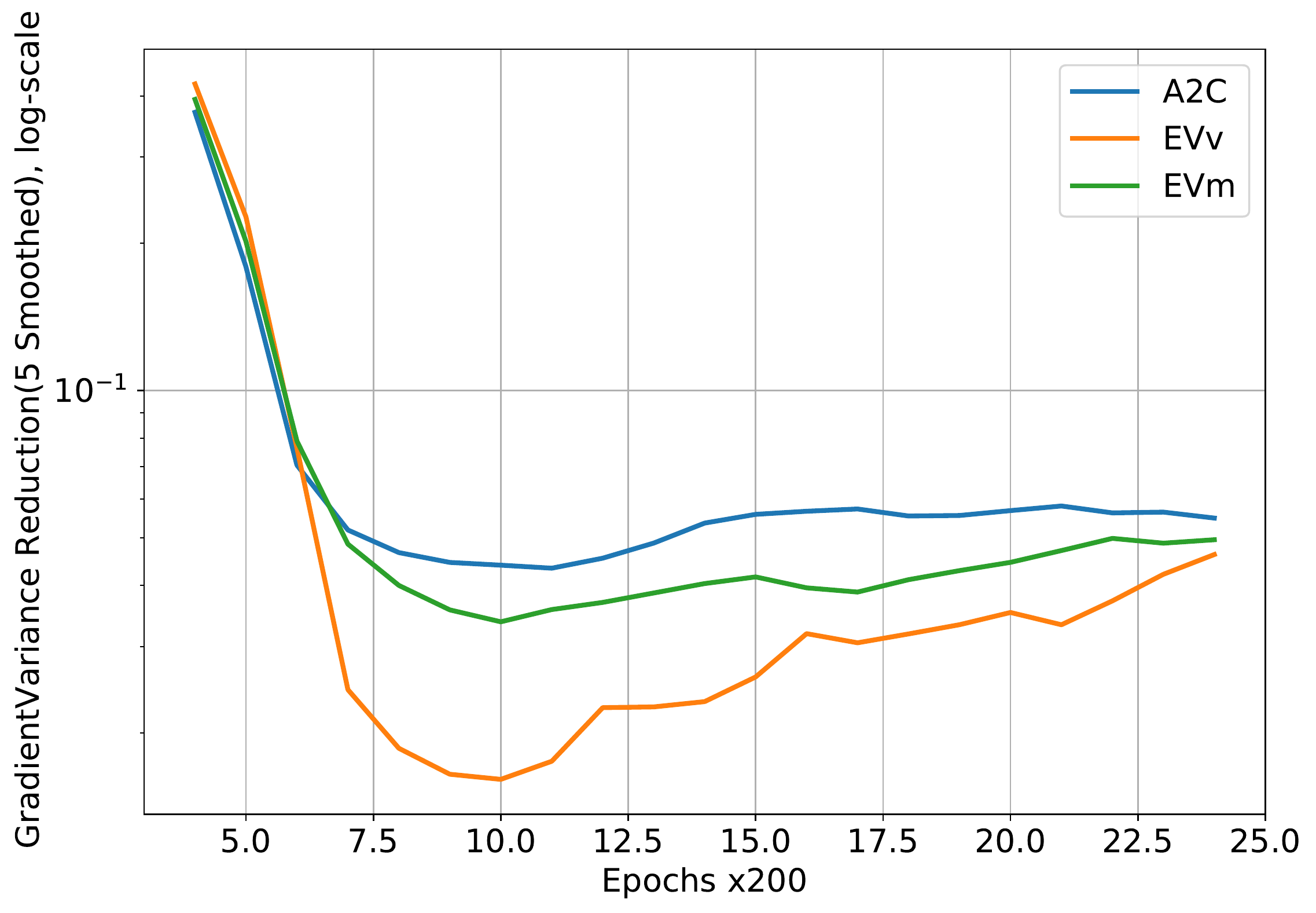} \\
    \end{tabular}
    \caption{The charts representing the results for \texttt{Acrobot} environment: (a) depicts mean rewards, (b) shows the standard deviations of the rewards and (c) displays the gradient variance reduction ratios.}
    \label{fig:plotsAcrobot}
\end{figure}

Experiments in  \texttt{Acrobot} (see Fig. \ref{fig:plotsAcrobot}(a)) show that EV-algorithms can give better speed-up in the training. In the beginning EVm allows to learn faster but in the end the performance is the same as A2C. One of the reasons of such behavior can be the fact that learning rate becomes small and the agent already reaches the ceiling.\\

\begin{figure}[h!]
    \centering
    \begin{tabular}{lcr}
    (a) \includegraphics[scale=.2]{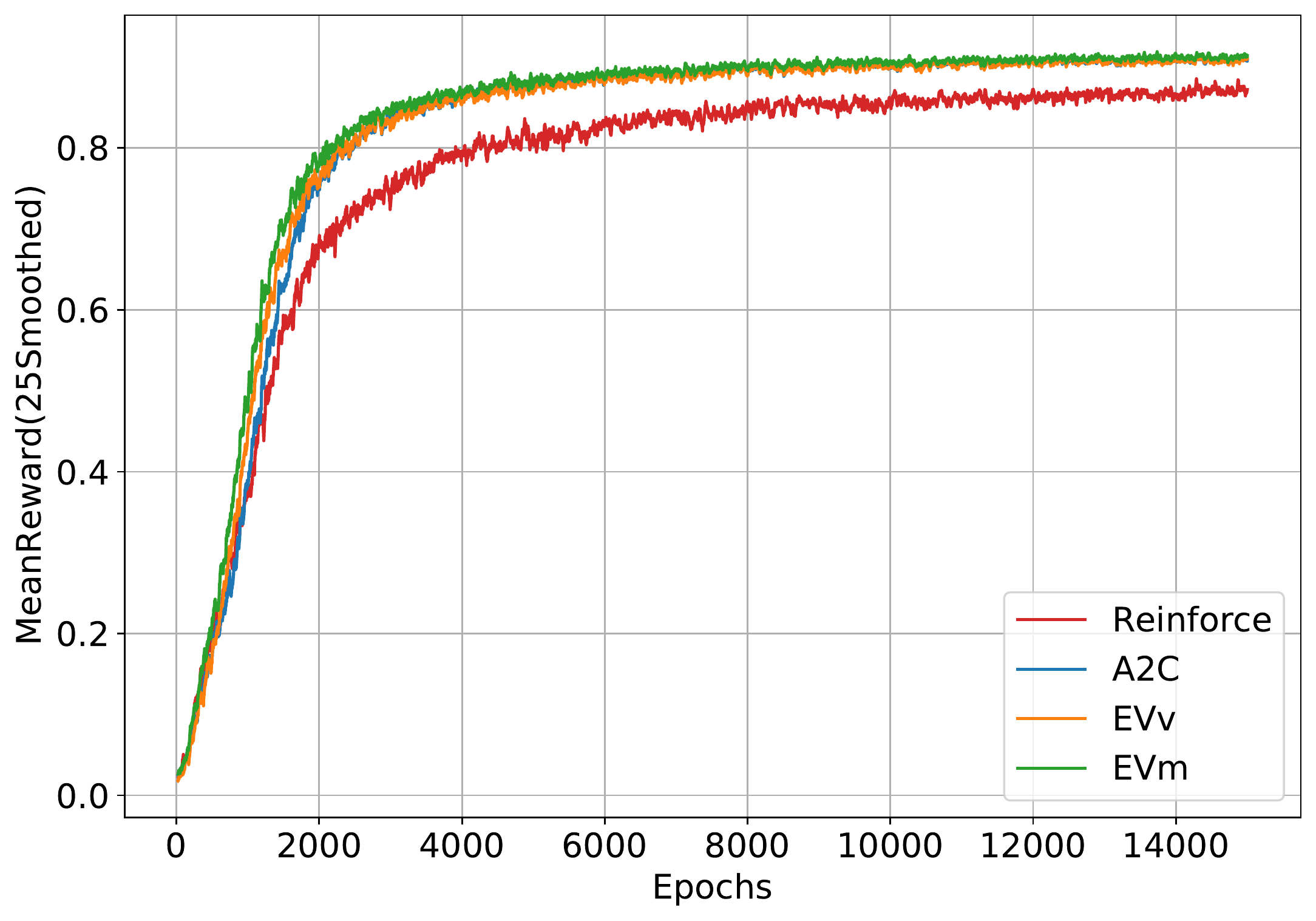} & (b) \includegraphics[scale=.2]{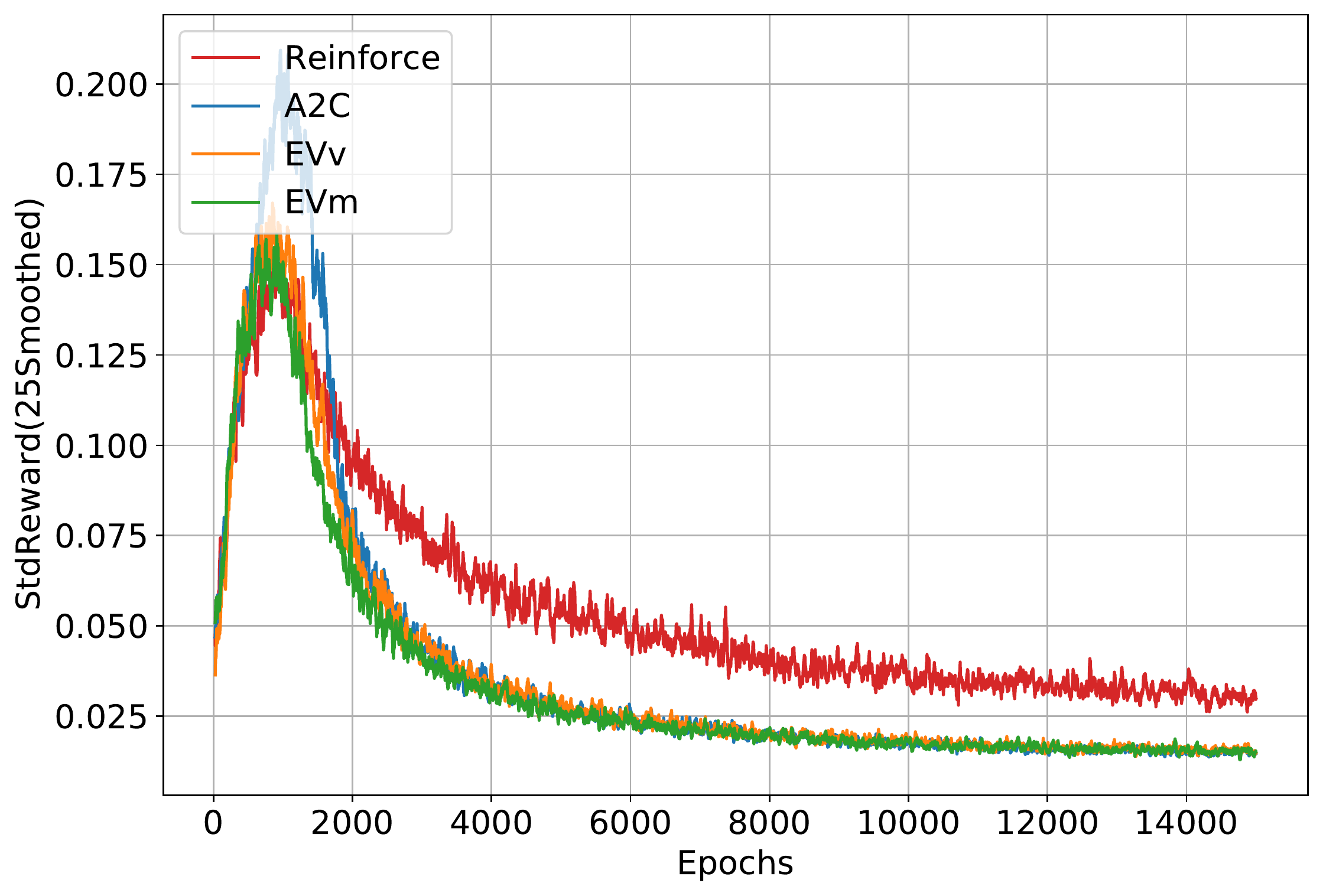} & (c) \includegraphics[scale=.2]{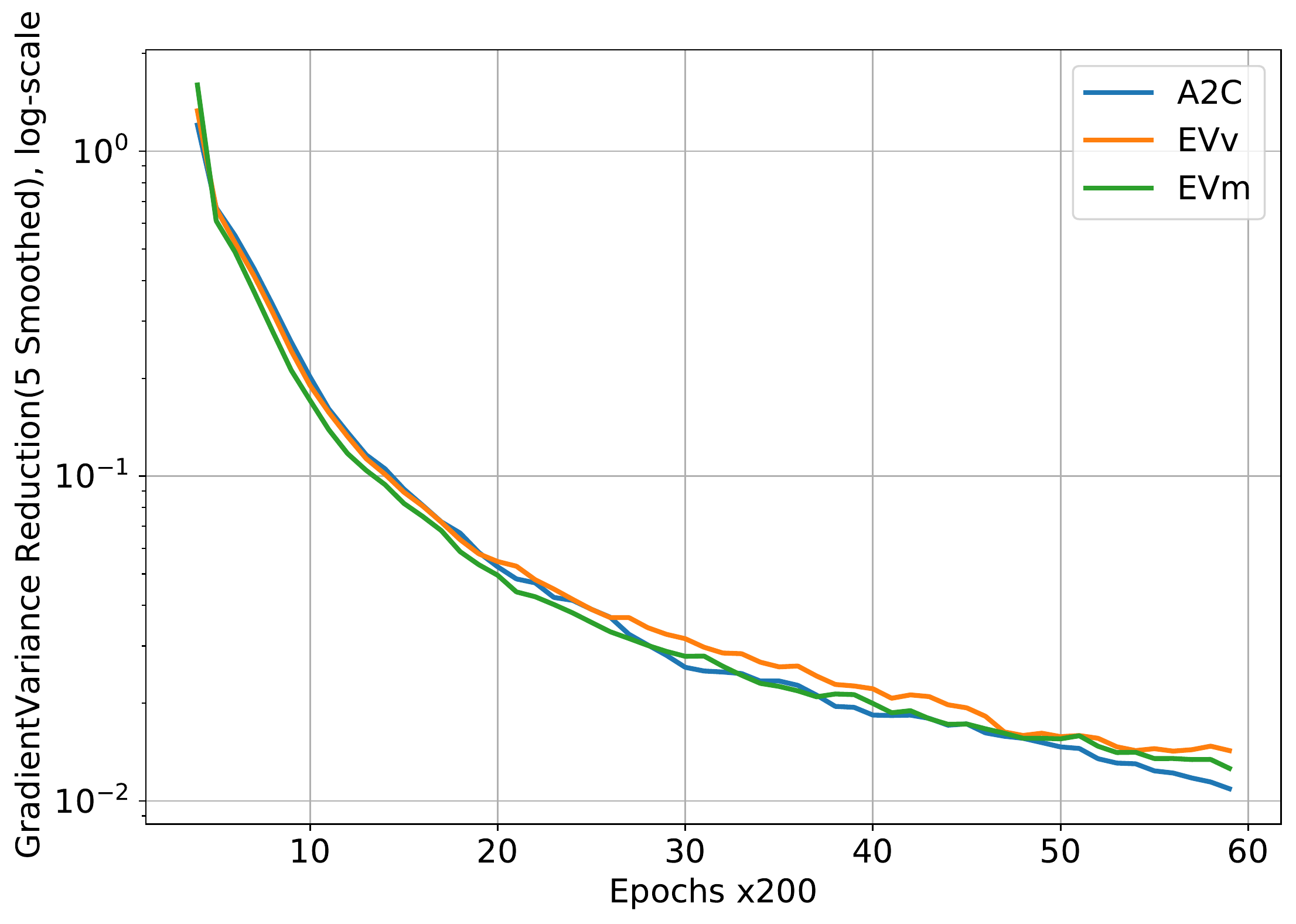} \\
    \end{tabular}
    \caption{The charts representing the results for \texttt{Unlock} environment: (a) depicts mean rewards, (b) shows the standard deviations of the rewards and (c) displays the gradient variance reduction ratios.}
    \label{fig:plotsUnlock}
\end{figure}

\texttt{Unlock} (Fig. \ref{fig:plotsUnlock}(a)) is the example of the environments where all algorithms work similarly: in terms of rewards we cannot see significant improvement even over Reinforce. In \texttt{Unlock}, however, there is a difference presented but very small. 

\subsection{Stability of Training}

When we study the charts for standard deviation of the rewards (Fig. \ref{fig:plotsCartpole}(b,e),\ref{fig:plotsAcrobot}(b),\ref{fig:plotsUnlock}(b)), we can see that EV-methods are better in terms of stability of the training, the algorithm more rarely has drops than that of A2C. This is greatly illustrated by \texttt{CartPole} in Fig. \ref{fig:plotsCartpole}(b,e) where the standard deviation is about 2 times less than in case of A2C. This holds for both configurations.\\

Fig. \ref{fig:plotsAcrobot} illustrating the experiments with \texttt{Acrobot} show that until the ceiling is reached EV methods still can have lower variance. In \texttt{Unlock} presented in Fig. \ref{fig:plotsUnlock}(b) we have not observed a significant difference in reward variance.

\subsection{Gradient Variance and its Influence}

The first thing we can notice reviewing the gradient variance is that A2C and EV reduce the variance similarly in \texttt{Unlock}. \texttt{CartPole} (see Fig. \ref{fig:plotsCartpole}(c,f)), however, gives an example of the case where EV works completely differently to A2C, it reduces the variance almost 100-1000 times in both policy configurations. Similar picture we can observe in all \texttt{CartPole} experiments. \\

We can see that in \texttt{Unlock} showed in Fig. \ref{fig:plotsUnlock} the variance can also be reduced approximately 10-100 times, however, we see very little gain in rewards. It shows that in some environments training does not respond to the variance reduction; as a reason, it can be just not enough to give the improvement.\\

As answer to the discussion \cite{tuckerMirage2018} about whether variance reduction helps in training we would note the following phenomenon. In all cases we have confirmed variance reduction but not everywhere we have seen different performance of A2C, Reinforce and EV. Note that we designed our experiment in such a way that the only thing differentiating the agents is the goal function for baseline training. Some environments due to their specific setting and structure just do not respond to this variance reduction. In some cases (like in \texttt{Acrobot}) we can see that there are moments in training where variance reduction helps and where it does not change anything or even make training slower. It is natural to suppose that all these specific features should be addressed by some training algorithm which would combine in a clever way several variance reduction techniques or would decide that variance reduction is not needed at all. The last thing can be vital for exploration properties. Hence, we would conclude that variance reduction is the technique for improvement but how and when to apply it during the training is an interesting open question.\\

The last thing we would like to note is that reward variance measured in previous sub-section is not an indicator of variance reduction since we have shown gradient variance reduction in all cases. Reward variance is decreased in relation to Reinforce, however, only in \texttt{CartPole} environement. Therefore, it cannot be used as a key metric for studying variance reduction in RL. The connection between reward variance and gradient variance seems to be an unanswered question in the literature.

%% file: conclusion.tex

In conclusion, we would like to state that variance reduction is a useful technique which can be used to improve the quality of policy-gradient algorithms. Sometimes, the desired effect cannot be reached due to the specific nature of the environment. However, as we have seen above, it has the potential to influence the training process in a good way.\\

As a new method for constructing variance reduction goals we suggested to use empirical variance which in turn resulted in EV-methods. Their motivation is more about actual variance reduction than in case of A2C and their performance is at least as good as A2C in terms of variance reduction and rewards. For them we also have suggested a probabilistic bound for the variance of the gradient estimate under some mild assumptions showing that when more trajectories are available the reduction can be better. Finally, EV-algorithms can be more stable in training which can allow to make sudden drops during the training less frequent.\\

We also have for the first time presented the study of actual gradient variance reduction in classic benchmark problems. Our results have shown that variance reduction can help in the training but sometimes the environment's specific features do not allow to achieve gain in rewards. Therefore, variance reduction technique needs to be used during the training but the exact circumstances in which it helps are yet to be discovered.

%% file: acknowledgements.tex
This research was supported with the computational resources of the HPC facilities at HSE University.

%% file: supplement.tex
%

%


\begin{Huge} \textbf{Supplementary Materials} \end{Huge}

In Supplementary Materials we provide all missing proofs, additional experiments and their details. The code used is available on GitHub \cite{github}.

\section{Proofs}

\subsection{Verification of the Assumptions}

\subsubsection{Proof of Proposition 3}

\input{prop3Proof}

\subsubsection{Proof of Proposition 4}

\input{prop4Proof}

\subsection{Proof of Proposition 5: A2C as an Upper Bound for EV}

\input{prop5Proof.tex}

\subsection{Proof of the Main Theorem}

\input{mainTheoremProof.tex}

\subsection{Proposition: Unbiasedness of S-Baseline}

\input{sbaselineUnbiased}

\subsection{Why Variance Reduction Matters (for Local Convergence)}

\input{reinfConvergence.tex}

\clearpage
\newpage
\section{Additional Experiments and Implementation Details}

Here we present additional experimental results. The detailed config-files can be found on GitHub page \cite{github}.

\input{suppMinigrid.tex}

\clearpage
\subsection{OpenAI Gym: Cartpole-v1}
\input{suppCartPole}

\clearpage
\subsection{OpenAI Gym: LunarLander-v2}
\input{suppLunarLander}

\clearpage
\subsection{OpenAI Gym: Acrobot-v1}
\input{suppAcrobot}

\clearpage
\subsection{Time Complexity Discussion}
In Figures \ref{fig:sup_GoToDoor_elapsedtime} - \ref{fig:sup_CartPole_time_fromConfigId} we demonstrate how training time depends on processed transitions for all environments. One can clearly see that EV algorithms are more time-consuming and sensitive to the growth of the sample size $K$ but its excessive training time mostly can be explained by the implementation. PyTorch-compatibility requires that we have to make $K$ extra backpropagations in order to compute empirical variance. We believe that this part of computations can be optimized and, therefore, accelerated in practice. \\

Scatter plots demonstrate how consumed time depends on the number of processed data. This allows better understanding of the processing cost of one transition from the simulated trajectory. We also provide the measured execution times per transition in box plots to see the difference between all considered algorithms regardless of the trajectory length. We used high-performance computing units with the same computation powers for each run inside one environment, so that these measurements were accurate and comparable. \\

Summing up, considering all the advantages of EV algorithms, they have higher time costs (see also Figures \ref{fig:sup_GoToDoor_time_fromK} - \ref{fig:sup_CartPole_time_fromConfigId}) and demand more specific implementation allowing faster computation of many gradients which currently cannot be easily developed in the framework of PyTorch. PyTorch allows great flexibility and very general models for the approximations of policy and baseline; if these are more specific, our algorithm can be implemented to be more effective.
\input{charts_sup_time}










%% file: prop3Proof.tex
\begin{proposition}
If there exist constants $C_L>0$ and $C_R>0$ such that 
\begin{align*}
\forall \theta \in \Theta, ~a\in \mathcal{A}, s\in \mathcal{S}, b_\phi \in \mathcal{B}_\Phi \quad &\Vert \nabla_\theta \log \pi(a \vert s)\Vert \leq C_L, \\
&\vert R(s,a)\vert \leq C_R,\\
&\vert b_\phi(s,a)\vert \leq C_R,
\end{align*}
then Assumption 1 is satisfied.
\label{prop:boundedrewgrad}
\end{proposition}

$\triangleright$ Note that the class of estimators $\mathcal{H}$ in the gradient scheme consists of the maps 
\[
\widetilde{\nabla}^{b_\phi}_{\theta} J: X \mapsto \sum_{t=0}^{T-1} \gamma^t (G_t-b_\phi(S_t,A_t)) \nabla_\theta \log \pi(A_t \vert S_t).
\]
Therefore,
\[
\Vert \widetilde{\nabla}^{b_\phi}_{\theta} J(X)\Vert \leq \sum_{t=0}^{T-1} \gamma^t \vert G_t(X) - b_\phi(S_t,A_t)\vert \Vert \nabla_\theta \log \pi(A_t \vert S_t)\Vert \leq \frac{2 C_R C_L}{1-\gamma}
\]
and in the case $\gamma=1$
\[
\Vert \widetilde{\nabla}^{b_\phi}_{\theta} J(X)\Vert \leq 2 C_R C_L T.
\]
$\square$

%% file: prop4Proof.tex
\begin{proposition}
Suppose that Assumption 3 holds for $\mathcal{B}_\phi$, i.e.
\[
\mathcal{N}(\epsilon, \mathcal{B}_\Phi, \Vert \cdot \Vert_{L^2(P_K)}) \leq \left( \frac{c}{\epsilon}\right)^\alpha
\]
for some $c,\alpha>0$. If there exist constant $C_L>0$ such that 
\begin{align*}
\forall \theta \in \Theta, ~a\in \mathcal{A}, s\in \mathcal{S} \quad &\Vert \nabla_\theta \log \pi(a \vert s)\Vert \leq C_L,
\end{align*}
then Assumption 3 holds also for $\mathcal{H}$ with the same constant $\alpha'=\alpha$ and constant $c'=c C_L \sqrt{2/(1-\gamma^2)}$.
\end{proposition}

$\triangleright$ Let us fix $\epsilon'>0$ and consider two estimators from $\mathcal{H}$: $\widetilde{\nabla}^{b_\phi}J$ and $\widetilde{\nabla}^{b_{\phi'}}J$ which is a member of the $\epsilon'$-net of $\mathcal{B}_\Phi$, in other words, such that 
\[
\Vert b_{\phi} - b_{\phi'} \Vert_{L^2(P_K)}:= \sqrt{P_K(b_{\phi} - b_{\phi'})^2} \leq \epsilon'.
\]
Recall that
\[
\left\Vert \widetilde{\nabla}^{b_\phi}J - \widetilde{\nabla}^{b_{\phi'}}J\right\Vert_{L^2(P_K)} = \sqrt{ P_K \left\Vert \widetilde{\nabla}^{b_\phi}J - \widetilde{\nabla}^{b_{\phi'}}J\right\Vert_2^2}
\]
and let us bound $\left\Vert \widetilde{\nabla}^{b_\phi}J(X_i) - \widetilde{\nabla}^{b_{\phi'}}J(X_i)\right\Vert_2$ for some arbitrary $i=1,..,K$. We could derive
\[
\left\Vert \widetilde{\nabla}^{b_\phi}J(X_i) - \widetilde{\nabla}^{b_{\phi'}}J(X_i)\right\Vert_2 \leq  C_L \sum_{t=0}^{T-1} \gamma^t \left\vert b_\phi(S_t^{(i)},A_t^{(i)}) - b_{\phi'}(S_t^{(i)},A_t^{(i)})\right\vert
\]
and, thus, it leads to
\begin{align*}
&P_K \left\Vert \widetilde{\nabla}^{b_\phi}J - \widetilde{\nabla}^{b_{\phi'}}J\right\Vert_2^2 \leq 2C_L^2 \frac{1}{K} \sum_{i=1}^K \sum_{t=0}^{T-1}\gamma^{2t} \left\vert b_\phi(S_t^{(i)},A_t^{(i)}) - b_{\phi'}(S_t^{(i)},A_t^{(i)})\right\vert^2\leq\\
&\leq 2C_L^2  \sum_{t=0}^{T-1}\gamma^{2t} \frac{1}{K} \sum_{i=1}^K \left\vert b_\phi(S_t^{(i)},A_t^{(i)}) - b_{\phi'}(S_t^{(i)},A_t^{(i)})\right\vert^2 \leq \frac{2C_L^2 \epsilon'^2 }{1-\gamma^2}.
\end{align*}
This allows us to use the $\epsilon'$-net for $\mathcal{B}_\Phi$ to construct $\epsilon$-net for $\mathcal{H}$. Hence, Assumption \ref{ass:polycover} is satisfied with $\alpha'=\alpha$ and $c'= c C_L \sqrt{2/(1-\gamma^2)}$.
$\square$\\

Let us briefly remark that the Proposition allows transferring any covering assumption for baselines to the vector setting and so one could use the assumptions for baselines which are much easier to check in practice.

%% file: prop5Proof.tex
\begin{proposition}
If there exist constant $C_L>0$ such that 
\begin{align*}
\forall \theta \in \Theta, ~a\in \mathcal{A}, s\in \mathcal{S} \quad &\Vert \nabla_\theta \log \pi(a \vert s)\Vert \leq C_L,
\end{align*}
then for all $K\geq 2$ A2C goal function $V^{A2C}_K(\phi)$ is an upper bound (up to a constant) for EV goal functions:
\[
V^{EVm}_K(\phi) \leq 2 C_L^2 V^{A2C}_K(\phi), \quad V^{EVv}_K(\phi) \leq 2 C_L^2 V^{A2C}_K(\phi).
\]

\end{proposition}

$\triangleright$ First, note that for all $\phi$, by Jensen's inequality, $V^{EVv}_K(\phi) \leq V^{EVm}_K(\phi)$, so we could work with the bound for EVm. Secondly, $K=1$ simply does not allow using EVv-criterion, but the bound for EVm remains valid. Via Young's inequality we get
\[
V^{EVm}_K(\phi) \leq \frac{2}{K}\sum_{k=1}^K \sum_{t=0}^{T-1} \gamma^{2t} \left(G_t(X^{(k)})-b_\phi(S_t^{(k)},A_t^{(k)}) \right)^2 \left\Vert \nabla_\theta \log \pi\left(A_t^{(k)} \vert S_t^{(k)}\right) \right\Vert_2^2 \leq 2C_L^2 V^{A2C}_K(\phi).
\]
$\square$

%% file: mainTheoremProof.tex
Suppose we are given sample $X,X_1,...,X_K$ of random vectors taking values in $\mathcal{X} \subset \mathds{R}^d$ and $\mathcal{H}:= \left\lbrace h: \mathcal{X} \to \mathds{R}^D ~~s.t.~~ \mathds{E}[h(X)]=\mathcal{E}\right\rbrace$. Also denote $\Vert \cdot \Vert := \Vert \cdot \Vert_2$ for shorter notation, when applied to function $h: \mathcal{X} \to \mathds{R}^D$, $\Vert h \Vert := \sup_{x \in \mathcal{X}} \Vert h(x) \Vert$ by default. The brackets $(\cdot,\cdot)$ denote the standard inner product.\\

Our goal is to find
\[
h_* \in \text{argmin}_{h \in \mathcal{H}} V(h)
\]
with variance functional defined as
\[
V(h):=\mathds{E}\left[ \Vert h(X) - \mathcal{E} \Vert_2^2\right].
\]

In order to tackle this problem we consider the simpler one  called \textit{Empirical Variance(EV)} and calculate
\[
h_K \in \text{argmin}_{h \in \mathcal{H}} V_K(h)
\]
with
\[
V_K(h):=\frac{1}{K-1}\sum_{k=1}^K \Vert h(X_k) - P_K h\Vert^2,
\]
where $P_K$ is the empirical measure based on $X_1,..,X_K$, so $P_K h = \frac{1}{K} \sum_{k=1}^K h(X_k)$. In what follows we will operate with several key assumptions about the problem at hand.
\begin{assum}
Class $\mathcal{H}$ consists of bounded functions: 
\[
\forall h \in \mathcal{H}~~ \sup_{x\in \mathcal{X}}\Vert h(x) \Vert\leq b.
\] 
\label{ass:bounded}
\end{assum}
\begin{assum}
The solution $h_*$ is unique and $\mathcal{H}$ is star-shaped around $h_*$: 
\[
\forall h \in \mathcal{H},~\alpha \in [0,1] \quad \alpha h + (1-\alpha)h_* \in \mathcal{H}.
\]
\label{ass:starshape}
\end{assum}

Star-shape assumption replaces the assumption of the convexity of $\mathcal{H}$ which is stronger and yet this replacement does not change much in the analysis.

\begin{assum}
Class $\mathcal{H}$ has covering of polynomial size: there are $\alpha \geq 2$ and $c>0$ such that for all $u \in (0,b]$
\[
\mathcal{N}( \mathcal{H}, \Vert \cdot \Vert_{L^2(P_K)},u) \leq \left( \frac{c}{u}\right)^\alpha ~a.s.
\]
where the norm is defined as
\[
\Vert h \Vert_{L^2(P_K)}= \Vert h \Vert_{(K)}  := \sqrt{ P_K \Vert h\Vert_2^2}
\]
\label{ass:polycover}
\end{assum}

The basis of the analysis lies in usage of
\begin{lemma}
(Lemma 4.1 in \cite{belomEVM2017}) Let $\left\lbrace \phi(\delta):~\delta \geq 0 \right\rbrace$ be non-negative r.v. indexed by $\delta \geq 0$ such that a.s. $\phi(\delta) \leq \phi(\delta')$ if $\delta \leq \delta'$. Define $\left\lbrace  \beta(\delta,t): \delta \geq 0, t \geq 0 \right\rbrace$, deterministic real numbers such that
\[
\mathds{P}(\phi(\delta)\geq \beta(\delta,t)) \leq e^{-t}.
\]
Set for all non-negative $t$
\[
\beta(t):= \inf \left\lbrace \tau >0 ~: ~ \sup_{\delta \geq \tau} \frac{\beta(\delta,t\delta/\tau)}{\delta} \leq \frac{1}{2} \right\rbrace.
\]
If $\hat{\delta}$ is a non-negative random variable which is a priori bounded and such that almost surely $\hat{\delta} \leq \phi(\hat{\delta})$, then for all $t \geq 0$
\[
\mathds{P}\left( \hat{\delta} \geq \beta(t) \right) \leq 2e^{-t}.
\]
\label{lemm:koltch}
\end{lemma}

We would like to stress out that the main idea of the proof remains the same but on the way there must be done some changes to fit it into the setting of vector estimation we consider.

\subsubsection{Bound for Functions with $\delta$-Optimal Variance}

The idea is to construct an upper bound with high probability for excess risk $V(h_K)-V(h_*)$ under assumptions $V(h)-V(h_*)\leq\delta$ and use that as $\phi$ in Lemma \ref{lemm:koltch}. This will give us the desired w.h.p. bound for excess risk in general. Let us start with the basic bound from which we obtain all further results. Essentially, the sequence $\phi(\delta)$ from the Lemma appears in the left part.

\begin{theorem}
Assume A\ref{ass:bounded}, A\ref{ass:starshape}. If $h \in \mathcal{H}(\delta):= \left\lbrace h \in \mathcal{H} ~\vert~ V(h)-V(h_*)<\delta \right\rbrace$, then with probability at least $1-e^{-t}$
\[
V(h_K)-V(h_*) \leq 2 \mathds{E} \phi_K^{(1)} (\delta) + 4 \left( \mathds{E}\sup_{h \in \mathcal{H}(\delta)}\Vert(P-P_K)h \Vert\right)^2 +  \frac{40b^2t+24b^2}{3K} + 12b \sqrt{\frac{\delta t}{K}} 
\]
with 
\begin{align*}
&\phi_K^{(1)}(\delta):= \sup_{h \in \mathcal{H}(\delta)} (P-P_K)l(h).
\end{align*}
\label{th:basicBound}
\end{theorem}
$\triangleright$ To begin with, add and subtract $V_K(h_K),V_K(h_*)$ to get
\[
V(h_K) - V(h_*) \pm V_K(h_K) \pm V_K(h_*) \leq V(h_K) - V(h_*) - (V_K(h_K)-V_K(h_*)).
\]
the last terms can be represented as
\[
V_K(h)= P_K \Vert h-\mathcal{E} \Vert_2^2 - \frac{1}{K(K-1)} \sum_{i\neq j =1}^K \left( h(X_i) - \mathcal{E}, h(X_j) - \mathcal{E} \right)
\]
giving us
\begin{align*}
&V(h_K) - V(h_*) - (V_K(h_K)-V_K(h_*))=\\
&= (P-P_K)\left( \Vert h_K- \mathcal{E} \Vert_2^2 - \Vert h_*- \mathcal{E} \Vert_2^2 \right) +\\
&+\frac{1}{K(K-1)}\sum_{i \neq j =1}^K \left( h(X_i) - \mathcal{E}, h(X_j) - \mathcal{E} \right) - \left( h_*(X_i) - \mathcal{E}, h_*(X_j) - \mathcal{E} \right)=\\
&= T_K(h_K) + W_K(h_K)
\end{align*}
which introduces
\begin{align*}
&T_K(h_K):=(P-P_K)\left( \Vert h_K- \mathcal{E} \Vert_2^2 - \Vert h_*- \mathcal{E} \Vert_2^2 \right),\\
&W_K(h_K):= w(h_K) - w(h_*), ~ w(h)=\frac{1}{K(K-1)}\sum_{i \neq j =1}^K \left( h(X_i) - \mathcal{E}, h(X_j) - \mathcal{E} \right).
\end{align*}
Since $h \in \mathcal{H}(\delta)$ it is true that
\[
V(h_K)-V(h_*) \leq \sup_{h \in \mathcal{H}(\delta)} T_K(h) + W_K(h) \leq \sup_{h \in \mathcal{H}(\delta)} T_K(h) + \sup_{h \in \mathcal{H}(\delta)} W_K(h)= \phi_K^{(1)}(\delta) + \phi_K^{(2)}(\delta).
\]
\textbf{Bound for $\phi_K^{(1)}$.} Firstly, let us introduce
\[
l(h)= \Vert h - \mathcal{E}\Vert_2^2 - \Vert h_* - \mathcal{E}\Vert_2^2.
\]
We can exploit the same Talagrand's inequality as in \cite[p.12]{belomEVM2017}. Recall that functions $h \in \mathcal{H}$ are bounded, therefore $\vert l(h)\vert \leq 4b^2$ and, hence, with probability at least $1-e^{-t}$
\[
\phi_K^{(1)}(\delta) \leq \mathds{E}\phi_K^{(1)}(\delta) + \sqrt{\frac{2t}{K}\left( \sigma^2(\delta) + 8b^2 \mathds{E}\phi_K^{(1)}(\delta)\right)} + \frac{4b^2 t}{3K},
\]
where 
\[
\sigma^2(\delta):= \sup_{h \in \mathcal{H}(\delta)} P l(h)^2.
\]
Let us bound this quantity. In order to proceed, notice that for all $h_1,h_2\in \mathcal{H}$
\[
l(h_1) - l(h_2) = (h_1,h_1) - (h_2,h_2) +2(h_2-h_1,\mathcal{E})=(h_2-h_1,h_2-h_1) + 2(h_1-h_2,h_2-\mathcal{E})
\]
and so for all $x \in \mathcal{X}$
\begin{equation}
    \vert l(h_1)(x) - l(h_2)(x) \vert \leq 6b \Vert h_2(x) - h_1(x) \Vert
\label{eq:lip:l}
\end{equation}
is obtained with Cauchy-Schwarz inequality. This results particularly in
\[
P l(h)^2 \leq 36b^2 P \Vert h-h_* \Vert_2^2.
\]
Since $l(h)$ has very specific form involving square norms, we could state that
\[
P\Vert h-h_*\Vert_2^2 = 2Pl(h) - 4Pl\left(\frac{h+h_*}{2}\right) \leq 2Pl(h)
\]
implying
\[
Pl(h)^2 \leq 72b^2 Pl(h) \leq 72b^2 \delta
\]
by definition of $\mathcal{H}(\delta)$.\\

With this and $\sqrt{u+v} \leq \sqrt{u} +\sqrt{v},~2\sqrt{uv} \leq u+v$ the bound can be simplified to
\[
\phi_K^{(1)}(\delta) \leq 2\mathds{E}\phi_K^{(1)}(\delta) + 12b\sqrt{\frac{\delta t}{K}} + \frac{16b^2 t}{3K}.
\]

\textbf{Bound for $\phi_K^{(2)}$.} This is much simpler, observe that
\[
w_K(h)= \frac{1}{K(K-1)}\left\lbrace \sum_{i,j=1}^K \left( h(X_i) - \mathcal{E}, h(X_j) - \mathcal{E} \right) - \sum_{i=1}^K \Vert h(X_i) - \mathcal{E}\Vert^2 \right\rbrace=
\]
or,
\[
= \frac{K}{K-1} \left(P_K(h-\mathcal{E}),P_K(h-\mathcal{E}) \right) - \frac{1}{K-1}P_K\Vert h - \mathcal{E}\Vert_2^2.
\]
So,
\[
W_K(h_K)=w_K(h_K)-w_K(h_*)\leq \frac{K}{K-1} \left(P_K(h-\mathcal{E}),P_K(h-\mathcal{E}) \right) + \frac{1}{K-1}P_K\Vert h_* - \mathcal{E}\Vert_2^2 \leq
\]
where the first inequality is due to negative terms, applying bound for $h$ now results in
\[
\leq \frac{K}{K-1} \left(P_K(h-\mathcal{E}),P_K(h-\mathcal{E}) \right) + \frac{4b^2}{K-1}.
\]
Finally,
\[
\phi_K^{(2)} \leq 2 \left( \sup_{h \in \mathcal{H}(\delta)} \Vert (P-P_K)h \Vert \right)^2 + \frac{4b^2}{K-1}
\]
and after adding and subtracting the expectation of the supremum and exploiting $2ab \leq a^2 + b^2$ together with $1/(K-1) \leq 2/K$ for $K\geq 2$ we arrive to
\[
\leq 4 \left( \sup_{h \in \mathcal{H}(\delta)} \Vert (P-P_K)h \Vert ~-~ \mathds{E}\sup_{h \in \mathcal{H}(\delta)} \Vert (P-P_K)h \Vert \right)^2 + 4 \left( \mathds{E}\sup_{h \in \mathcal{H}(\delta)} \Vert (P-P_K)h \Vert \right)^2 + \frac{8b^2}{K}.
\]
Apply now probabilistic inequality for bounded differences \cite[Th. 6.2]{BoucheronConcentration2013} to estimate the first term, with probability $\geq 1-e^{-t}$
\[
\left( \sup_{h \in \mathcal{H}(\delta)} \Vert (P-P_K)h \Vert ~-~ \mathds{E}\sup_{h \in \mathcal{H}(\delta)} \Vert (P-P_K)h \Vert \right)^2 \leq \frac{2b^2t}{K}.
\]
Therefore, with such probability
\[
\phi_K^{(2)} \leq \frac{8b^2t}{K} + 4 \left( \mathds{E}\sup_{h \in \mathcal{H}(\delta)} \Vert (P-P_K)h \Vert \right)^2 + \frac{8b^2}{K}.
\]

The resulting bound is now
\begin{align*}
&V(h_K)-V(h_*) \leq  2\mathds{E}\phi_K^{(1)}(\delta) + 4 \left( \mathds{E}\sup_{h \in \mathcal{H}(\delta)} \Vert (P-P_K)h \Vert \right)^2 + \frac{40b^2 t + 24b^2}{3K} + 12b\sqrt{\frac{\delta t}{K}}.
\end{align*}
with probability $\geq 1- e^{-t}$.
$\square$\\

\subsubsection{Bounding the Suprema}

To proceed further we need now to bound the two suprema in Theorem \ref{th:basicBound}. Lemma 5.3 in \cite{belomEVM2017} gives us a tool, it is stated as follows.

\begin{lemma}
Assume $X_1,...,X_K$ to be i.i.d. sample and $P_K$ be empirical measure. Let 
\[
\mathcal{H}:=\left\lbrace h: \mathcal{X} \to [-b,b]\right\rbrace
\]
and suppose that for all $u \in (0,b]$
\[
\mathcal{N}( \mathcal{H}, \Vert \cdot \Vert_{L^2(P_K)},u) \leq \left( \frac{c}{u}\right)^\alpha ~a.s.,
\]
then $\forall \sigma \in[\sigma_\mathcal{H},b]$
\[
\mathds{E} \sup_{h \in \mathcal{H}} (P-P_K)h \leq A \left( \sqrt{\frac{\alpha \sigma^2}{K} \log\frac{c}{\sigma}}
+\frac{\alpha b}{K} \log \frac{c}{\sigma}\right)
\]
where constants are explicitly given.
\label{lemma:expectation:polycover}
\end{lemma}

\begin{lemma}
Let A\ref{ass:bounded}, A\ref{ass:polycover} hold. Then
\[
\mathds{E}\phi_K^{(1)} \leq 2592 \left( \sqrt{\frac{72b^2\delta\alpha}{K}\log\frac{c}{6b\sqrt{2\delta}}} + \frac{\alpha b}{K}\log\frac{c}{6b\sqrt{2\delta}} \right).
\]
\label{lemma:ephik1}
\end{lemma}
$\triangleright$ Define \[
L(\delta):= \left\lbrace l(h) ~\vert~ h \in \mathcal{H}(\delta) \right\rbrace
\]
and note that in our case it also holds that
\[
\mathcal{N}(L(\delta), \Vert \cdot \Vert_{L^2(P_K)},u) \leq \mathcal{N}(\mathcal{H}(\delta), \Vert \cdot \Vert_{L^2(P_K)},u), 
\]
therefore, we could apply Lemma \ref{lemma:expectation:polycover} to $L(\delta)$ and get the result.
$\square$\\

The second supremum, fortunately, can be handled simpler.

\begin{lemma}
If A\ref{ass:bounded} is satisfied, it holds that
\[
\mathds{E}\sup_{h \in \mathcal{H}(\delta)} \Vert (P-P_K)h \Vert \leq \frac{2b}{\sqrt{K}}. 
\]
\label{lemma:directRademacher}
\end{lemma}
$\triangleright$ First note that by symmetrization
\[
\mathds{E}\sup_{h \in \mathcal{H}(\delta)} \Vert (P-P_K)h \Vert \leq \frac{2}{K} \mathds{E} \sup_{h \in \mathcal{H}(\delta)} \mathds{E}_\xi\Vert \sum_{k=1}^K \xi_k h(X_k) \Vert
\]
where $\xi_k$ are i.i.d. Rademacher's random variables.
 Expand the norm, apply Jensen's inequality to the square root and get
\begin{align}
&\mathds{E}_\xi \Vert \sum_{k=1}^K \xi_k h(X_k) \Vert  =  \mathds{E}_\xi \sqrt{ \sum_{k=1}^K \Vert h(X_k)\Vert^2 + 2\sum_{d=1}^D \sum_{1\leq i<j\leq K} \xi_i h(X_i) \xi_j h(X_j) } \leq \\
&\leq \sqrt{ \sum_{k=1}^K \Vert h(X_k)\Vert^2 + 2 \mathds{E}_\xi \sum_{d=1}^D \sum_{1\leq i<j\leq K} \xi_i h(X_i) \xi_j h(X_j) } = \sqrt{ \sum_{k=1}^K \Vert h(X_k)\Vert^2} \leq b \sqrt{K}.
\end{align}
$\square$\\

With Theorem \ref{th:basicBound},  Lemma \ref{lemma:ephik1} and Lemma \ref{lemma:directRademacher}  we make a conclusion.
\begin{theorem}
Let A\ref{ass:bounded}, A\ref{ass:starshape}, A\ref{ass:polycover} hold. If $h \in \mathcal{H}(\delta)$ then
\begin{align*}
&V(h_K)-V(h_*) \leq 5184 \left( \sqrt{\frac{72b^2\delta\alpha}{K}\log\frac{c}{6b\sqrt{2\delta}}} + \frac{\alpha b}{K}\log\frac{c}{6b\sqrt{2\delta}} \right) + \\
& + \frac{40b^2 t + 72b^2}{3K} + 12b\sqrt{\frac{\delta t}{K}}
\end{align*}
with probability $\geq 1- e^{-t}$.
\end{theorem}

\subsubsection{Proof of the Main Theorem}

Finally, we apply Lemma \ref{lemm:koltch}. What remains is to carefully compute $\beta(t)$ and obtain

\begin{theorem} (Theorem 2 in the main text)
 It holds that
\[
V(h)-V(h_*) \leq \max_j \beta^{(j)}(t)
\]
with probability at least $1-4e^{-t}$, $\beta^{(j)}(t)$ are defined in the proof.
\end{theorem}
$\triangleright$ We have bounded with probability $\geq 1-e^{-t}$ the excess risk of $\mathcal{H}(\delta)$,  so that 
\[
\beta_K(\delta,t) = C_0 \left( \sqrt{\frac{b^2\delta\alpha}{K}\log\frac{c}{6b\sqrt{2\delta}}} + \frac{\alpha b}{K}\log\frac{c}{6b\sqrt{2\delta}} \right) + \frac{40b^2 t + 72b^2}{3K} + 12b\sqrt{\frac{\delta t}{K}}.
\]
Now compute for $\tau>0$
\begin{align*}
&\sup_{\delta \geq \tau} \frac{\beta_K(\delta,t\delta/\tau)}{\delta} = C_0 \left( \sqrt{\frac{b^2\alpha}{\tau K}\log\frac{c}{6b\sqrt{2\tau}}} + \frac{\alpha b}{\tau K}\log\frac{c}{6b\sqrt{2\tau}} \right) + \\
& + \frac{40b^2 t + 72b^2}{3K \tau} + 12b\sqrt{\frac{\delta t}{K \tau}}.
\end{align*}
Finally, observe that
\[
\beta_K(t) := \inf\left\lbrace \tau >0~:~ \sup_{\delta \geq \tau} \frac{\beta_K(\delta,t\delta/\tau)}{\delta} \leq \frac{1}{2} \right\rbrace \leq \max_j \beta^{j}(t)
\]
where
\begin{align*}
& \beta^{1}(t) = \inf\left\lbrace \tau>0: 72 \sqrt{\frac{32b^2 \alpha}{K \tau}\log\frac{c}{4b\sqrt{2\tau}}} \leq \frac{1}{8} \right\rbrace \leq C_1 \frac{ \log K}{K},\\
& \beta^{2}(t) = \inf\left\lbrace \tau>0: 2592\frac{\alpha b}{K\tau}\log\frac{c}{4b\sqrt{2\tau}} \leq \frac{1}{8} \right\rbrace \leq C_2 \frac{ \log K}{K},\\
&\beta^{3}(t) = \inf\left\lbrace \tau>0: \frac{40b^2 t + 72b^2}{3K \tau} \leq \frac{1}{8} \right\rbrace =  \frac{ 8 (40b^2t +72b^2)}{3K},\\
&\beta^{4}(t) = \inf\left\lbrace \tau>0: 12b\sqrt{\frac{t}{K \tau}} \leq \frac{1}{8} \right\rbrace =  \frac{ 9216 b^2t}{K}.
\end{align*}
It holds with probability $1-4e^{-t}$
$\square$

%% file: sbaselineUnbiased.tex
S-baseline is known to result in unbiased estimate, here for the sake of completeness we give a proof.

\begin{proposition} For all $b_\phi: \mathcal{S} \to \mathds{R}$ the expected value
\[
\mathds{E}\left[ \sum_{t=0}^{\infty} \gamma^t b_\phi(S_t)\nabla_\theta \log \pi_\theta(A_t\vert S_t) \right]=0.
\]
\label{prop:sbaseline}
\end{proposition}
$\triangleright$ Let us consider one term of the sum and note that we can use tower property of conditional expectation:
\[
\mathds{E}\left[ \gamma^t b_\phi(S_t)\nabla_\theta\log \pi_\theta (A_t \vert S_t) \right] = \gamma^t\mathds{E}\left\lbrace \mathds{E}\left[ b_\phi(S_t)\nabla_\theta\log \pi_\theta (A_t \vert S_t) ~\vert~ S_t\right] \right\rbrace.
\]
Now note that $b_\phi(S_t)$ is measurable in the inner expectation, so,
\[
= \gamma^t \mathds{E}\left\lbrace b_\phi(S_t) \mathds{E}\left[ \nabla_\theta\log \pi_\theta (A_t \vert S_t) ~\vert~ S_t \right]\right\rbrace.
\]
Finally, with the help of the log derivative we show that
\[
\mathds{E}\left[ \nabla_\theta\log \pi_\theta (A_t \vert S_t) ~\vert~ S_t\right] =0
\]
and the result follows.
$\square$\\

%% file: reinfConvergence.tex
We base our proof on some techniques of \cite{XuSVRPGConv2019} where SVRPG algorithm is considered but the proof we need has the same structure with $b=m=1$ and  some adjustments.

Let $\widetilde{\nabla}J : (\mathcal{S} \times \mathcal{A} \times \mathds{R})^T \to \mathds{R}^D$ be an unbiased gradient estimate (with baseline or just REINFORCE). Our gradient algorithm reads as
\[
\theta_{n+1} = \theta_n + \alpha_n \frac{1}{K}\sum_{k=1}^K \widetilde{\nabla}J(X_n^{(k)}),
\]
where $\theta_n \in \Theta \subset \mathds{R}^D$ are policy parameters at iteration $n$ and $X_n^{(k)} \in (\mathcal{S} \times \mathcal{A} \times \mathds{R})$ is the trajectory data at iteration $n$ of which there are $K$ independent samples. Let us for shorter notation set $\widetilde{\nabla}J_{n}^{K}:= \frac{1}{K}\sum_{k=1}^K \widetilde{\nabla}J(X_n^{(k)})$. The Lemma below is the in the core of non-convex smooth optimization.

\begin{lemma}
If $\forall \theta \in \Theta ~~ \norm{\nabla^2 J(\theta)}{2}\leq L$, then for all $n\in \mathds{Z}_{>0}$
\begin{equation}
J(\theta_{n+1}) \geq J(\theta_n) -\frac{3\alpha_n}{4}\norm{\nabla J(\theta_n) - \widetilde{\nabla}J_n^{K}}{2}^2 + \left(\frac{1}{4\alpha_n}-\frac{L}{2}\right)\norm{\theta_{n+1}-\theta_n}{2}^2 + \frac{\alpha_n}{8} \norm{\nabla J(\theta_n)}{2}^2,
\end{equation}
where $v_n = \alpha_n $
\label{lemm:lowqbound}
\end{lemma}

$\triangleright$ It can be obtained by applying lower quadratic bound:
\begin{equation}
J(\theta_{n+1}) \geq J(\theta_n) + \langle \nabla J(\theta_n), \theta_{n+1}-\theta_n\rangle - \frac{L}{2} \norm{\theta_{n+1}-\theta_n}{2}^2.
\end{equation}
Next, notice that $\alpha_n \widetilde{\nabla}J_n^K = \theta_{n+1}-\theta_n$ and add and subtract $ \widetilde{\nabla}J_n^K$ in the left entry of the second term:
\begin{equation}
J(\theta_{n+1}) \geq J(\theta_n) + \langle \nabla J(\theta_n)-\widetilde{\nabla}J_n^K, \alpha_n\widetilde{\nabla}J_n^K \rangle + \alpha_n \norm{\widetilde{\nabla}J_n^K}{2}^2- \frac{L}{2} \norm{\theta_{n+1}-\theta_n}{2}^2.
\end{equation}
Now apply Young's polarization inequality ($ab \geq -(a^2+b^2)/2$) to the same term and arrive to
\begin{equation}
 J(\theta_{n+1}) \geq J(\theta_n) -\frac{\alpha_n}{2} \norm{ \nabla J(\theta_n)-\widetilde{\nabla}J_n^K}{2}^2 + \frac{\alpha_n}{2} \norm{\widetilde{\nabla}J_n^K}{2}^2- \frac{L}{2} \norm{\theta_{n+1}-\theta_n}{2}^2.   
\end{equation}
Observe that the second and the third term can be bounded further using
\begin{equation}
    \norm{\nabla J (\theta_{n+1})}{2}^2 \leq 2 \norm{\widetilde{\nabla} J_n^K}{2}^2 + 2\norm{\nabla J (\theta_{n+1}) - \widetilde{\nabla}J_n^K}{2}^2,
\end{equation}
which results in
\begin{equation}
J(\theta_{n+1}) \geq J(\theta_n) -\frac{3\alpha_n}{4}\norm{\nabla J(\theta_n) - \widetilde{\nabla}J_n^{K}}{2}^2 + \left(\frac{1}{4\alpha_n}-\frac{L}{2}\right)\norm{\theta_{n+1}-\theta_n}{2}^2 + \frac{\alpha_n}{8} \norm{\nabla J(\theta_n)}{2}^2.
\end{equation}
$\square$\\

With this Lemma we can prove a variety of different convergence results, we would rather refer here to \cite{XuSVRPGConv2019,koppelGlobalConv2020}. And yet, to illustrate the need for the variance reduction, consider the following theorem.\\

\begin{theorem}
There is a constant $C_R>0$ such that for all $k>0$ and $N\leq k$ the following bound holds assuming non-increasing step sizes $\alpha_n \leq 2/L$:
\begin{equation}
\frac{1}{k}\sum_{n=k-N}^k \mathds{E} \left\Vert \nabla J(\theta_n) \right\Vert_2^2 \leq \frac{16 C_R}{k\alpha_k} + \frac{1}{k}\sum_{n=k-N}^k \mathds{E}\left\Vert \nabla J(\theta_n)-\widetilde{\nabla}J_n^K\right\Vert_2^2.
\label{eq:convBound1}
\end{equation}
In particular, when $N=k-1$, one gets
\begin{equation}
\frac{1}{k}\sum_{n=1}^k \mathds{E} \left\Vert \nabla J(\theta_n) \right\Vert_2^2 \leq \frac{16 C_R}{k\alpha_k} + \frac{1}{k}\sum_{n=1}^k \mathds{E}\left\Vert \nabla J(\theta_n)-\widetilde{\nabla}J_n^K\right\Vert_2^2.
\label{eq:convBound2}
\end{equation}
\end{theorem}
$\triangleright$ Introduce quantity $U(\theta):=J(\theta^*)-J(\theta)$. Let us use Lemma \ref{lemm:lowqbound}, divide both parts by $\alpha_n$ and sum them from $n=k-N$ to $k$ with $k,N$ satisfying $N\leq k$, then take the expectation:
\begin{equation}
\sum_{n=k-N}^k \E{ \norm{\nabla J(\theta_n)}{2}^2}  \leq 8\sum_{n=k-N}^k \frac{1}{\alpha_n} \E{U(\theta_n)-U(\theta_{n+1})} +6 \sum_{n=k-N}^k \E{ \norm{\nabla J(\theta_n) - \widetilde{\nabla}J_n^{K}}{2}^2} .
\end{equation}
Notice that we used $\alpha_n< 2/L$ to drop the term with $\norm{\theta_{n+1}-\theta_n}{2}^2$. One could rewrite the first sum on the right to get
\begin{align}
\sum_{n=k-N}^k \E{ \norm{\nabla J(\theta_n)}{2}^2 }  &\leq 8\sum_{n=k-N}^k \left( \frac{1}{\alpha_n}-\frac{1}{\alpha_{n-1}}\right) \E{U(\theta_n)} -\frac{8}{\alpha_k}\E{U(\theta_{k+1})} + \frac{8}{\alpha_{k-N-1}}\E{U(\theta_{k-N})}+\\
&+6 \sum_{n=k-N}^k \E{ \norm{\nabla J(\theta_n) - \widetilde{\nabla}J_n^{K}}{2}^2} .
\end{align}
Since the rewards are bounded, there is $C_R$ such that for all $\theta$ the difference $U(\theta)\leq C_R$; secondly, the step sizes are non-increasing; finally, we can discard the second term which is non-positive. Thus,
\begin{align}
\sum_{n=k-N}^k \E{ \norm{\nabla J(\theta_n)}{2}^2 } &\leq 8C_R\sum_{n=k-N}^k \left( \frac{1}{\alpha_n}-\frac{1}{\alpha_{n-1}}\right)  + \frac{8C_R}{\alpha_{k-N-1}}+ 6 \sum_{n=k-N}^k \E{ \norm{\nabla J(\theta_n) - \widetilde{\nabla}J_n^{K}}{2}^2} \leq \\
&\leq \frac{8C_R}{\alpha_n}  + \frac{8C_R}{\alpha_{k-N-1}}+ 6 \sum_{n=k-N}^k \E{ \norm{\nabla J(\theta_n) - \widetilde{\nabla}J_n^{K}}{2}^2}.
\end{align}
We could again use the fact that $\alpha_n$ are non-increasing to simplify the first two terms, then divide both parts by $k$:
\begin{align}
\frac{1}{k}\sum_{n=k-N}^k \E{ \norm{\nabla J(\theta_n)}{2}^2} \leq \frac{16C_R}{\alpha_n} + \frac{6}{k} \sum_{n=k-N}^k \E{ \norm{\nabla J(\theta_n) - \widetilde{\nabla}J_n^{K}}{2}^2}.
\end{align}
$\square$\\

This result shows, that the convergence of the gradient to zero is influenced by the variance of the gradient estimator. In practice, however, the variance reduction ratio is very low and therefore it slightly but not dramatically improves the algorithm. Theory of SVRPG \cite{XuSVRPGConv2019}, however, suggests that in terms of rates with the accurate design of the step sizes the rate can be slightly improved. Despite all this, variance reduction provably improves \textit{local} convergence but as to global convergence (which is more tricky to specify), the variance also may play a good role in avoiding local optima as shown by \cite{koppelGlobalConv2020}. This, we believe partially explains, why in practice the quality of the algorithms is not so strongly influenced by the variance reduction, as one might have thought.

%% file: suppMinigrid.tex
\subsection{Minigrid}

Minimalistic Gridworld Environment (MiniGrid) provides gridworld Gym environments that were designed to be simple and lightweight, therefore, ideally fitting for making experiments. In particular, we considered  GoToDoor and Unlock.\\

In both environments we have used 20 independent runs of the algorithms. All measurements of mean rewards and variance of the gradient estimator (measured each 250 epochs on a newly generated pool of 100 trajectories) are averaged over these runs. Standard deviations of the rewards are obtained as the sample standard deviation of the observed rewards reflecting the width of the confidence intervals of the mean reward curves. The exact config-files used for experiments can be found on attached GitHub.

\subsubsection{Go-To-Door-5x5}
This environment is a room with four doors, one on each wall. The agent receives a textual (mission) string as input, telling it which door to go to, (e.g: "go to the red door"). It receives a positive reward for performing the done action next to the correct door, as indicated in the mission string.
\input{charts_sup_gotodoor}

\clearpage
\subsubsection{Unlock}

\input{suppMinigridUnlock}


%% file: charts_sup_gotodoor.tex

\begin{figure}[h!]
    \centering
    \begin{tabular}{l}
    (a) \includegraphics[scale=.25]{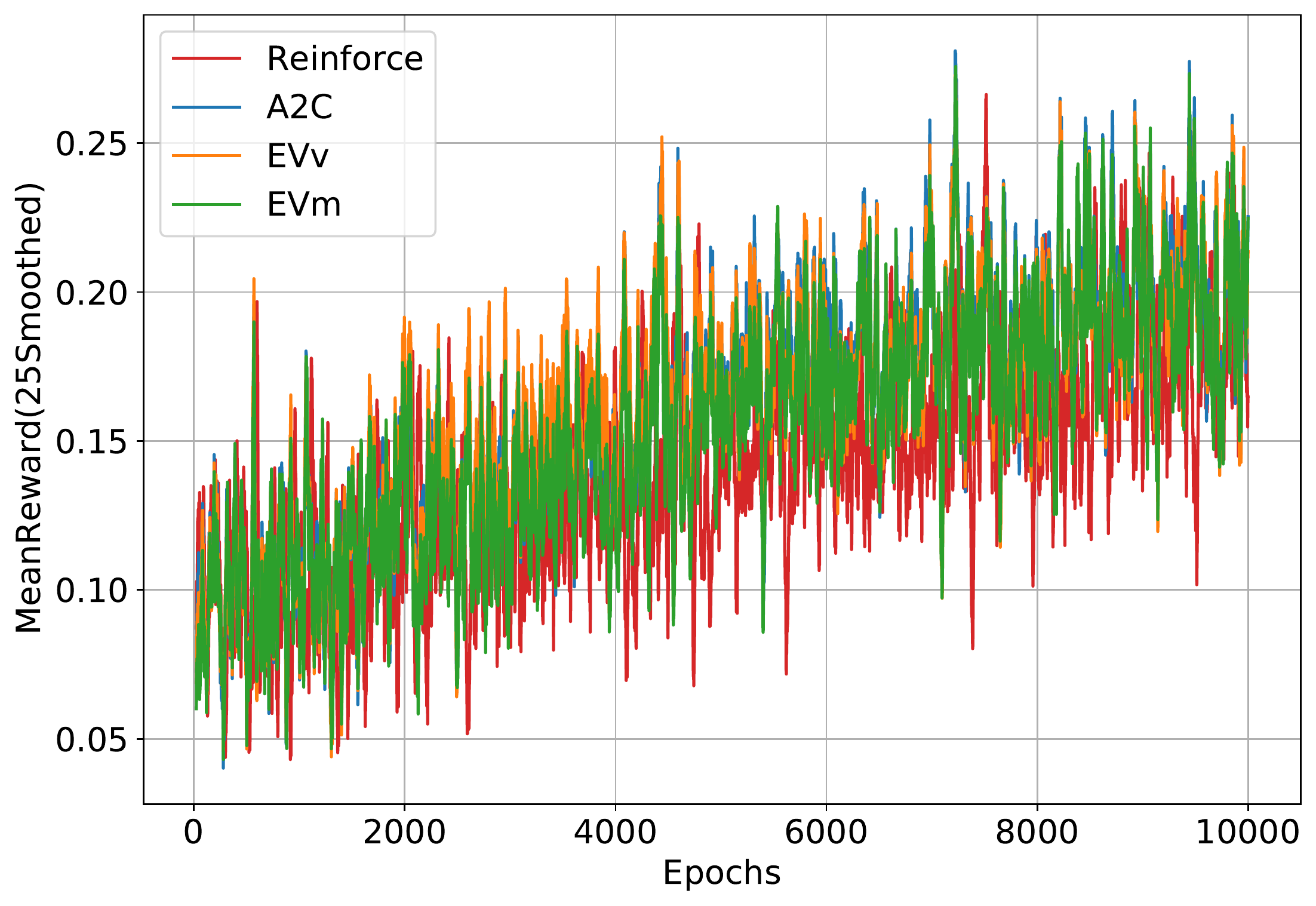} 
    (b) \includegraphics[scale=.25]{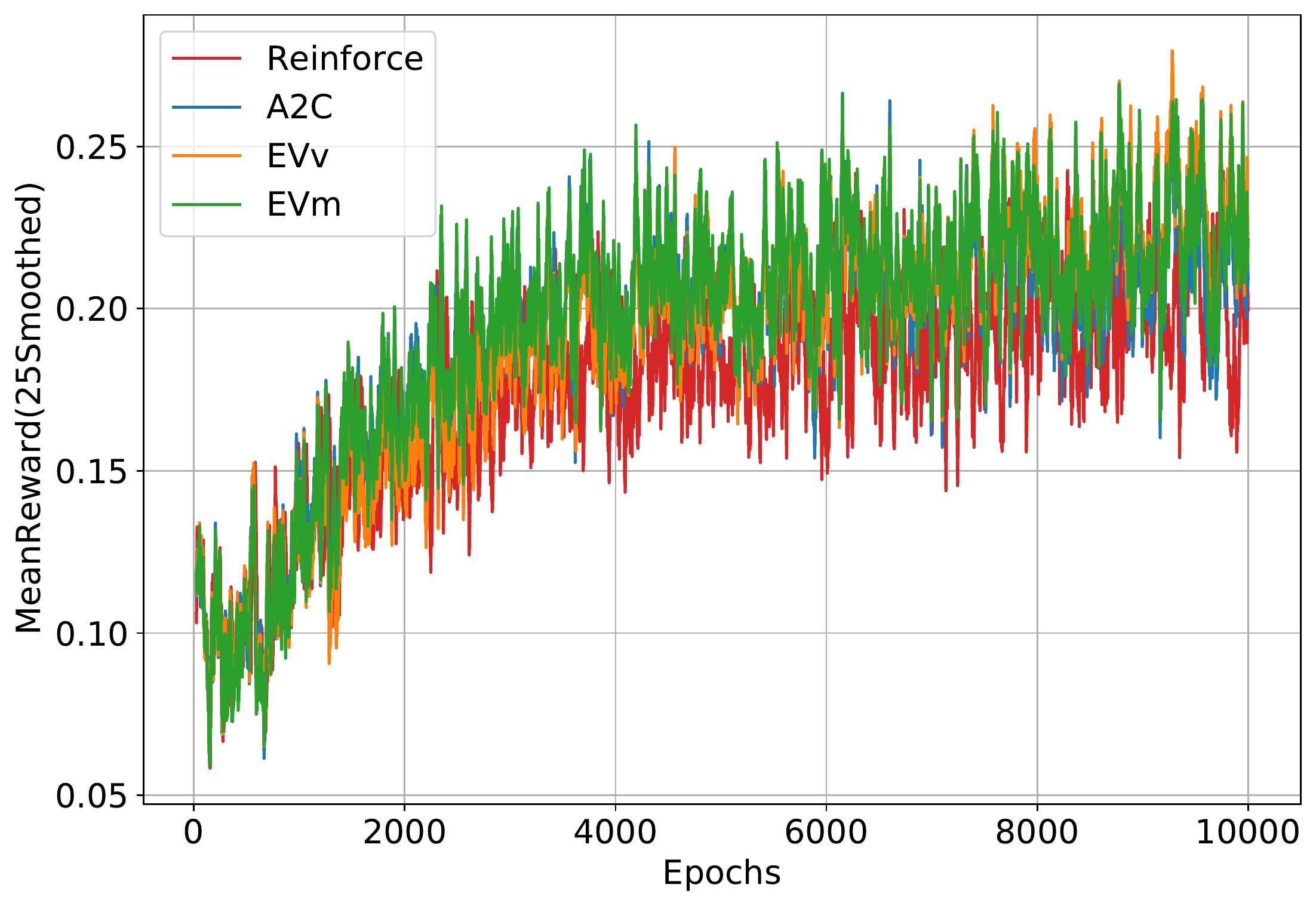} \\ 
    (c) \includegraphics[scale=.25]{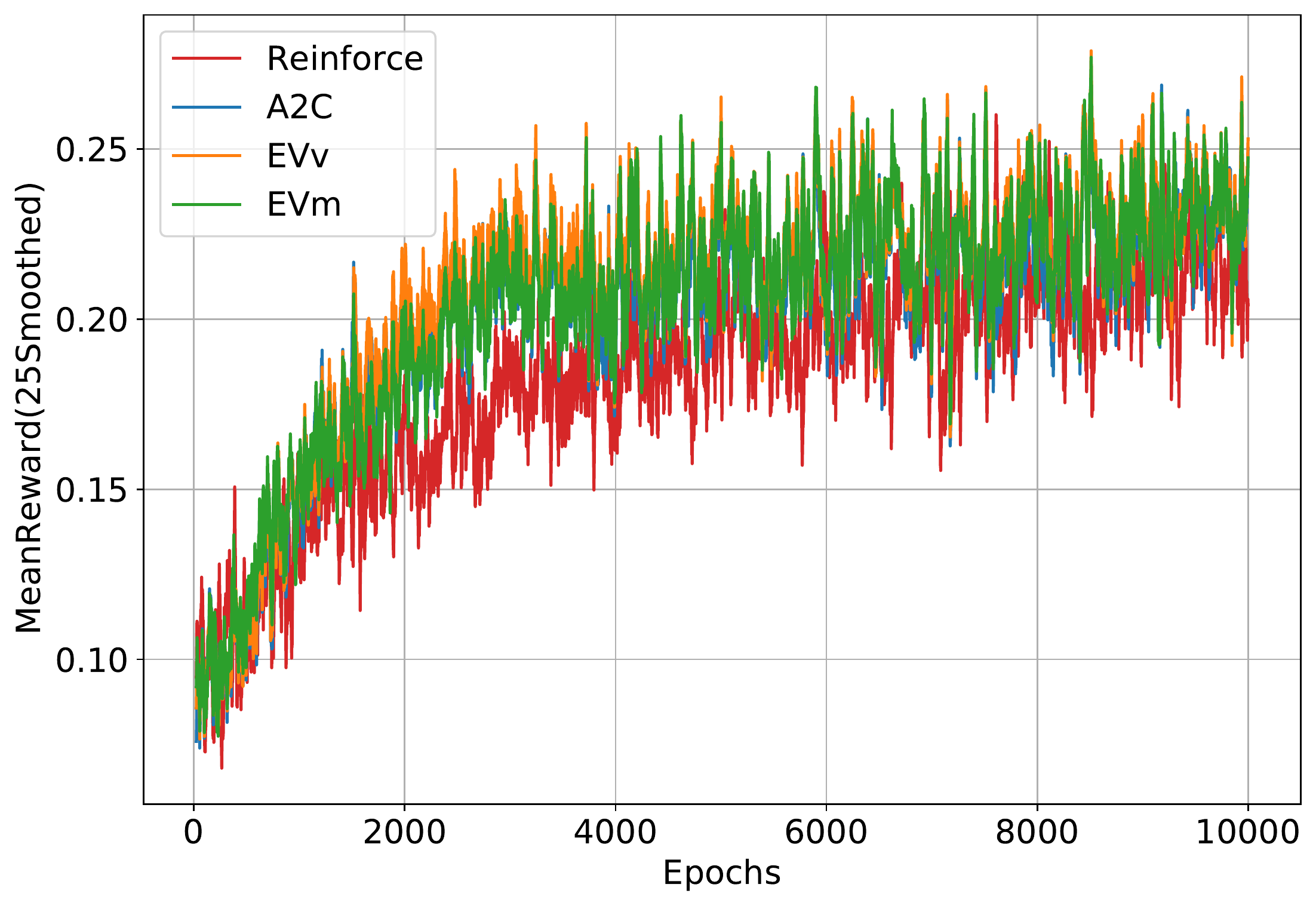} 
    (d) \includegraphics[scale=.25]{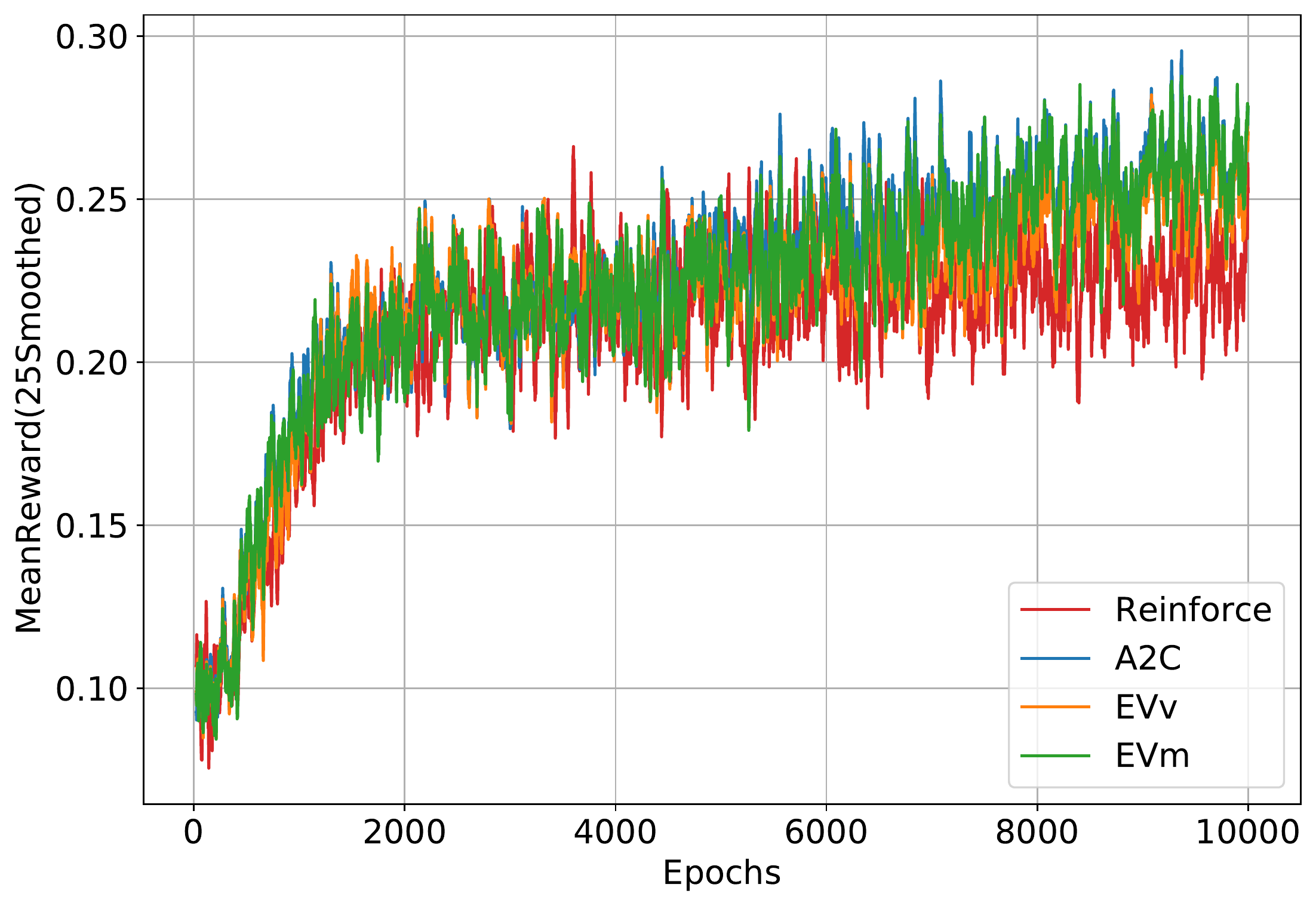}
    \end{tabular}
    \caption{The charts representing mean rewards in GoToDoor environment, standing for absolute values for cases $K=5$(a), $K=10$(b), $K=15$(c), $K=20$(d). The results are averaged over 20 runs. The resulting curves are smoothed with sliding window of size 25.}
    \label{fig:sup_GoToDoor_rew}
\end{figure}

\begin{figure}[h!]
    \centering
    \begin{tabular}{l}
    (a) \includegraphics[scale=.25]{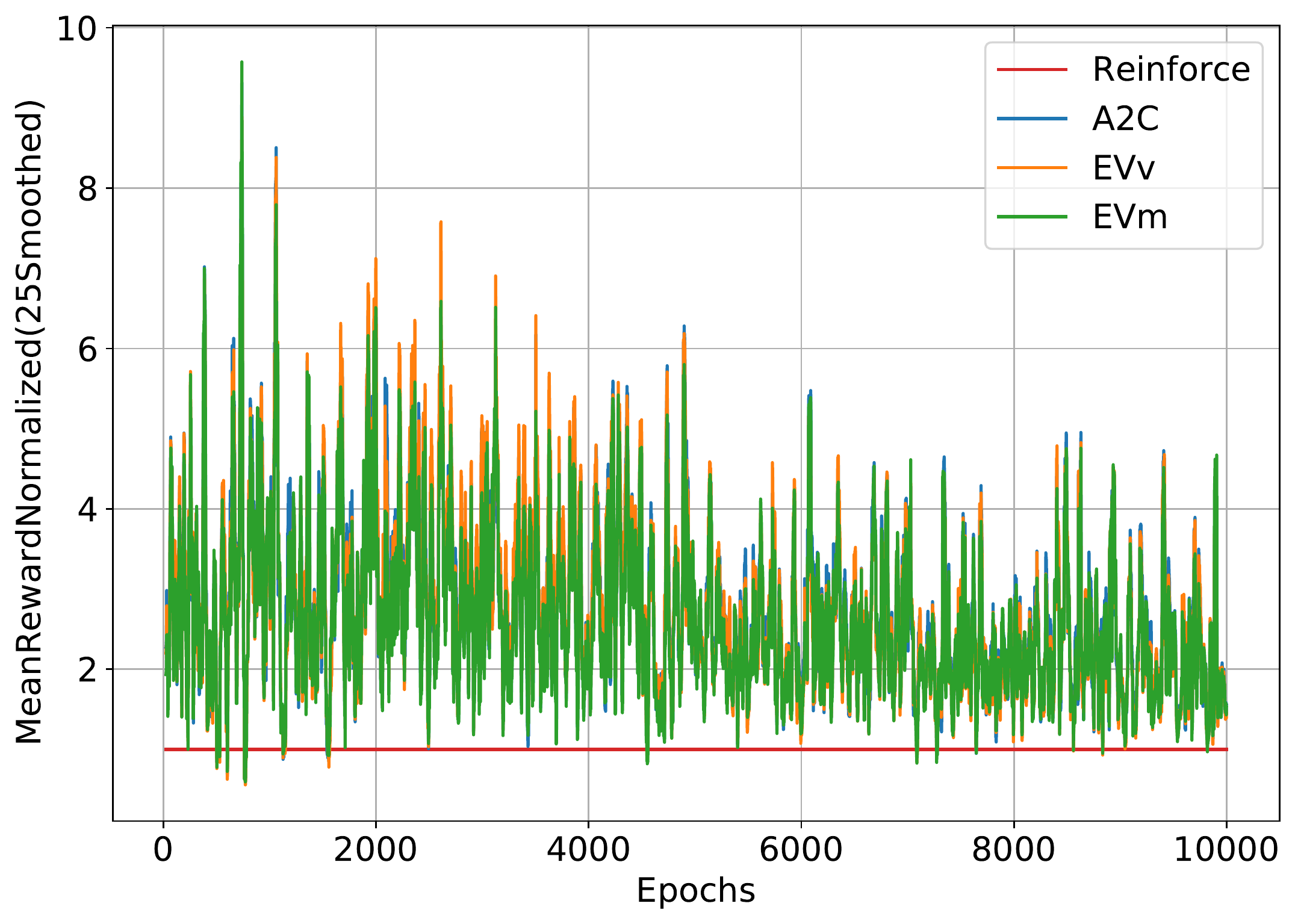} 
    (b) \includegraphics[scale=.25]{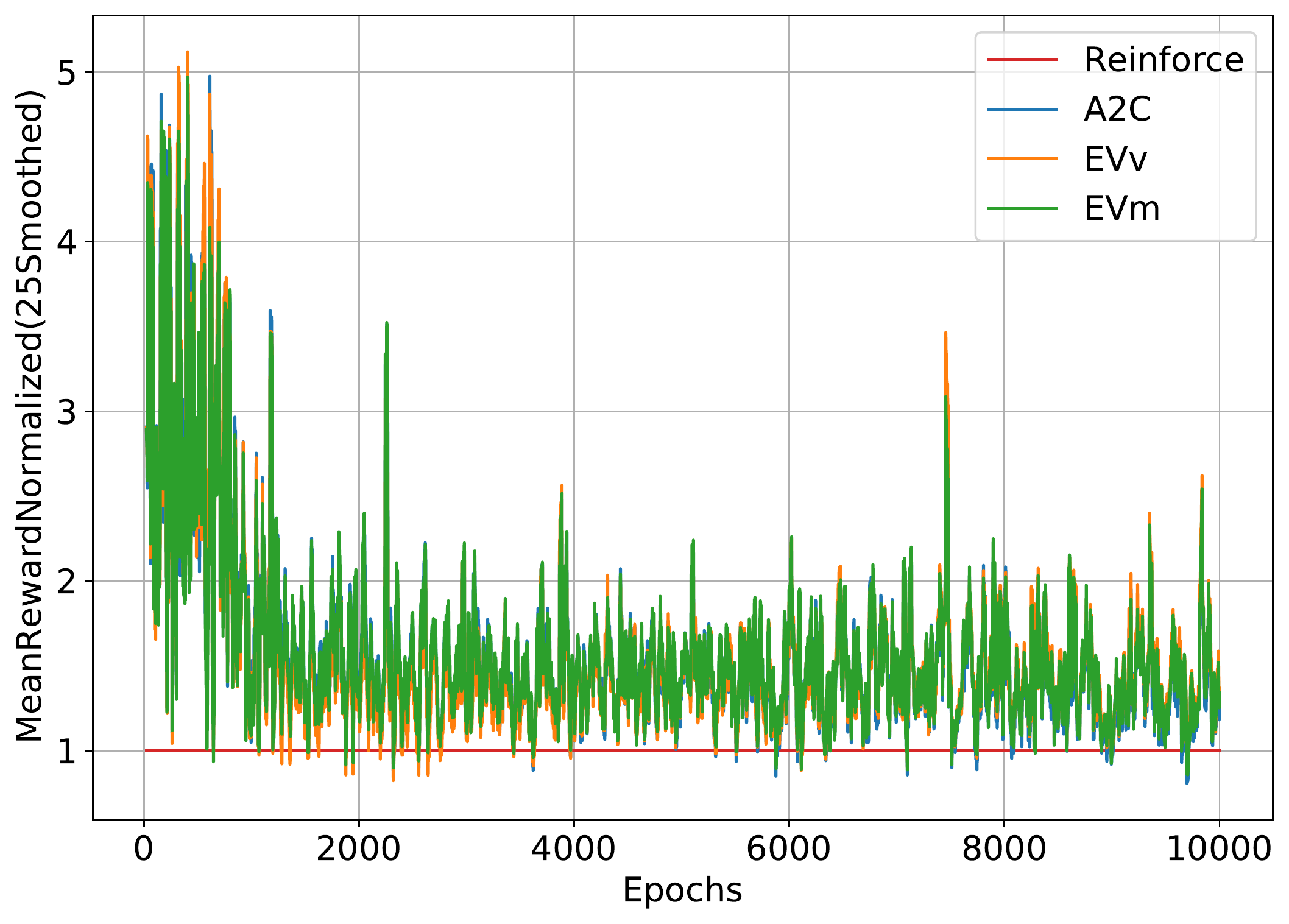} \\ 
    (c) \includegraphics[scale=.25]{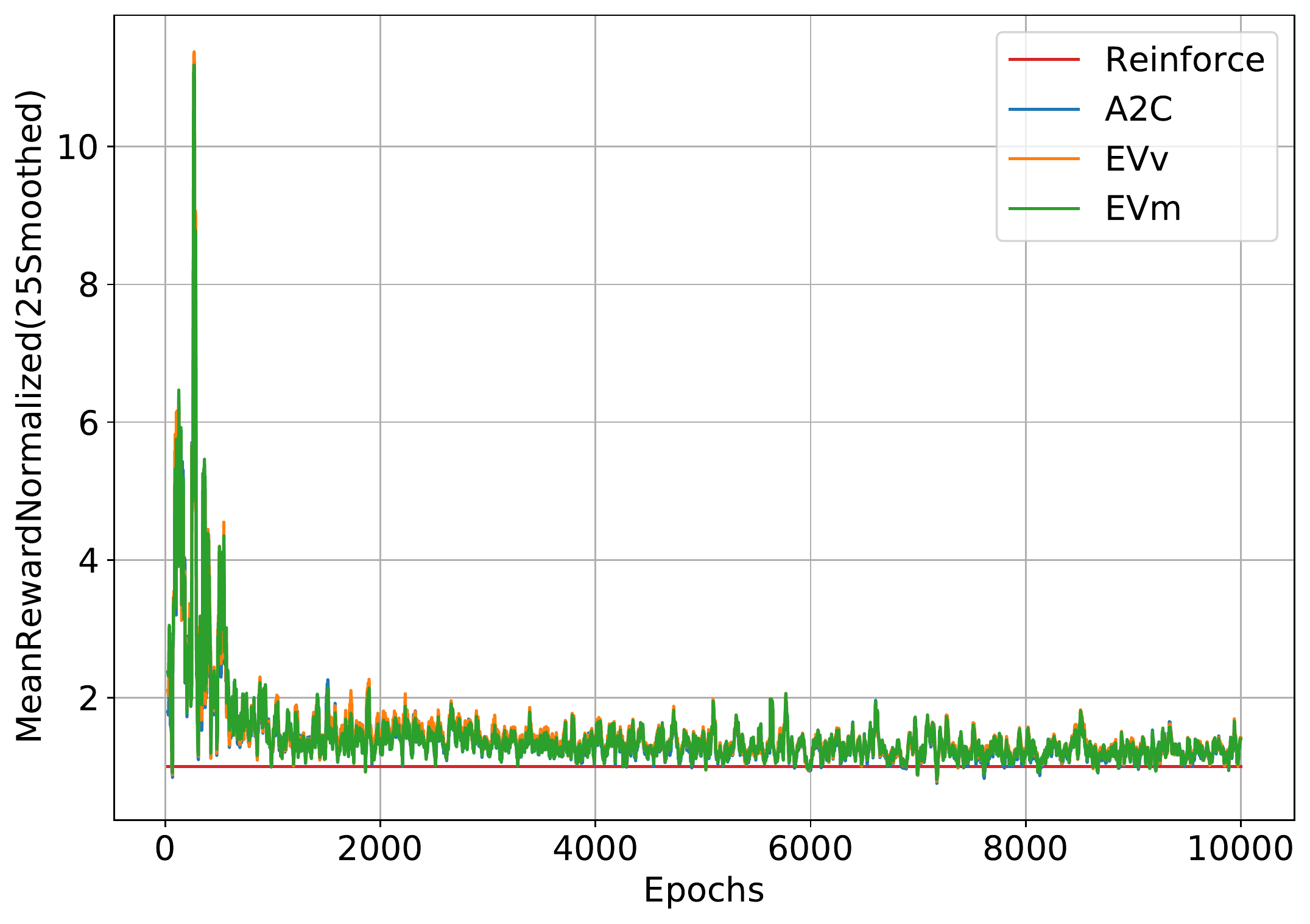} 
    (d) \includegraphics[scale=.25]{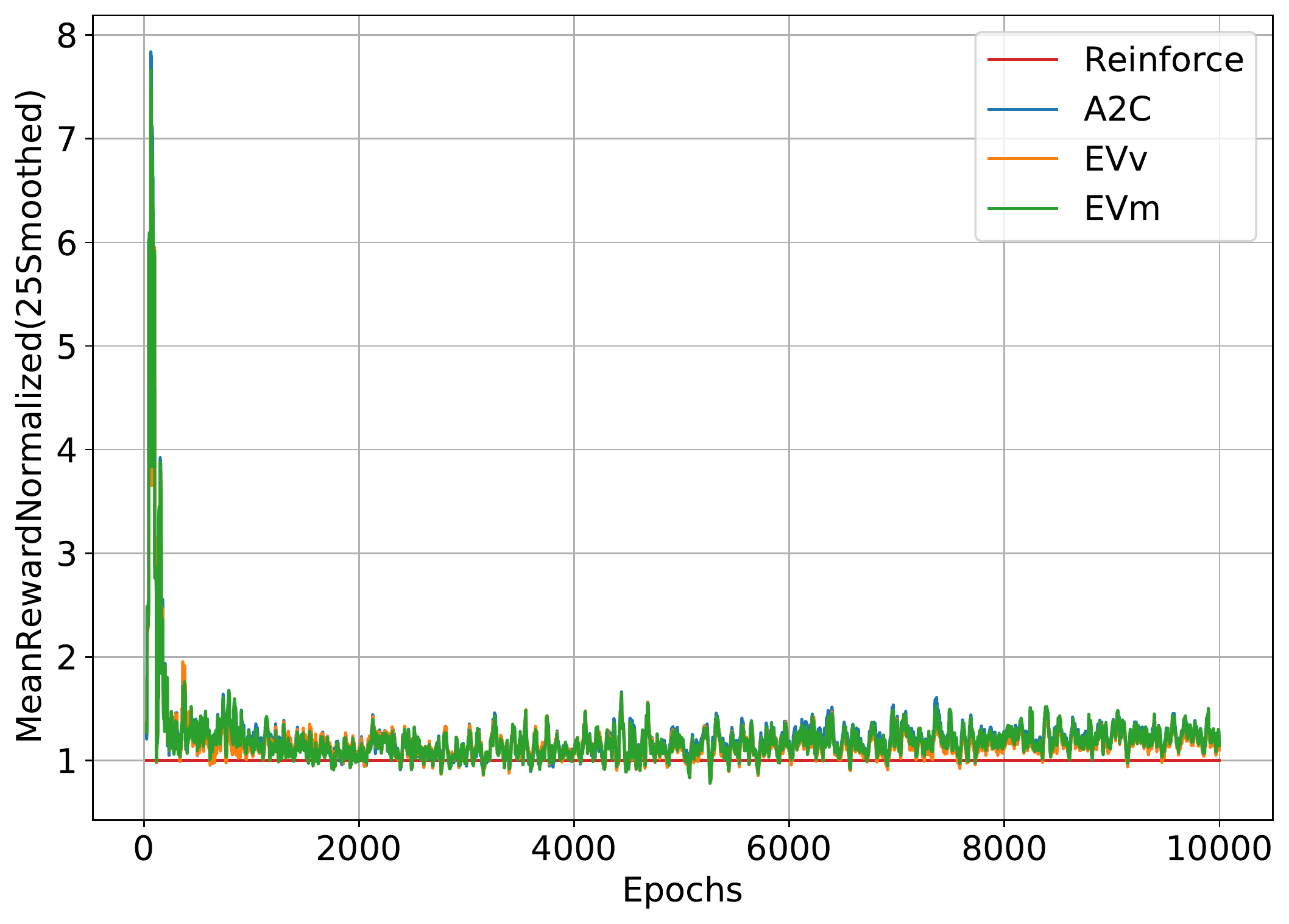}
    \end{tabular}
    \caption{The charts representing mean rewards in GoToDoor environment, the curves normalized by the mean reward of the REINFORCE. for cases K=5(a), K=10(b), K=15(c), K=20(d). The results are averaged over 20 runs. The resulting curves are smoothed with sliding window of size 25}
    \label{fig:sup_GoToDoor_rew_normalized}
\end{figure}

In GoToDoor environment we can clearly see that the EV-agent is at least as good as A2C at lower number of samples ($K=5$), and the more samples are available during training, the better performance we can observe from EV-agents. We can see mean rewards in absolute values in Fig. \ref{fig:sup_GoToDoor_rew} and in relative scale (normalized by the results of REINFORCE) in Fig. \ref{fig:sup_GoToDoor_rew_normalized} to see improvements over the results of REINFORCE algorithm.

\begin{figure}[h!]
    \centering
    \begin{tabular}{l}
    (a) \includegraphics[scale=.25]{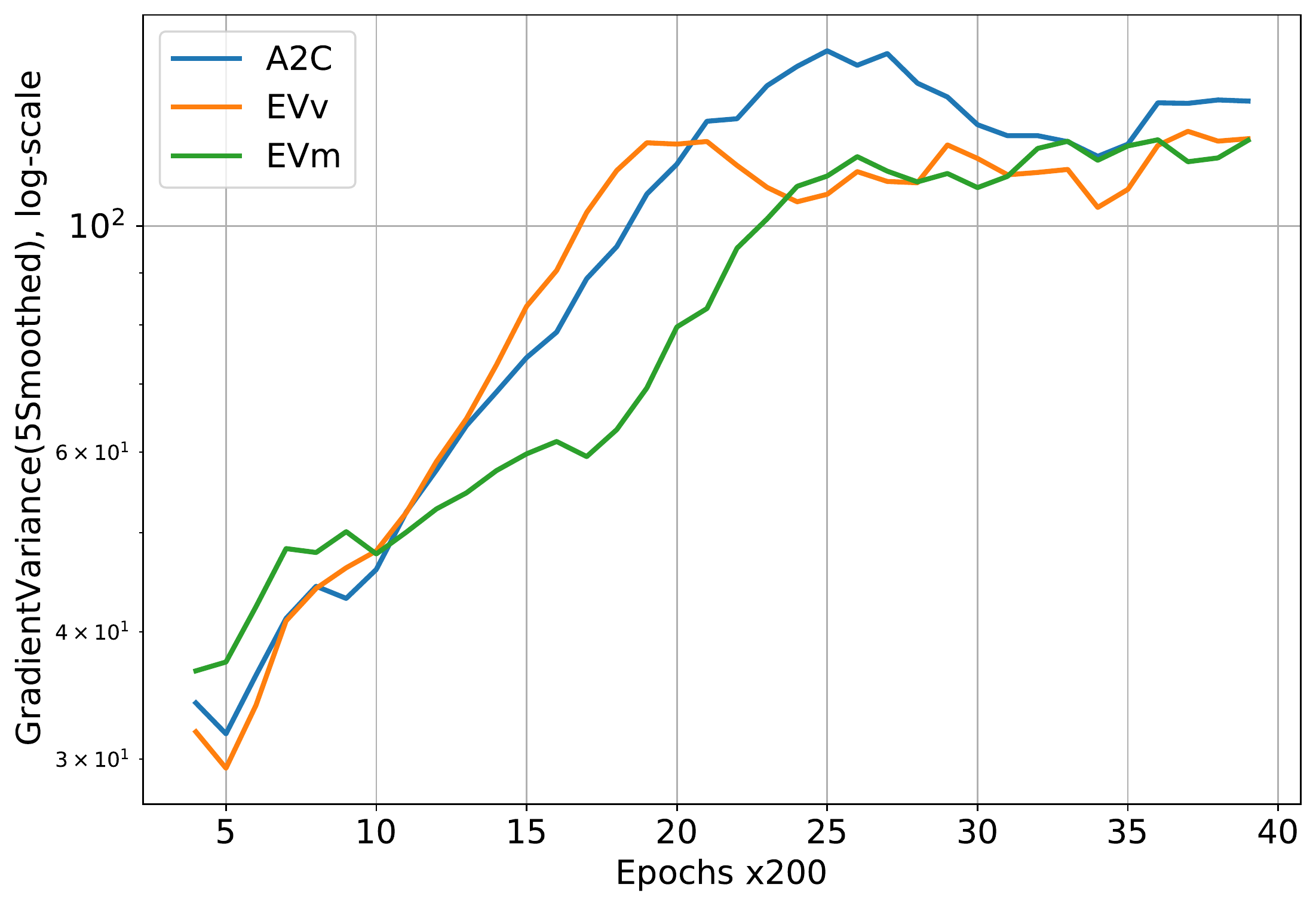} 
    (b) \includegraphics[scale=.25]{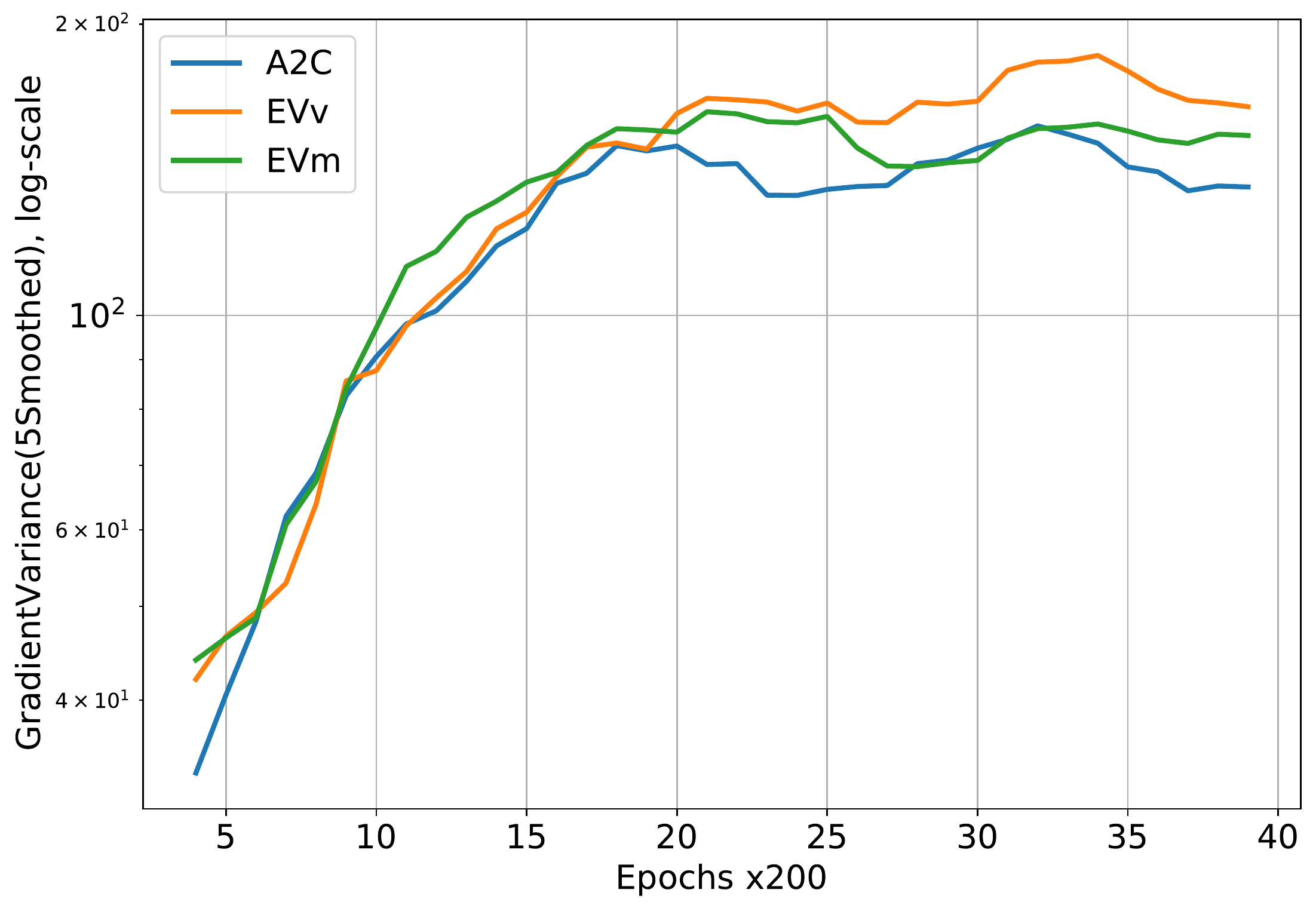} \\ 
    (c) \includegraphics[scale=.25]{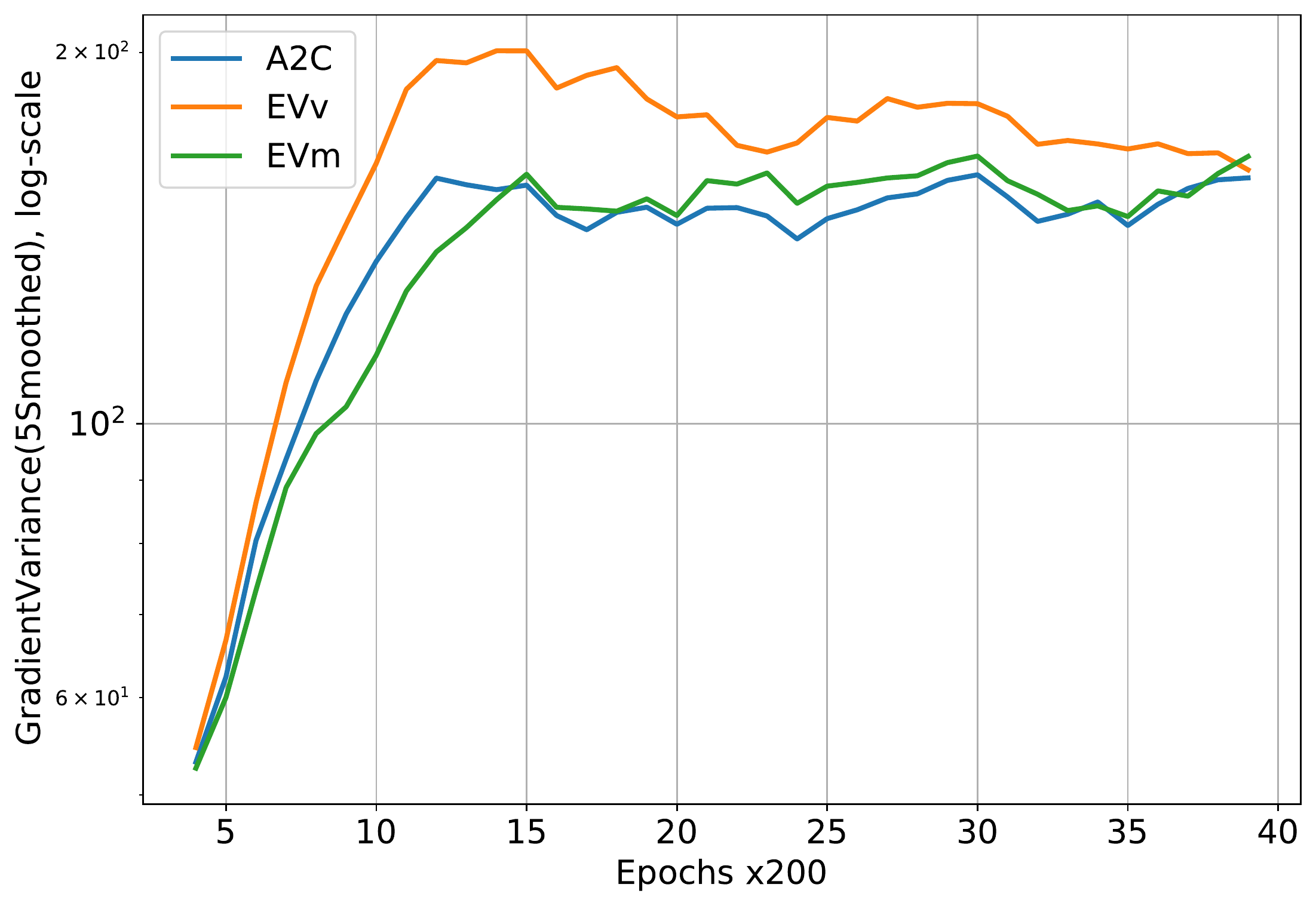} 
    (d) \includegraphics[scale=.25]{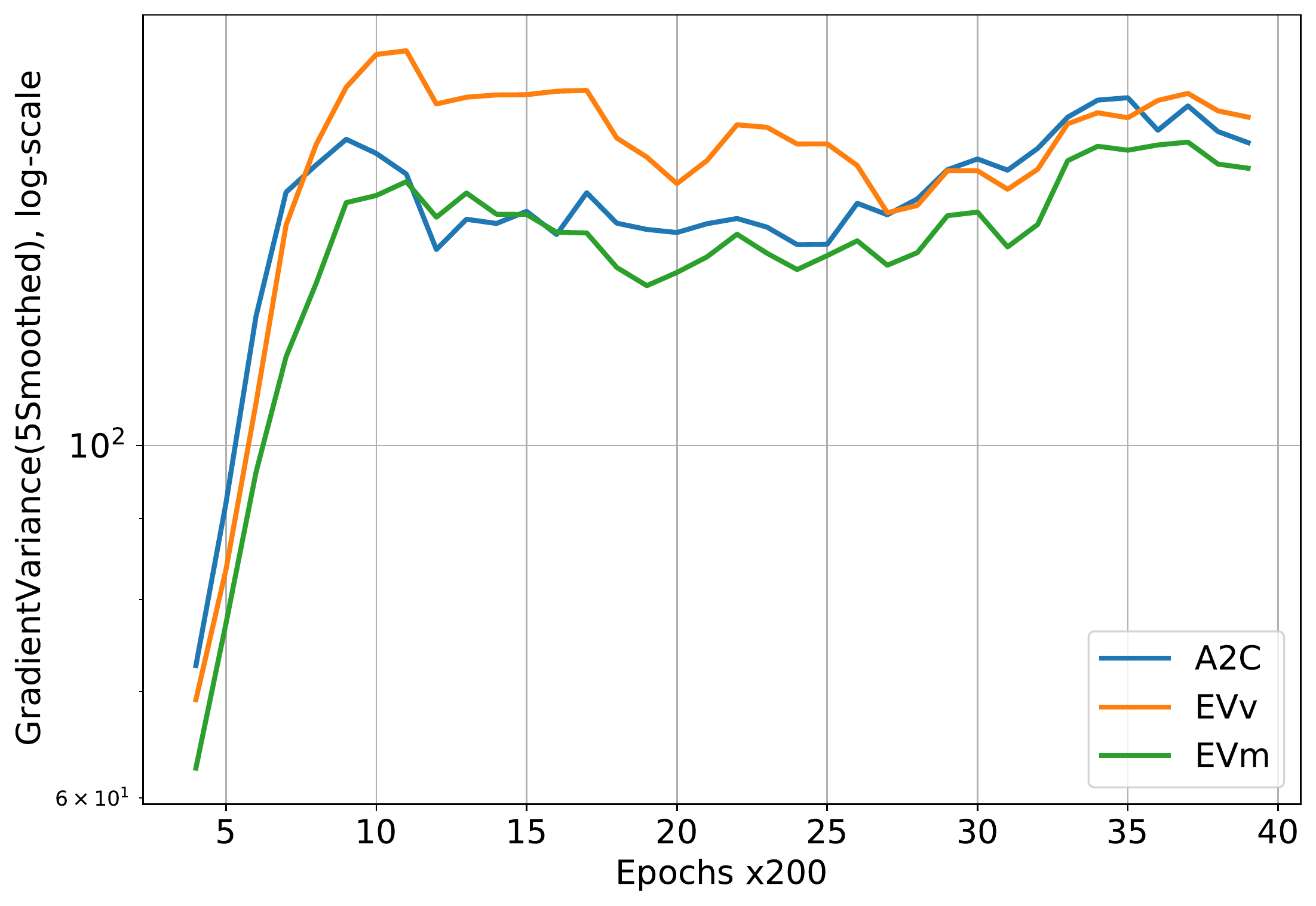}
    \end{tabular}
    \caption{The charts representing the variance of the gradient estimator in absolute values for cases $K=5$(a), $K=10$(b), $K=15$(c), $K=20$(d). The results are averaged over 20 runs. The resulting curves are smoothed with sliding window of size 5.}
    \label{fig:sup_GoToDoor_gradvar}
\end{figure}

\begin{figure}[h!]
    \centering
    \begin{tabular}{l}
    (a) \includegraphics[scale=.25]{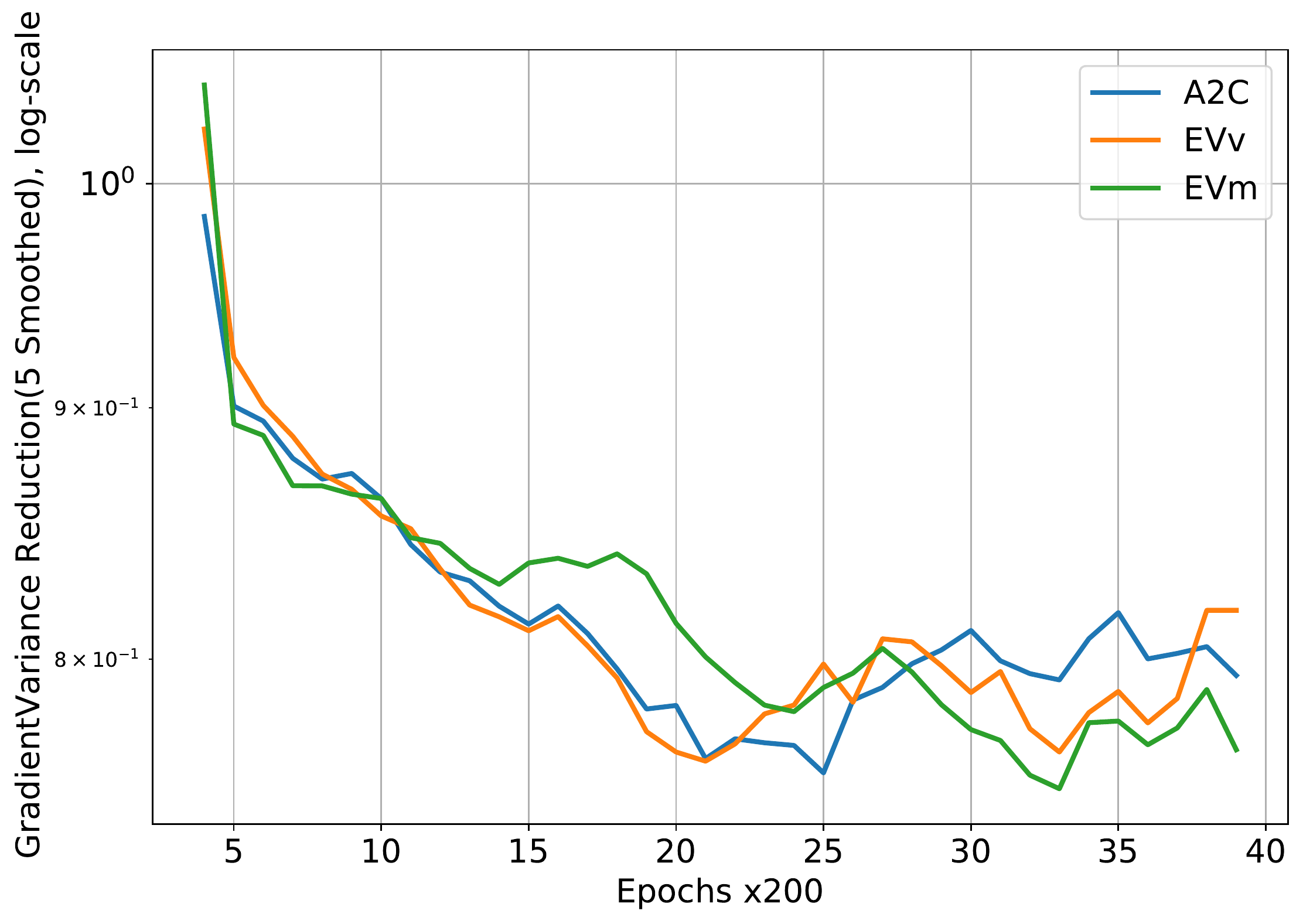} 
    (b) \includegraphics[scale=.25]{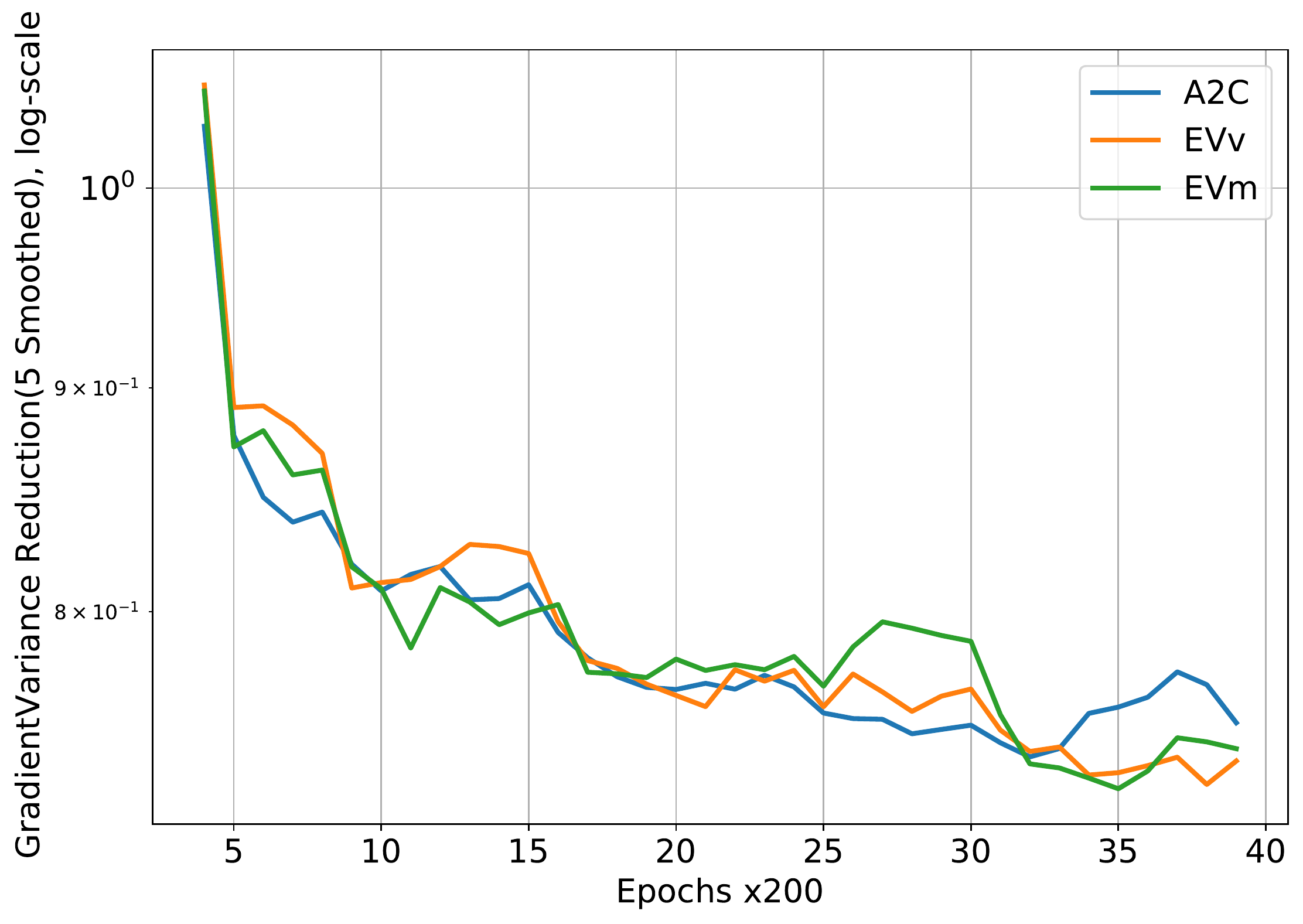} \\ 
    (c) \includegraphics[scale=.25]{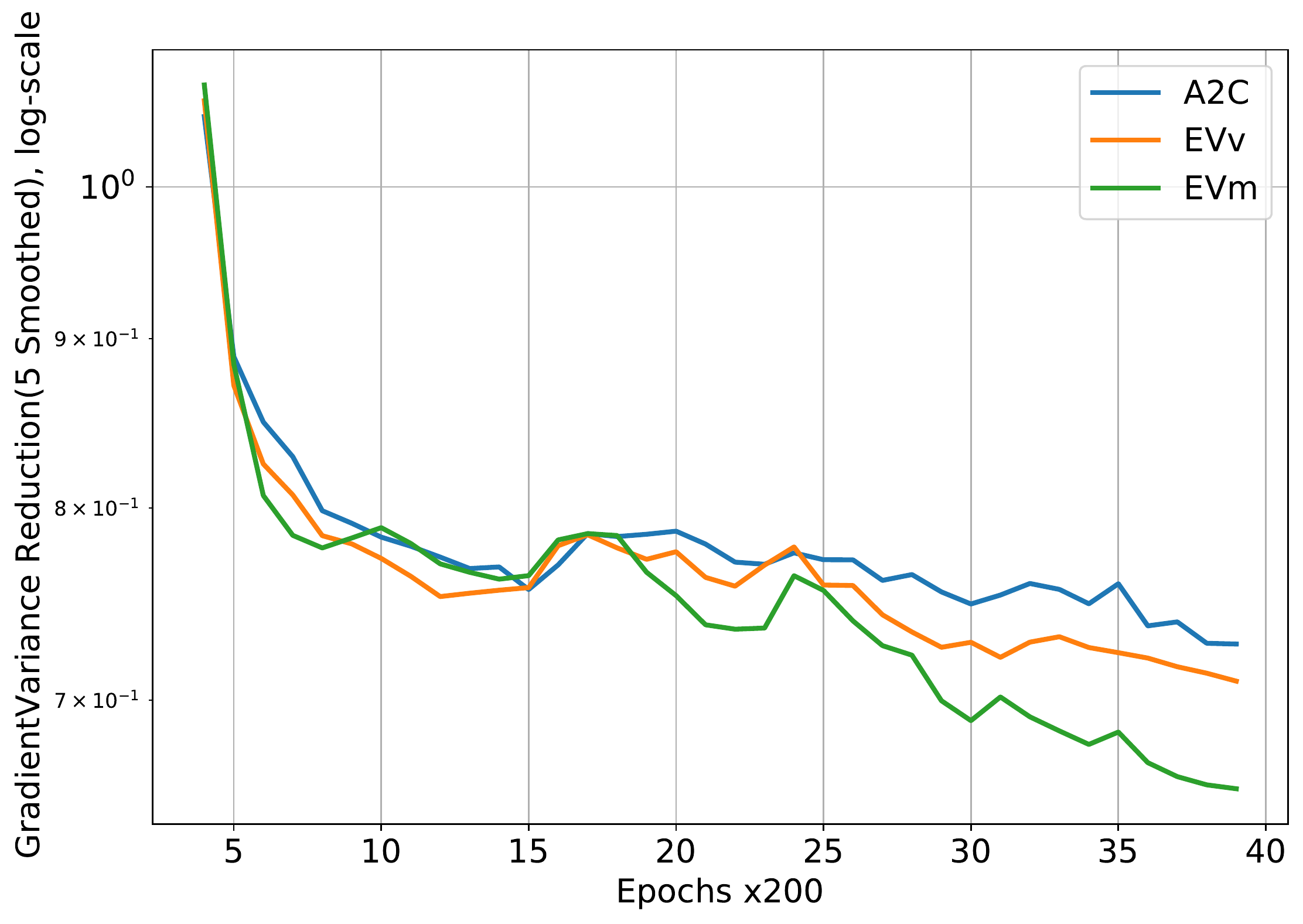} 
    (d) \includegraphics[scale=.25]{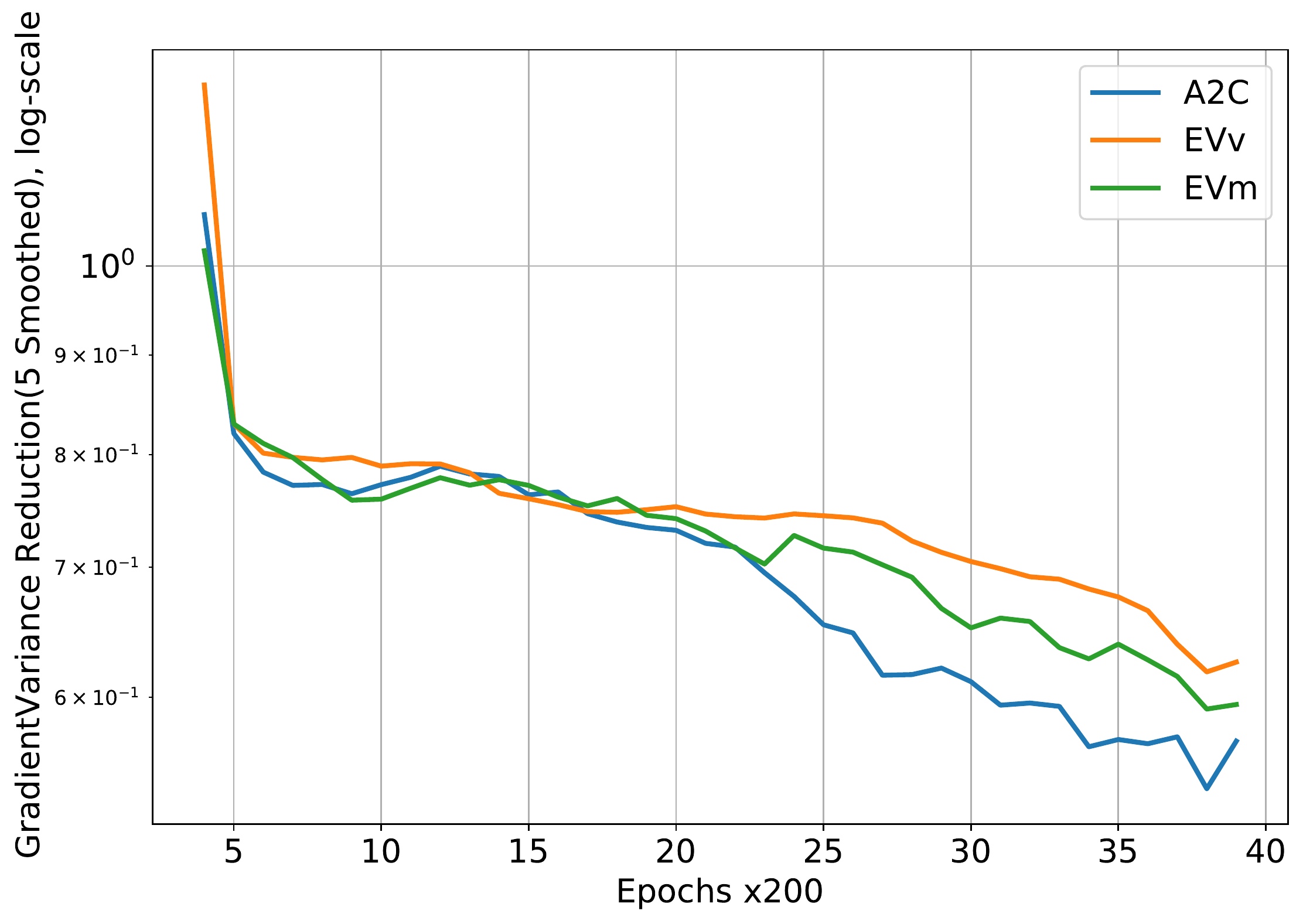}
    \end{tabular}
    \caption{The charts representing the variance of the gradient estimator normalized by the variance of REINFORCE for cases $K=5$(a), $K=10$(b), $K=15$(c), $K=20$(d). The numbers $<1$ indicate the relative reduction. The results are averaged over 20 runs. The resulting curves are smoothed with sliding window of size 5.}
    \label{fig:sup_GoToDoor_gradvar_normalized}
\end{figure}

We also address the effect of gradient variance reduction and its effect on the performance of the algorithm. We can see that variance reduction depends on the number of samples. It's negligible, most of the time even an increase is presented, when number of samples is small ($K=5$ and $K=10$). We see reduction happening with larger $K=15$ and $K=20$. It seems that variance reduction might speed-up the training process, but it is clearly not a key contributor. Variance reduction also seems to be useless at the start since EV-agents' and A2C's seem to have even higher variance than REINFORCE and better performance. However, later reduction might allow to increase final rewards. With $K=15, 20$ the algorithms are able to reduce REINFORCE gradient variance only by $30\%$. We give charts demonstrating gradient variance in absolute values in Fig. \ref{fig:sup_GoToDoor_gradvar} and in relative scale (normalized by the gradient variance of REINFORCE) in Fig. \ref{fig:sup_GoToDoor_gradvar_normalized}.

\clearpage

\begin{figure}[h!]
    \centering
    \begin{tabular}{l}
    (a) \includegraphics[scale=.25]{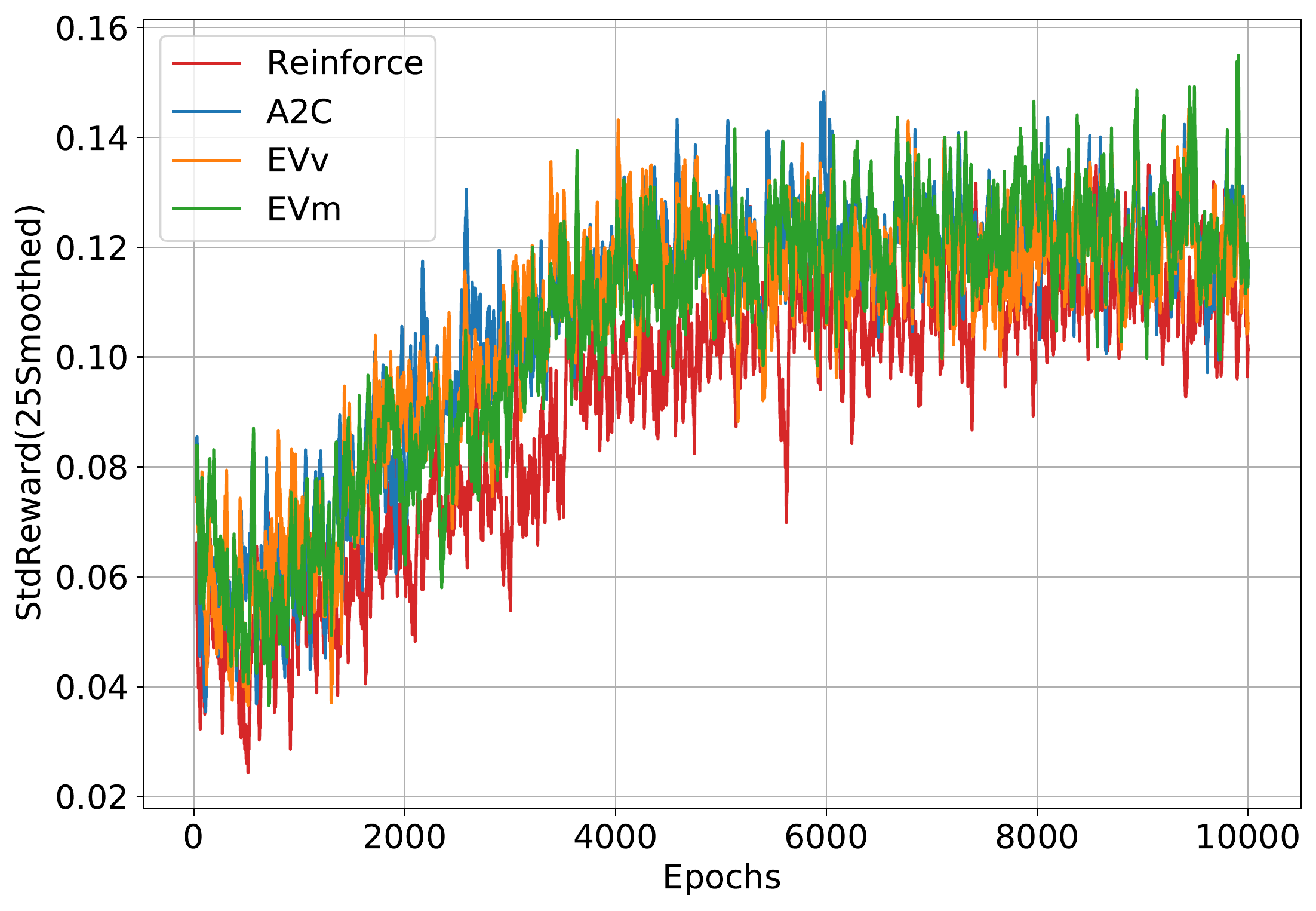} 
    (b) \includegraphics[scale=.25]{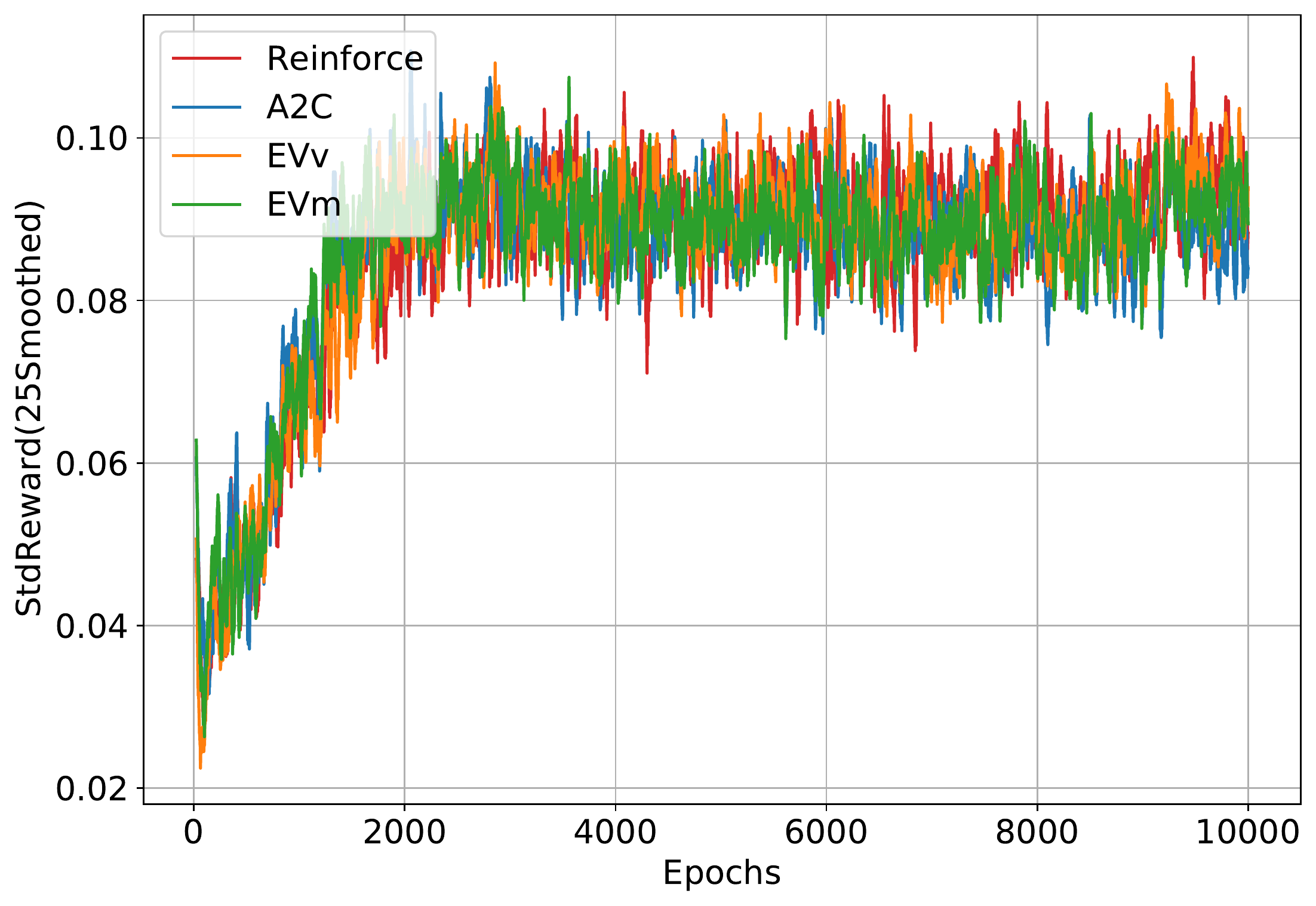} \\ 
    (c) \includegraphics[scale=.25]{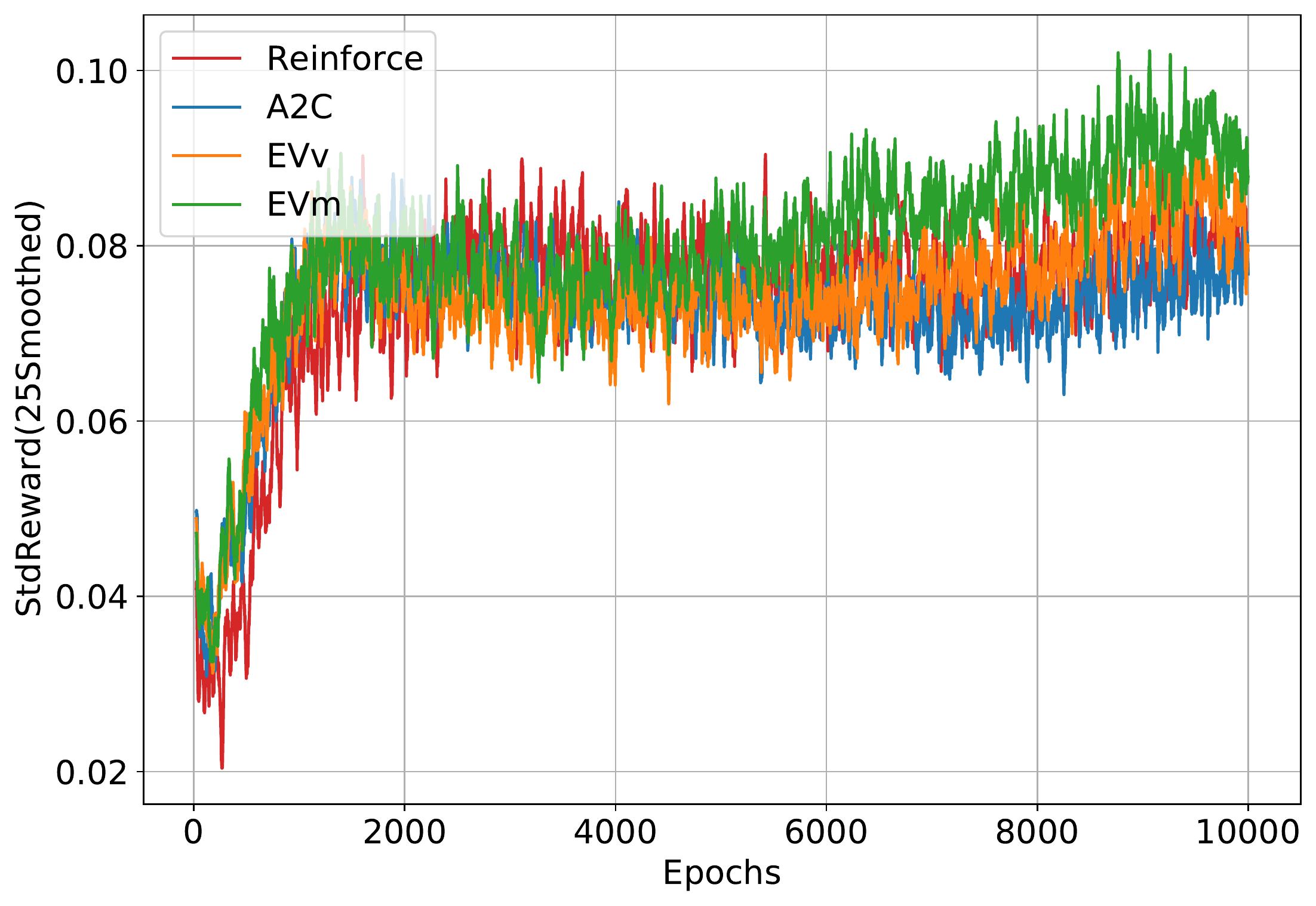} 
    (d) \includegraphics[scale=.25]{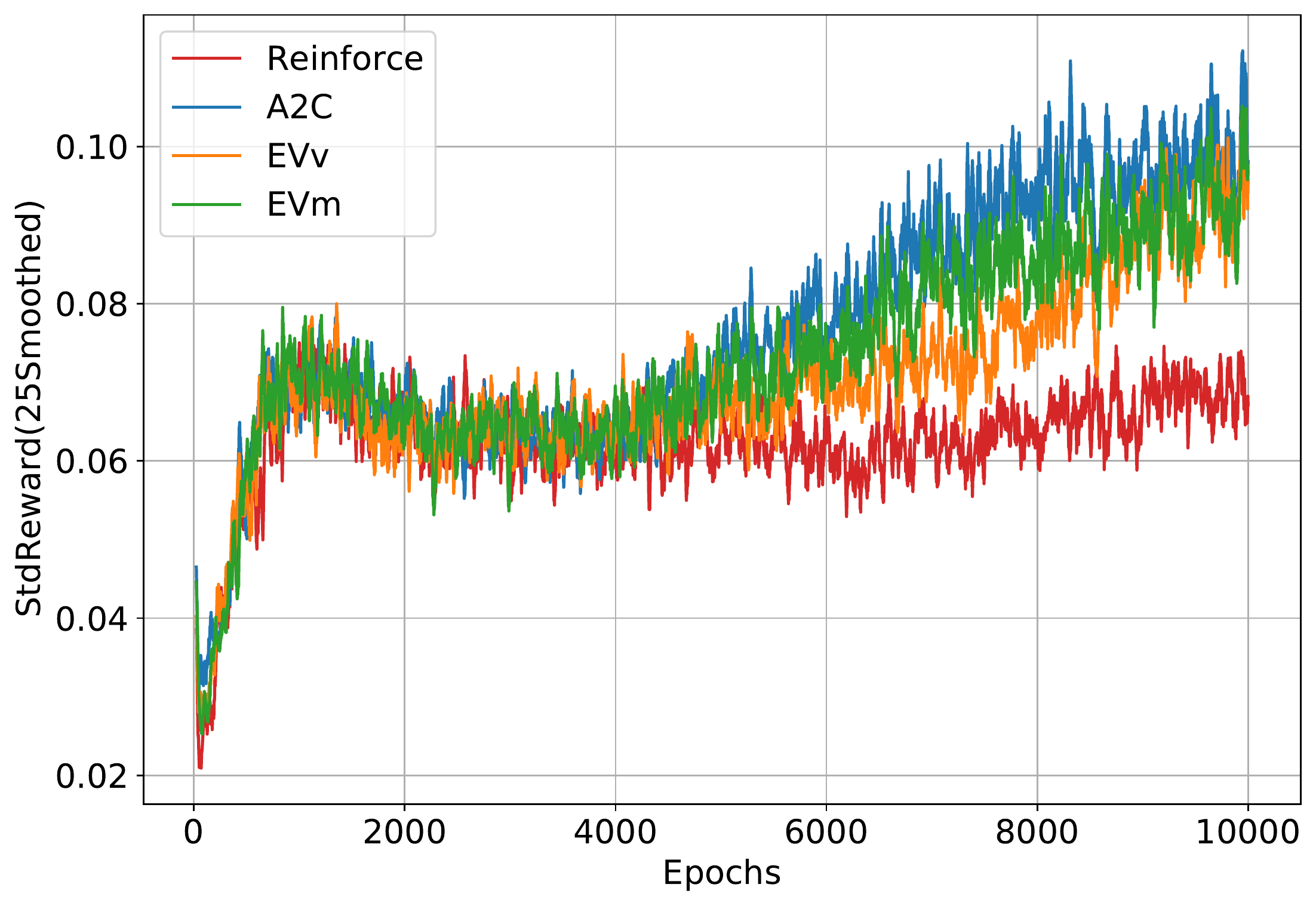}
    \end{tabular}
    \caption{The charts representing the standard deviation of the rewards for cases $K=5$(a), $K=10$(b), $K=15$(c), $K=20$(d). The results are averaged over 20 runs. The resulting curves are smoothed with sliding window of size 25.}
    \label{fig:sup_GoToDoor_stdrew}
\end{figure}

\begin{figure}[h!]
    \centering
    \begin{tabular}{l}
    (a) \includegraphics[scale=.25]{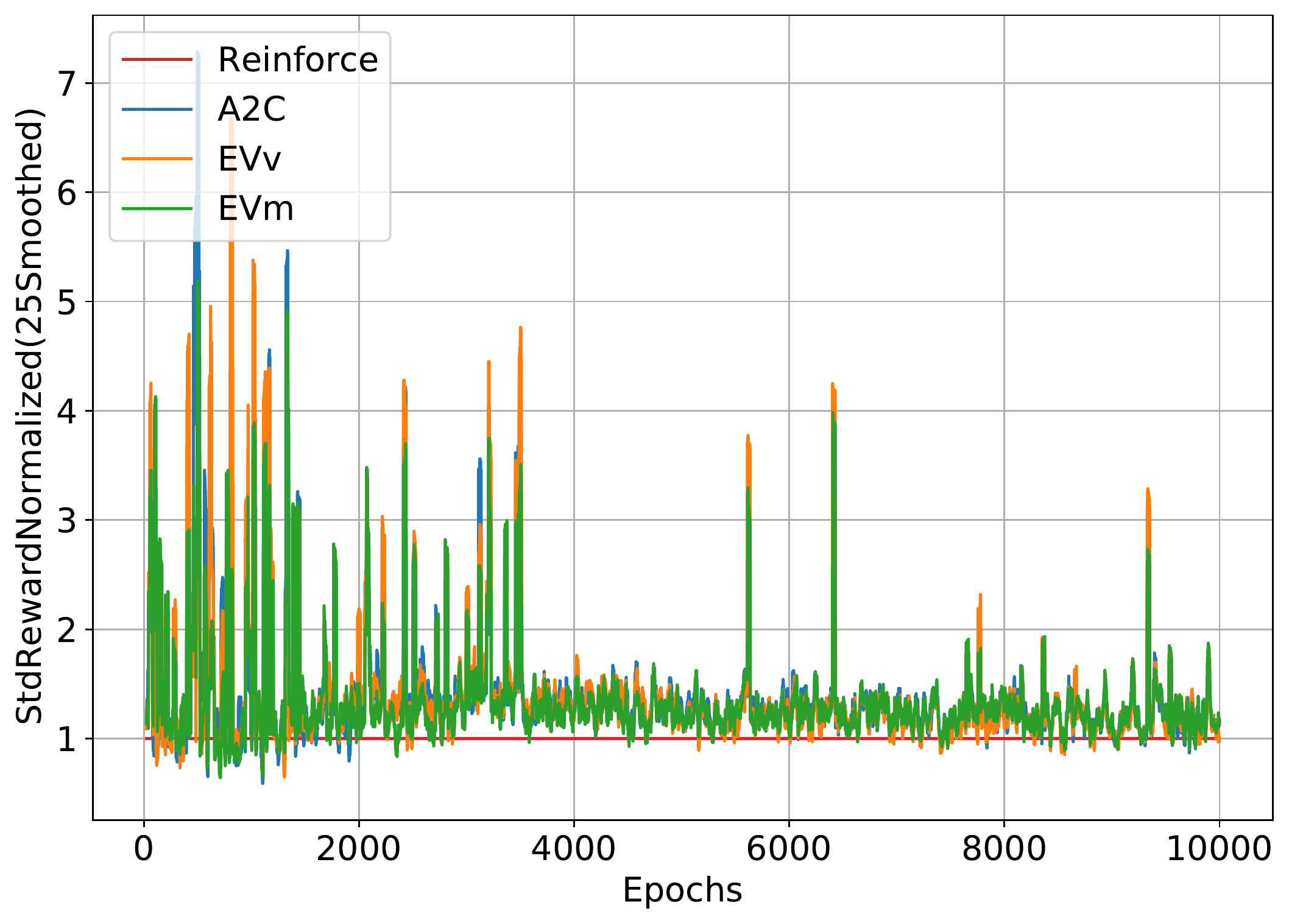} 
    (b) \includegraphics[scale=.25]{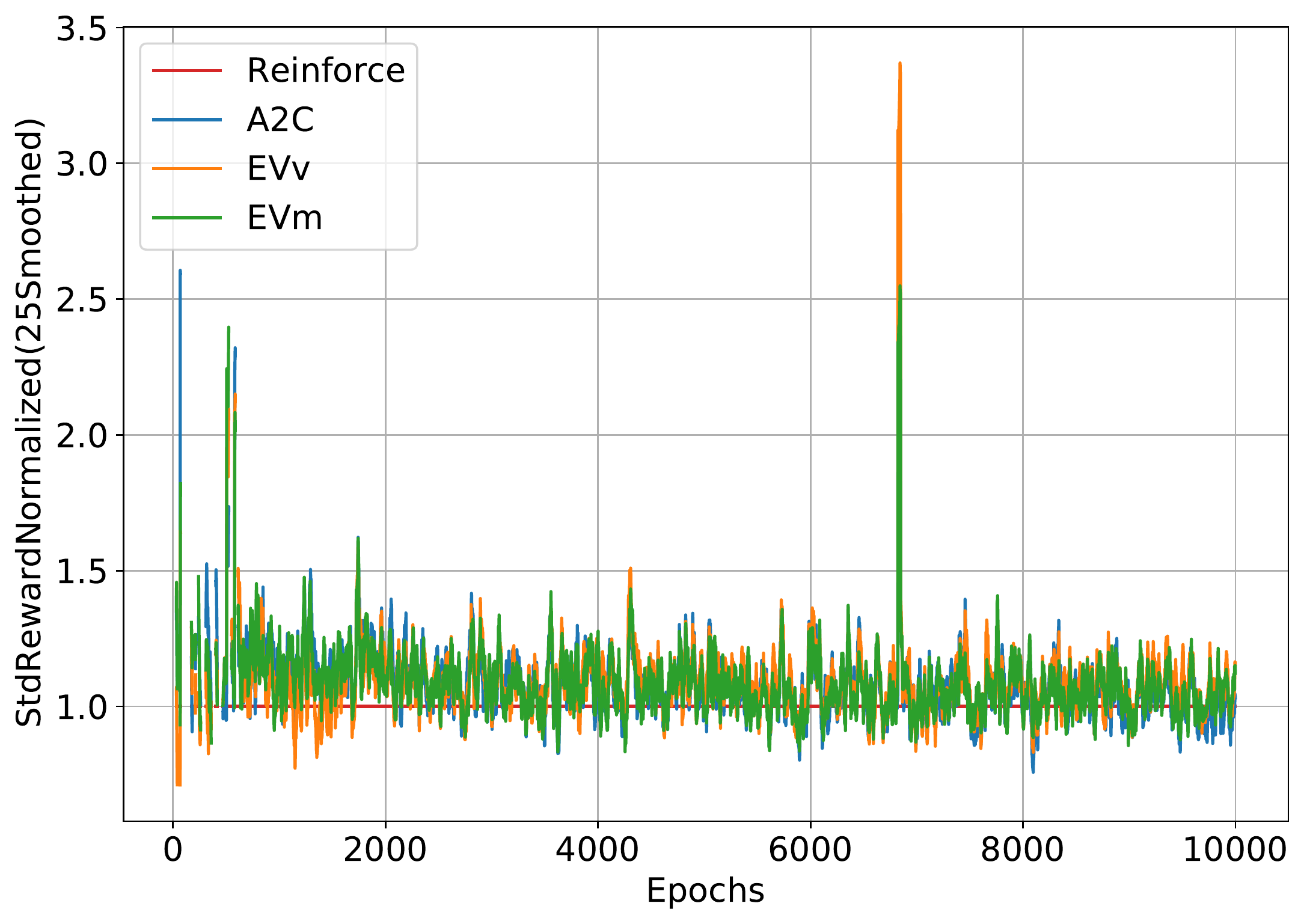}\\ 
    (c) \includegraphics[scale=.25]{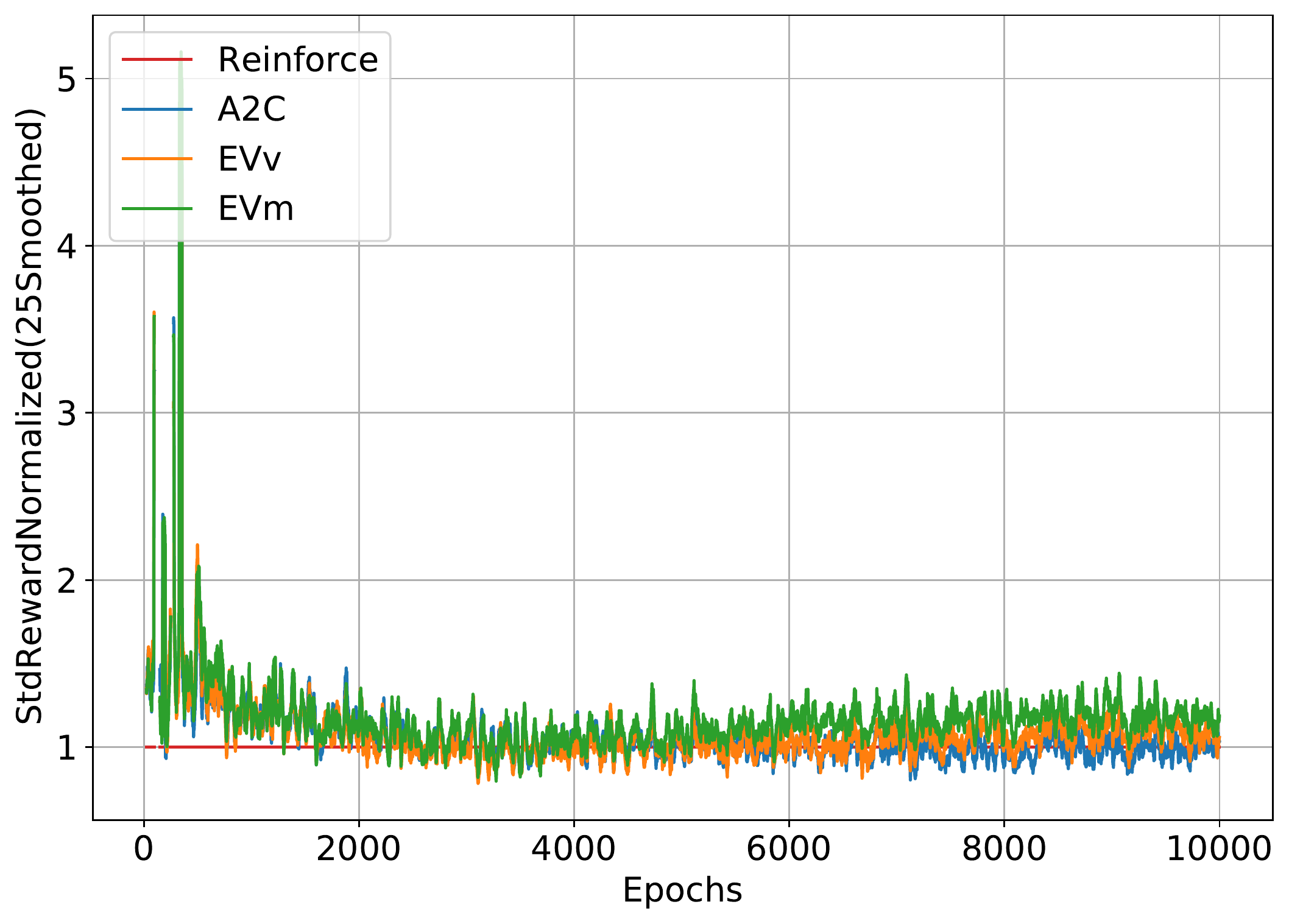} 
    (d) \includegraphics[scale=.25]{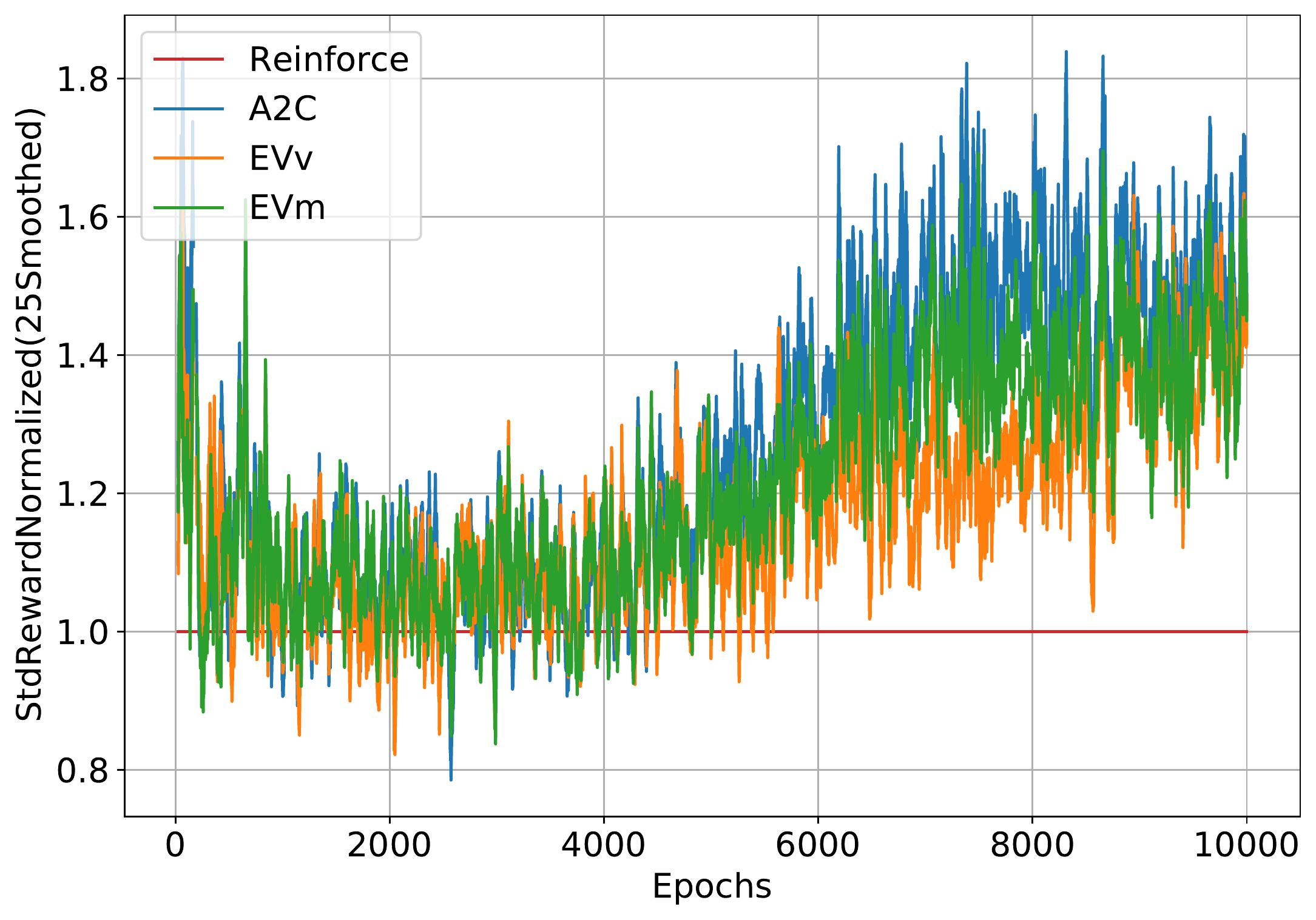}
    \end{tabular}
    \caption{The charts representing the standard deviation of the rewards normalized by the standard deviation of the REINFORCE for cases $K=5(a)$, $K=10$(b), $K=15$(c), $K=20$(d). The results are averaged over 20 runs. The resulting curves are smoothed with sliding window of size 5.}
    \label{fig:sup_GoToDoor_stdrew_normalized}
\end{figure}

One can also evaluate the algorithms by looking at the standard deviation of the rewards. Between the methods no significant difference is observed when the sample size is small ($K=5$ or $K=10$). It becomes considerable though in cases of $K=15$ and $K=20$. EV-agents turn out to have the biggest reward standard deviation among the algorithms. The standard deviation of the rewards is demonstrated in absolute values in Fig. \ref{fig:sup_GoToDoor_stdrew} and in relative scale (normalized by the standard deviation of REINFORCE) in Fig. \ref{fig:sup_GoToDoor_stdrew_normalized}. We note that this standard deviation does not at all reflect the variance reduction of the gradient estimator as follows from the comparison of the charts. In fact, REINFORCE with no variance reduced is slightly better in this regard than other methods.

%% file: suppMinigridUnlock.tex
The agent has to open a locked door. First, it has to find a key and then go to the door.\\

\begin{figure}[h!]
    \centering
    \begin{tabular}{l}
    (a) \includegraphics[scale=.25]{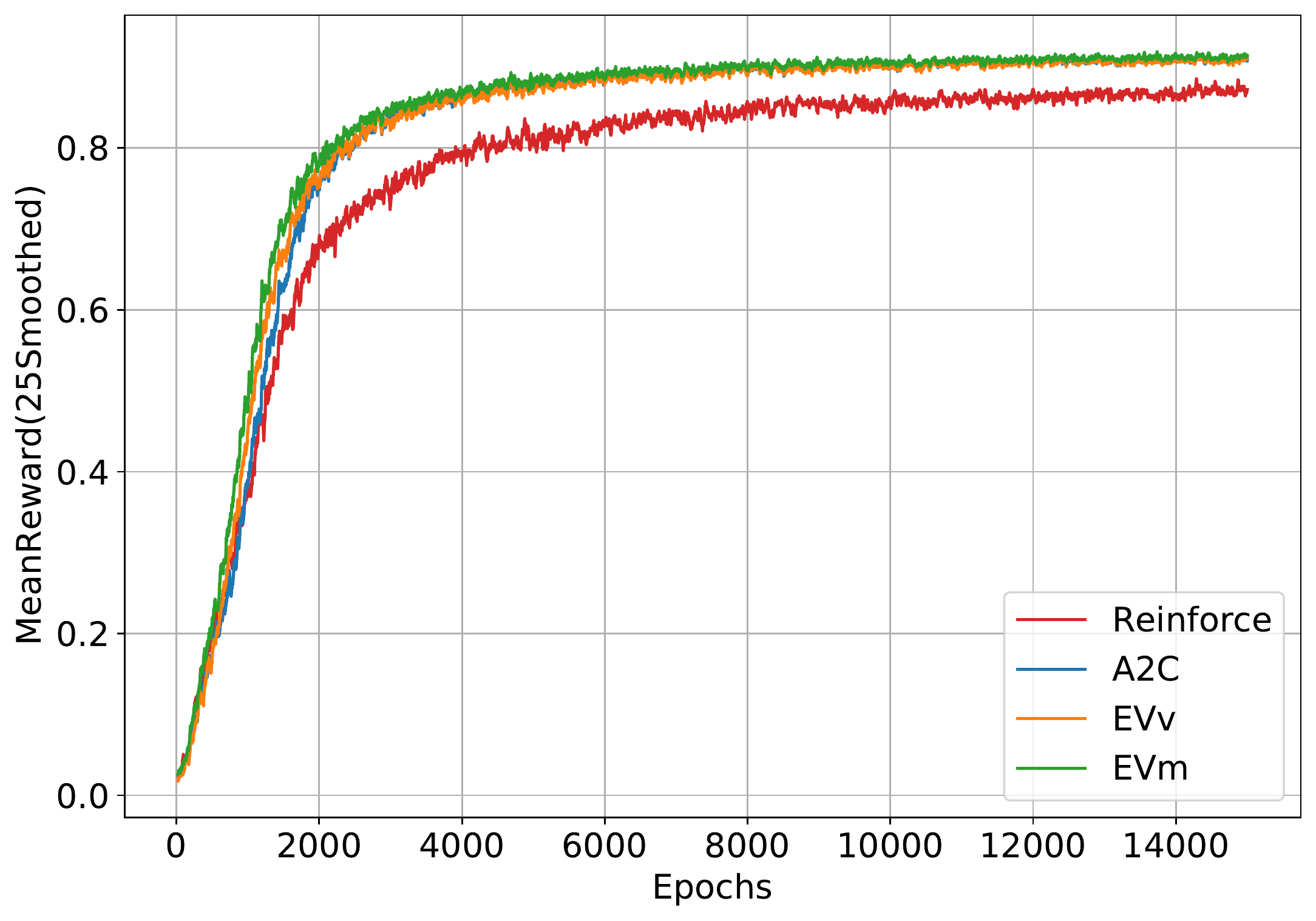} 
    (b) \includegraphics[scale=.25]{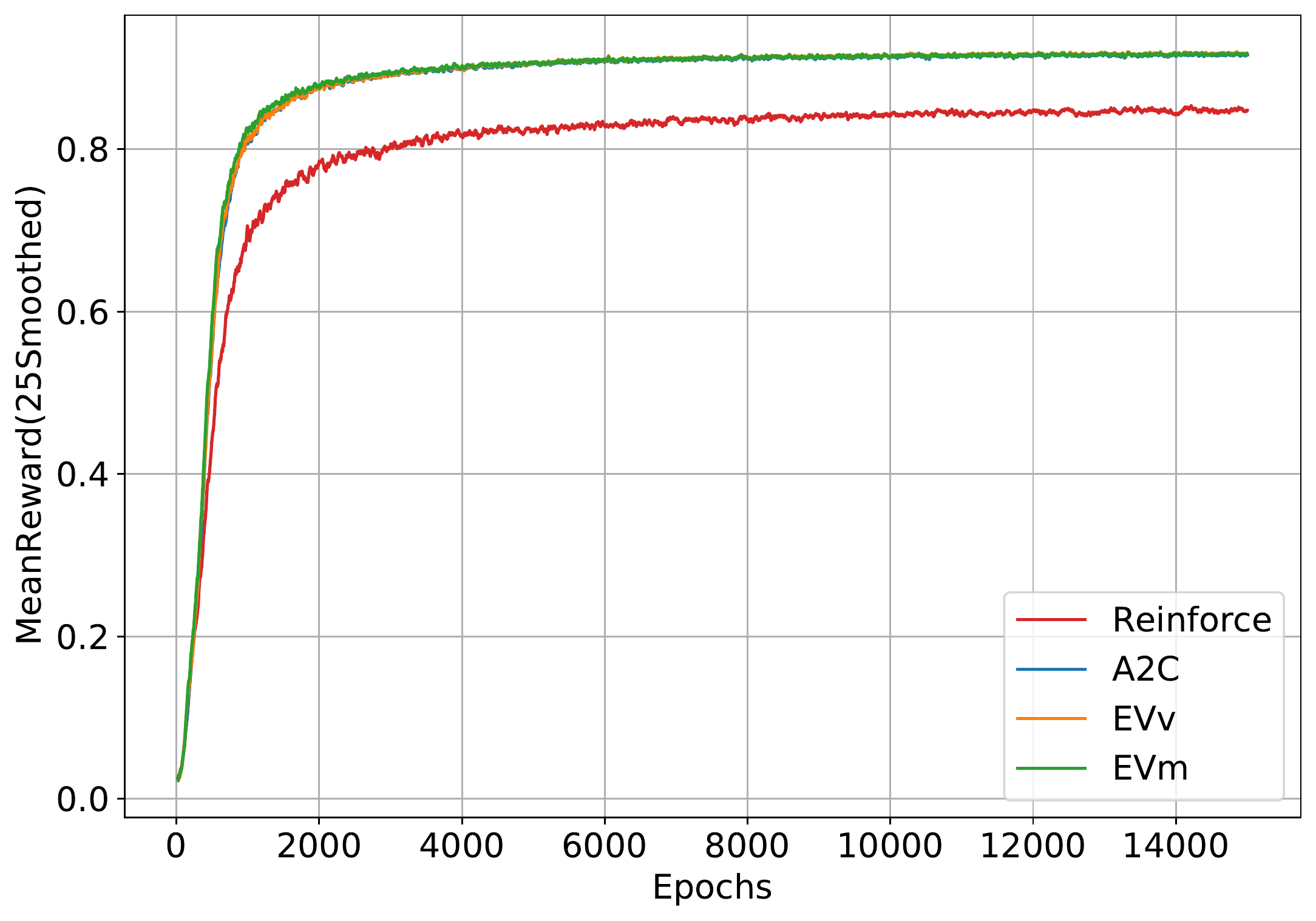} 
    \end{tabular}
    \caption{The charts representing mean rewards in Unlock environment, standing for absolute values. The results are averaged over 20 runs. The resulting curves are smoothed with sliding window of size 25.}
    \label{fig:sup_Unlock_rew}
\end{figure}

\begin{figure}[h!]
    \centering
    \begin{tabular}{l}
    (a) \includegraphics[scale=.25]{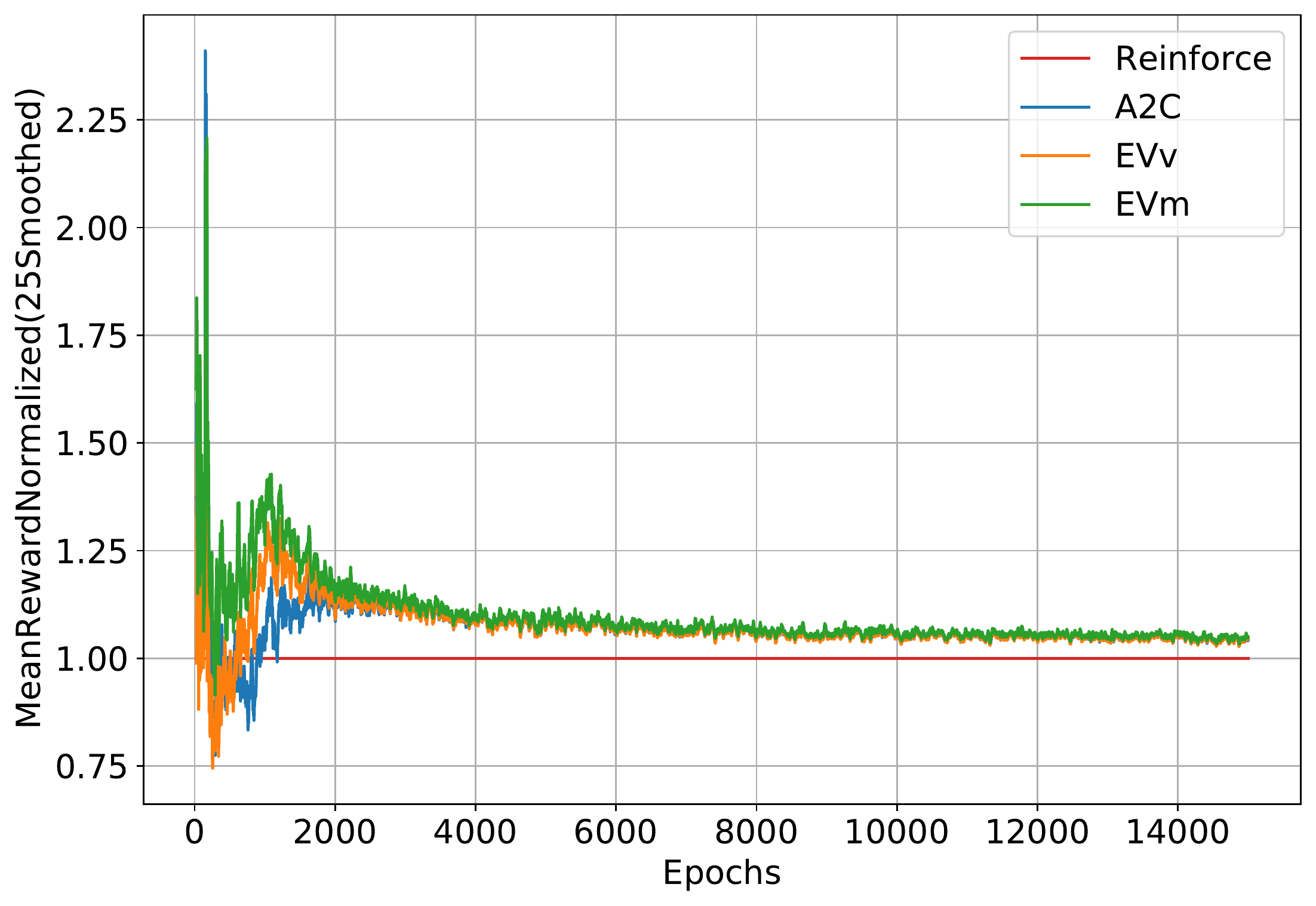} 
    (b) \includegraphics[scale=.25]{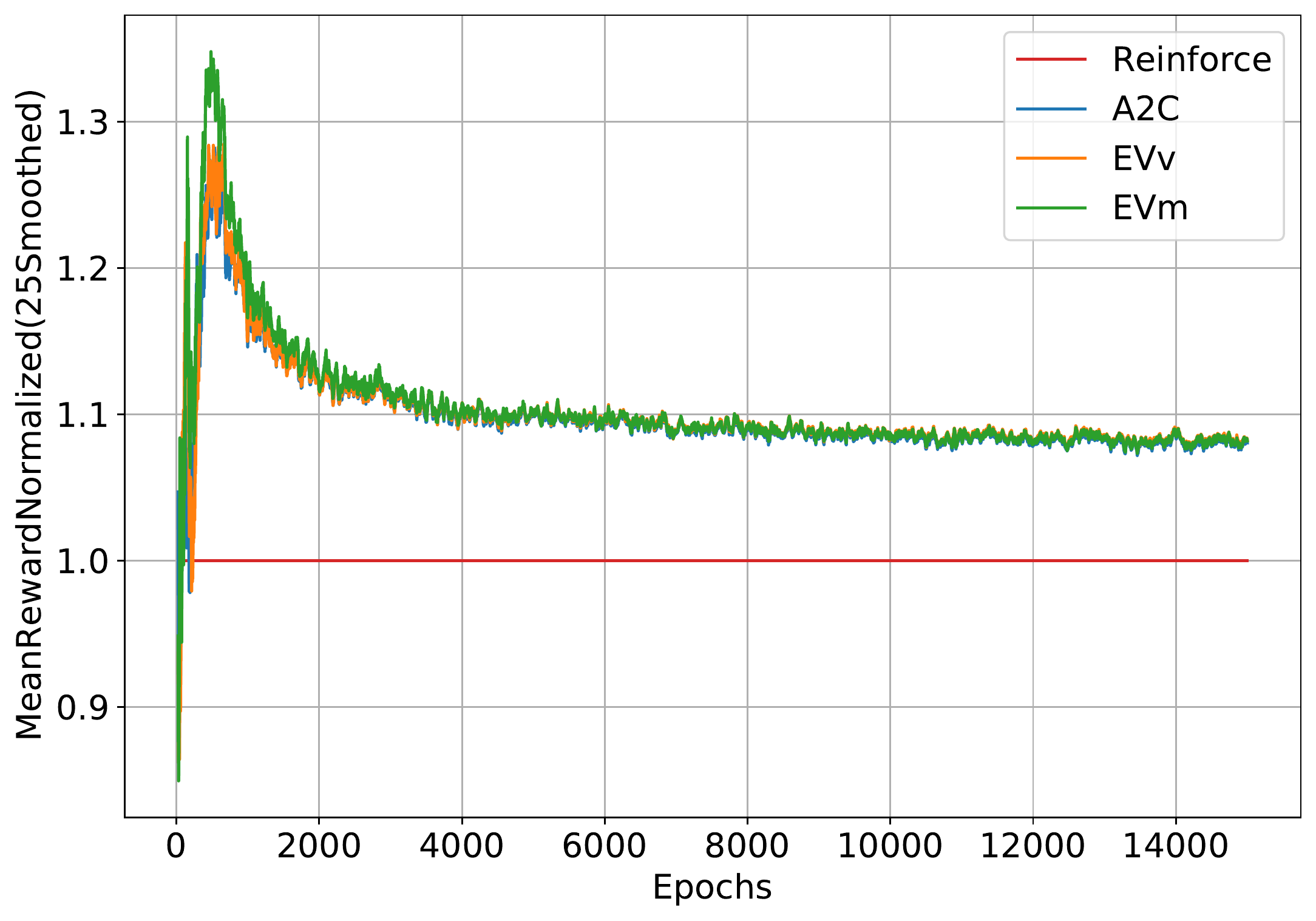} 
    \end{tabular}
    \caption{The charts representing mean rewards in Unlock environment, normalized by the mean reward of the REINFORCE. The results are averaged over 20 runs. The resulting curves are smoothed with sliding window of size 25.}
    \label{fig:sup_Unlock_rew_normalized}
\end{figure}

\begin{figure}[h!]
    \centering
    \begin{tabular}{l}
    (a) \includegraphics[scale=.25]{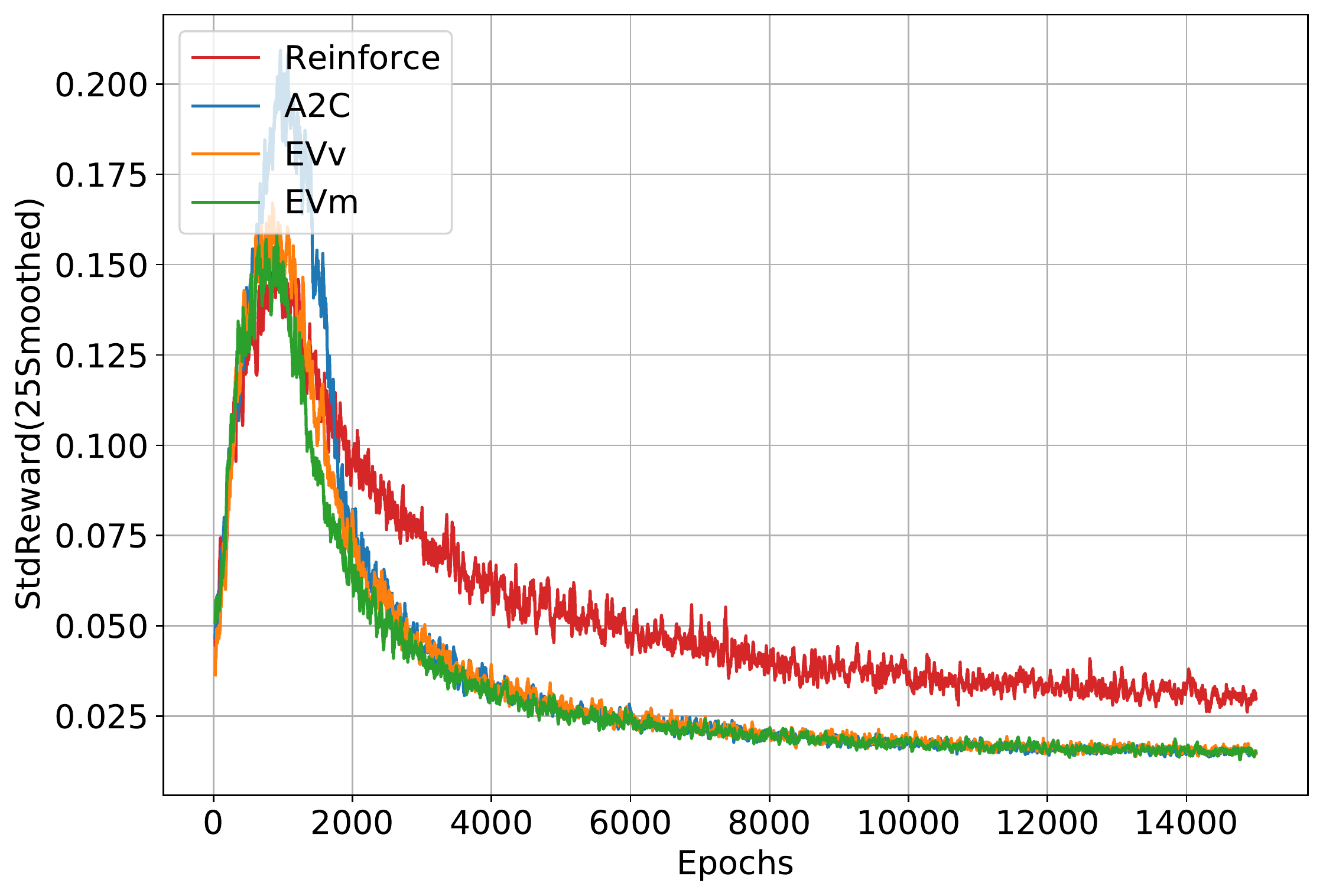} 
    (b) \includegraphics[scale=.25]{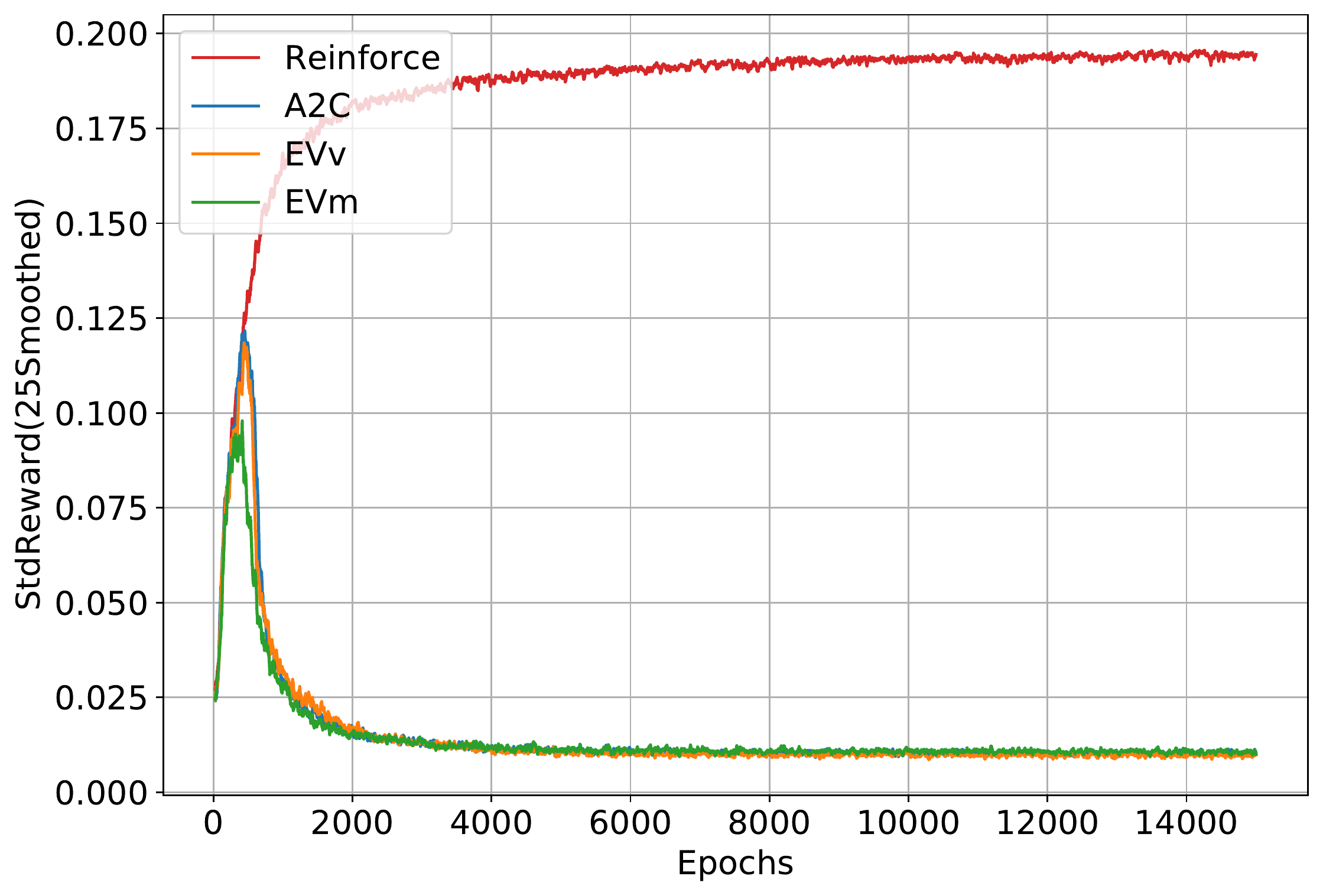} 
    \end{tabular}
    \caption{The charts representing the standard deviation of the rewards. The results are averaged over 20 runs. The resulting curves are smoothed with sliding window of size 25.}
    \label{fig:sup_Unlock_stdrew}
\end{figure}

\begin{figure}[h!]
    \centering
    \begin{tabular}{l}
    (a) \includegraphics[scale=.25]{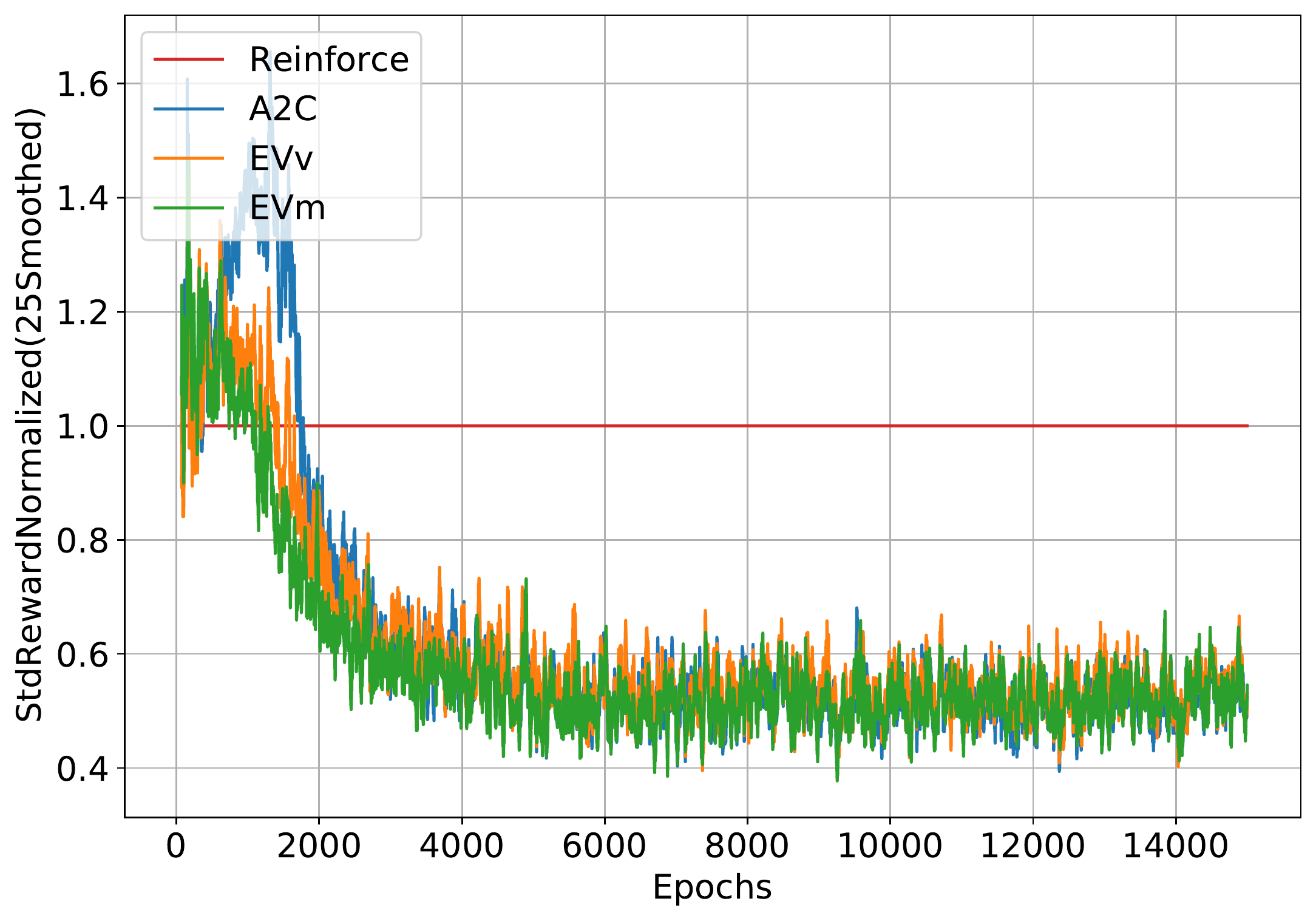} %
    (b) \includegraphics[scale=.25]{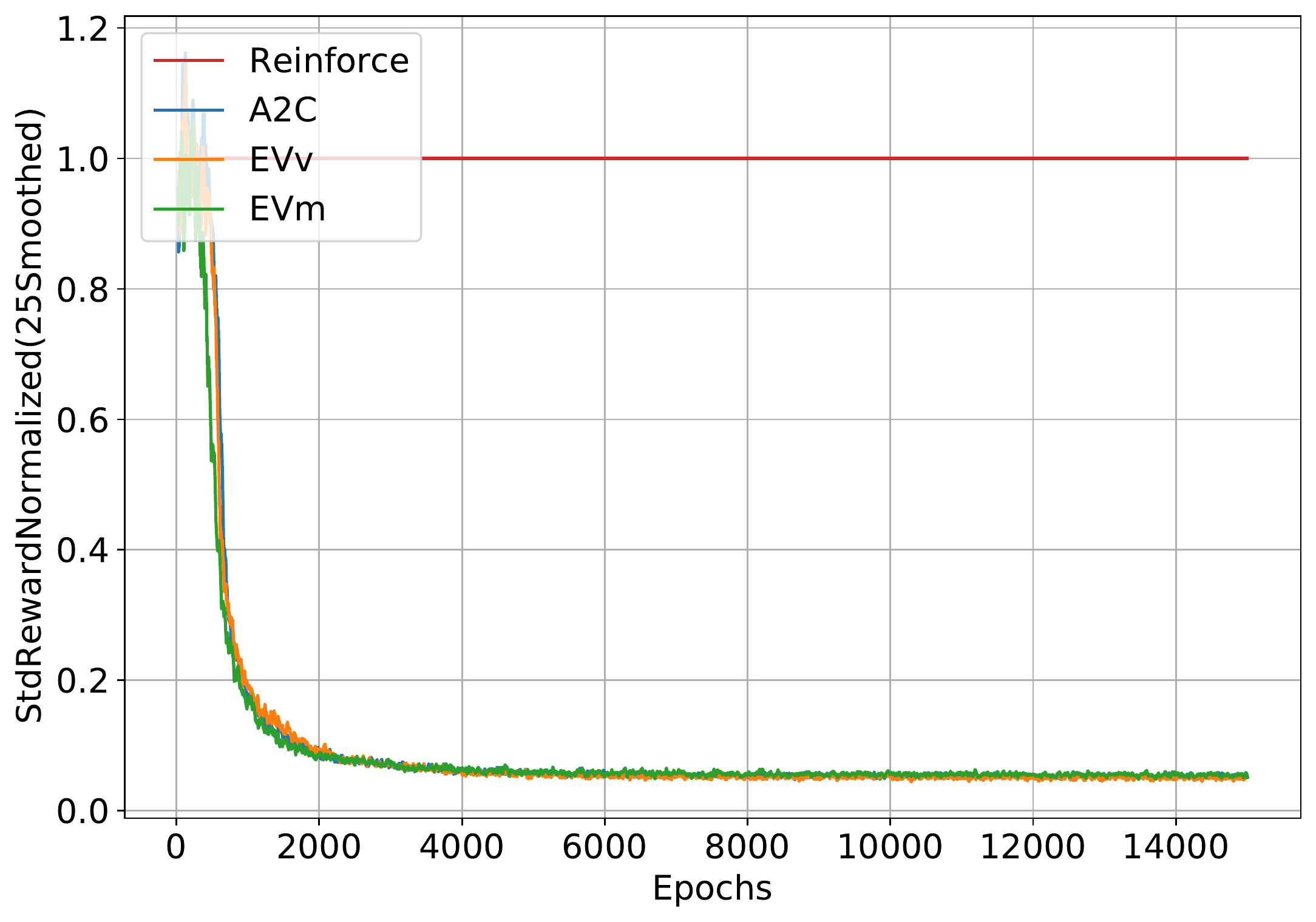} 
    \end{tabular}
    \caption{The charts representing the standard deviation of the rewards normalized by the standard deviation of the REINFORCE. The results are averaged over 20 runs. The resulting curves are smoothed with sliding window of size 25.}
    \label{fig:sup_Unlock_stdrew_normalized}
\end{figure}

In this environment we considered two different sample sizes: $K=5$ and $K=20$. Here EV agents and A2C seem to converge to the same policy (see Fig. \ref{fig:sup_Unlock_rew} and Fig. \ref{fig:sup_Unlock_rew_normalized}). The charts on Fig. \ref{fig:sup_Unlock_gradvar} indicate that the variance is reduced 10-100 times similarly for A2C- and EV-algorithms. Considering mean rewards we clearly see that such reduction results in considerable gain of about 10-20\% and, what is important, it considerably adds to the stability which is displayed on Fig. \ref{fig:sup_Unlock_stdrew} and \ref{fig:sup_Unlock_stdrew_normalized}. 

Important thing to notice is that in the beginning (before approximately 2000 Epochs passed) we observe small gain of EV over A2C (especially in case of smaller sample with $K=5$) and it goes together with more stability which is indicated by the plots of standard deviation. Hence, a clever use of EV method instead of A2C sometimes can give an additional confident gain despite the fact that the gradient variance reduction is almost the same.

\begin{figure}[h!]
    \centering
    \begin{tabular}{l}
    (a) \includegraphics[scale=.25]{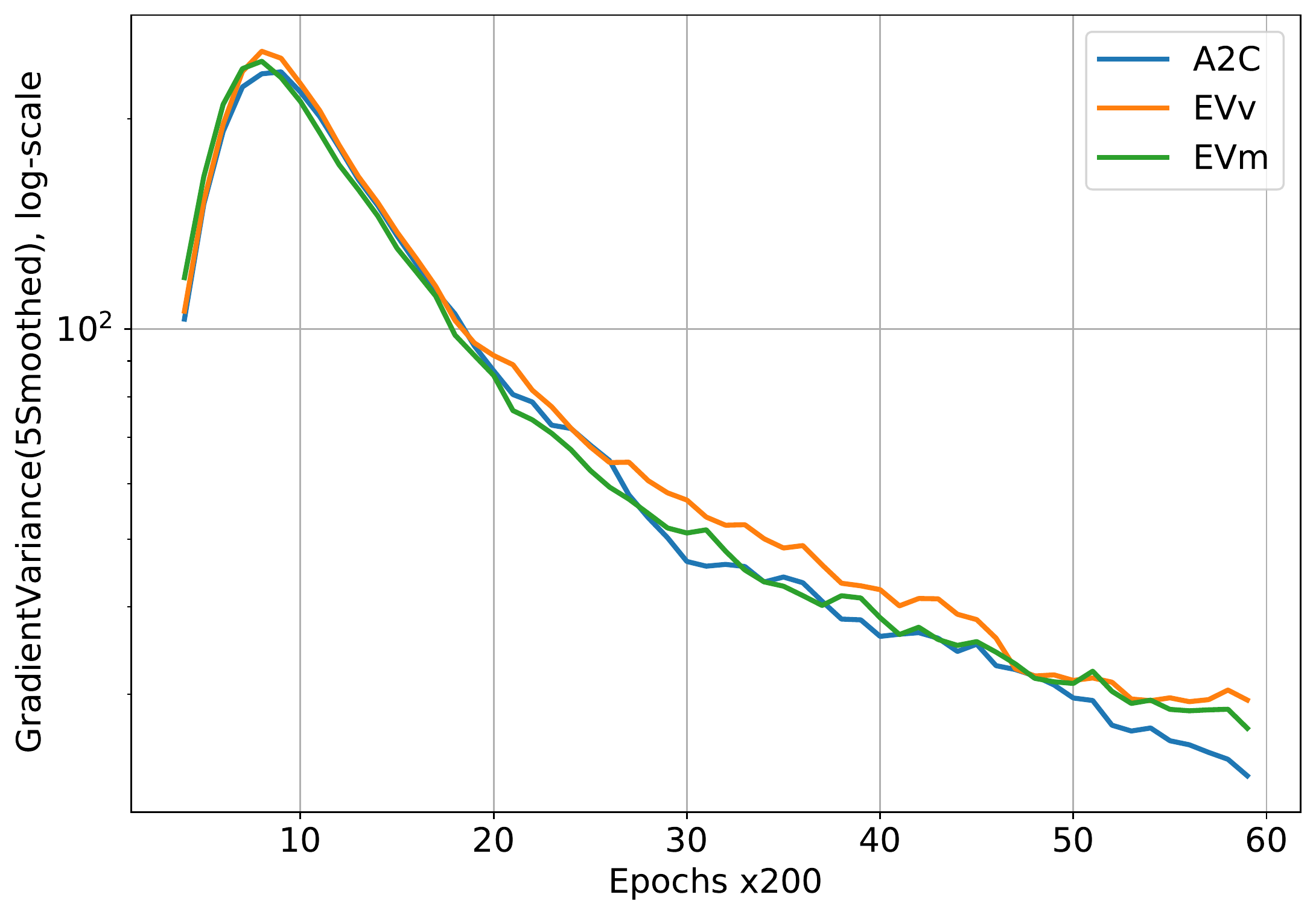} 
    (b) \includegraphics[scale=.25]{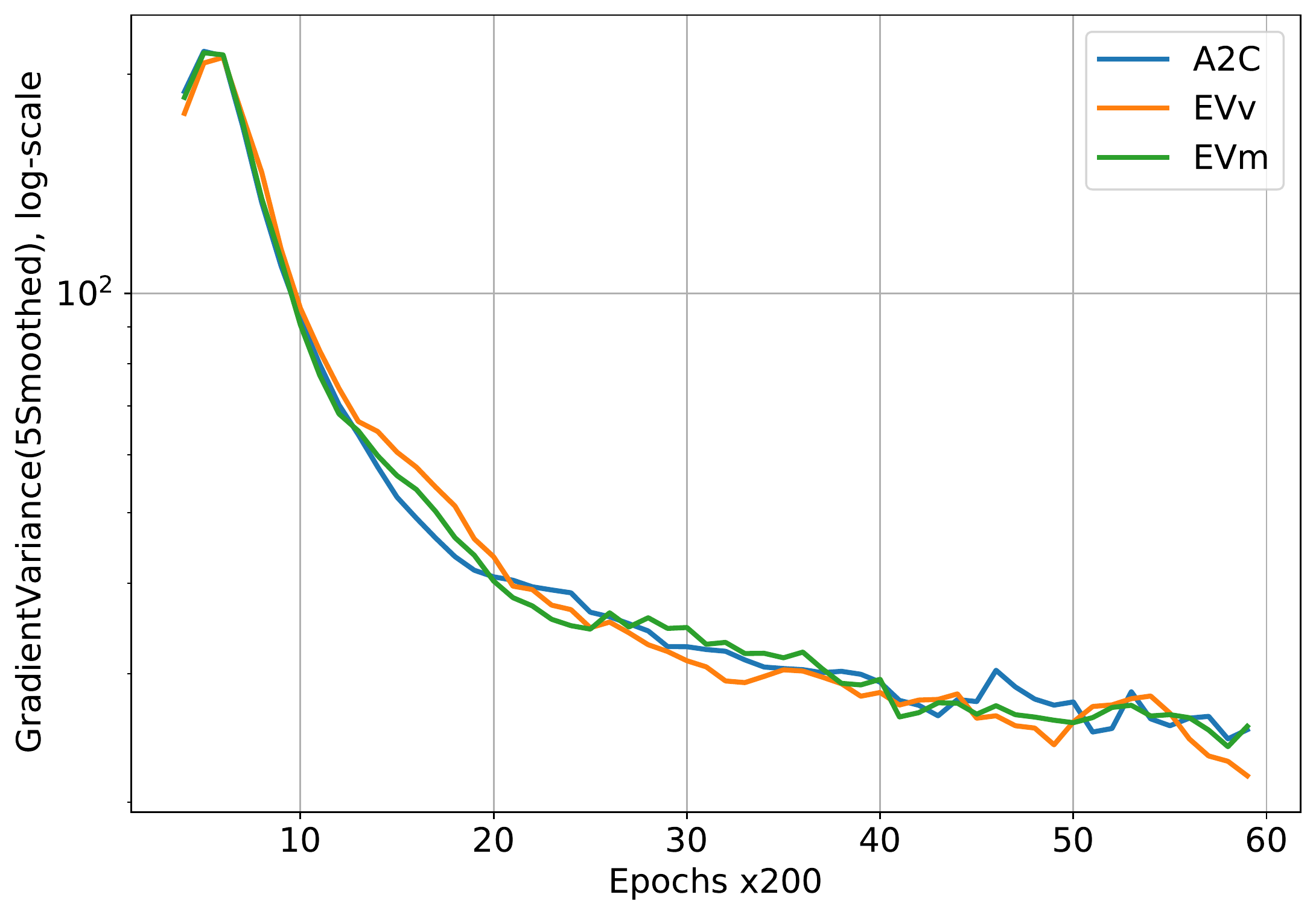} 
    \end{tabular}
    \caption{The charts representing the variance of the gradient estimator in absolute values. The results are averaged over 20 runs. The resulting curves are smoothed with sliding window of size 5.}
    \label{fig:sup_Unlock_gradvar}
\end{figure}

\begin{figure}[h!]
    \centering
    \begin{tabular}{l}
    (a) \includegraphics[scale=.25]{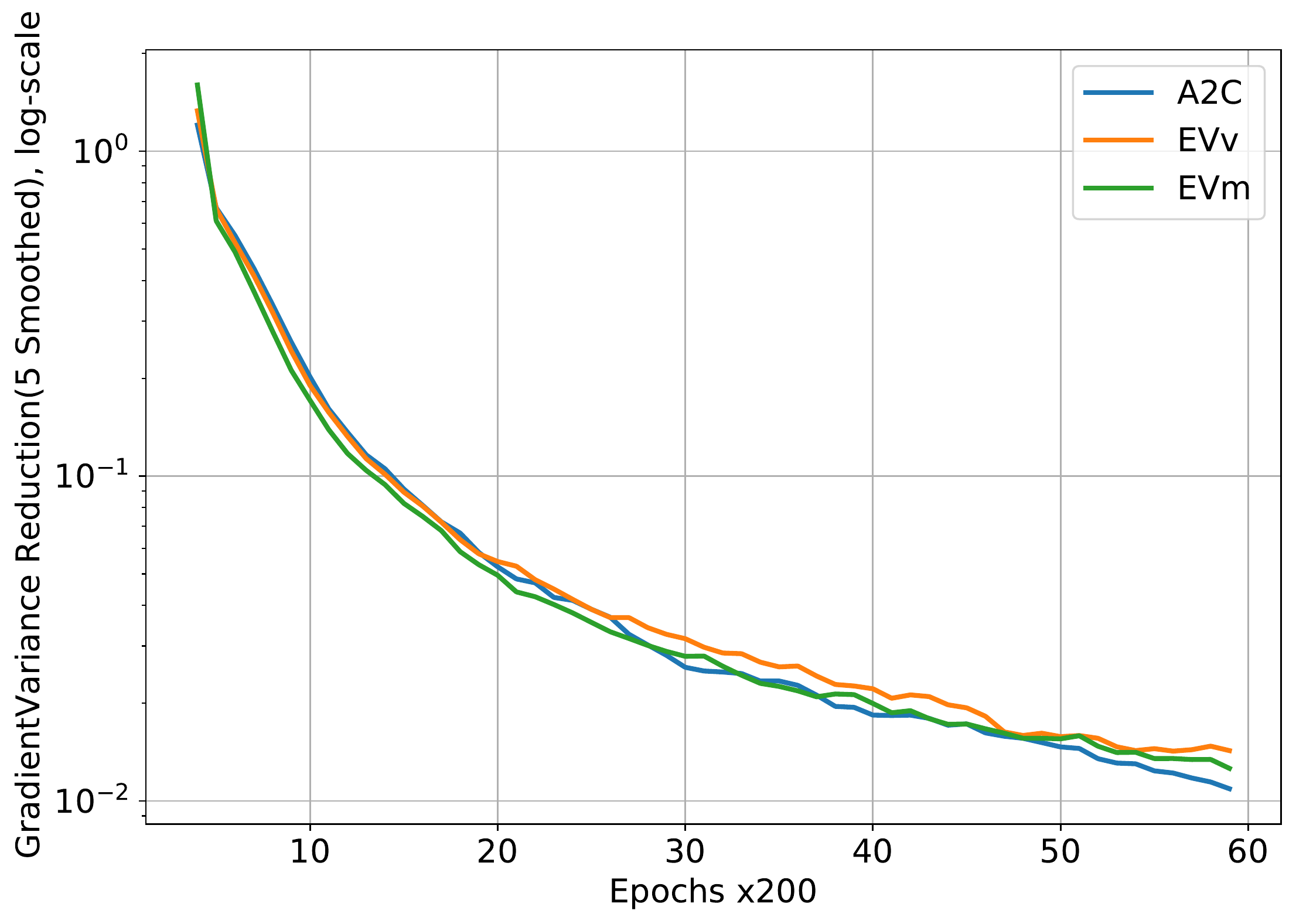} 
    (b) \includegraphics[scale=.25]{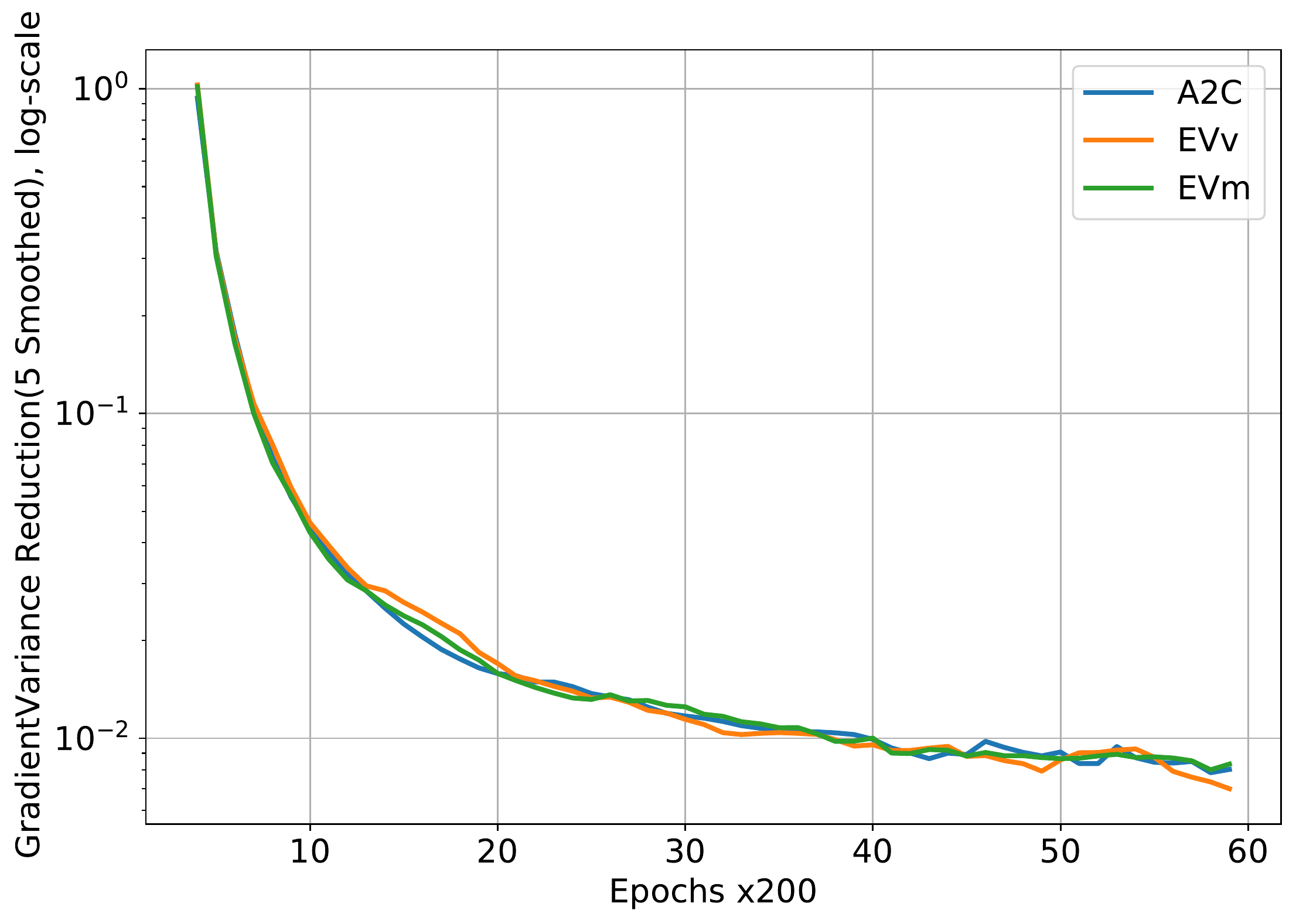} 
    \end{tabular}
    \caption{The charts representing the variance of the gradient estimator normalized by the gradient variance of the REINFORCE (log-scale is set up along y-axis). The results are averaged over 20 runs. The resulting curves are smoothed with sliding window of size 5.}
    \label{fig:sup_Unlock_gradvar}
\end{figure}

%% file: suppCartPole.tex
CartPole is a Gym environment where a pole is attached by a joint to a cart, which moves along x-axis. Agent can apply a force +1 or -1 to the cart making it move right or left. The pole starts upright and the agent has to keep it as long as possible preventing from falling. The agent receives +1 reward every timestamp that the pole remains upright. The episode ends when the pole is more than 15 degrees from vertical, or the cart moves more than 2.4 units from the center. \\

In this environment we demonstrate 5 configurations with different policy and baseline architectures to look how algorithms behave with changing policy and baseline configurations. The exact config-files can be found on GitHub \cite{github}. The measurements of mean rewards are averaged over 40 independent runs of the algorithms and reward variance is measured as the sample variance of the observed rewards in each epoch. We provide the charts relative to REINFORCE which are obtained by dividing the curves by the corresponding values of REINFORCE. These allow to see the improvements over REINFORCE more clearly.\\

\begin{figure}[h!]
    \centering
    \begin{tabular}{lcr}
    (a) \includegraphics[scale=.2]{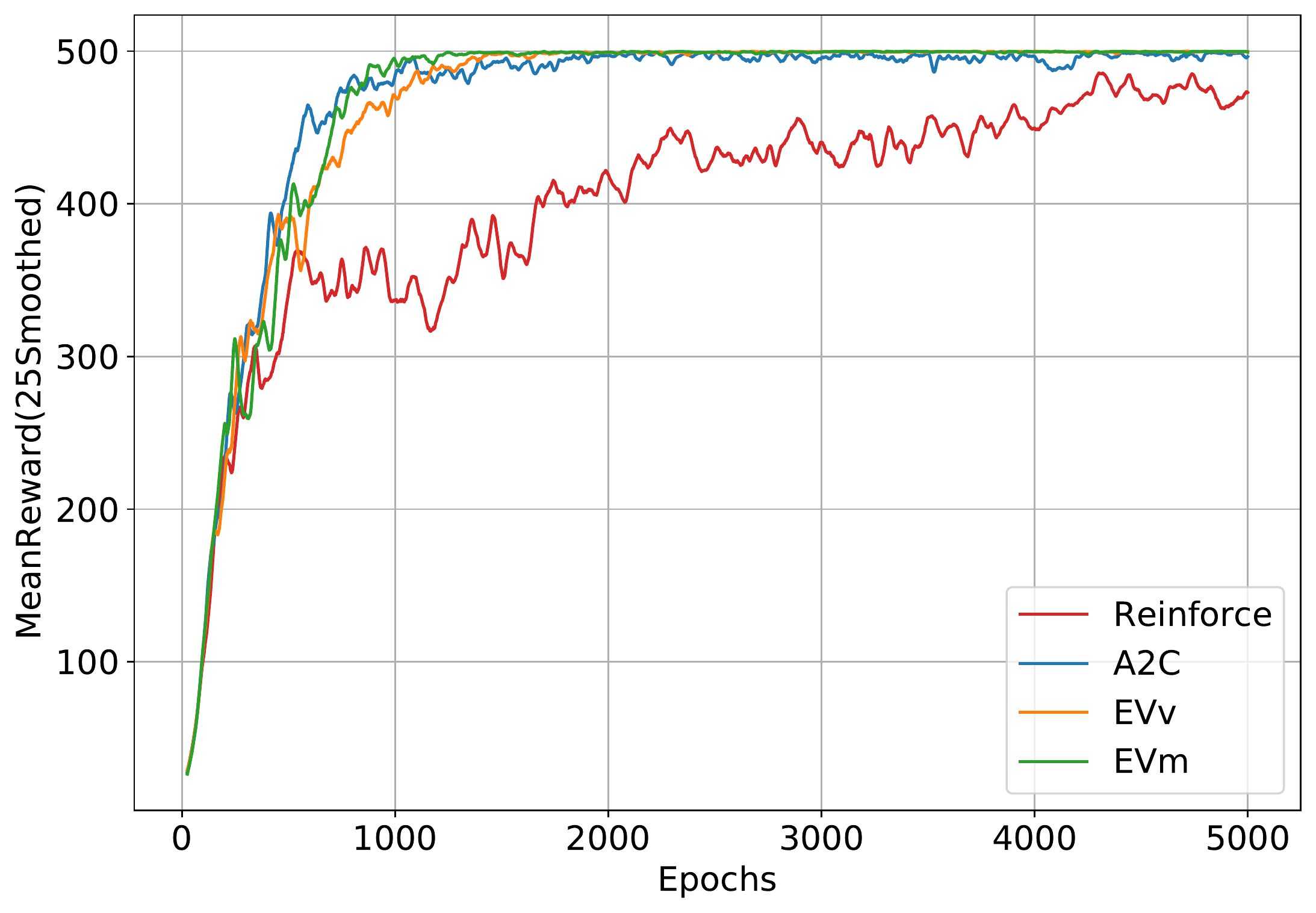} & (b) \includegraphics[scale=.2]{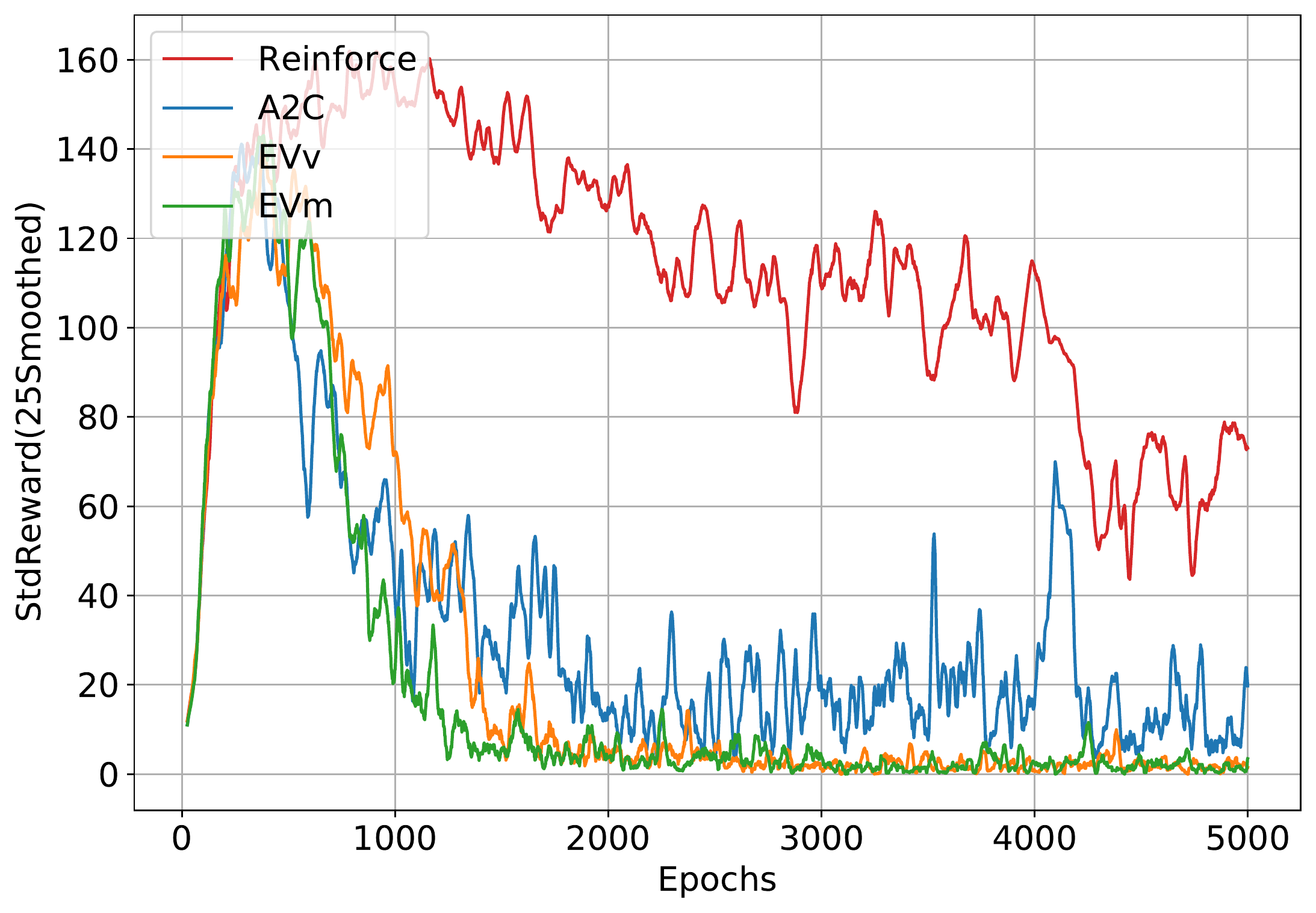} & (c)\includegraphics[scale=.2]{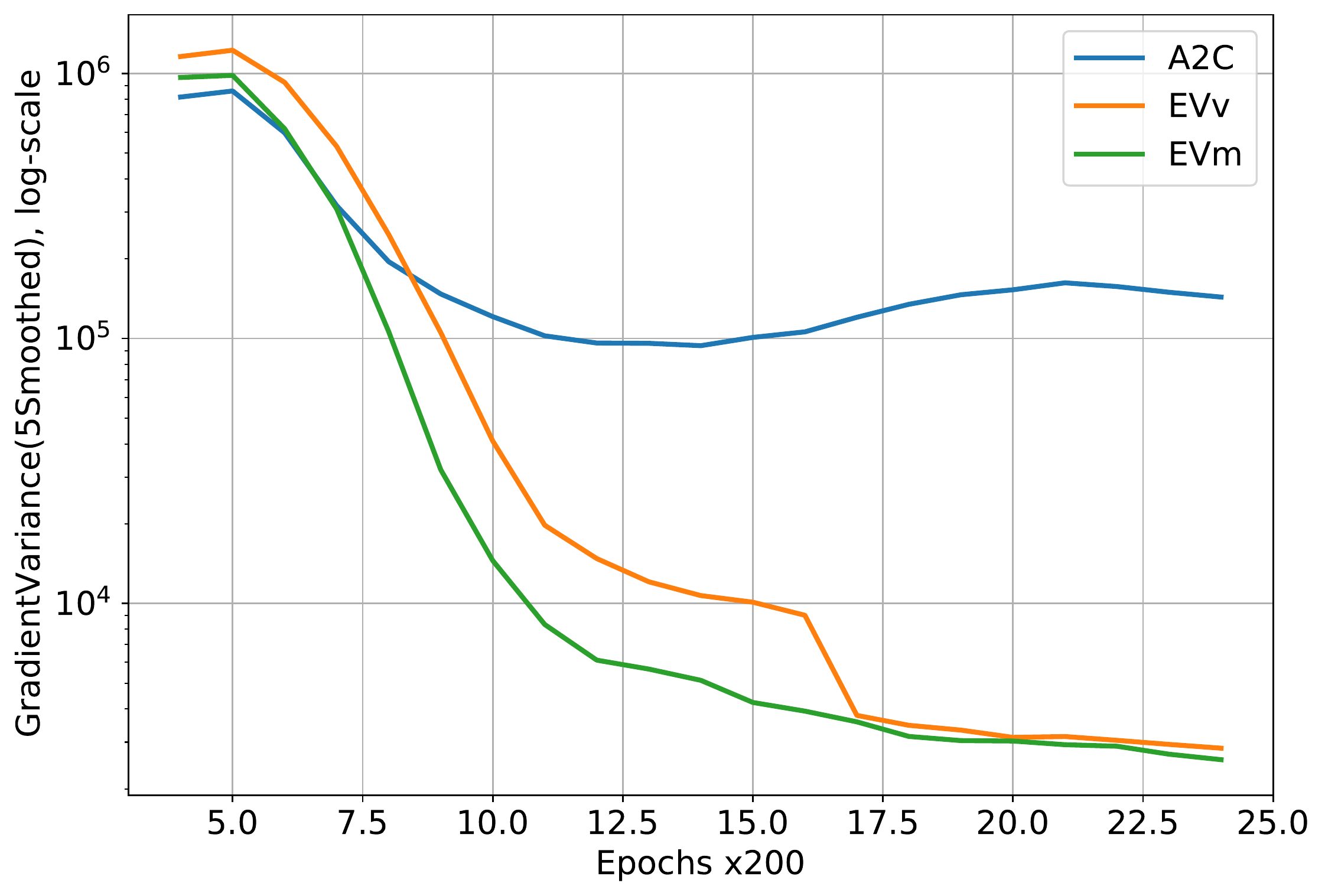} \\
    (d) \includegraphics[scale=.2]{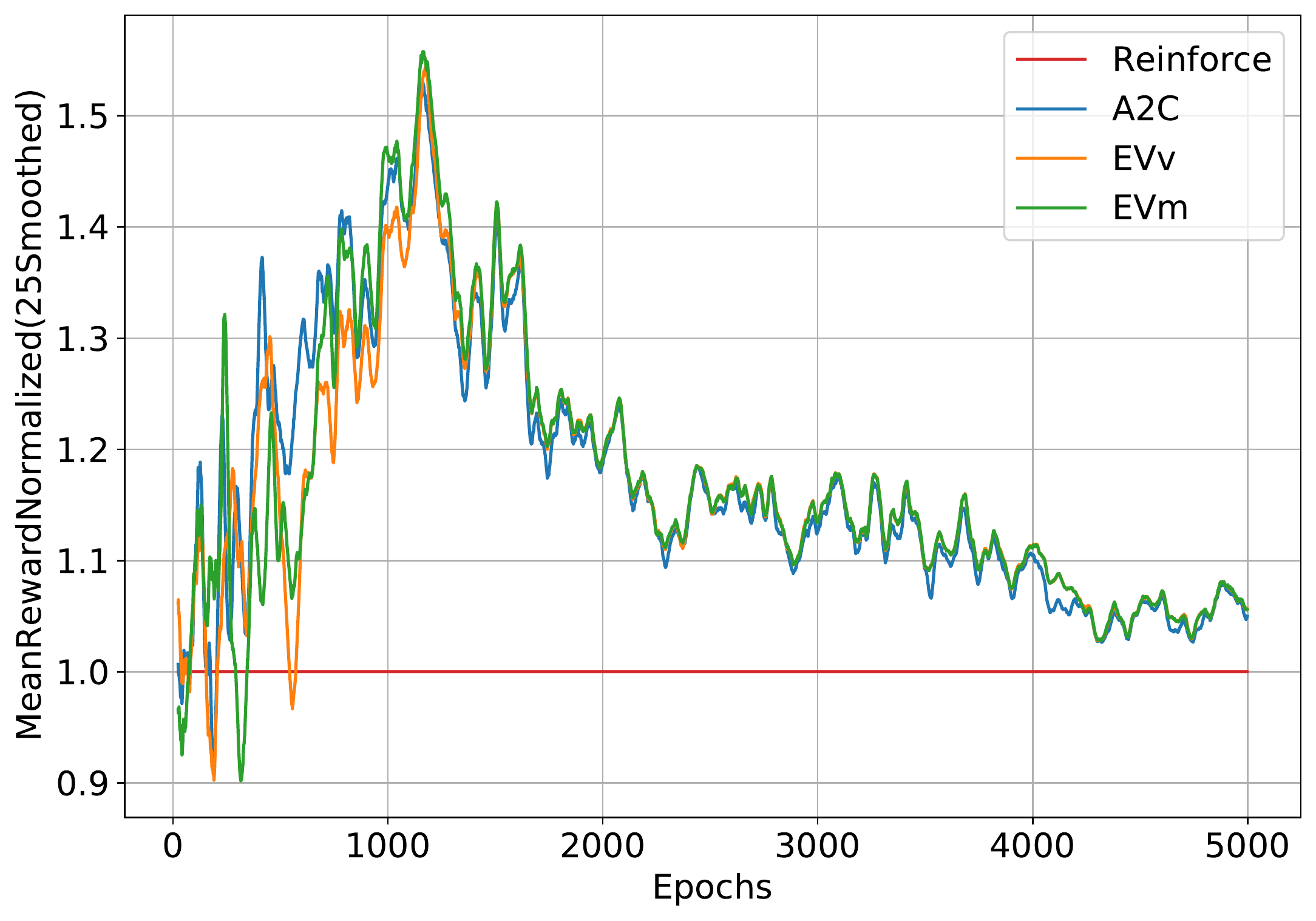} & (e) \includegraphics[scale=.2]{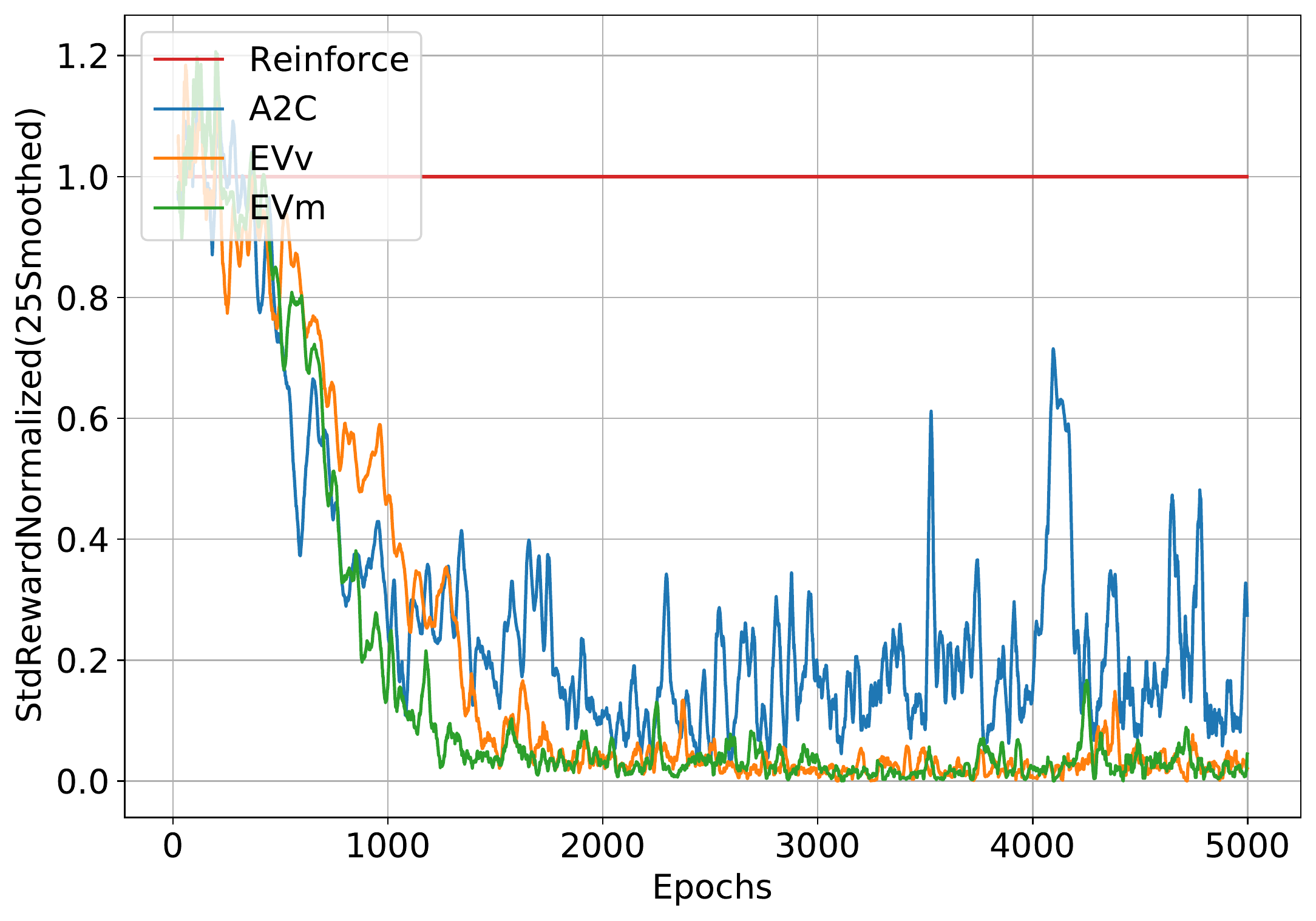} & (f)\includegraphics[scale=.2]{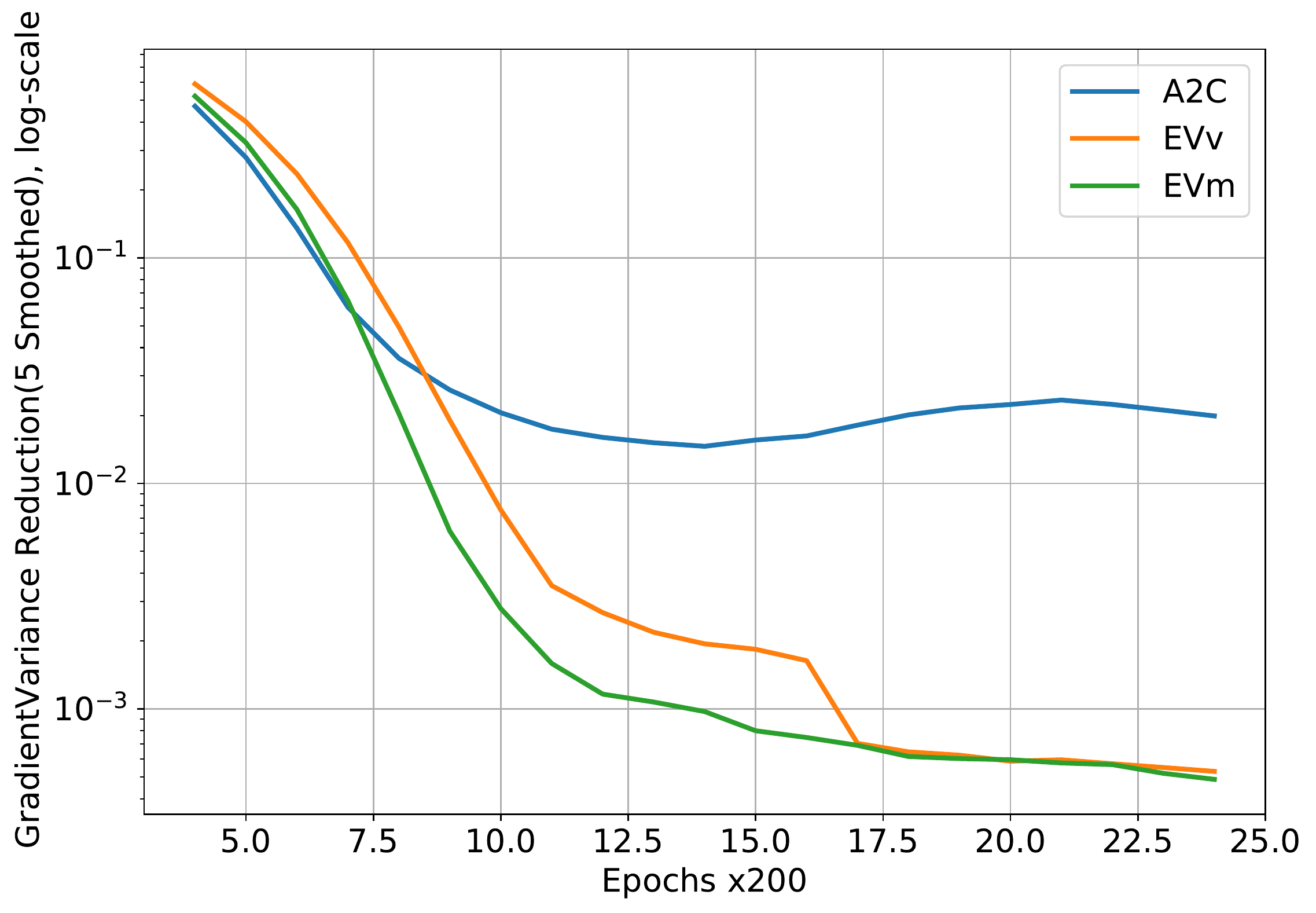} \\
    \end{tabular}
    \caption{The charts representing the results of the experiments in CartPole environment (config1): (a) displays mean rewards, (b) shows standard deviation of the rewards, (c) depicts gradient variance, in (d,e) the first two quantities are shown relative to REINFORCE and (f) shows gradient variance reduction ratio.}
    \label{fig:sup_Cartpole_config1}
\end{figure}

Cartpole config1 (see Fig.\ref{fig:sup_Cartpole_config1}) has two hidden layers in policy network with 128 neurons each and 1 hidden layer in baseline network with 128 neurons. We assume, that is a medium complexity setting for this environment. Both networks have ReLU activations. \\

We can observe here that even with simple configuration EV agents have similar or slightly higher rewards, achieving about 500 points and decrease rewards variance significantly showing that EV methods are more stable than A2C and do not have many deep falls during the training as A2C or REINFORCE. It is clearly an effect of the gradient variance which is reduced drastically: almost 100-1000 times.

\begin{figure}[h!]
    \centering
    \begin{tabular}{lcr}
    (a) \includegraphics[scale=.2]{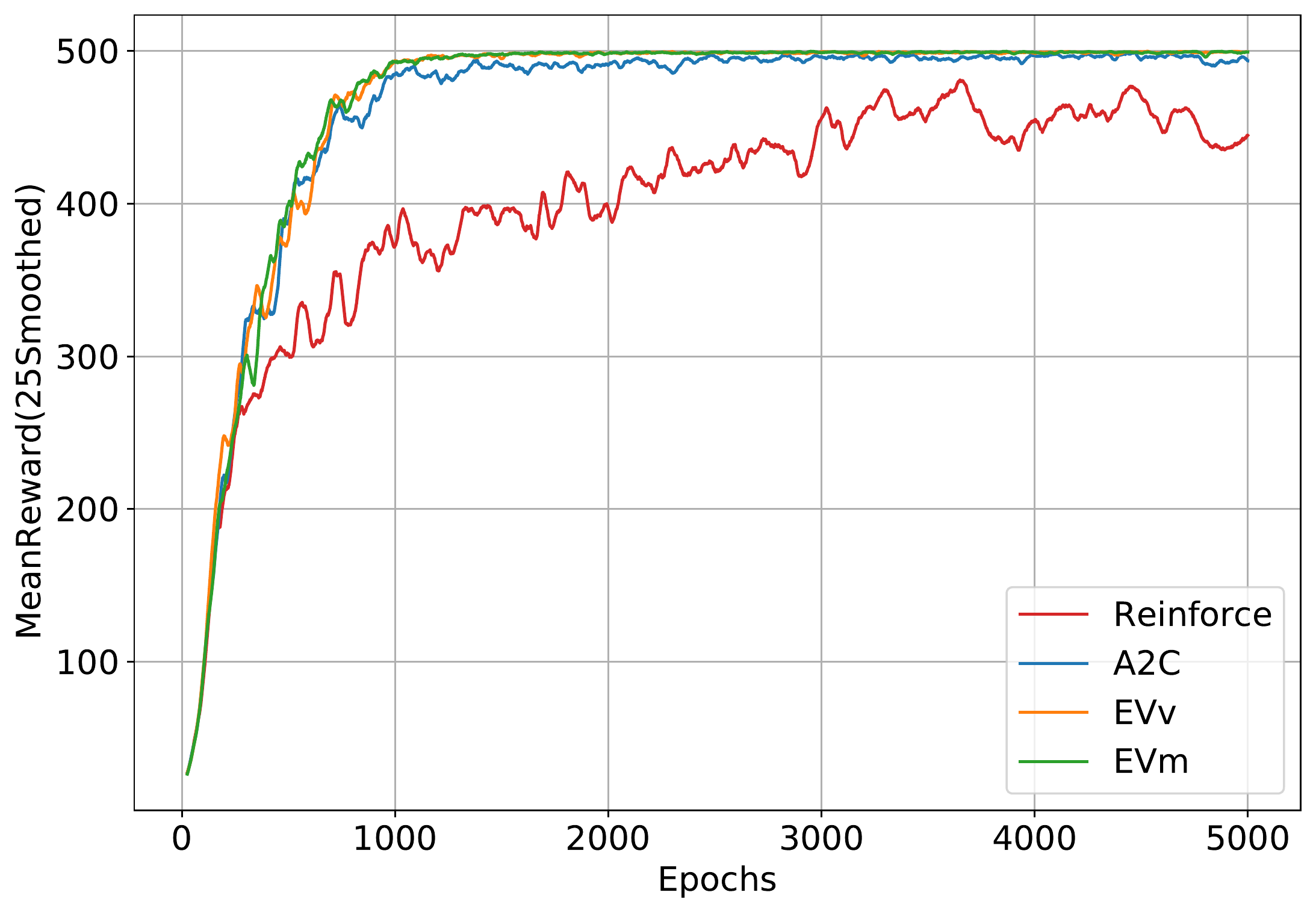} & (b) \includegraphics[scale=.2]{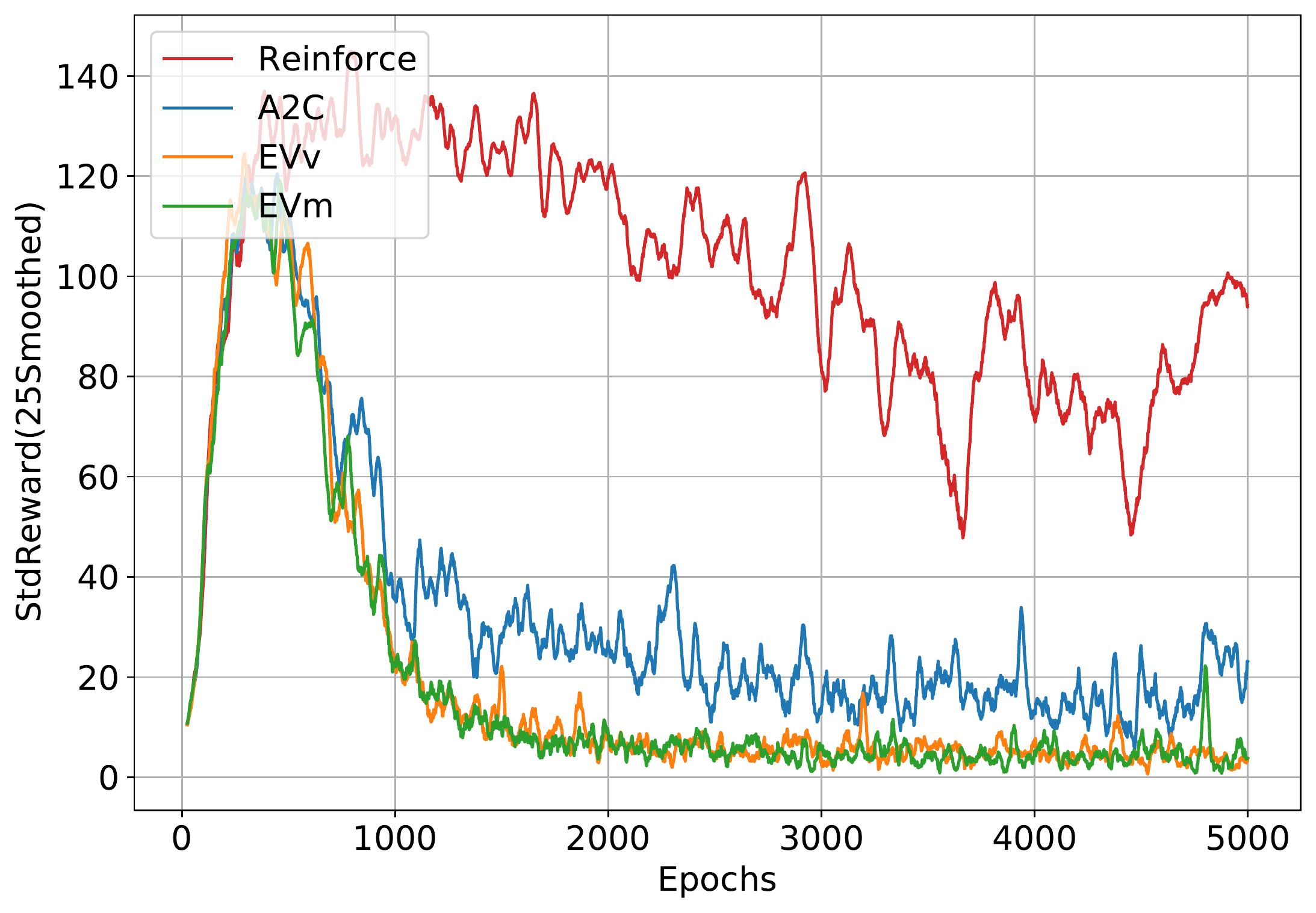} & (c)\includegraphics[scale=.2]{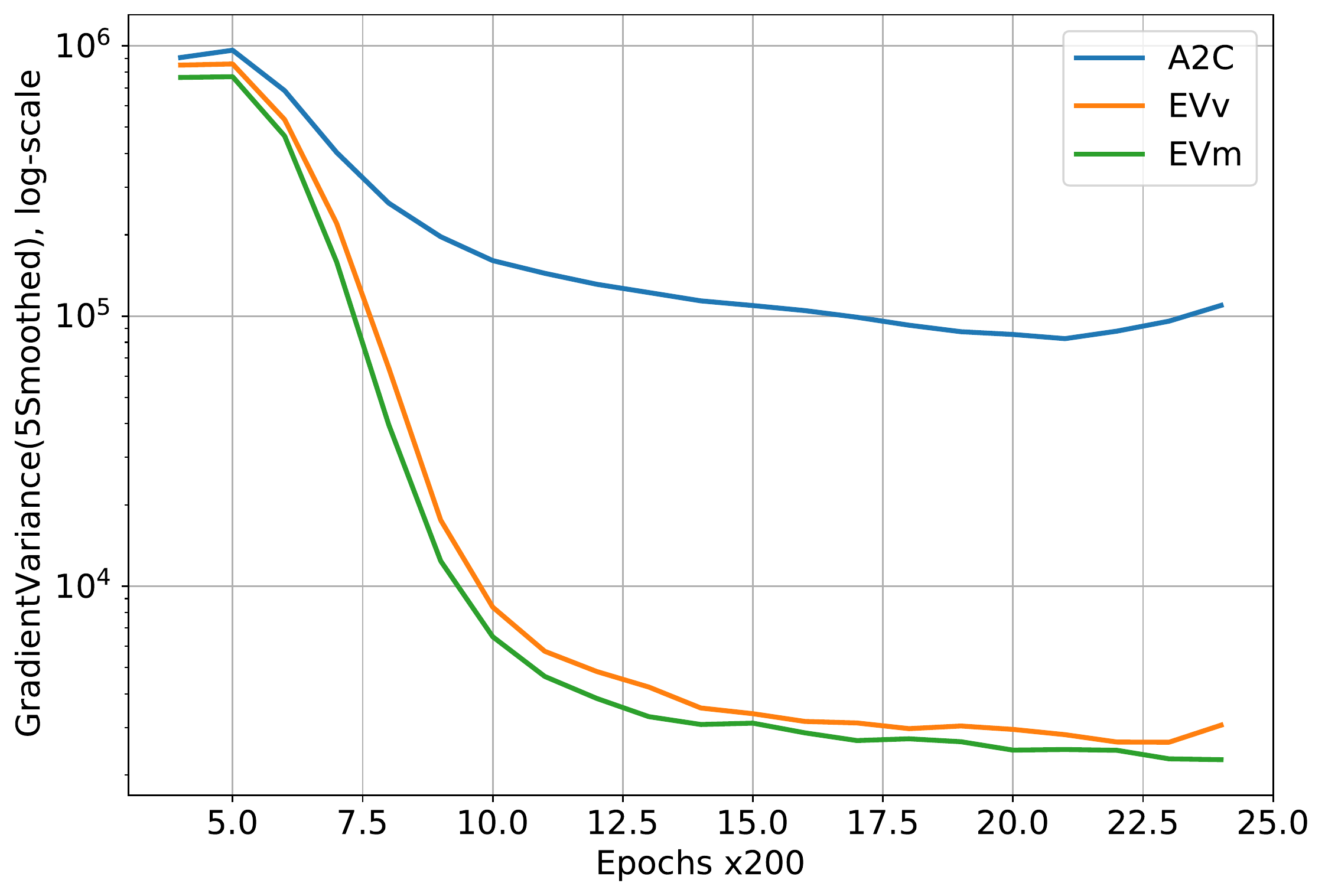} \\
    (d) \includegraphics[scale=.2]{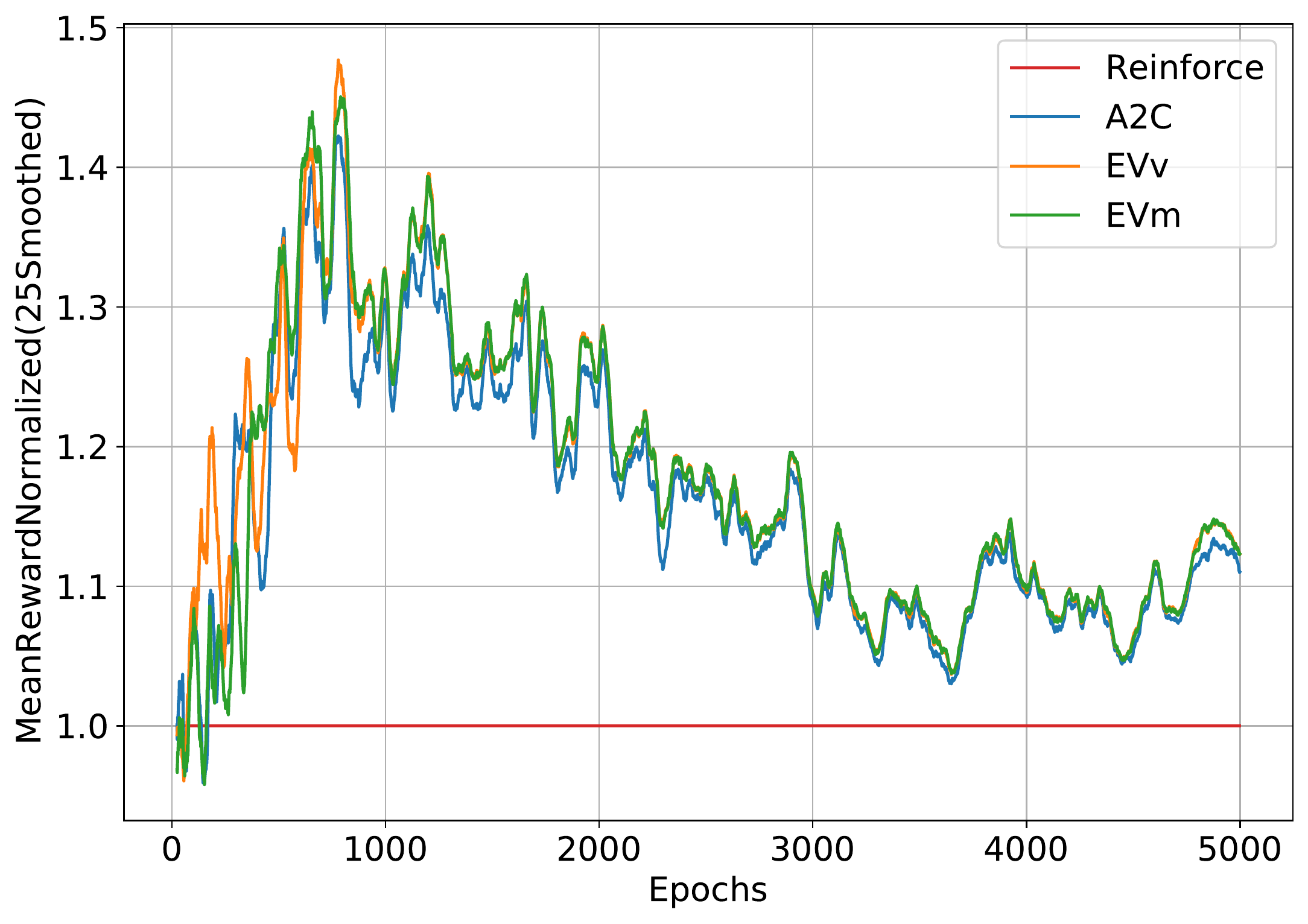} & (e) \includegraphics[scale=.2]{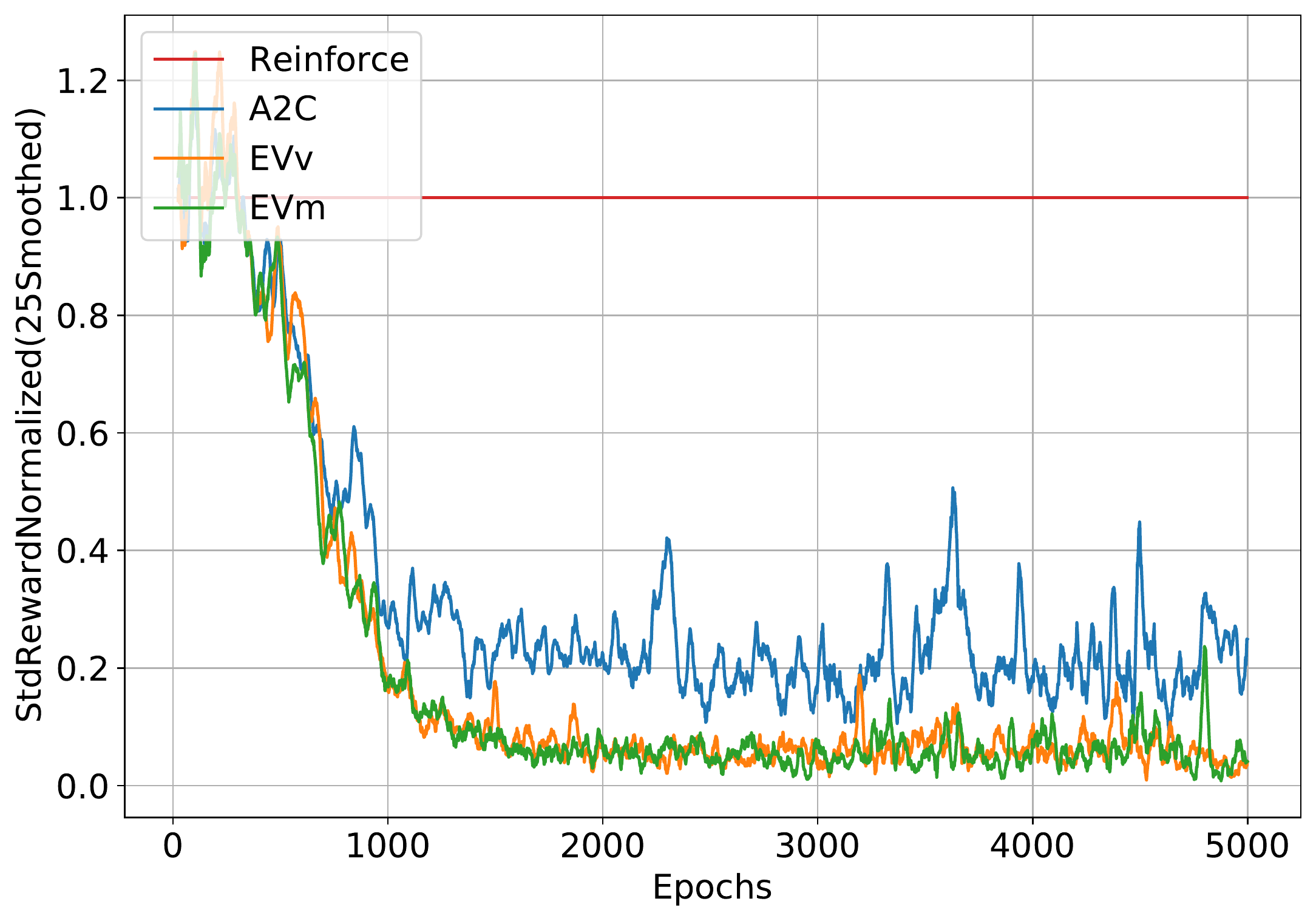} & (f)\includegraphics[scale=.2]{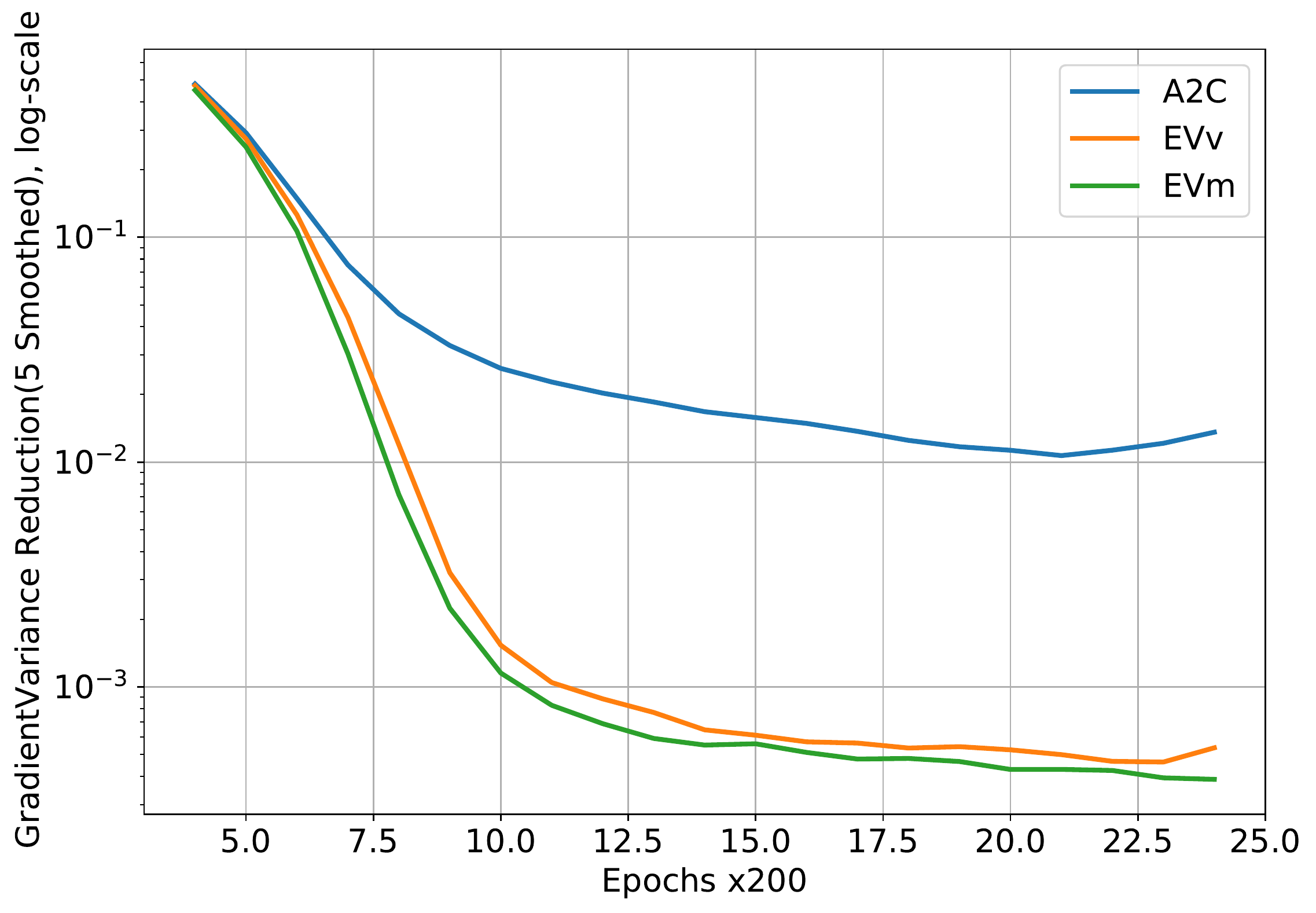} \\
    \end{tabular}
    \caption{The charts representing the results of the experiments in CartPole environment (config5): (a) displays mean rewards, (b) shows standard deviation of the rewards, (c) depicts gradient variance, in (d,e) the first two quantities are shown relative to REINFORCE and (f) shows gradient variance reduction ratio.}
    \label{fig:sup_Cartpole_config5}
\end{figure}

In config5 (see Fig.\ref{fig:sup_Cartpole_config5}) we keep the architecture from config1, but change the activation function with MISH. The results are almost the same: EV-agents show a little predominance over A2C, preserving the least reward variance and gradient variance reduction among all the algorithms.

\begin{figure}[h!]
    \centering
    \begin{tabular}{lcr}
    (a) \includegraphics[scale=.2]{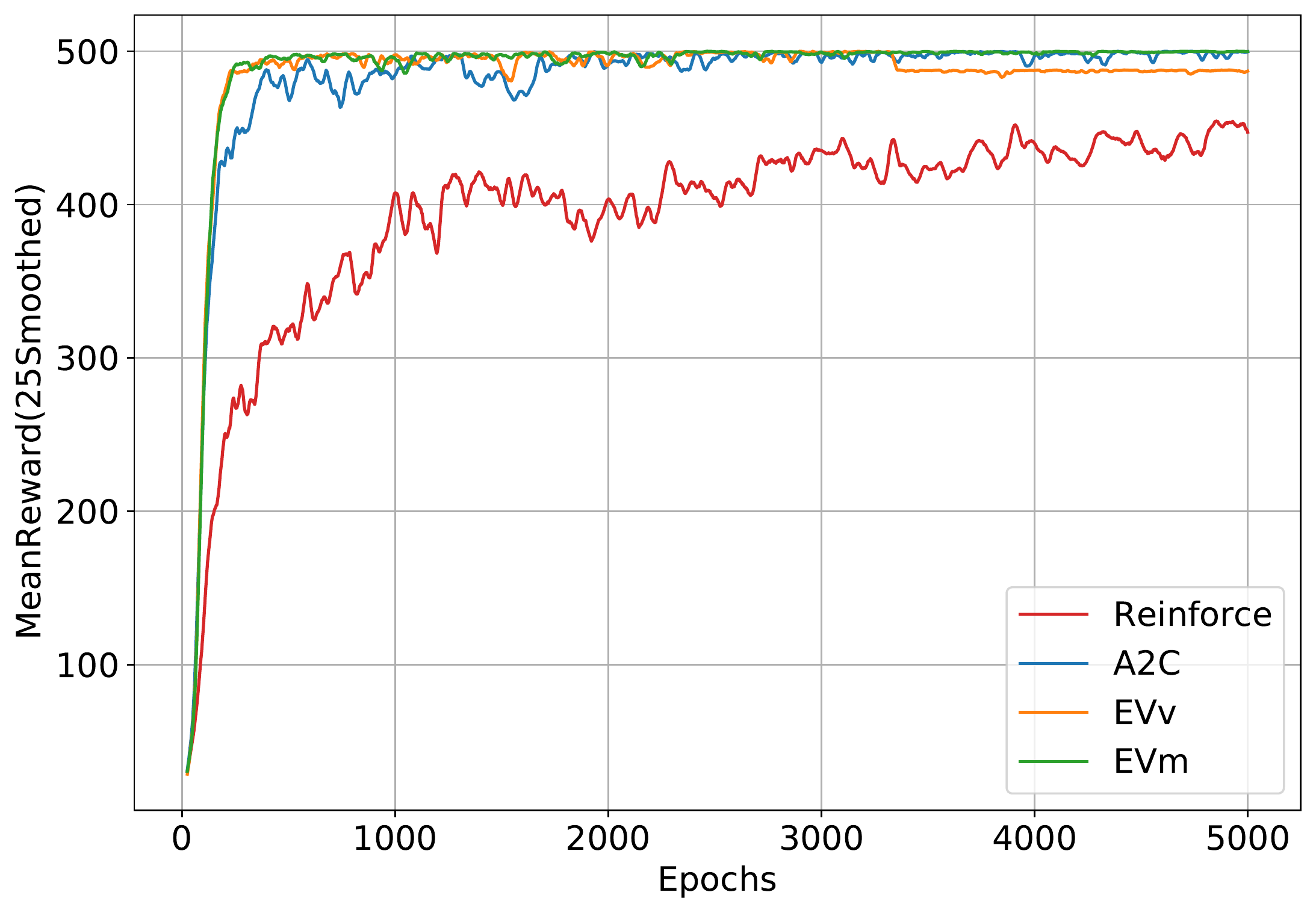} & (b) \includegraphics[scale=.2]{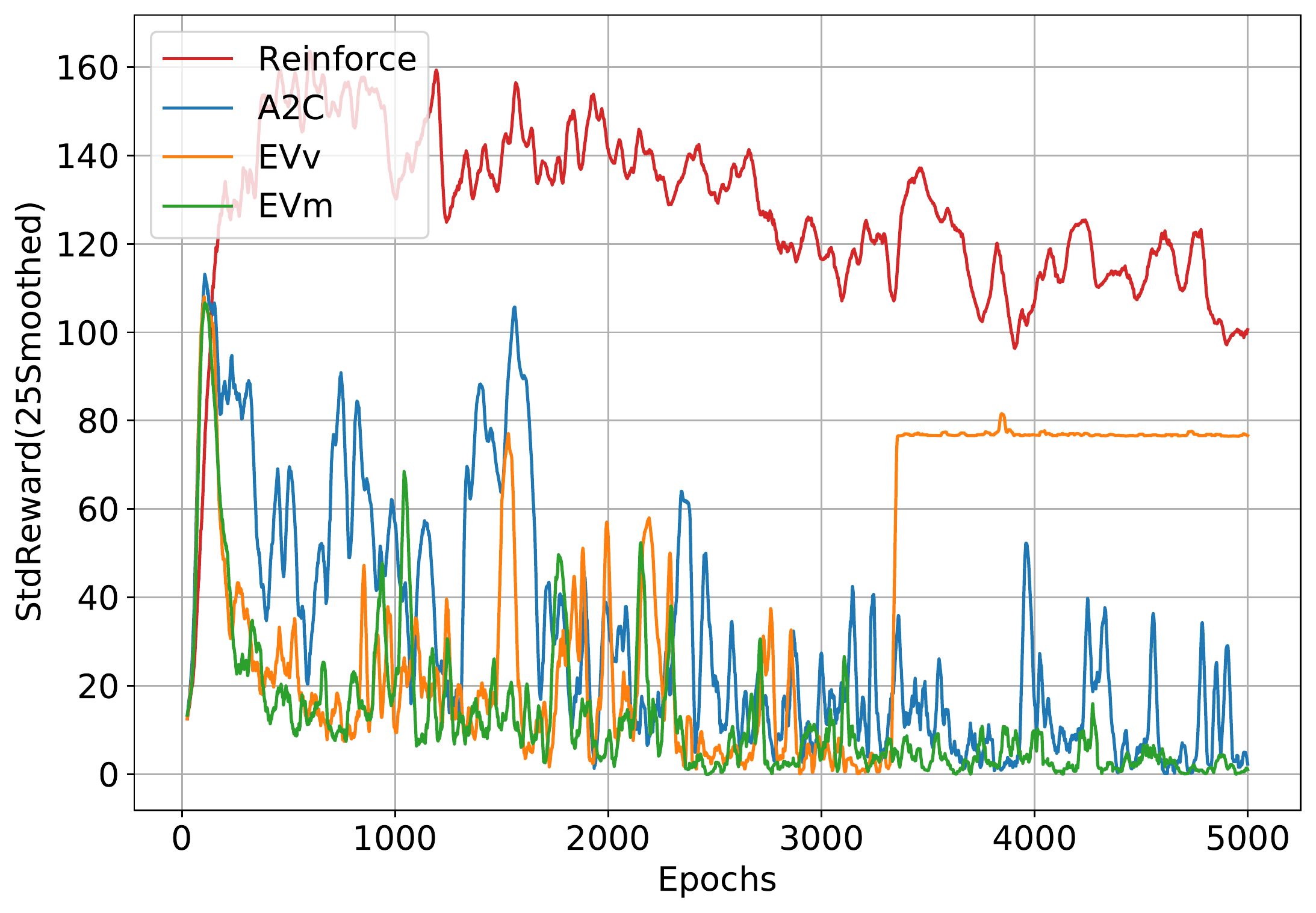} & (c)\includegraphics[scale=.2]{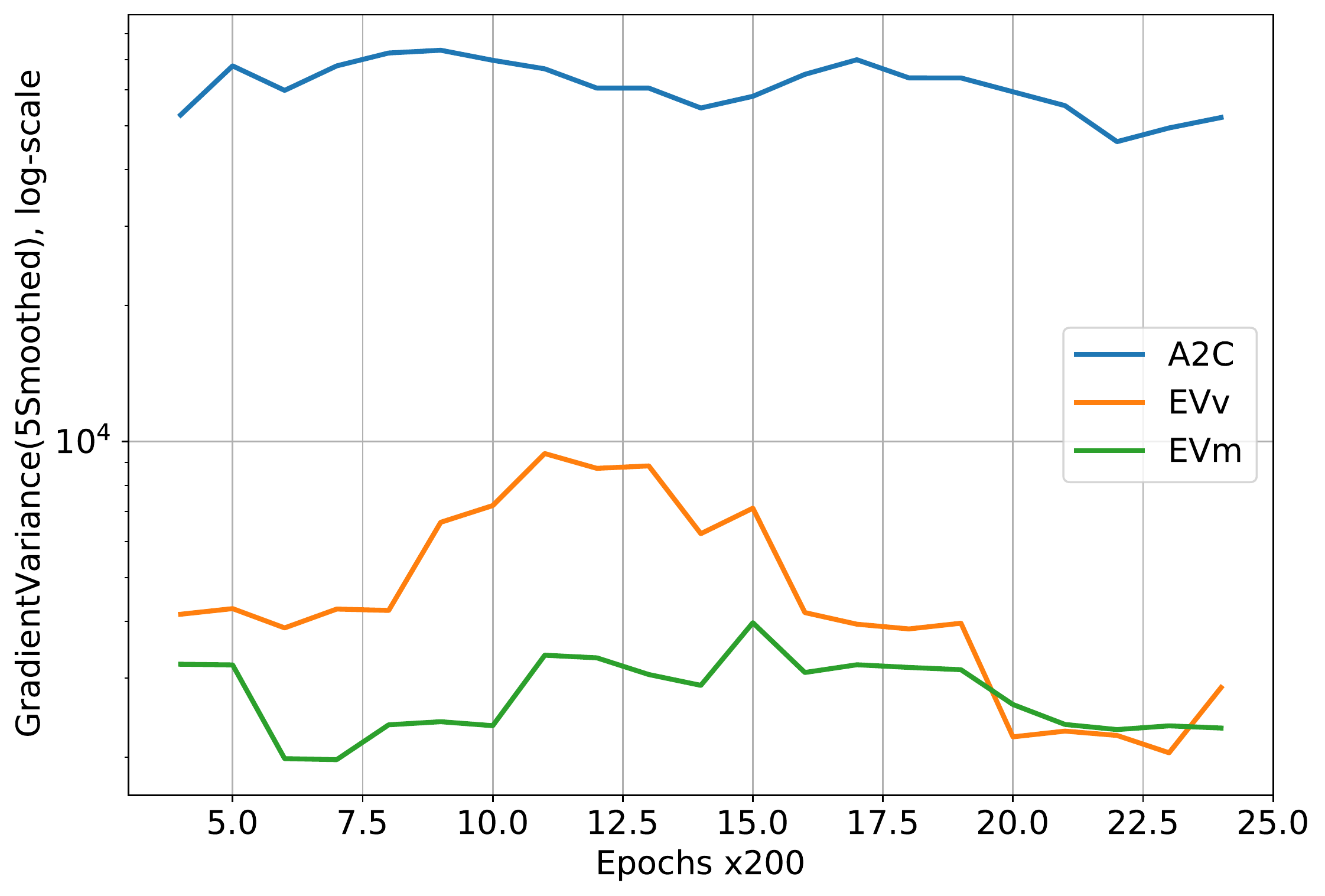} \\
    (d) \includegraphics[scale=.2]{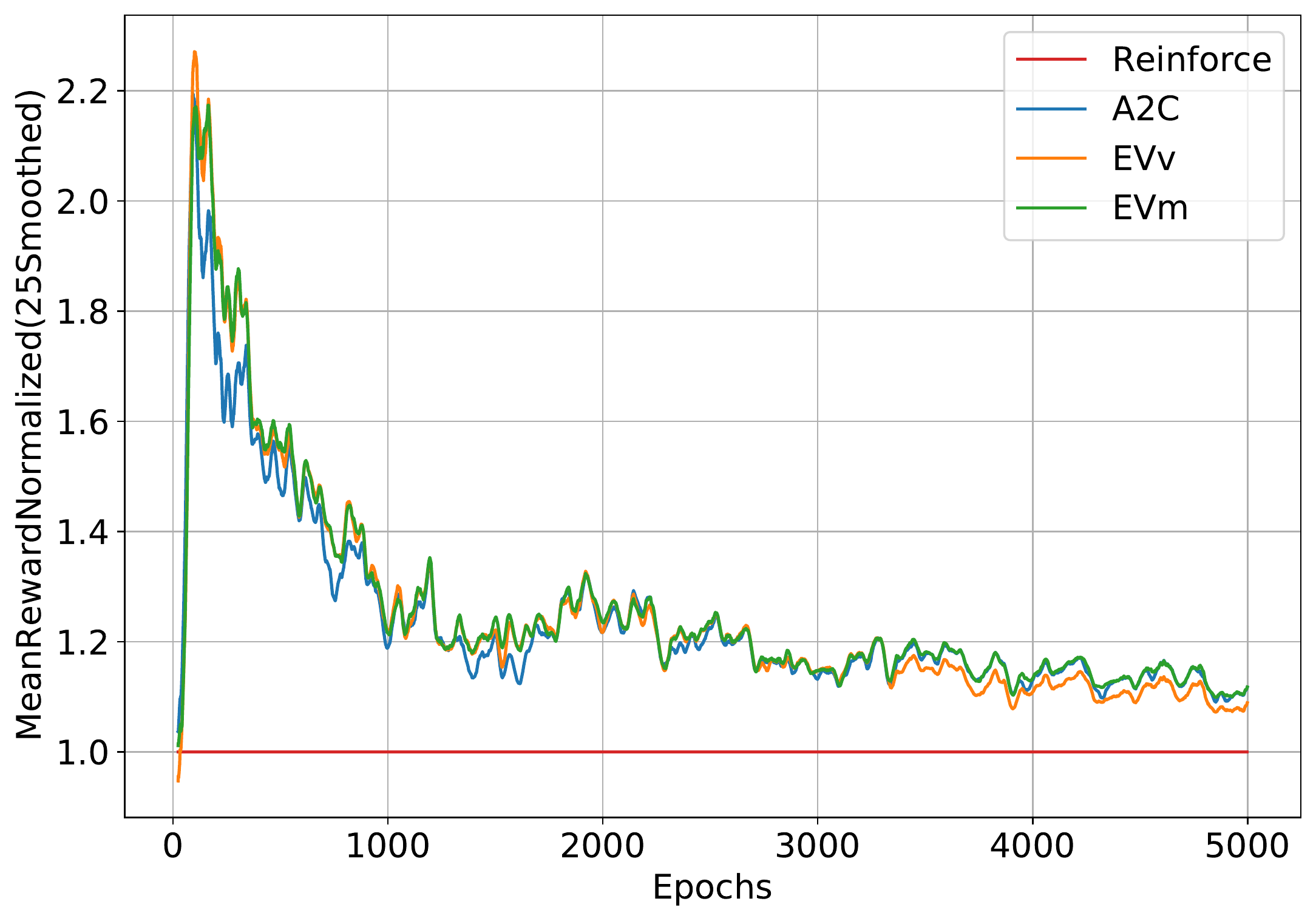} & (e) \includegraphics[scale=.2]{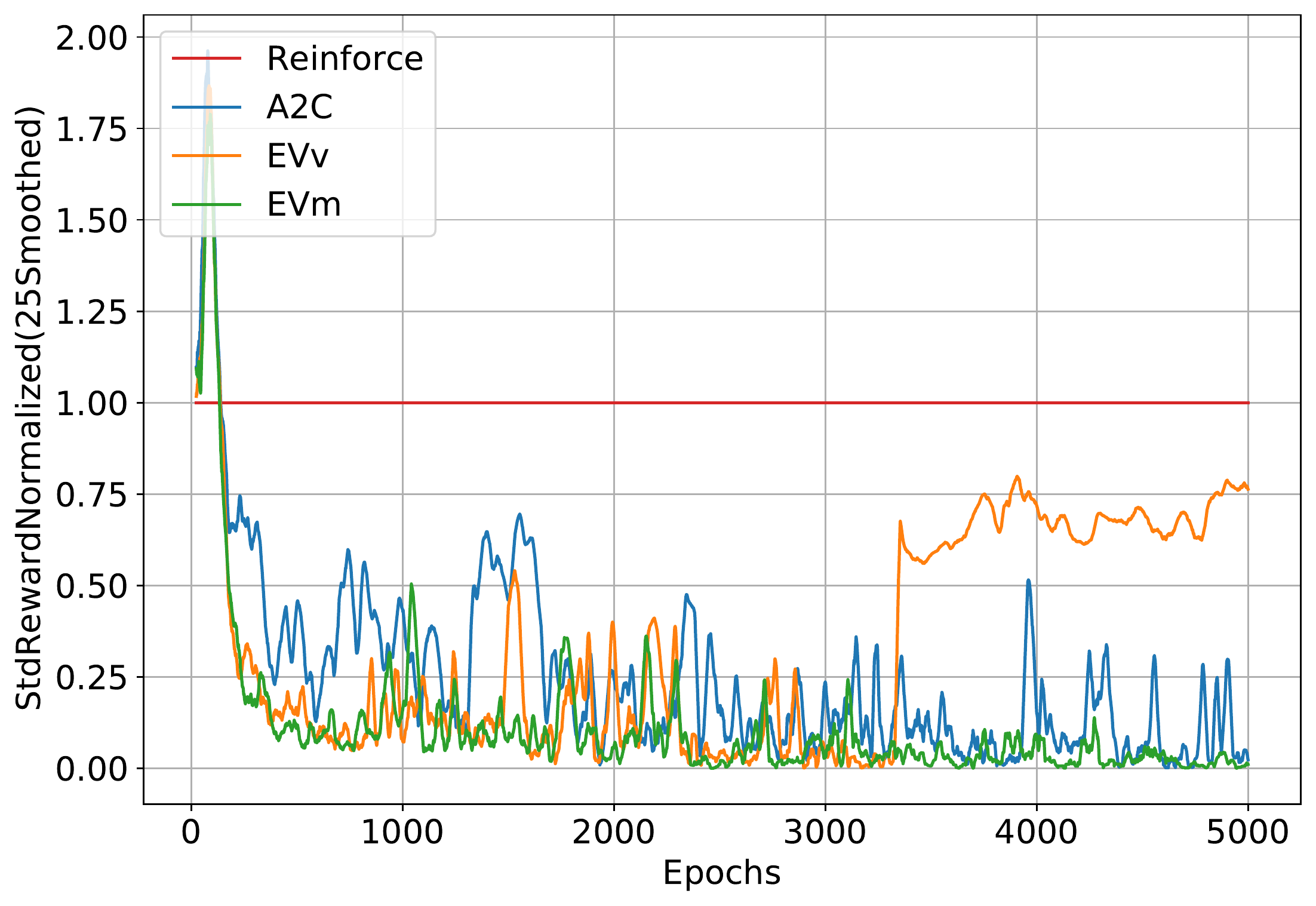} & (f)\includegraphics[scale=.2]{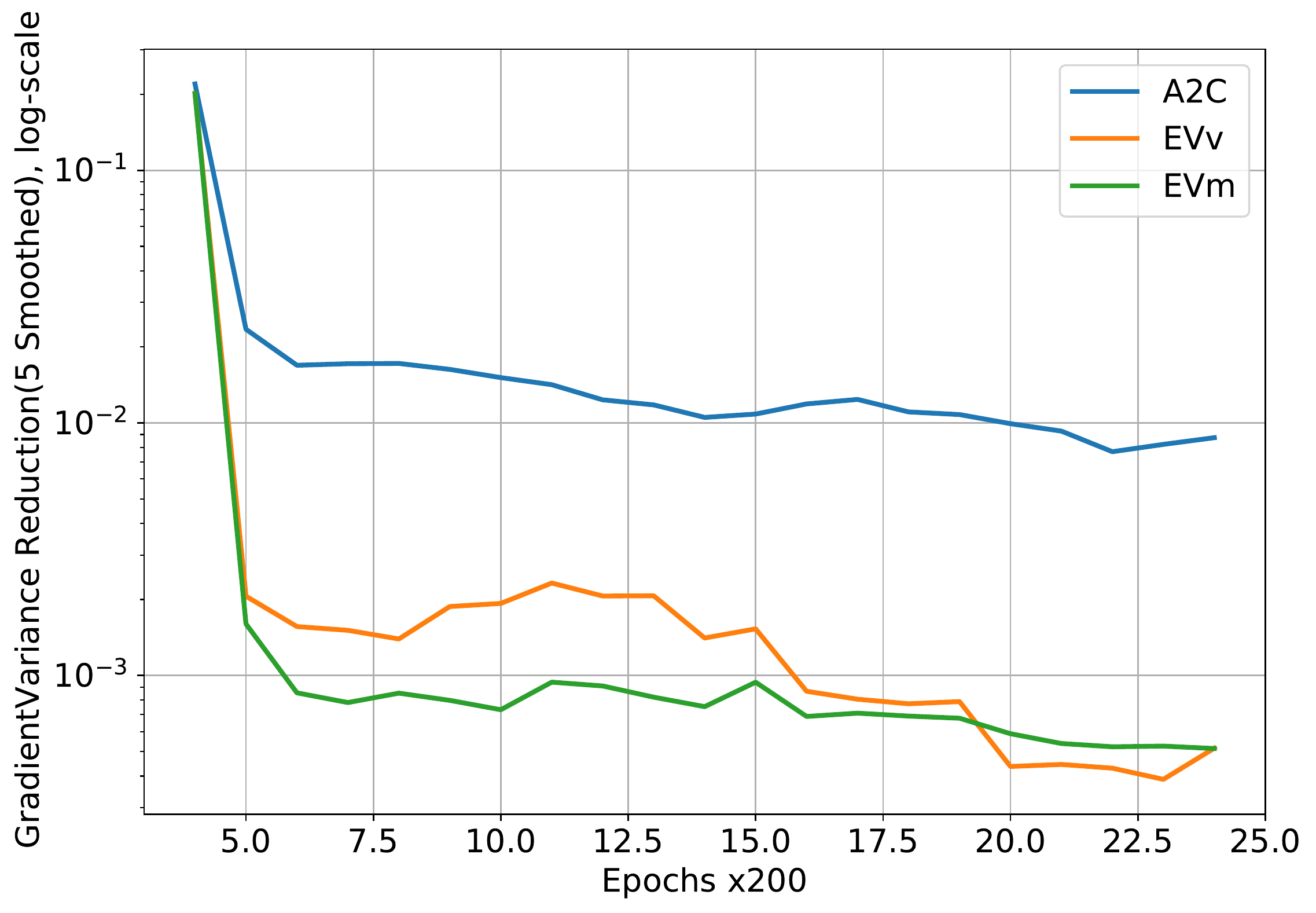} \\
    \end{tabular}
    \caption{The charts representing the results of the experiments in CartPole environment (config7): (a) displays mean rewards, (b) shows standard deviation of the rewards, (c) depicts gradient variance, in (d,e) the first two quantities are shown relative to REINFORCE and (f) shows gradient variance reduction ratio.}
    \label{fig:sup_Cartpole_config7}
\end{figure}

In config7 (see Fig.\ref{fig:sup_Cartpole_config7}) we move towards more complex architecture of baseline function: now it has 3 hidden layers of 128, 256, 128 neurons respectively. EV agents demonstrate better performance but this increment is rather small. Nevertheless, reward variance and gradient variance again remain the best in EV-methods.

\begin{figure}[h!]
    \centering
    \begin{tabular}{lcr}
    (a) \includegraphics[scale=.2]{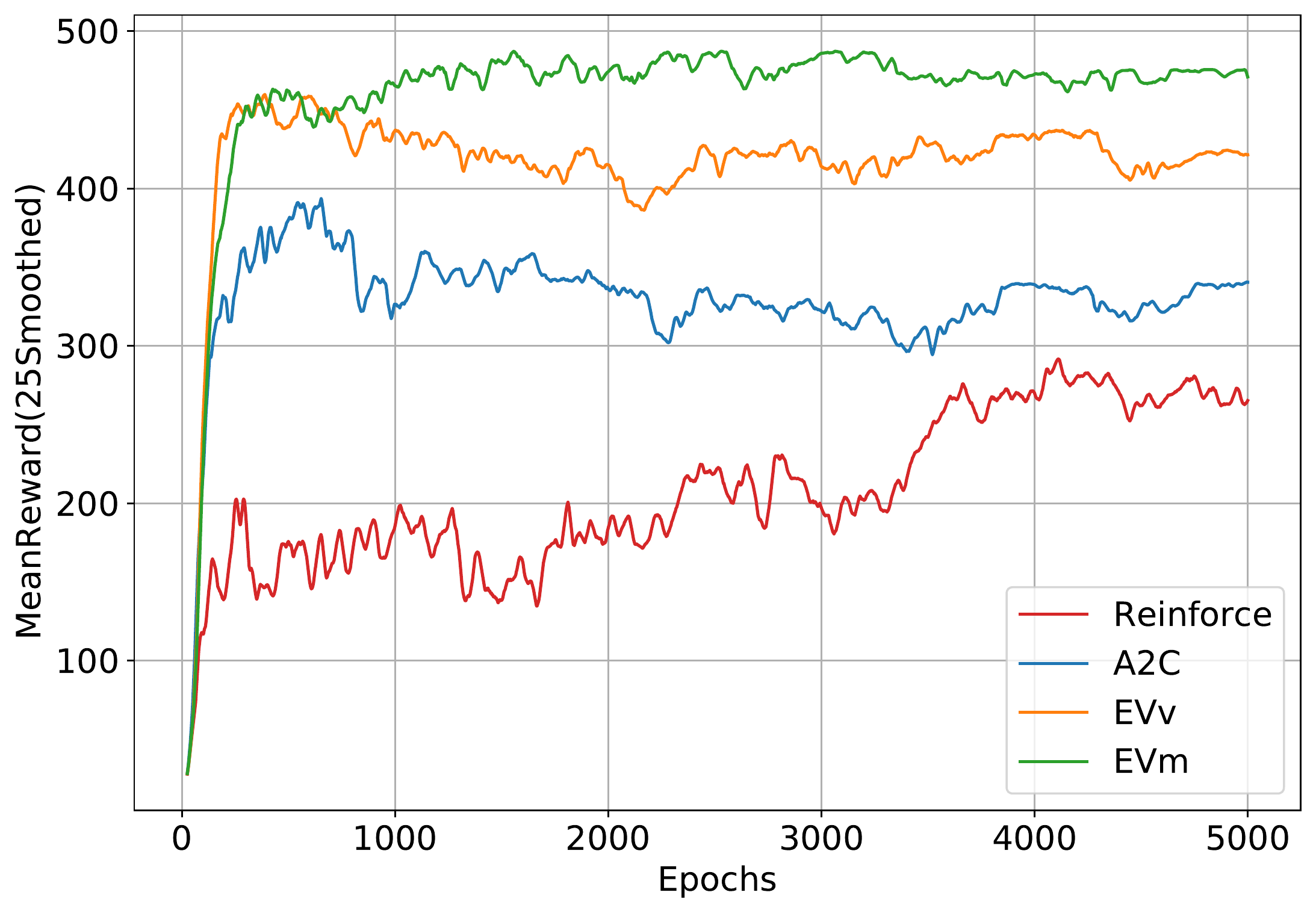} & (b) \includegraphics[scale=.2]{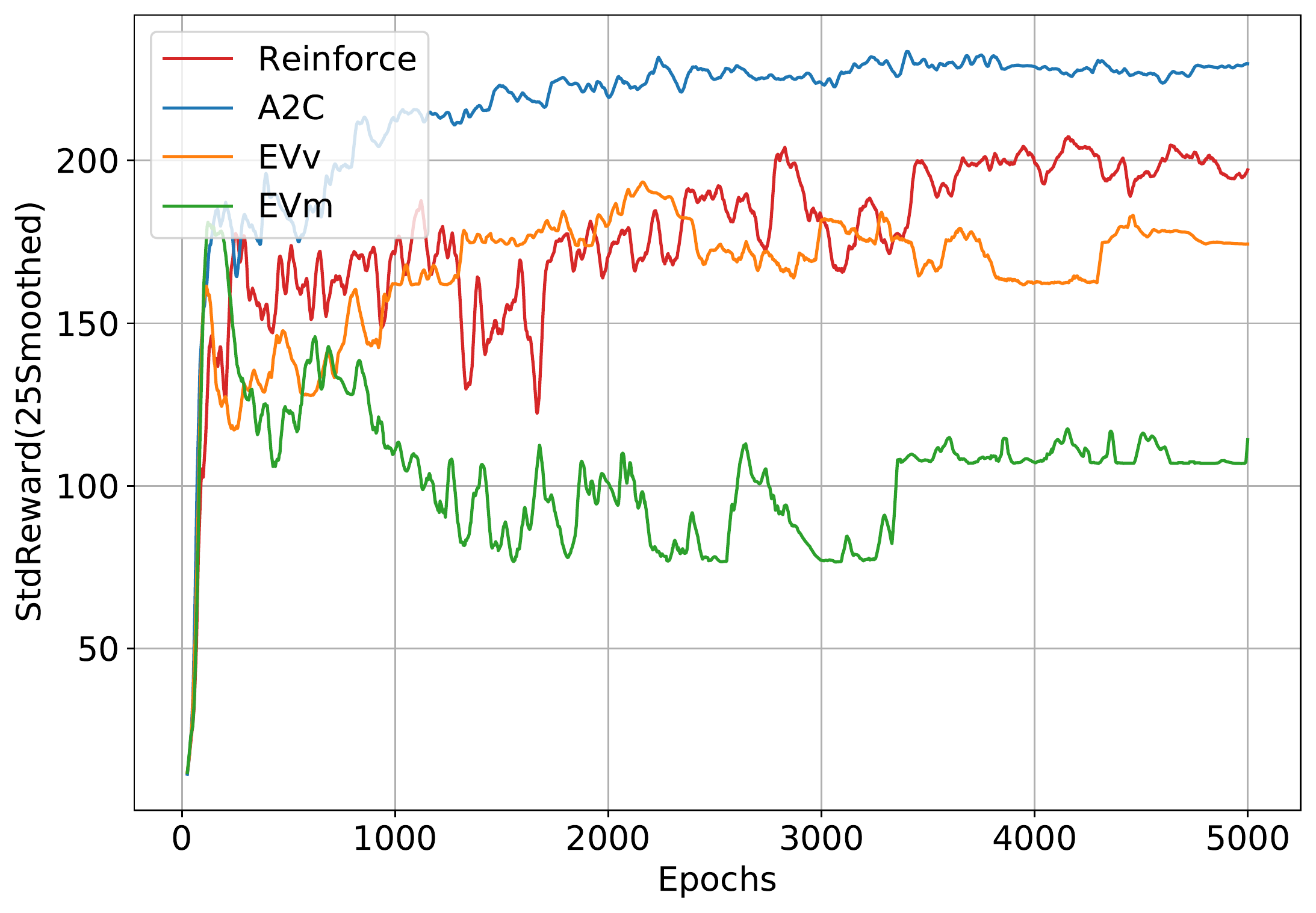} & (c)\includegraphics[scale=.2]{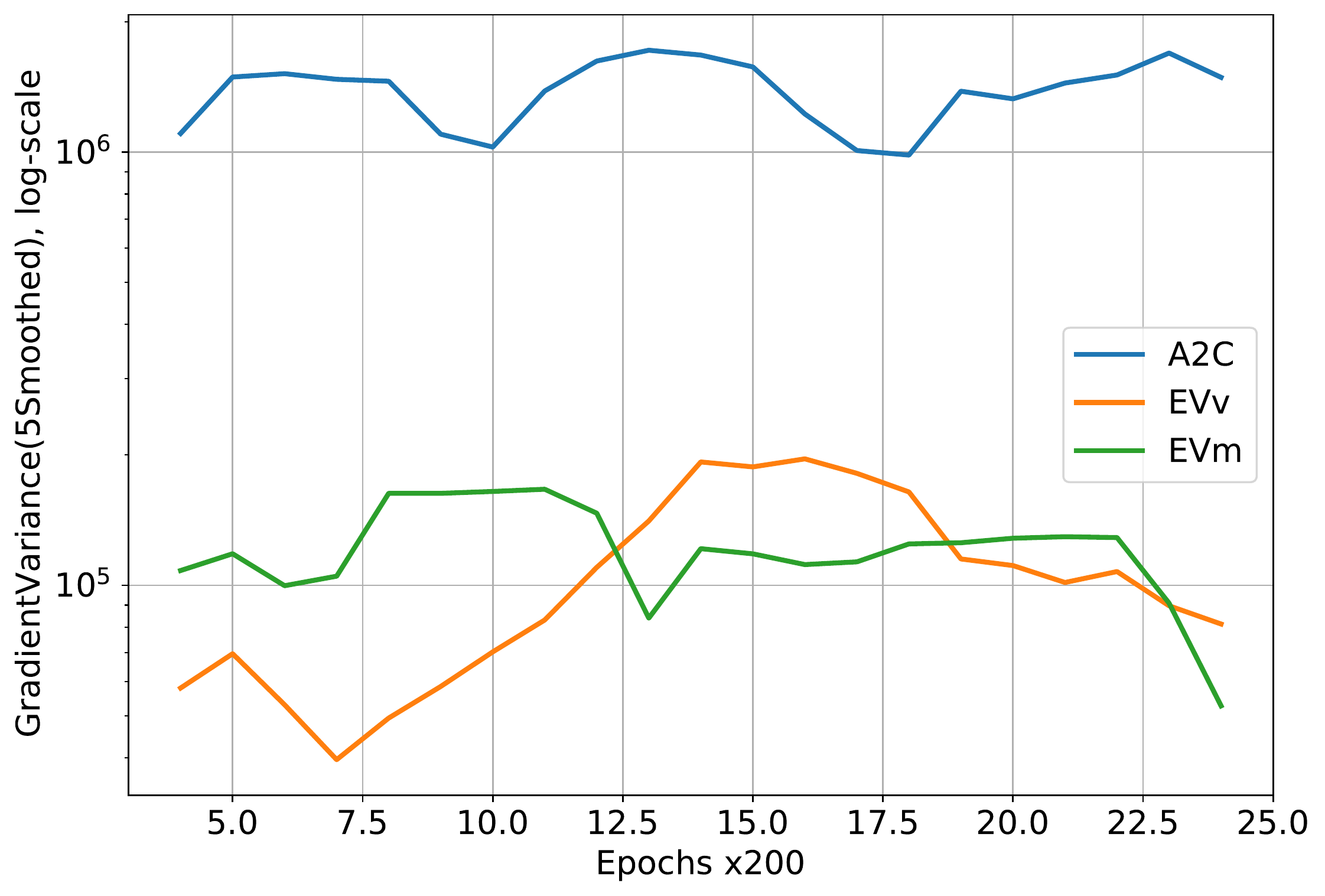} \\
    (d) \includegraphics[scale=.2]{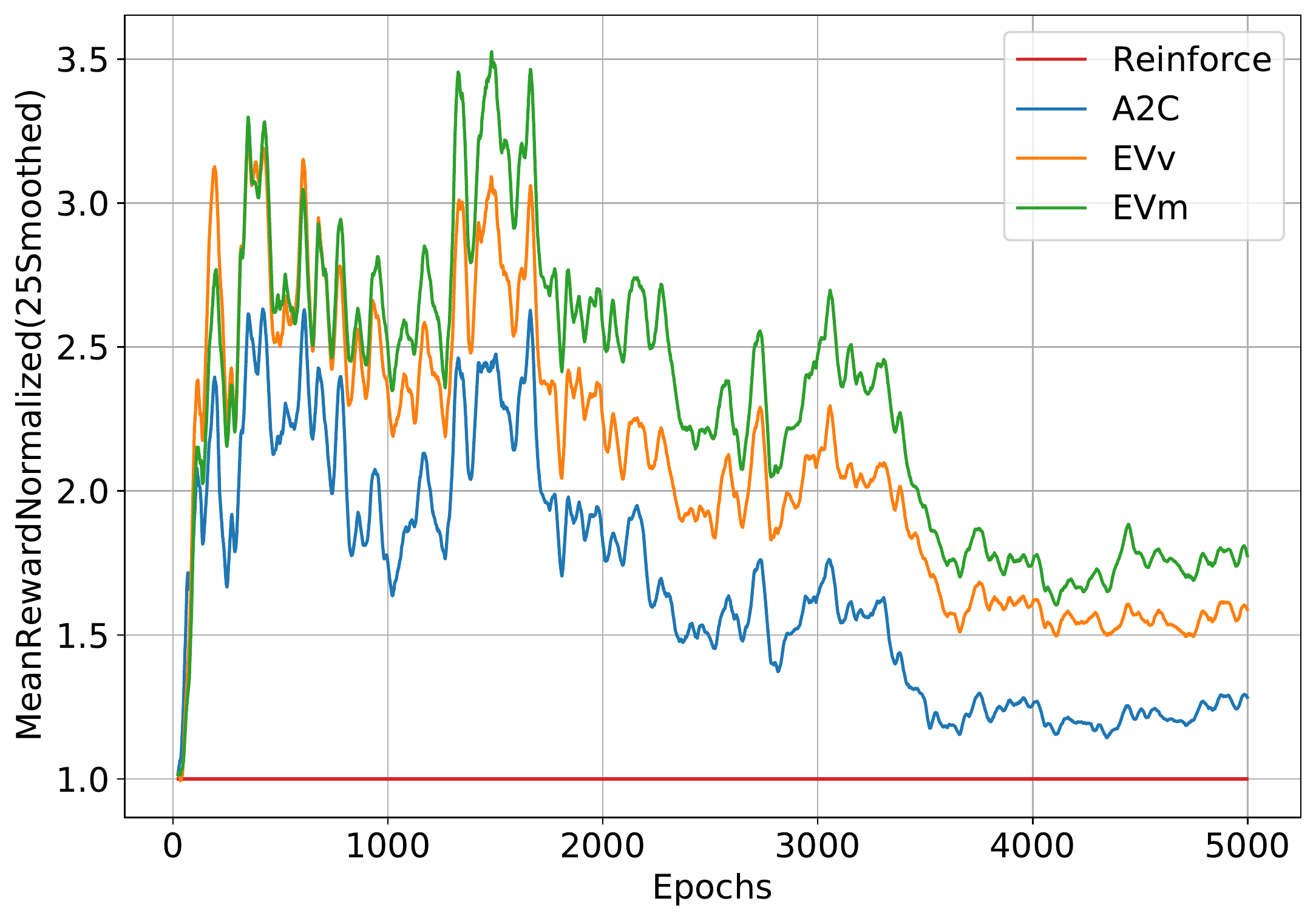} & (e) \includegraphics[scale=.2]{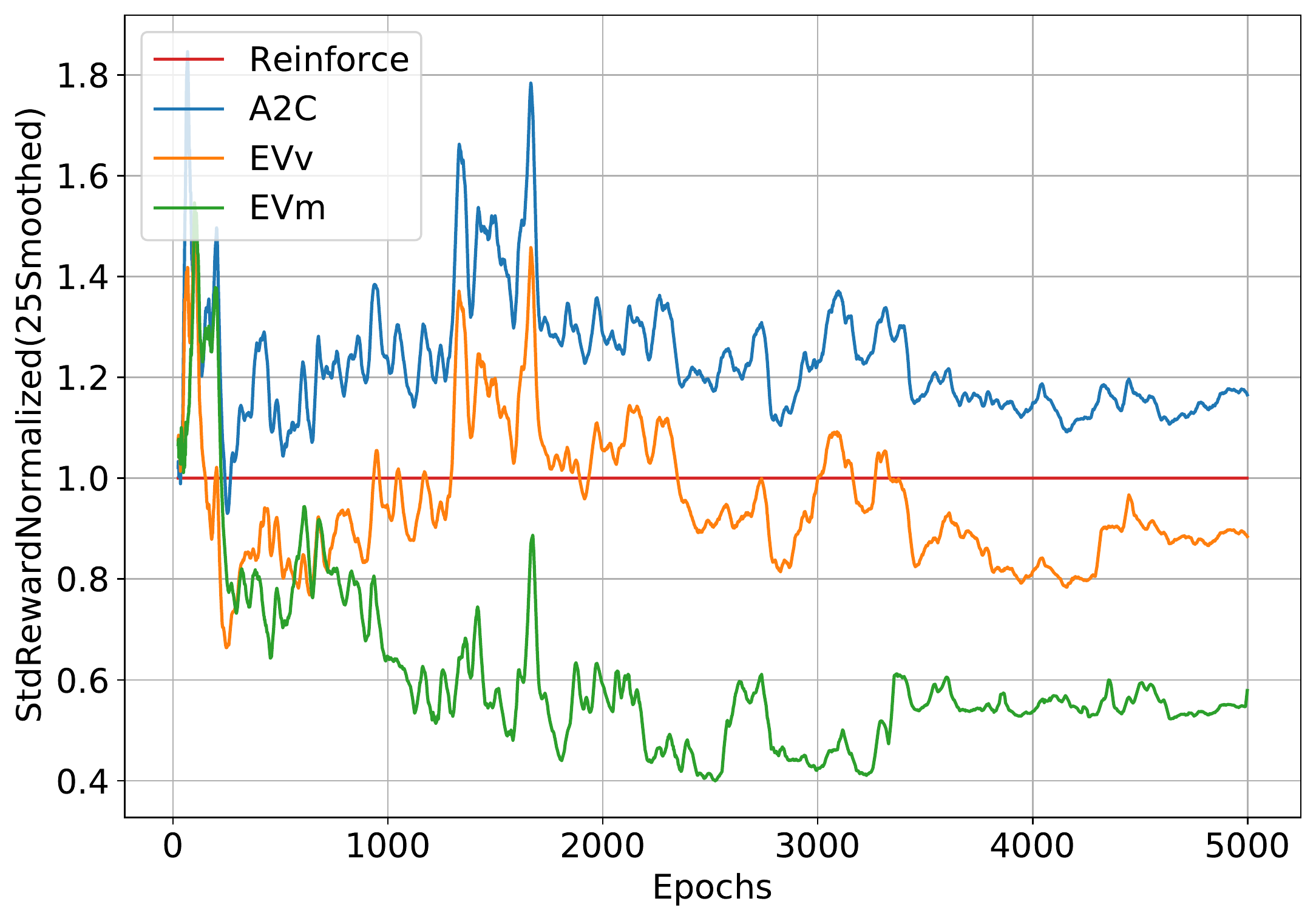} & (f)\includegraphics[scale=.2]{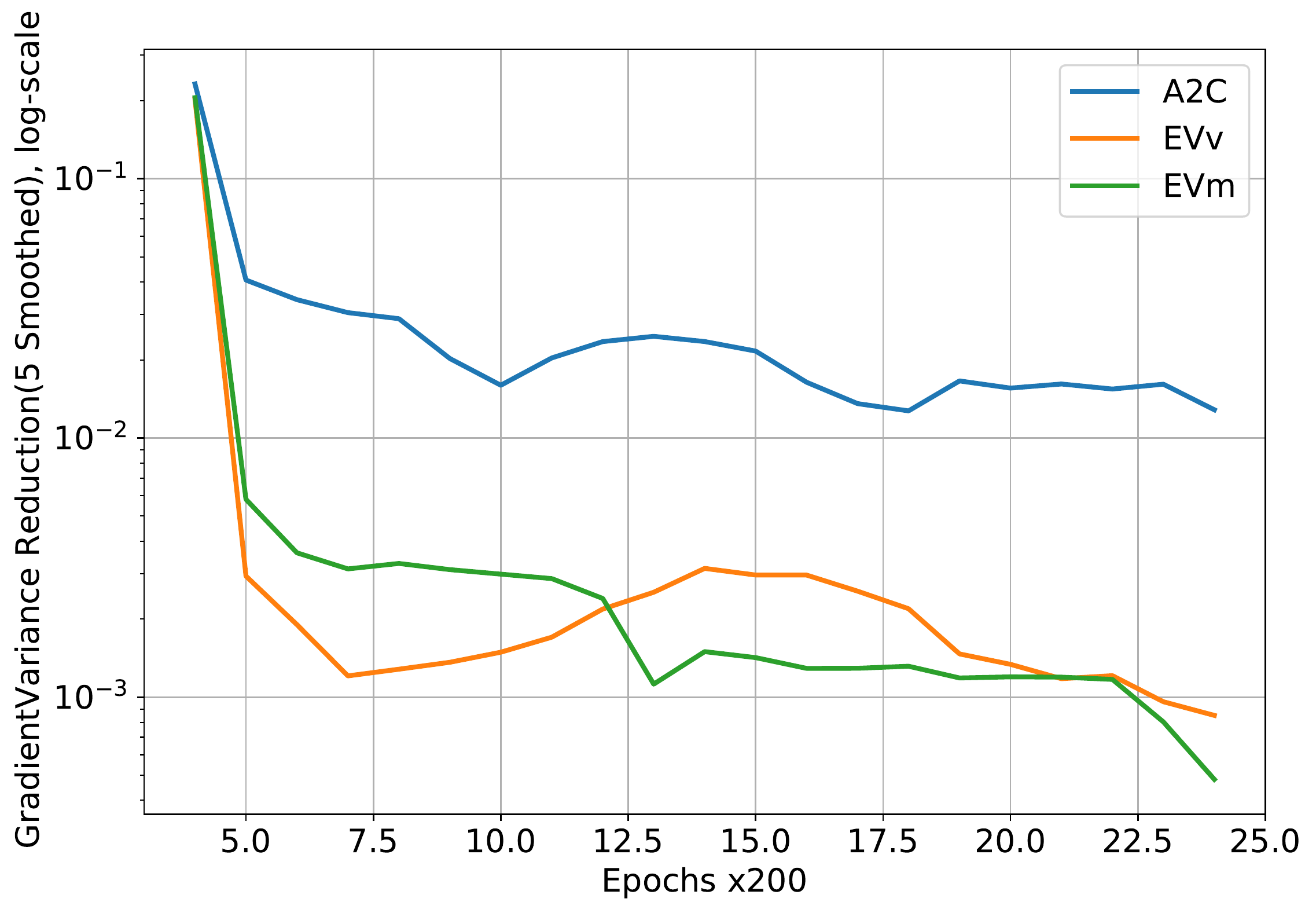} \\
    \end{tabular}
    \caption{The charts representing the results of the experiments in CartPole environment (config8): (a) displays mean rewards, (b) shows standard deviation of the rewards, (c) depicts gradient variance, in (d,e) the first two quantities are shown relative to REINFORCE and (f) shows gradient variance reduction ratio.}
    \label{fig:sup_Cartpole_config8}
\end{figure}

In config8 (see Fig.\ref{fig:sup_Cartpole_config8}) we address more complex setting of policy, adding two layers. The policy network has finally 3 hidden layers with MISH activation with 64, 128, 256 neurons respectively. This change greatly increases efficiency of EV algorithms enabling to achieve more than 400 points of reward and demonstrating a big dominance over A2C which is unable to train such a complex policy to have similar performance. At the same time, reward variance of EVs remains at the level of REINFORCE while A2C level exceeds it by almost 30-50\%.

If config9 we again preserve architecture settings changing only activation from MISH to ReLU. We observe a small difference in mean rewards but another activation function clearly helped in A2C training. Regardless, EV-methods are still predominant: more stable, with less gradient variance and with higher rewards achieved.\\

In conclusion, our experiments show that EV methods are sometimes considerably better in terms of mean rewards than A2C, or work at least as A2C. Study of the reward variance shows that EV-methods in CartPole are considerably more stable and do not have deep falls as in A2C or Reinforce. This study allows us to judge about the stability of the training process in case of EV algorithms and claim that they are able to perform better than A2C if more complex policies are used.

\begin{figure}[h!]
    \centering
    \begin{tabular}{lcr}
    (a) \includegraphics[scale=.2]{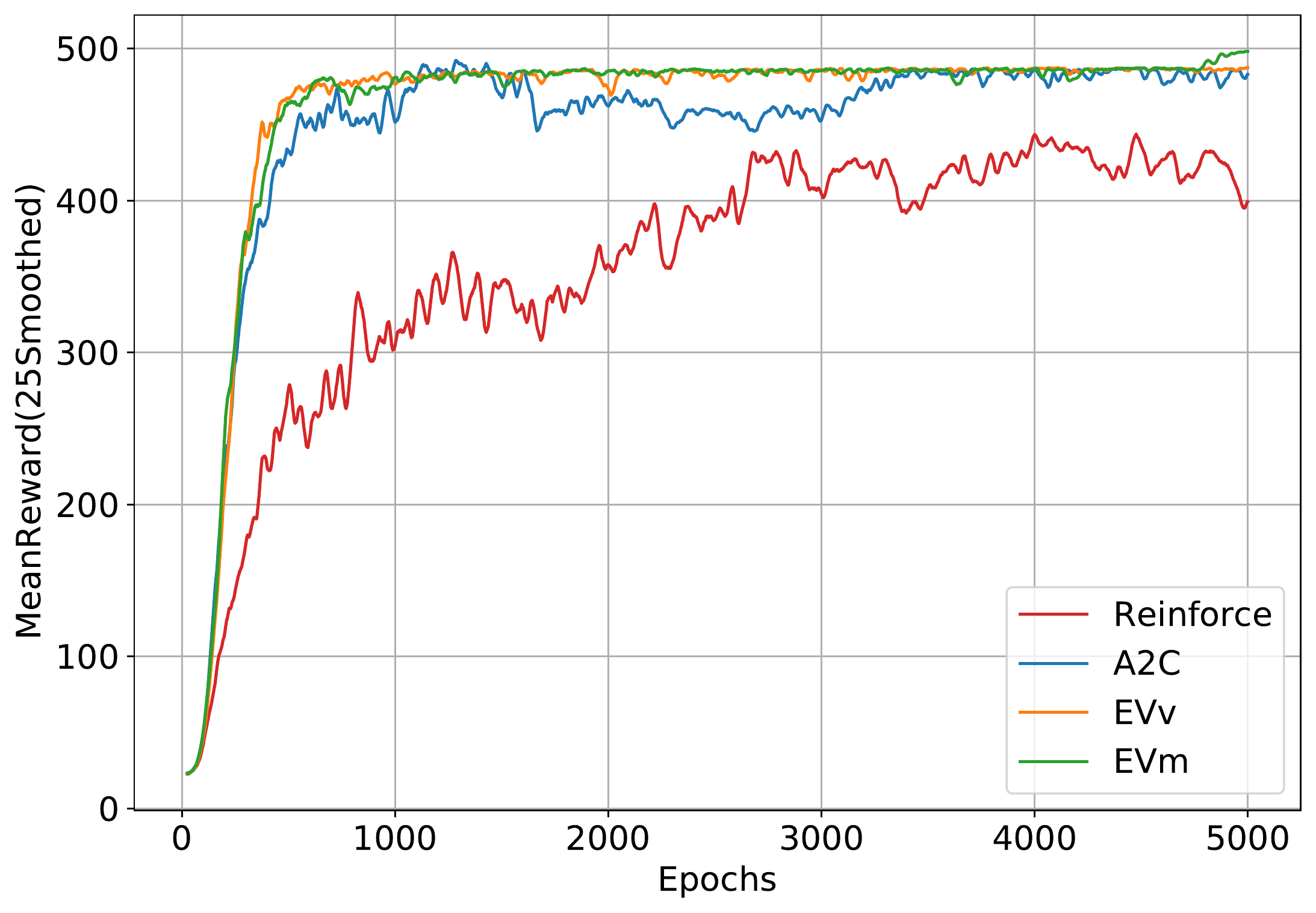} & (b) \includegraphics[scale=.2]{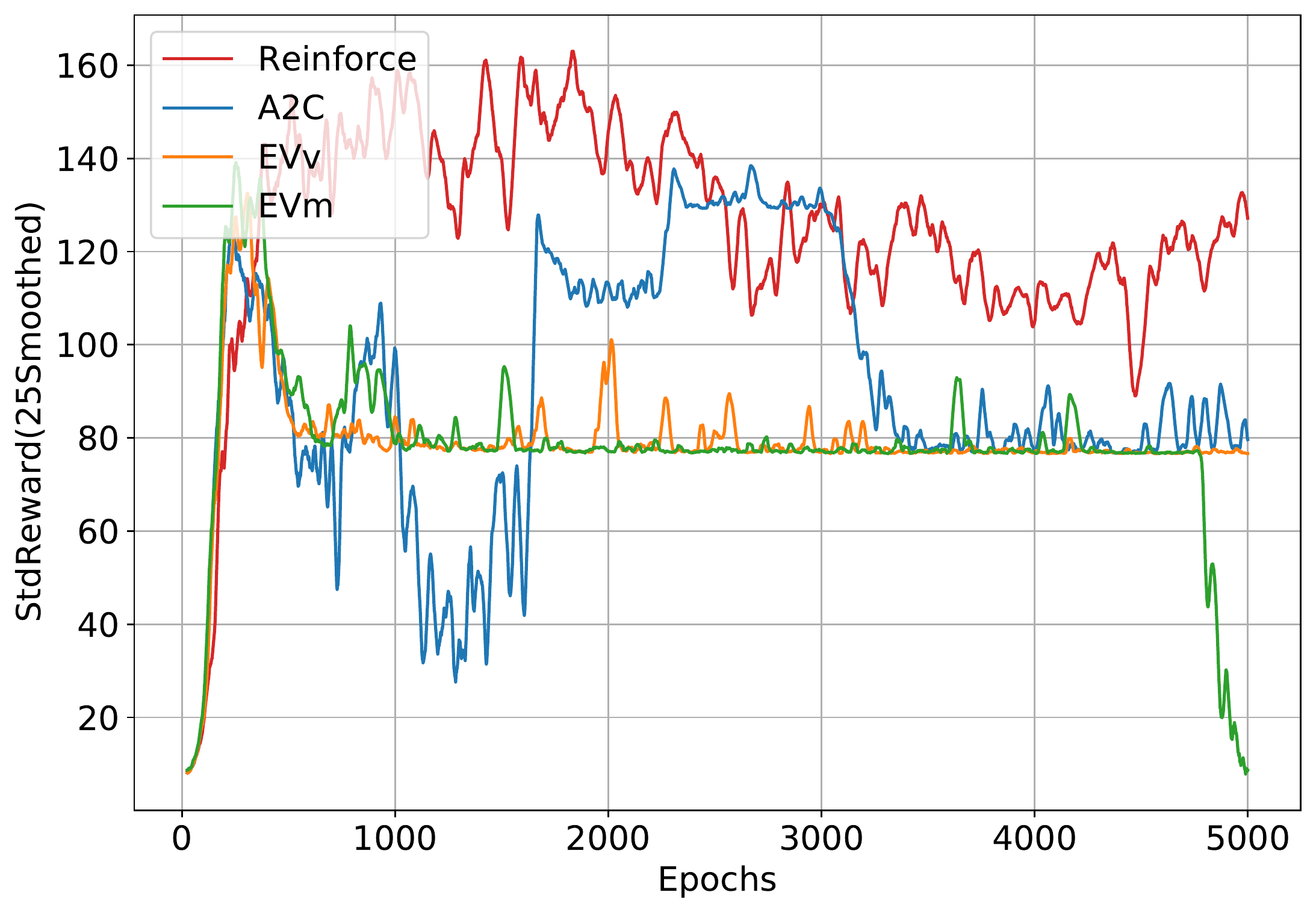} & (c)\includegraphics[scale=.2]{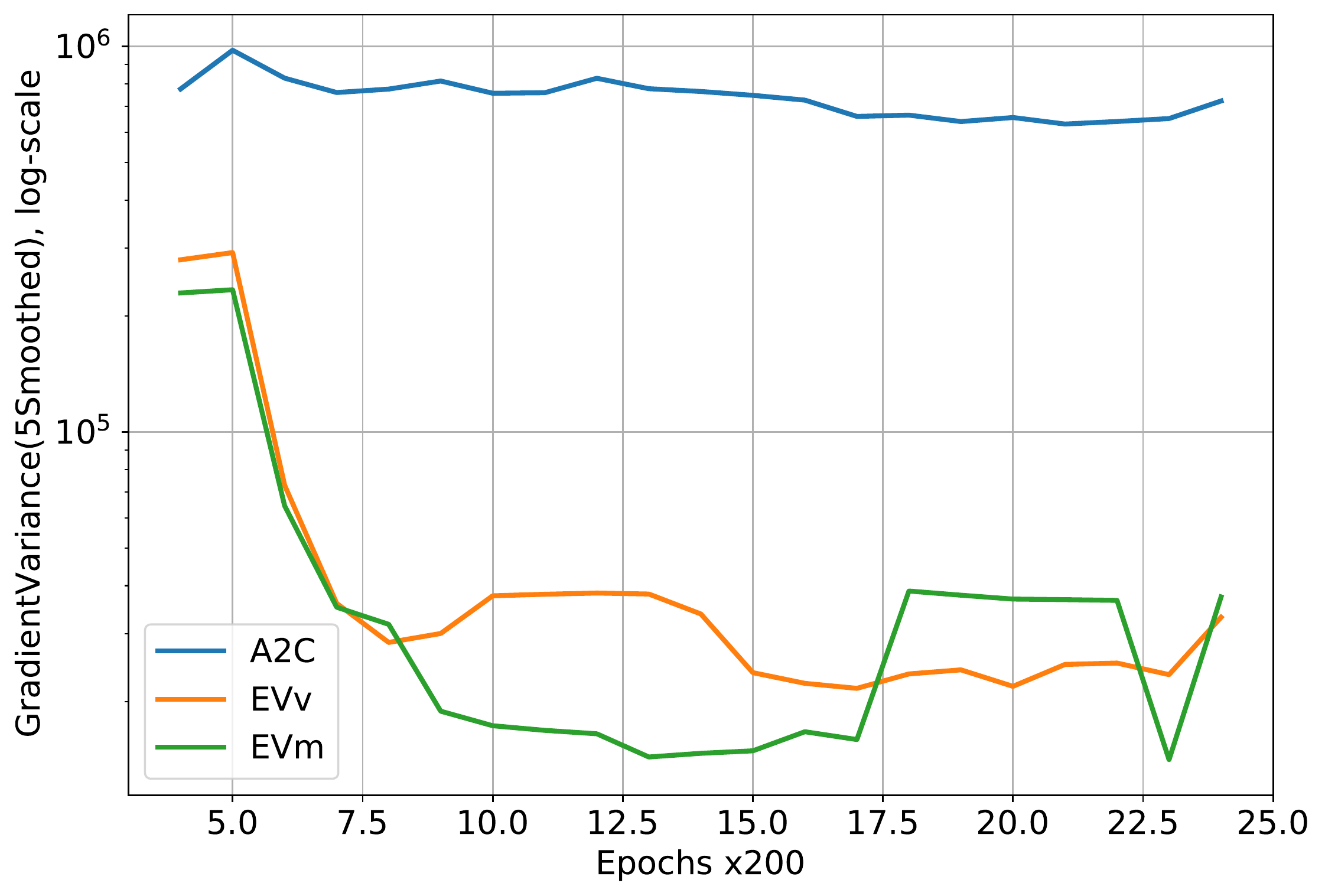} \\
    (d) \includegraphics[scale=.2]{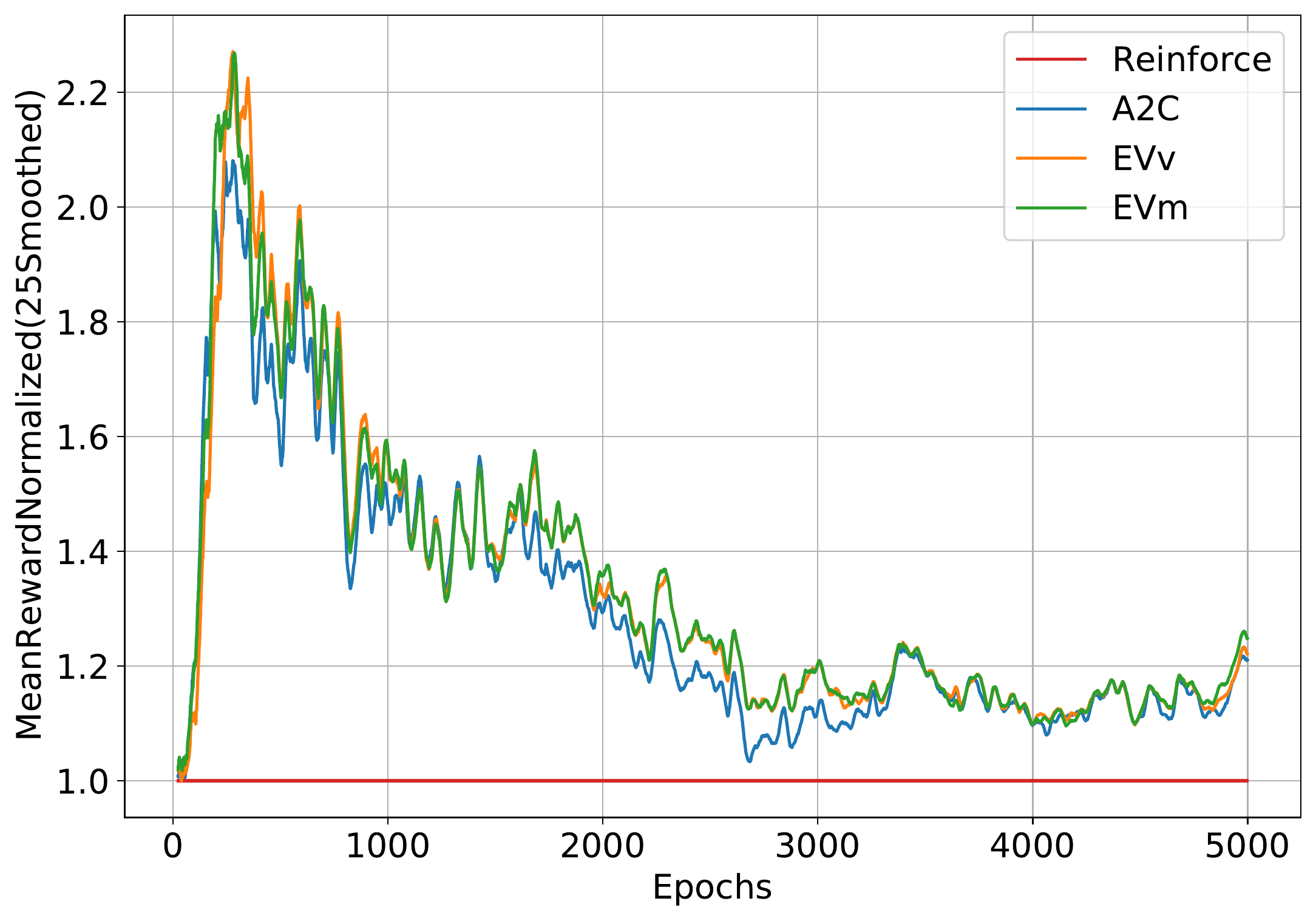} & (e) \includegraphics[scale=.2]{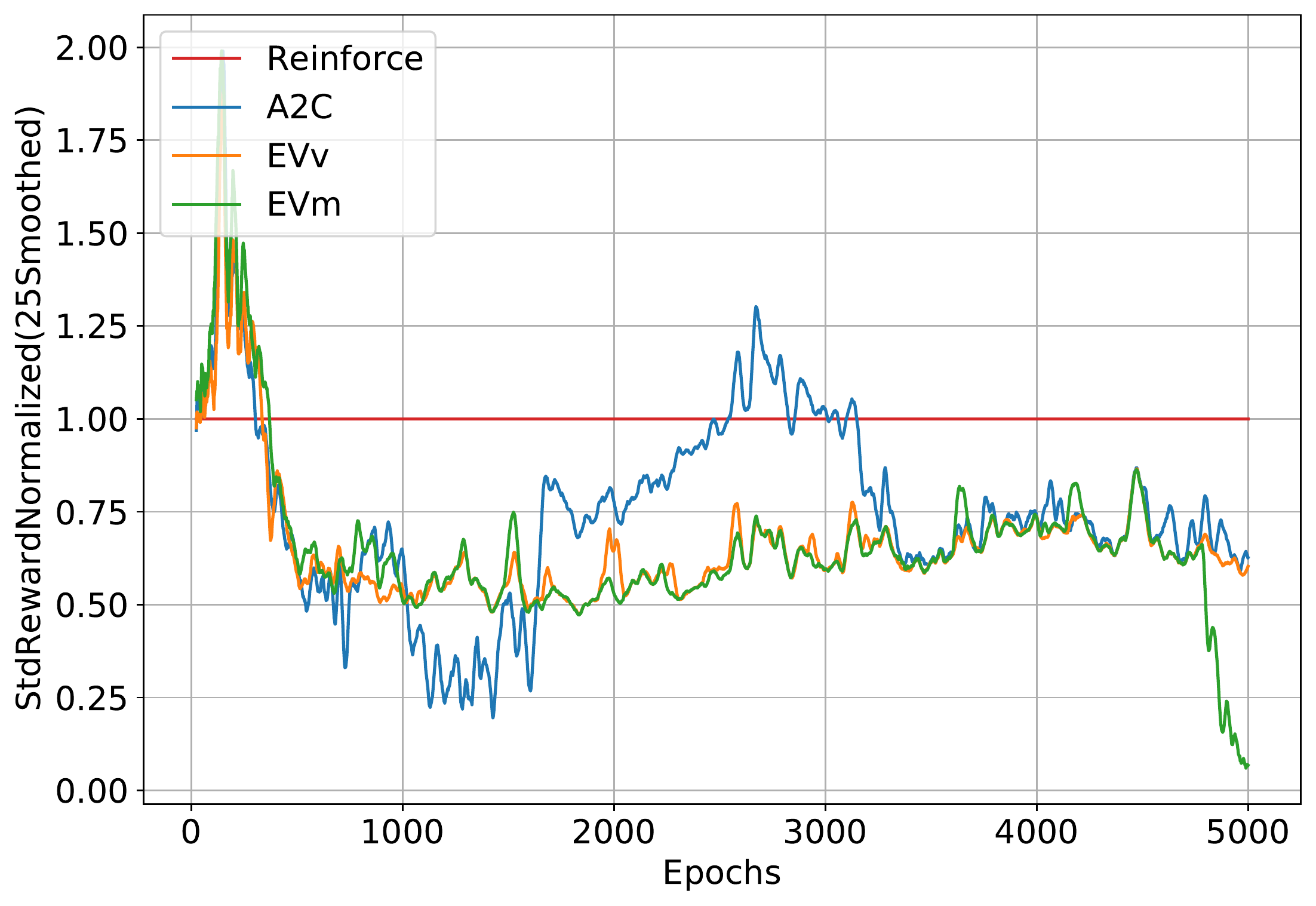} & (f)\includegraphics[scale=.2]{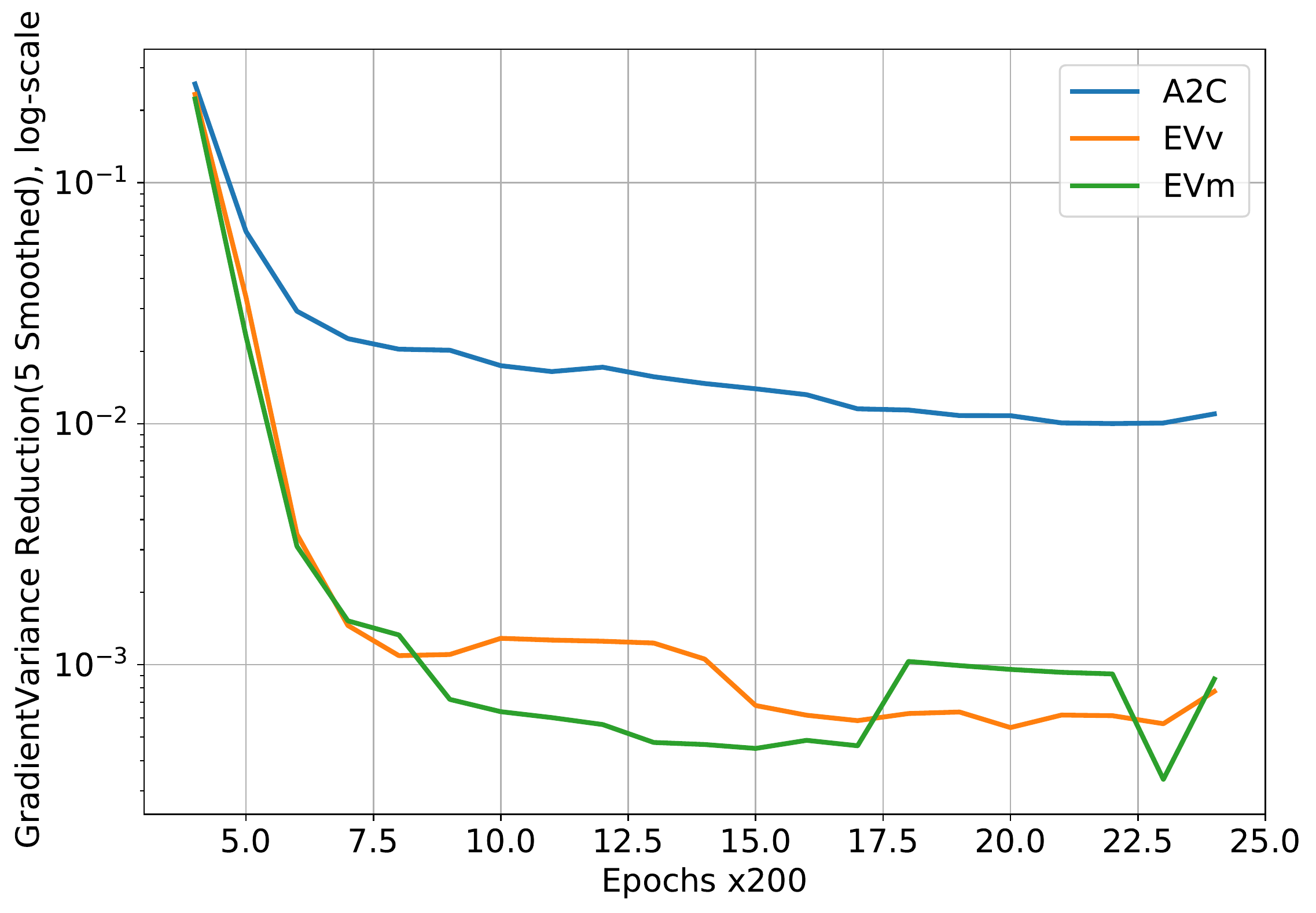} \\
    \end{tabular}
    \caption{The charts representing the results of the experiments in CartPole environment (config9): (a) displays mean rewards, (b) shows standard deviation of the rewards, (c) depicts gradient variance, in (d,e) the first two quantities are shown relative to REINFORCE and (f) shows gradient variance reduction ratio.}
    \label{fig:sup_Cartpole_config9}
\end{figure}

%% file: suppLunarLander.tex
LunarLander is a console-like game where the agent can observe the physical state of the system and decide which engine to fire (the primary one at the bottom or one of the secondaries, the left or right). There are 8 state variables: two coordinates of the lander, its linear velocities, its angle and angular velocity, and two boolean values that show whether each leg is in contact with the ground.\\

\begin{figure}[h!]
    \centering
    \begin{tabular}{lcr}
    (a) \includegraphics[scale=.2]{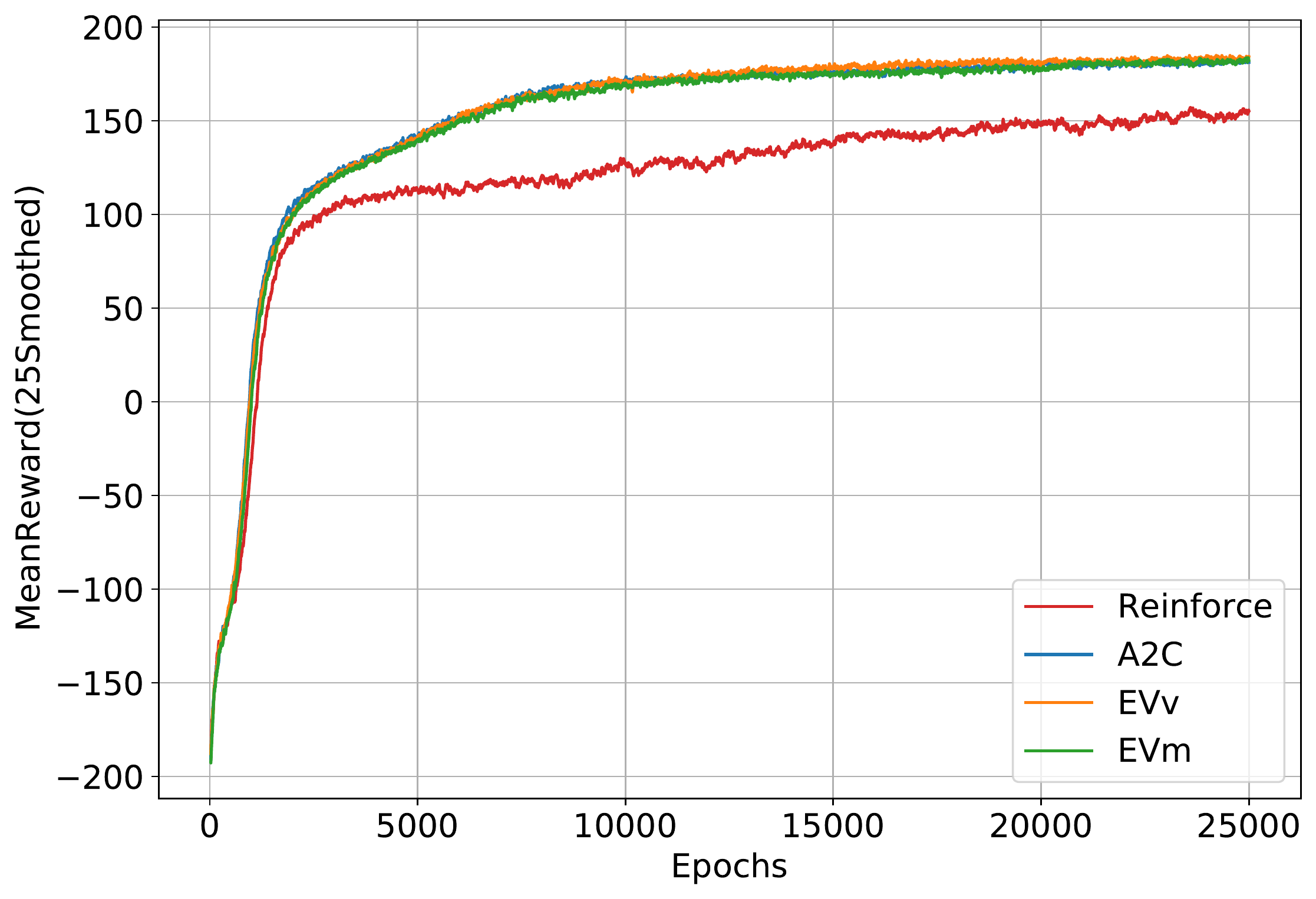} & (b) \includegraphics[scale=.2]{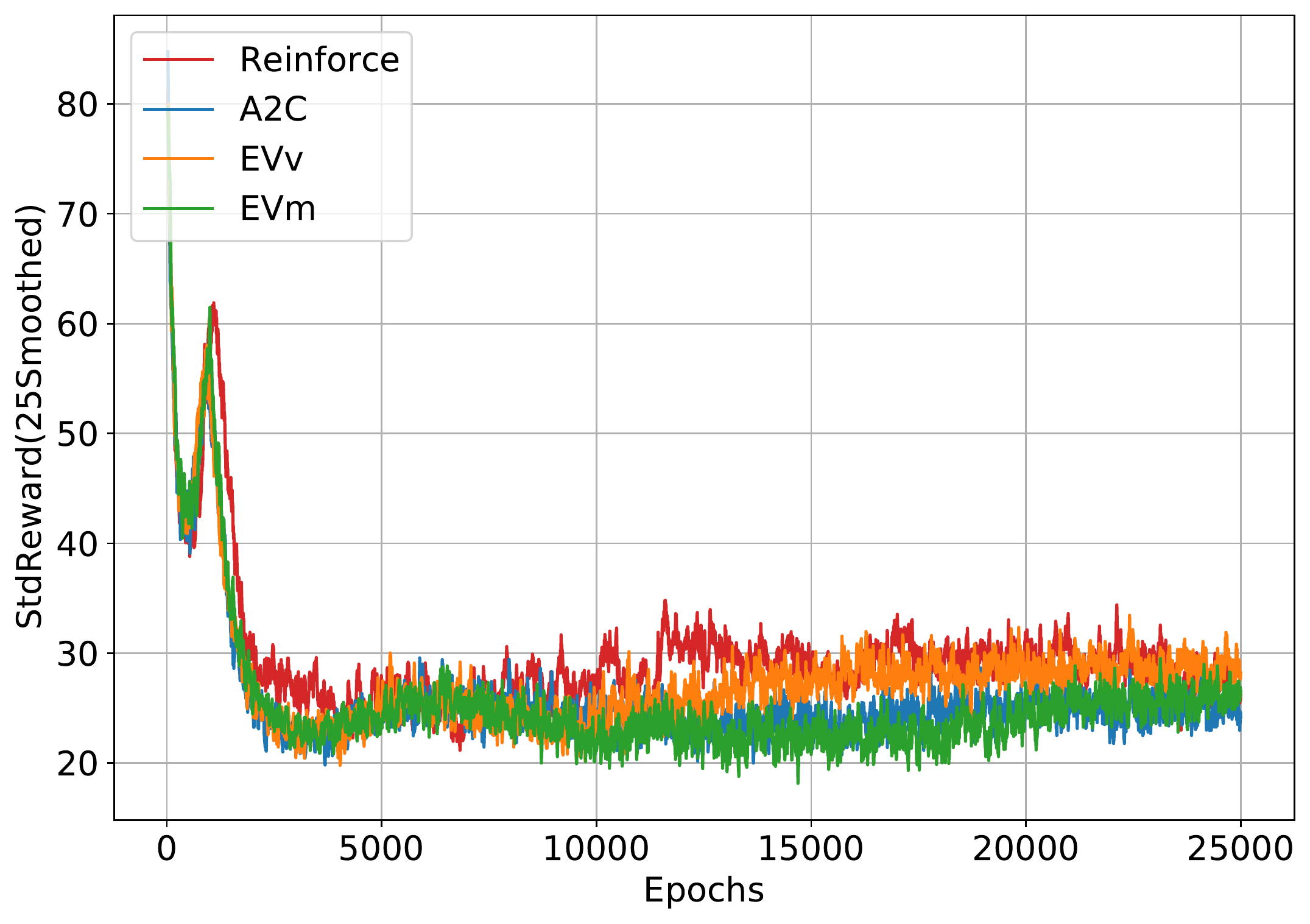} & (c)\includegraphics[scale=.2]{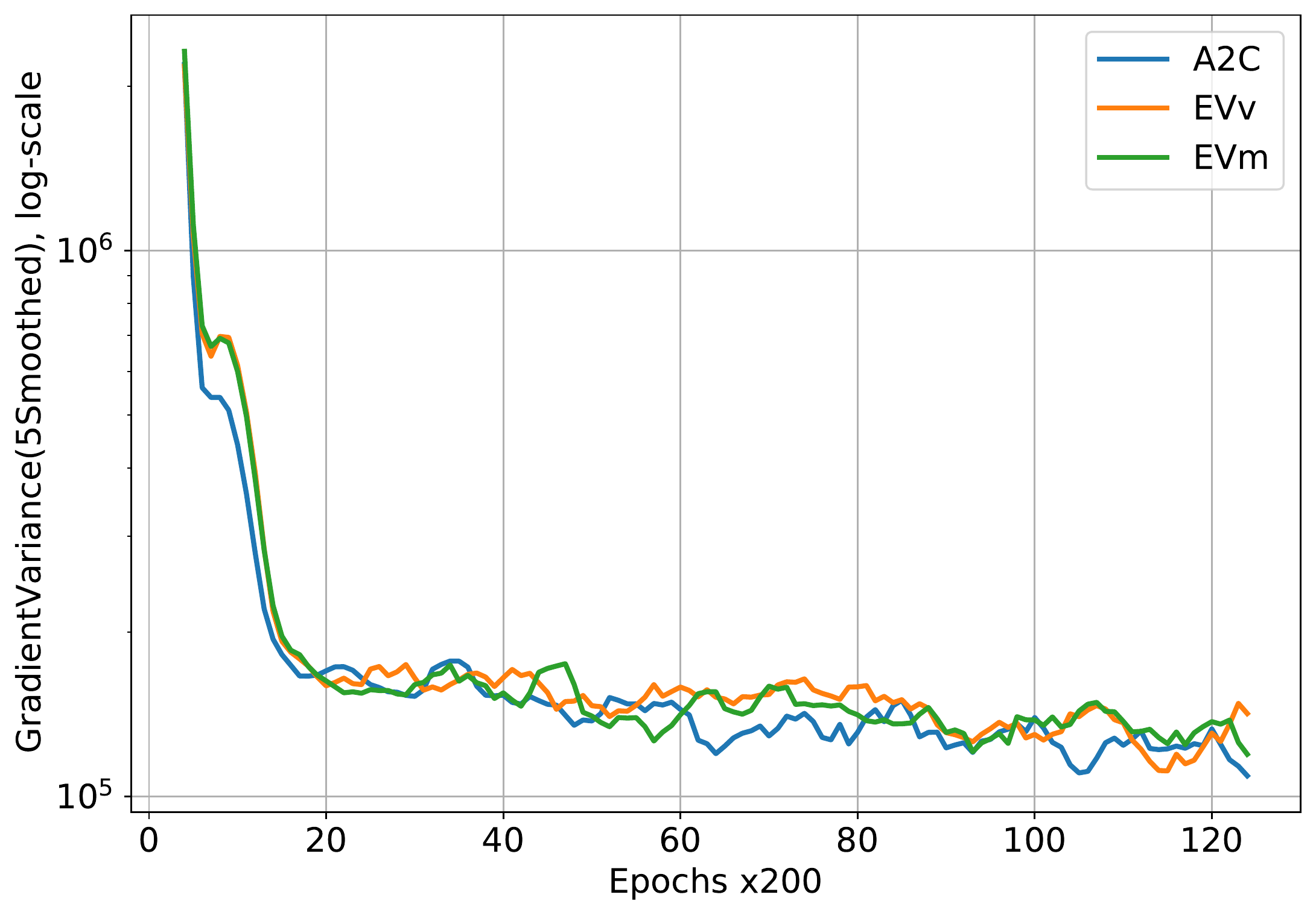} \\
    (d) \includegraphics[scale=.2]{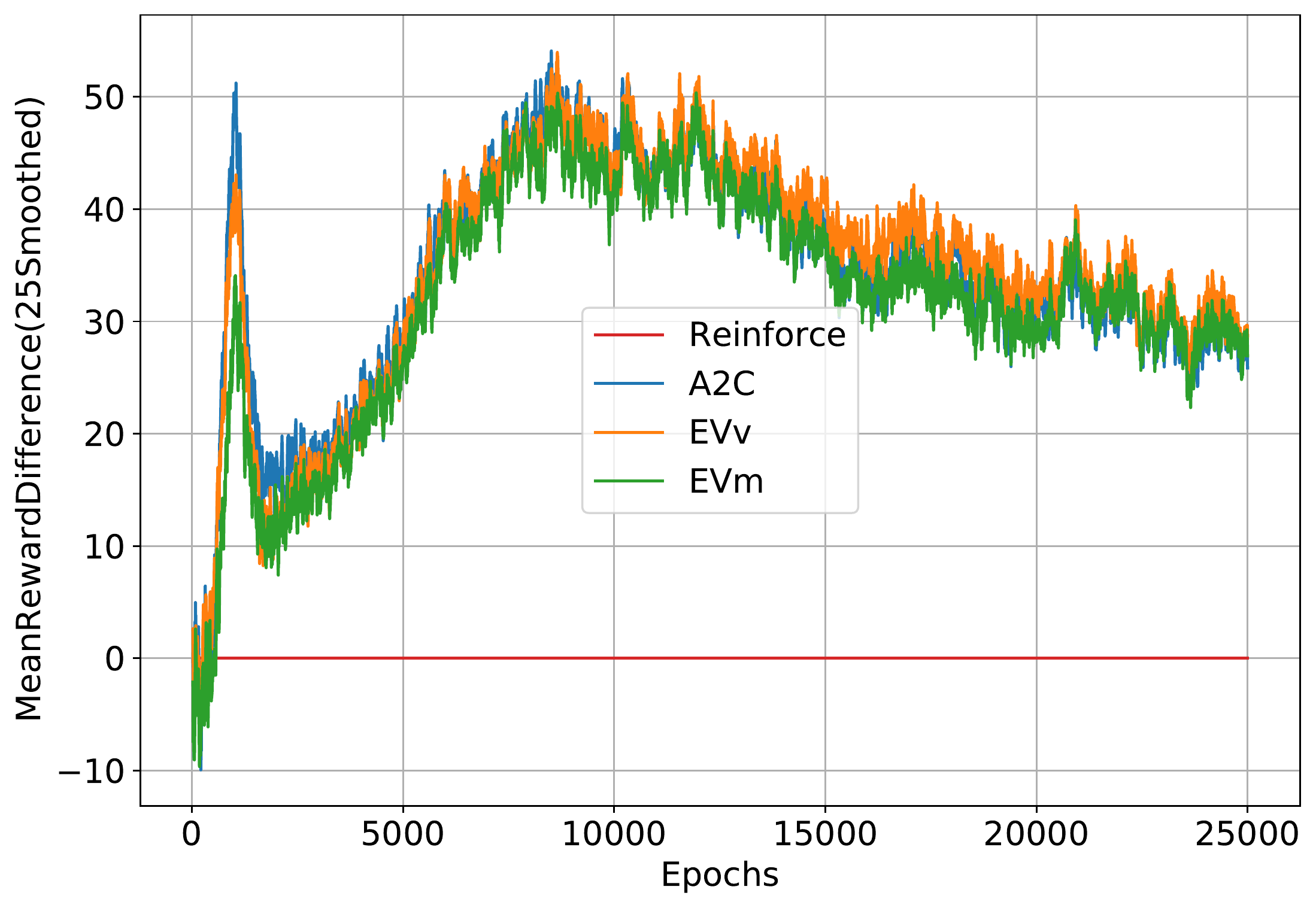} & (e) \includegraphics[scale=.2]{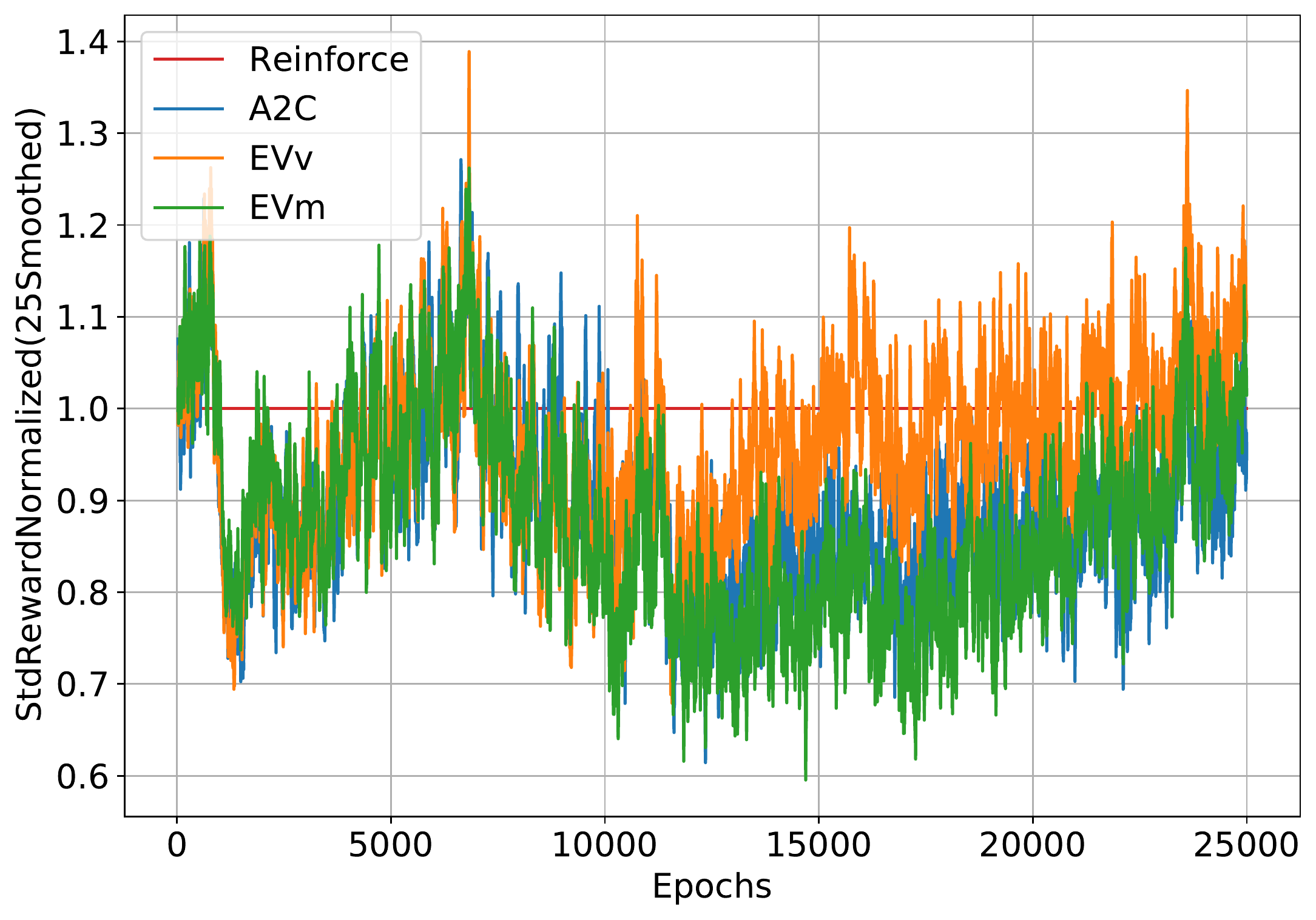} & (f)\includegraphics[scale=.2]{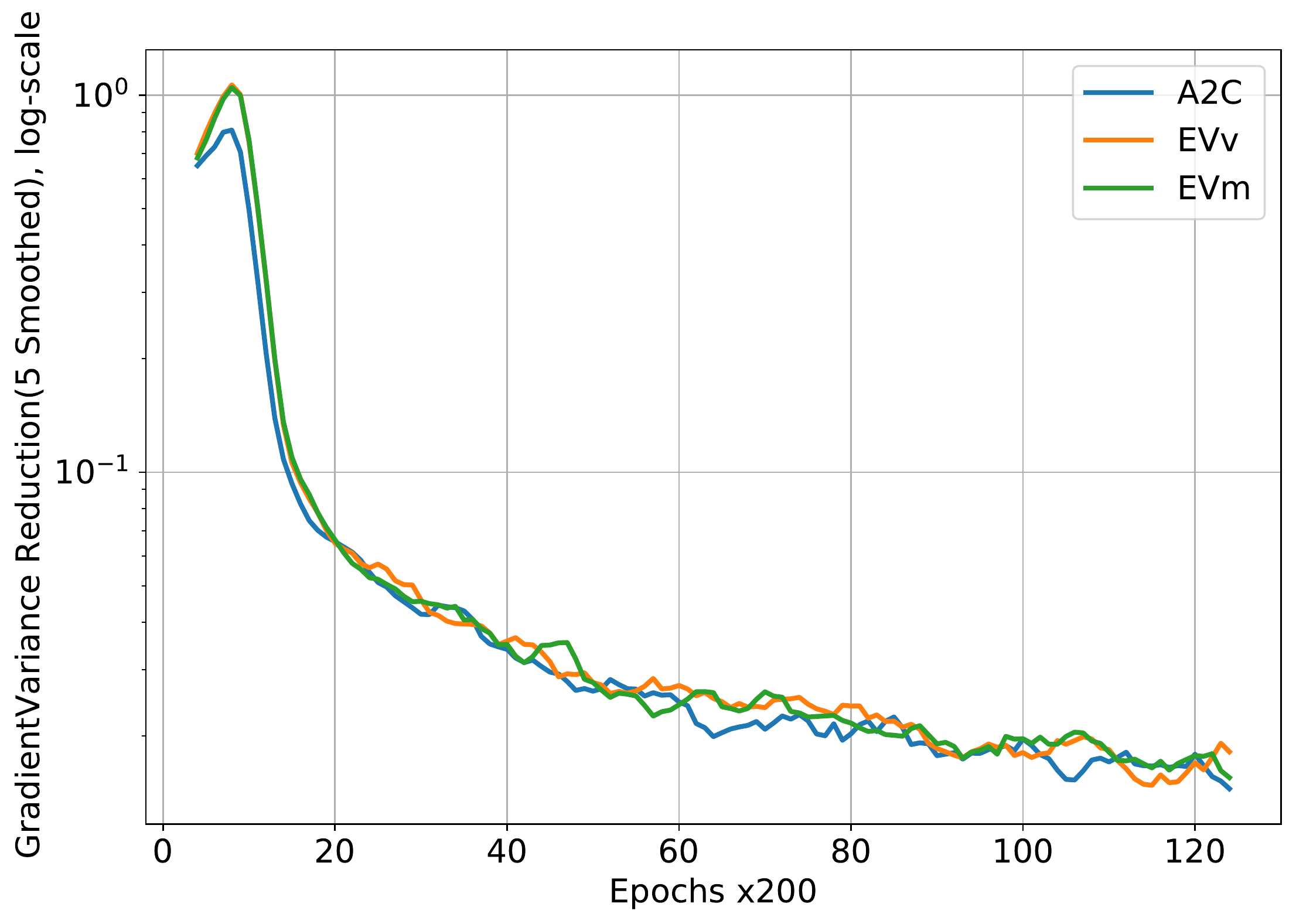} \\
    \end{tabular}
    \caption{The charts representing the results of the experiments in LunarLander environment (config1): (a) displays mean rewards, (b) shows standard deviation of the rewards, (c) depicts gradient variance, in (d) the difference between the algorithm and REINFORCE is shown, (e) shows the standard deviation of the rewards relative to REINFORCE and (f) shows gradient variance reduction ratio.}
    \label{fig:sup_LL}
\end{figure}

LunarLander (see Fig. \ref{fig:sup_LL}) is the example of the case where all algorithms work in the same way and there is no significant difference between A2C and EV. It happens regardless to the policy type we choose; the final performances are different among the configs but inside one config A2C and EV gave the same result. We see that all algorithms behave similarly in variance reduction as well, showing that EV-methods are still good but sometimes A2C works with the same result.

%% file: suppAcrobot.tex
The system consists of two links forming a chain, with one end of the chain fixed. The joint between the two links is actuated. The goal is to apply torques on the actuated joint to swing the free end of the linear chain above a given height while starting from the initial state of hanging downwards. The actions can be to apply $\pm1$ or $0$ torque to the joint and the goal is to have the free end reach a designated target height in as few steps as possible, and as such all steps that do not reach the goal incur a reward of -1.\\

\begin{figure}[h!]
    \centering
    \begin{tabular}{lcr}
    (a) \includegraphics[scale=.2]{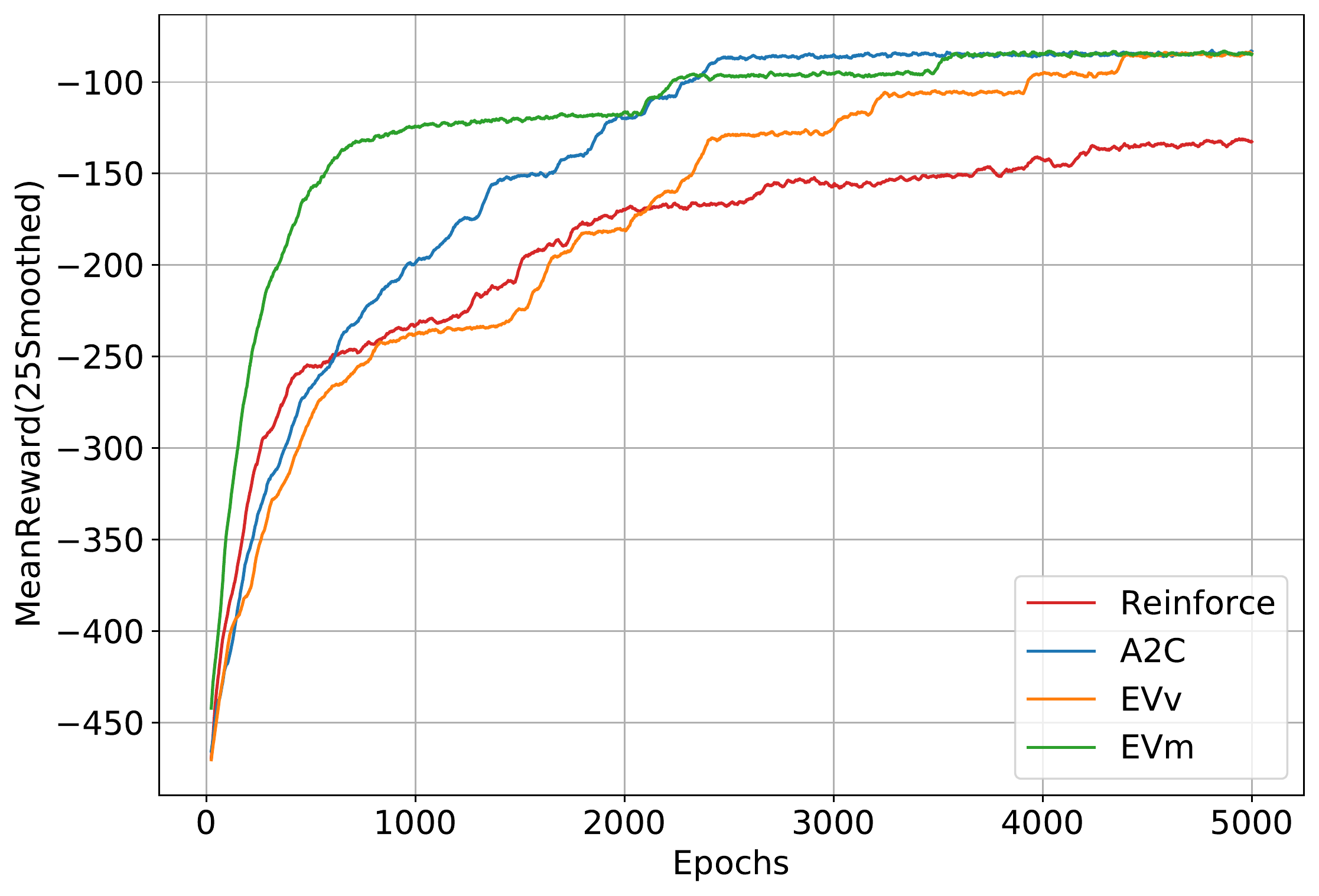} & (b) \includegraphics[scale=.2]{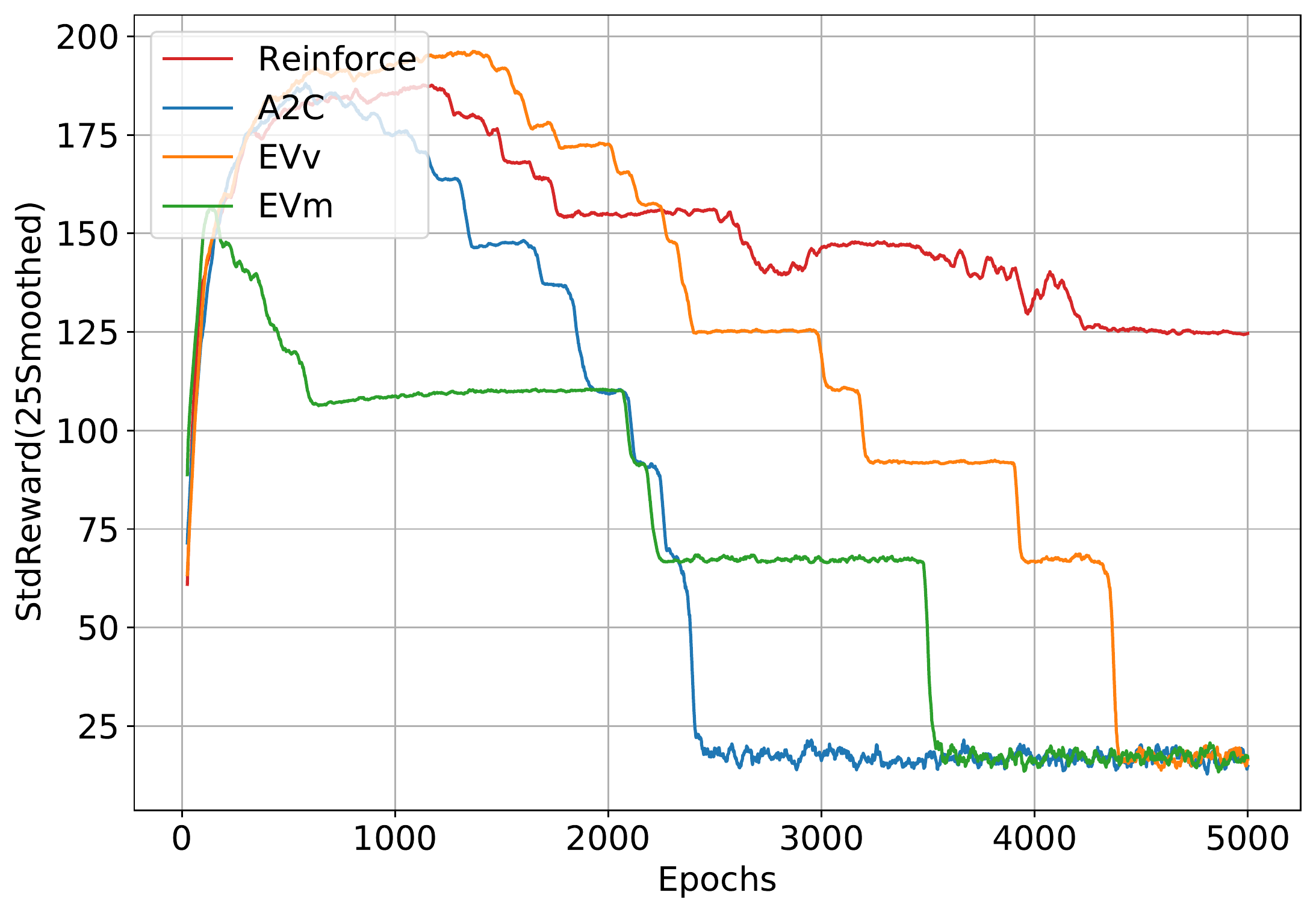} & (c)\includegraphics[scale=.2]{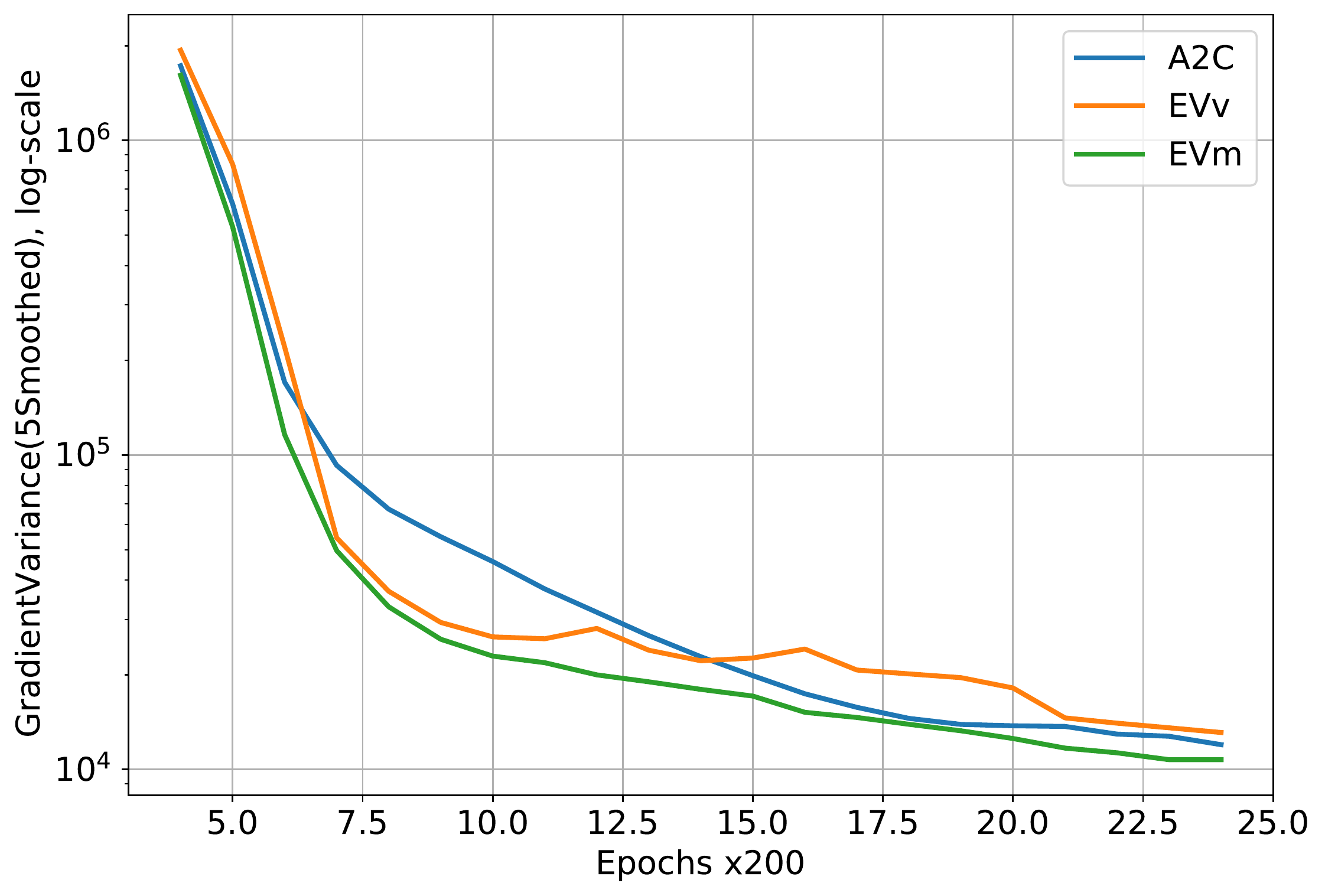} \\
    (d) \includegraphics[scale=.2]{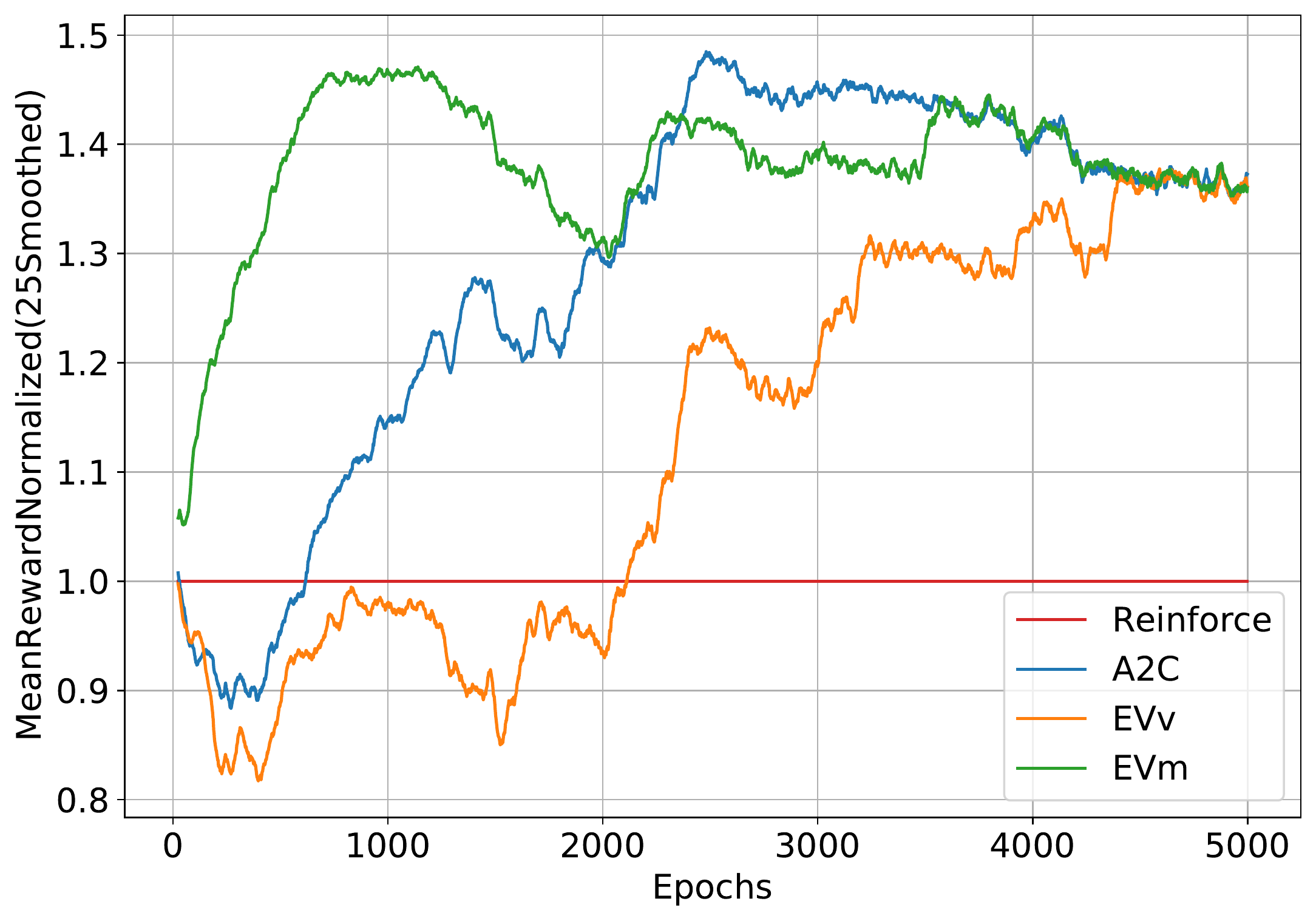} & (e) \includegraphics[scale=.2]{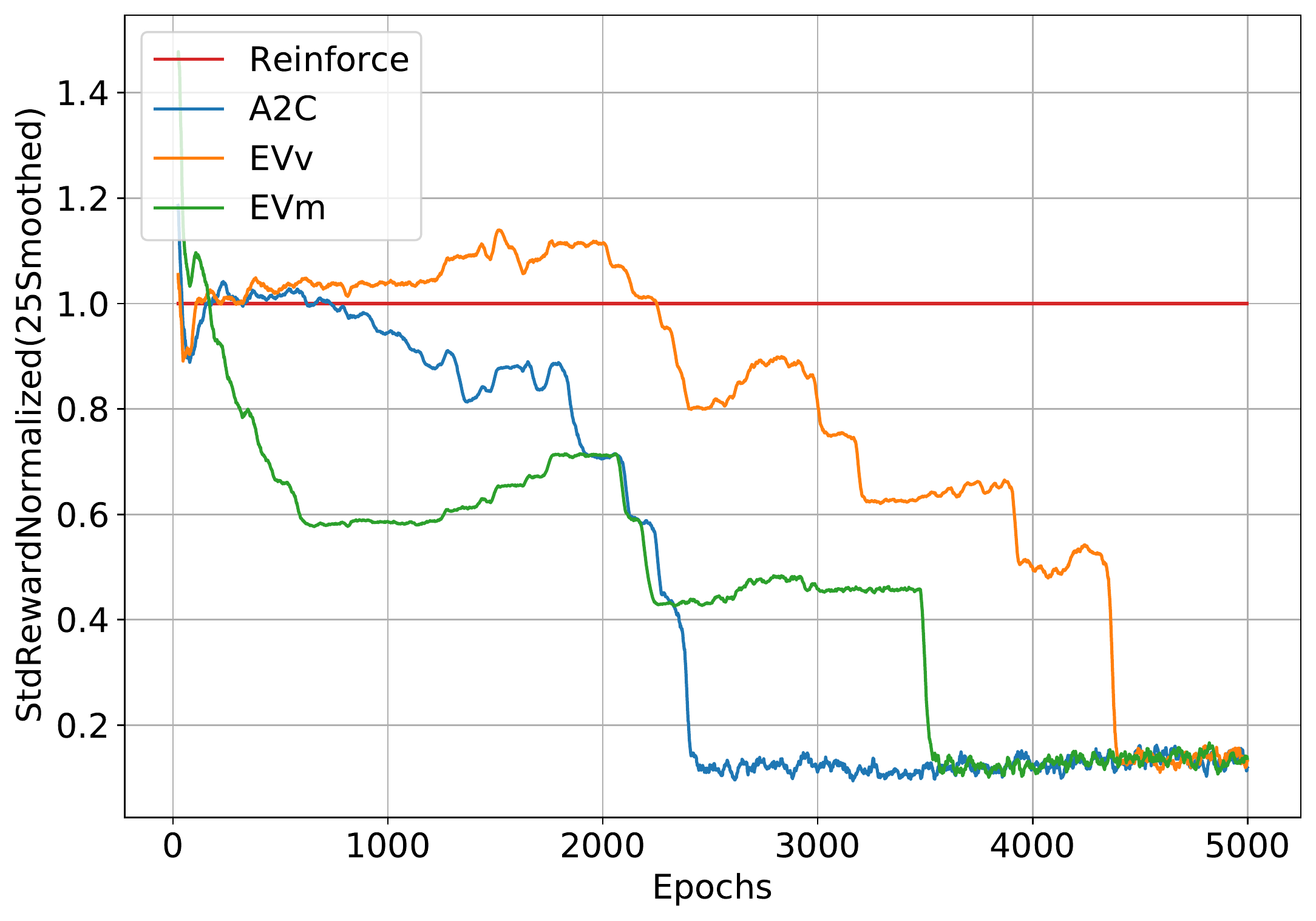} & (f)\includegraphics[scale=.2]{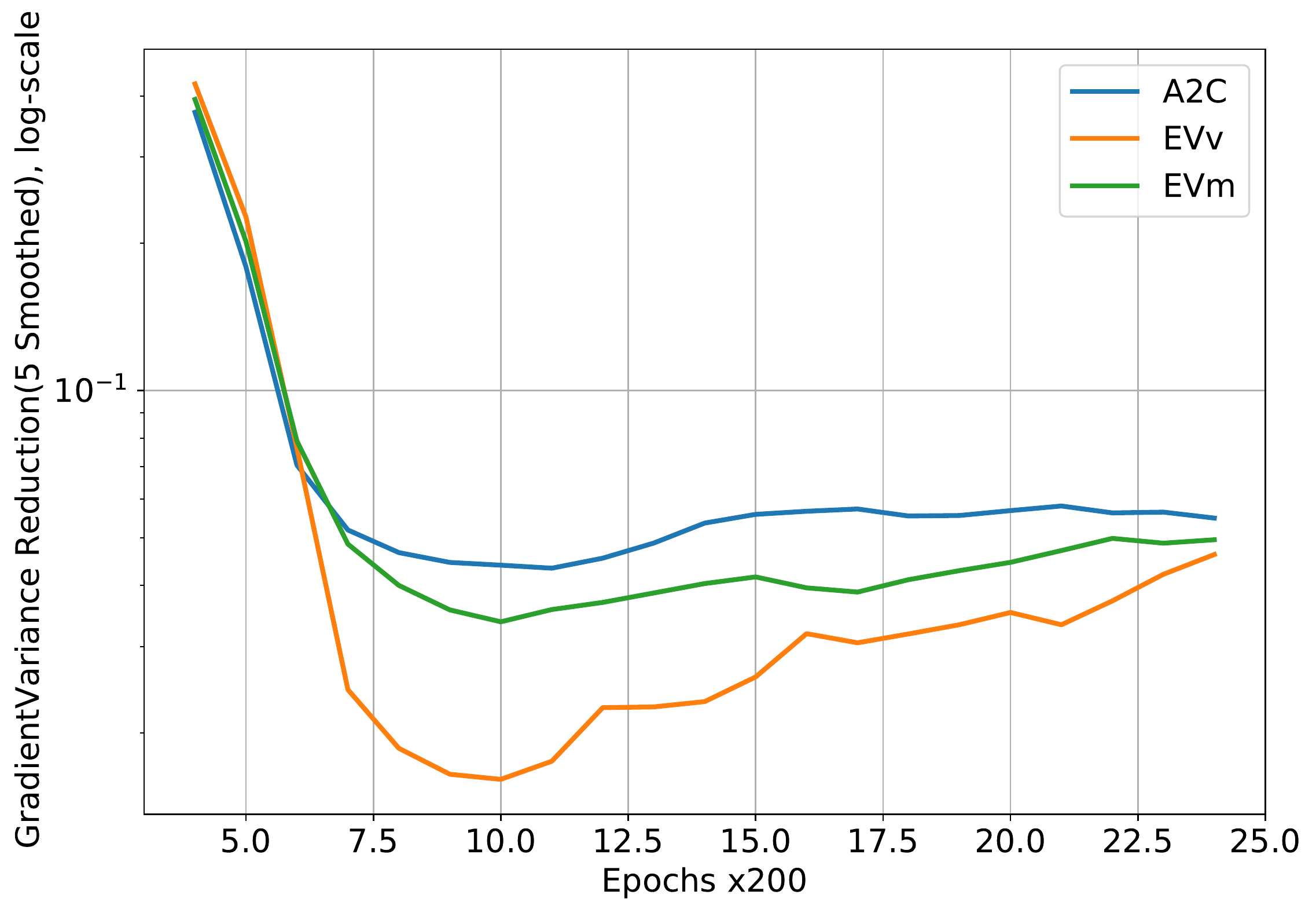} \\
    \end{tabular}
    \caption{The charts representing the results of the experiments in Acrobot environment (config1): (a) displays mean rewards, (b) shows standard deviation of the rewards, (c) depicts gradient variance, in (d) the difference between the algorithm and REINFORCE is shown, (e) shows the standard deviation of the rewards relative to REINFORCE and (f) shows gradient variance reduction ratio.}
    \label{fig:sup_Acb}
\end{figure}

The config we show here and in the main text (see Fig. \ref{fig:sup_Acb}) is an example where EV can boost training sometimes and that a clever combination of EV and A2C may result in even better algorithms than these three.  We can clearly see that until the agent reaches reward ceiling there is a clear predominance of EVm over EVv and A2C but in the end they result in the same policy. It can be seen that standard deviation of the rewards indicate positive effect in the same time. Still, it must be noted that variance reduction is the best in EVm and EVv until the ceiling is reached. Hence, the environment itself does not require so excessive variance reduction and there is still an open space for discussions about whether the variance reduction needed in such environment.

%% file: charts_sup_time.tex
\begin{figure}[h!]
    \centering
    \begin{tabular}{l}
    (a) \includegraphics[scale=.27]{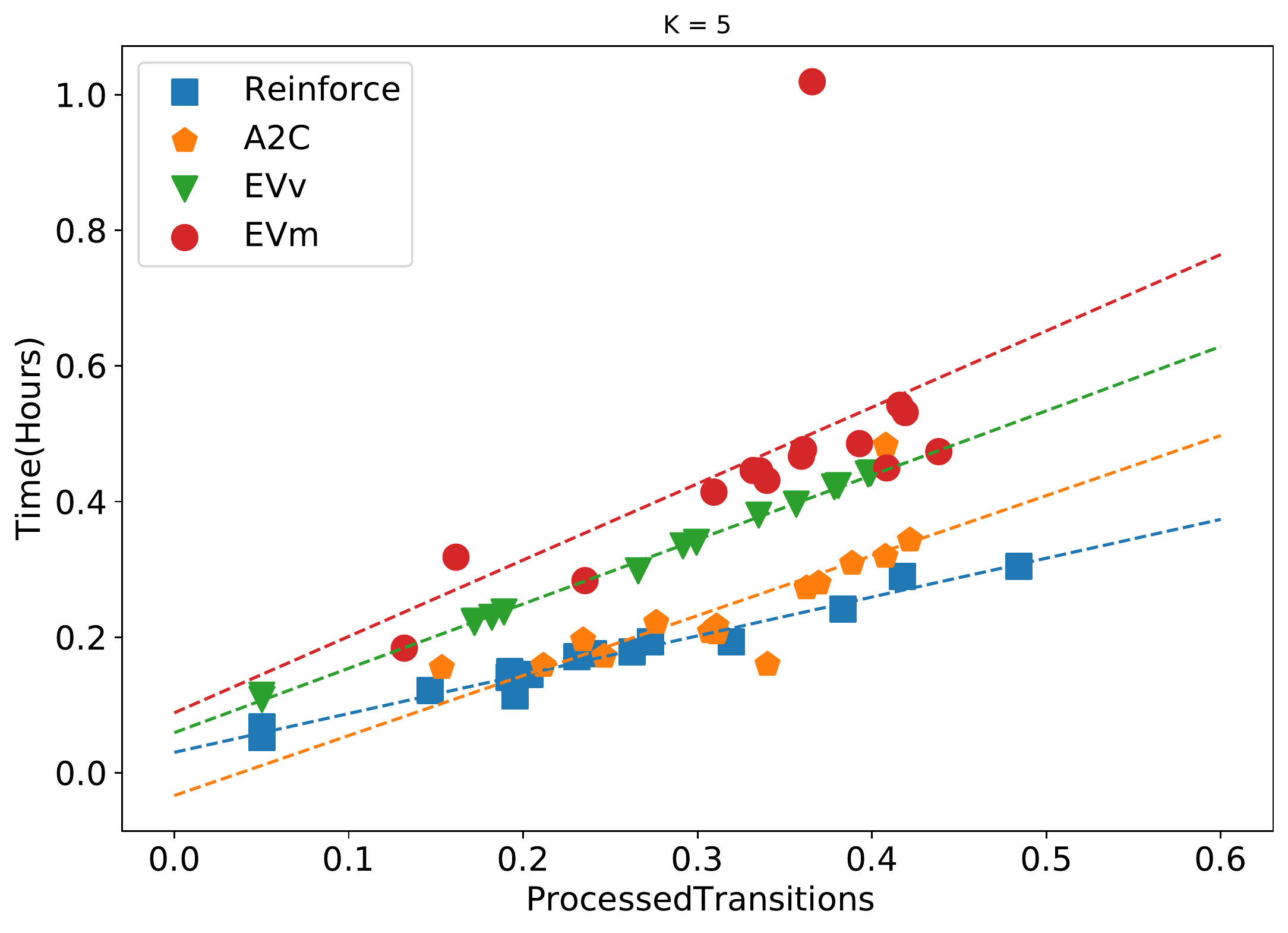}
    (b) \includegraphics[scale=.27]{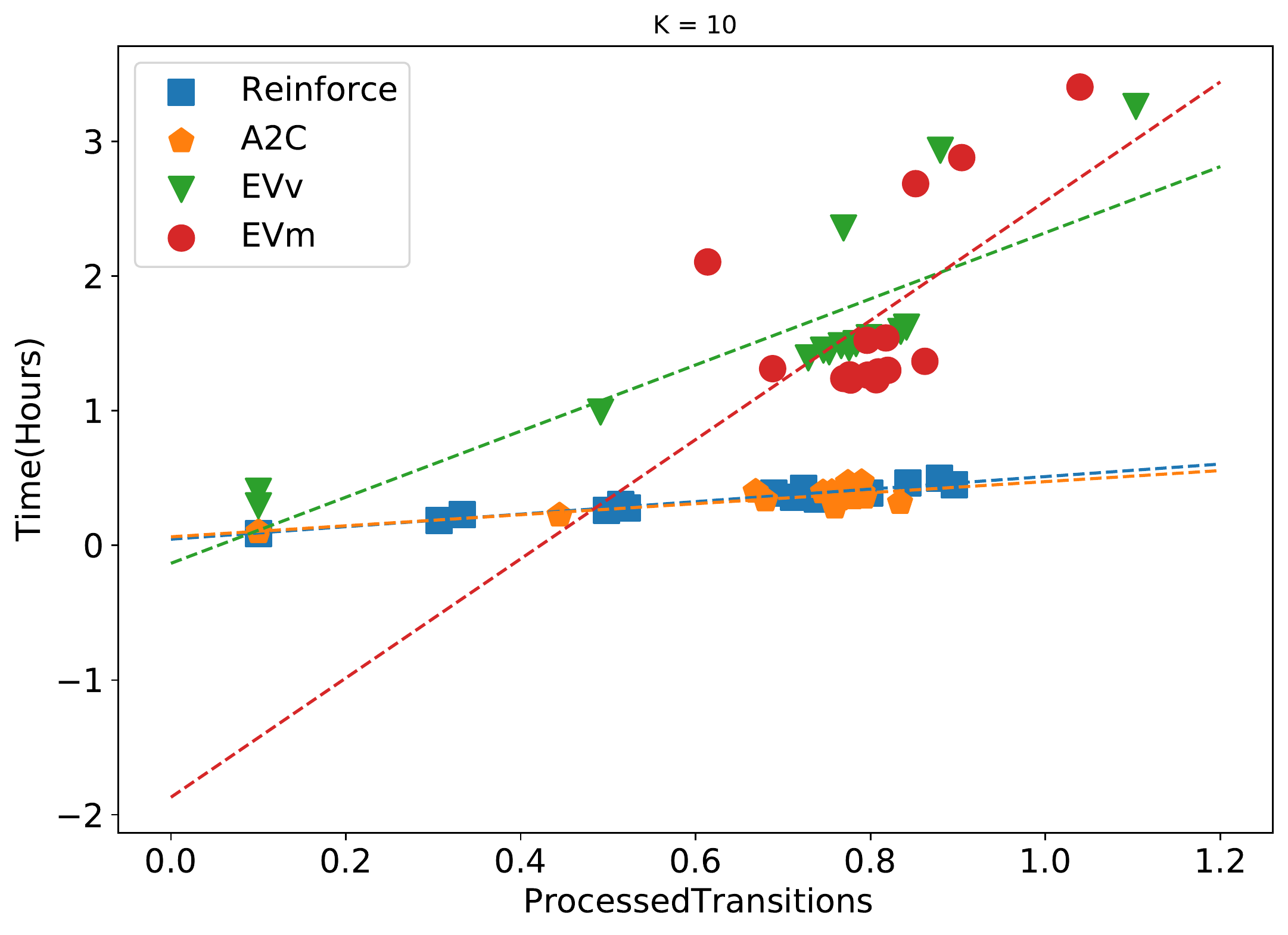} \\ 
    (c) \includegraphics[scale=.27]{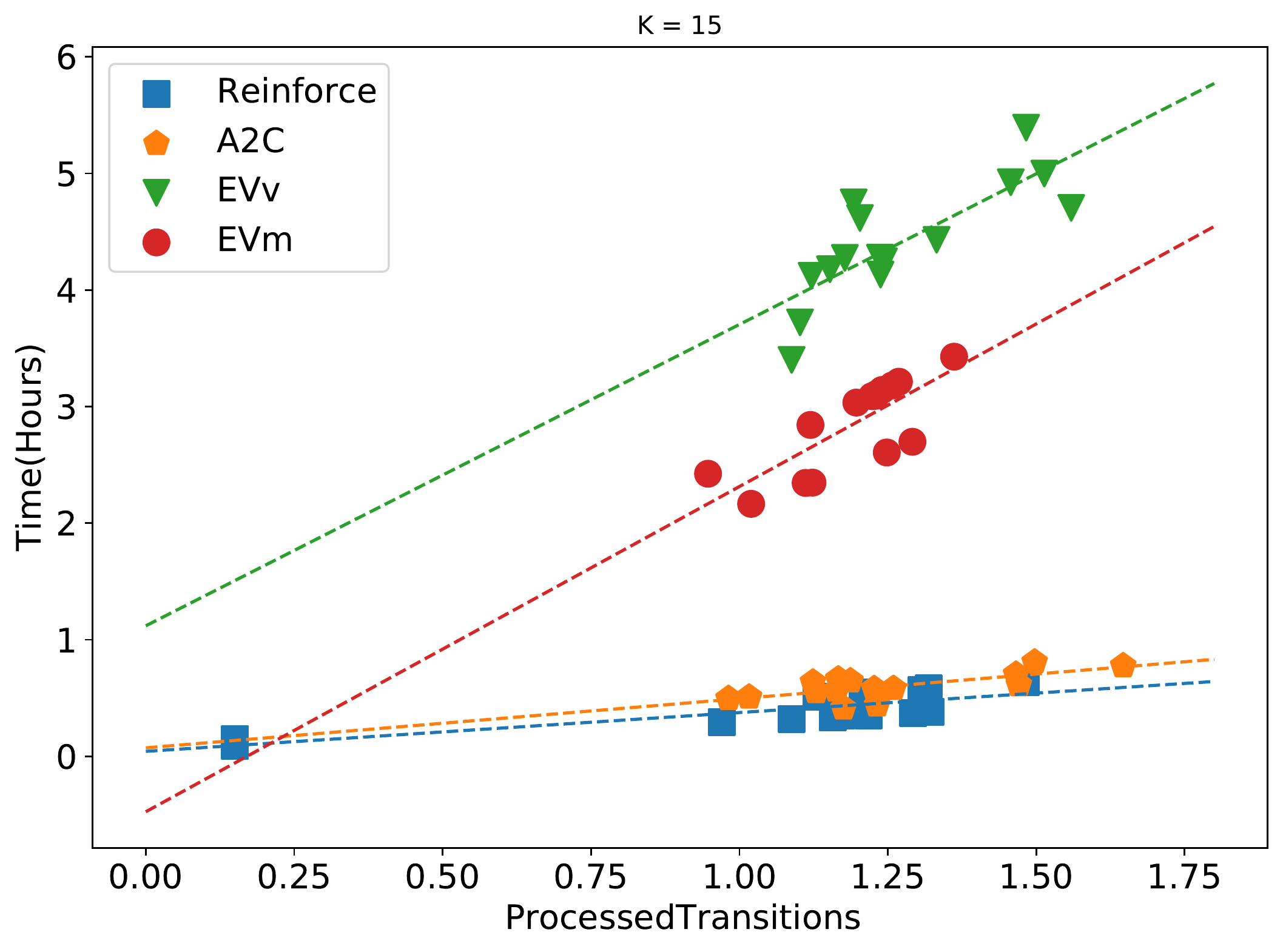}
    (d) \includegraphics[scale=.27]{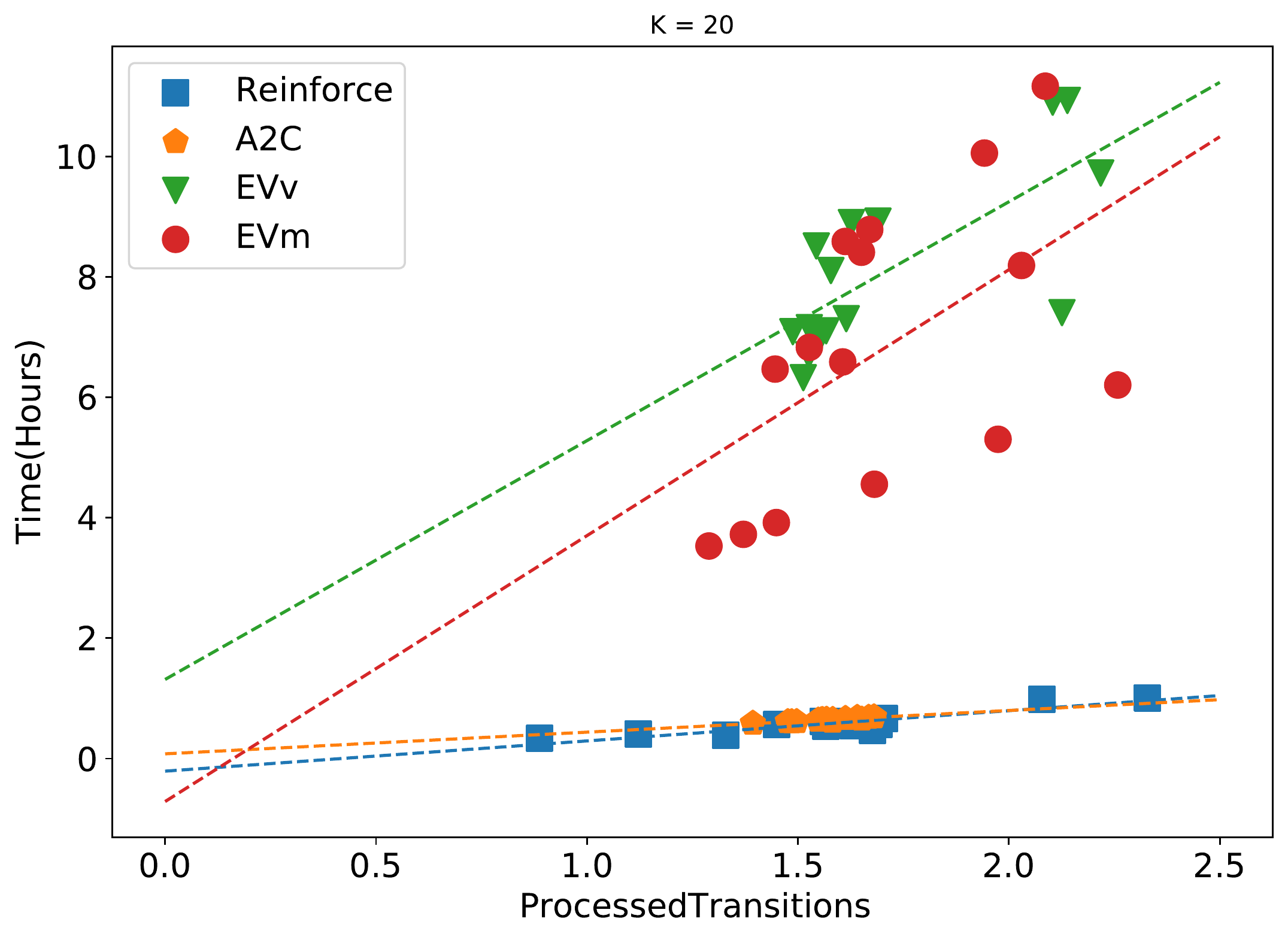} 
    \end{tabular}
    \caption{The charts representing dependency of training time from number of the processed transitions(scale of millions) for GoToDoor}
    \label{fig:sup_GoToDoor_elapsedtime}
\end{figure}

\begin{figure}[h!]
    \centering
    \begin{tabular}{l}
    (a) \includegraphics[scale=.27]{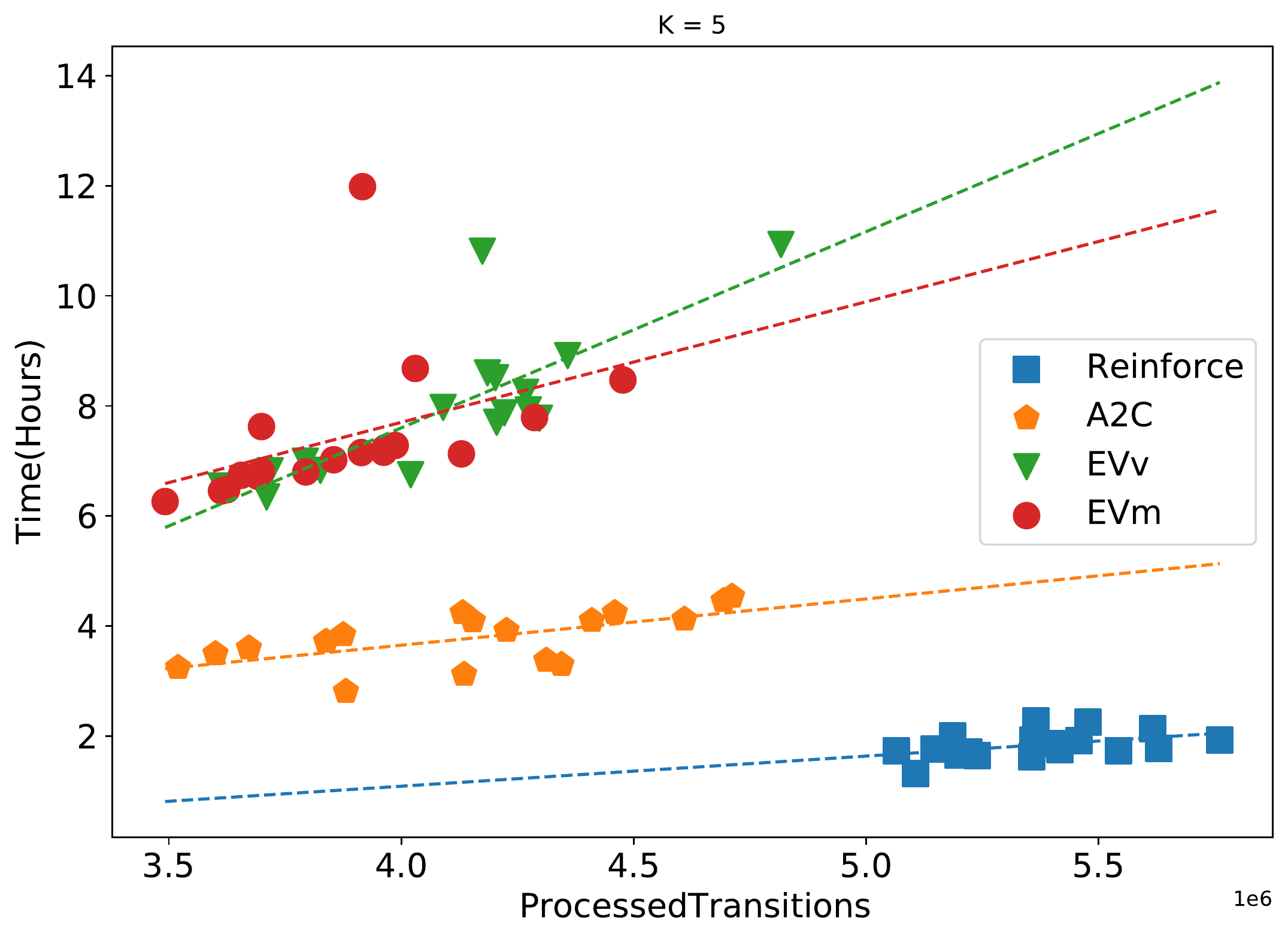}
    (b) \includegraphics[scale=.27]{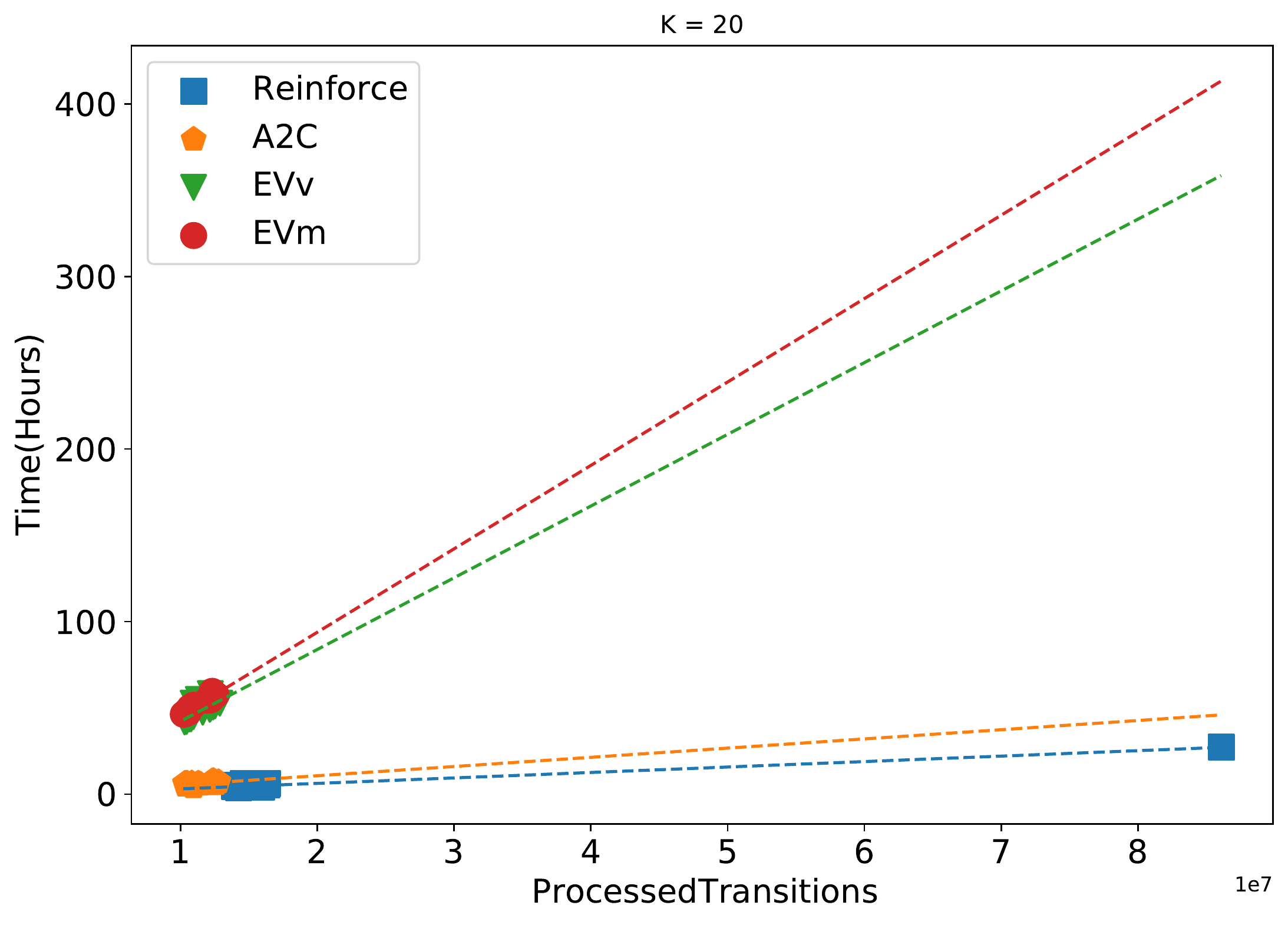} 
    \end{tabular}
    \caption{The charts representing dependency of training time from number of the processed transitions(scale of millions) for Unlock}
    \label{fig:sup_Unlock_elapsedtime}
\end{figure}

\begin{figure}[h!]
    \centering
    \begin{tabular}{l}
    (a) \includegraphics[scale=.27]{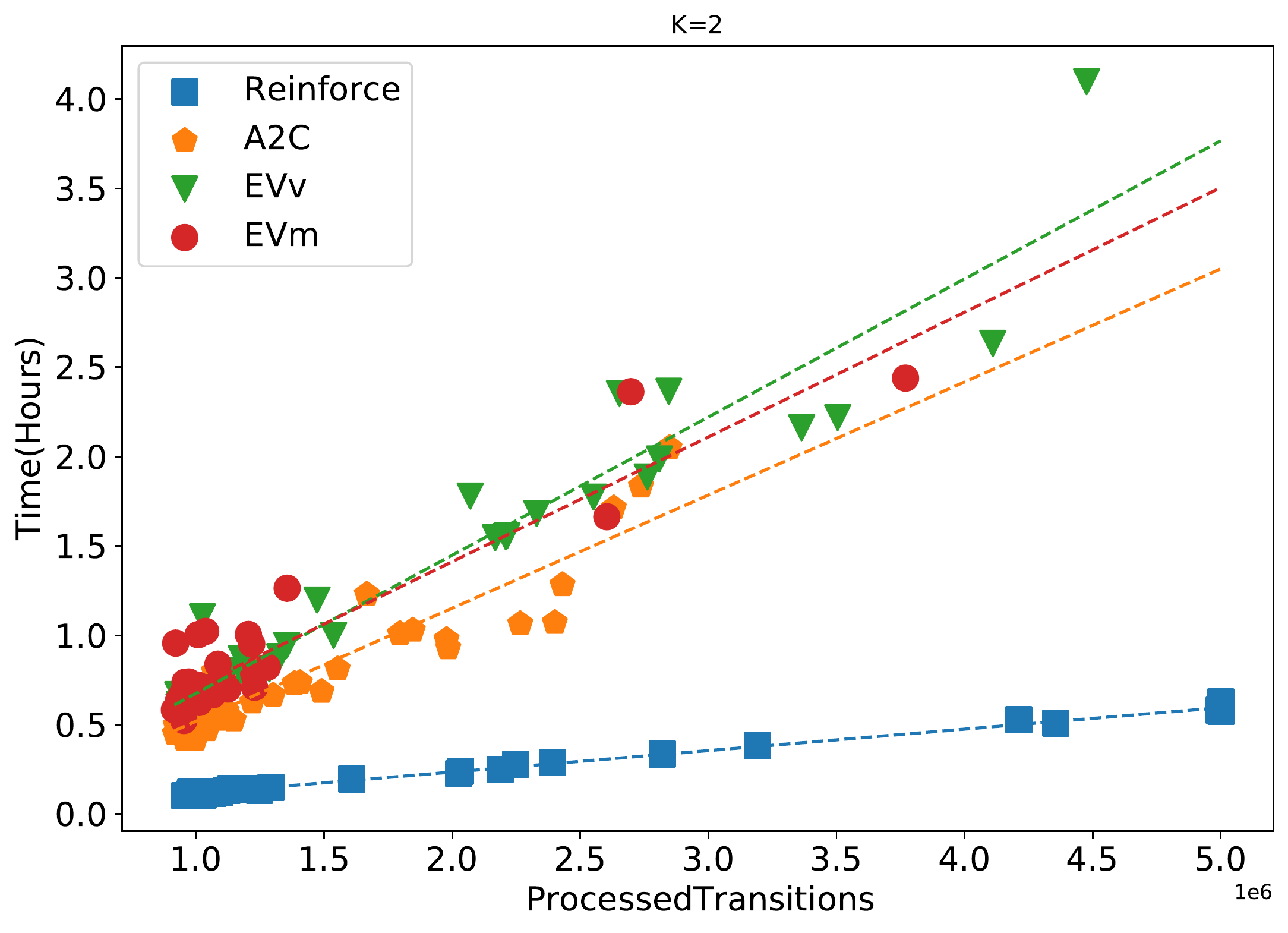}
    (b) \includegraphics[scale=.27]{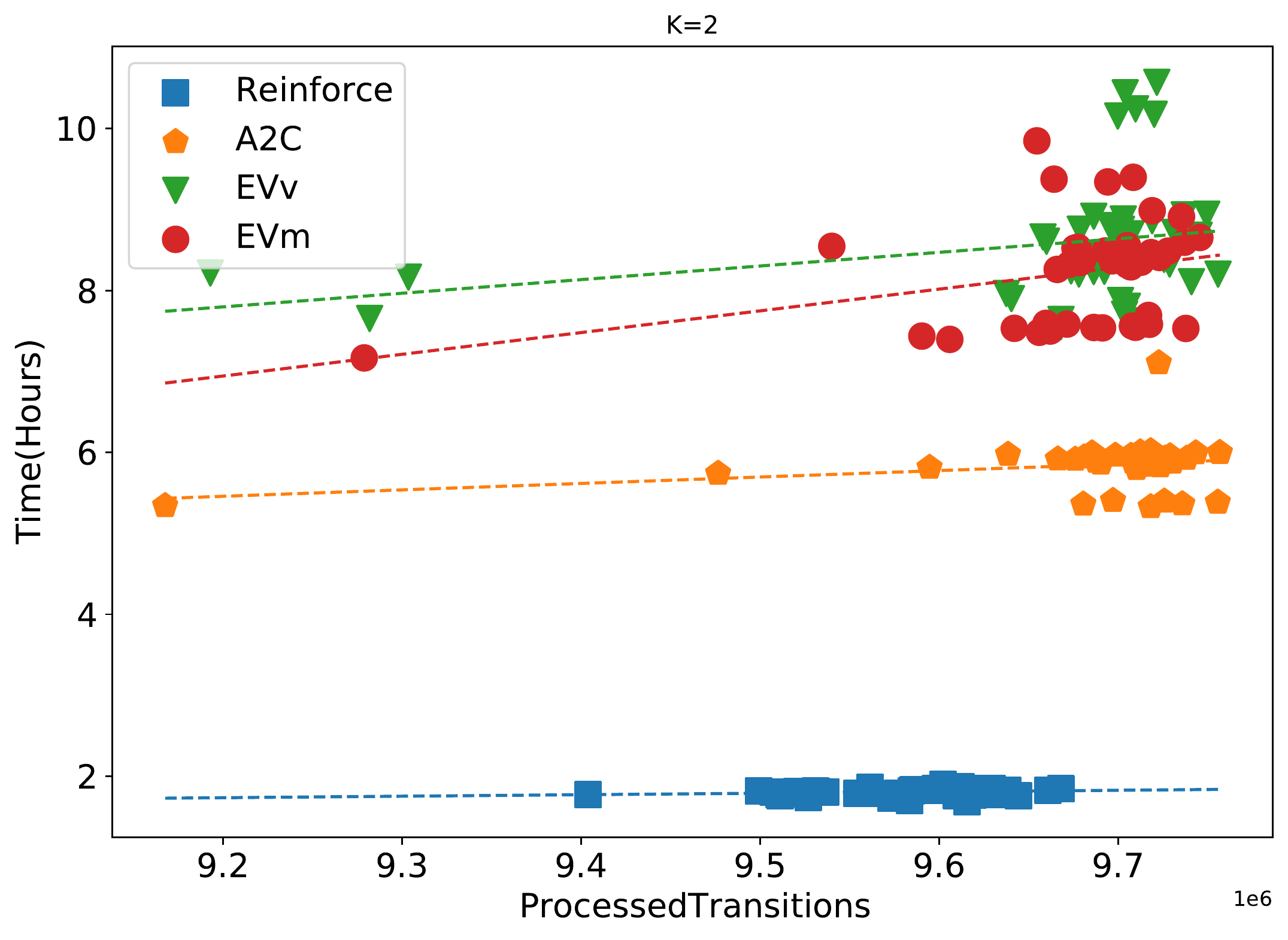}
    \end{tabular}
    \caption{The charts representing dependency of training time from number of the processed transitions(scale of millions) for Acrobot (a) and for LunarLander (b)}
\end{figure}

\begin{figure}[h!]
    \centering
    \begin{tabular}{l}
    (a) \includegraphics[scale=.27]{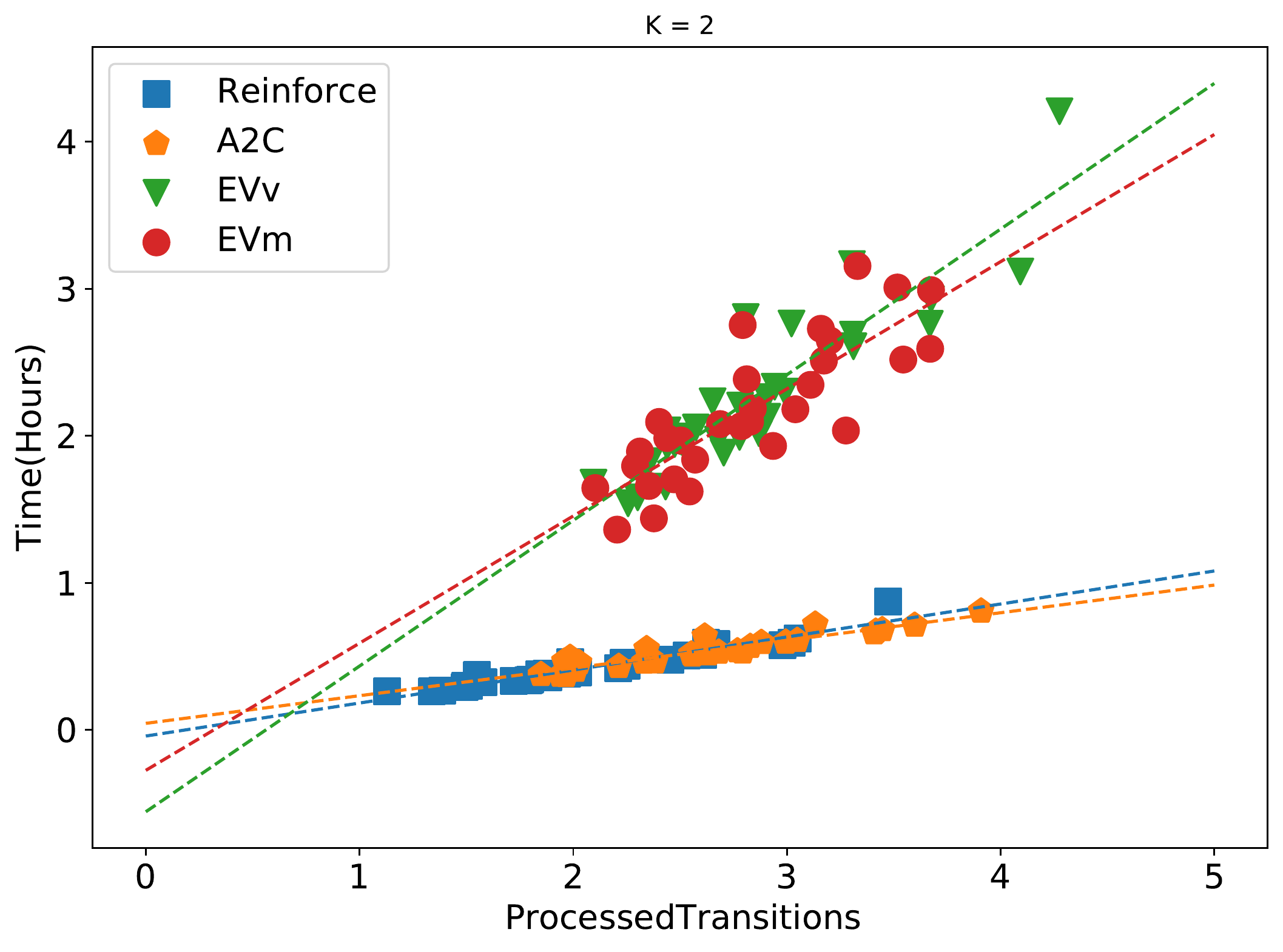}
    (b) \includegraphics[scale=.27]{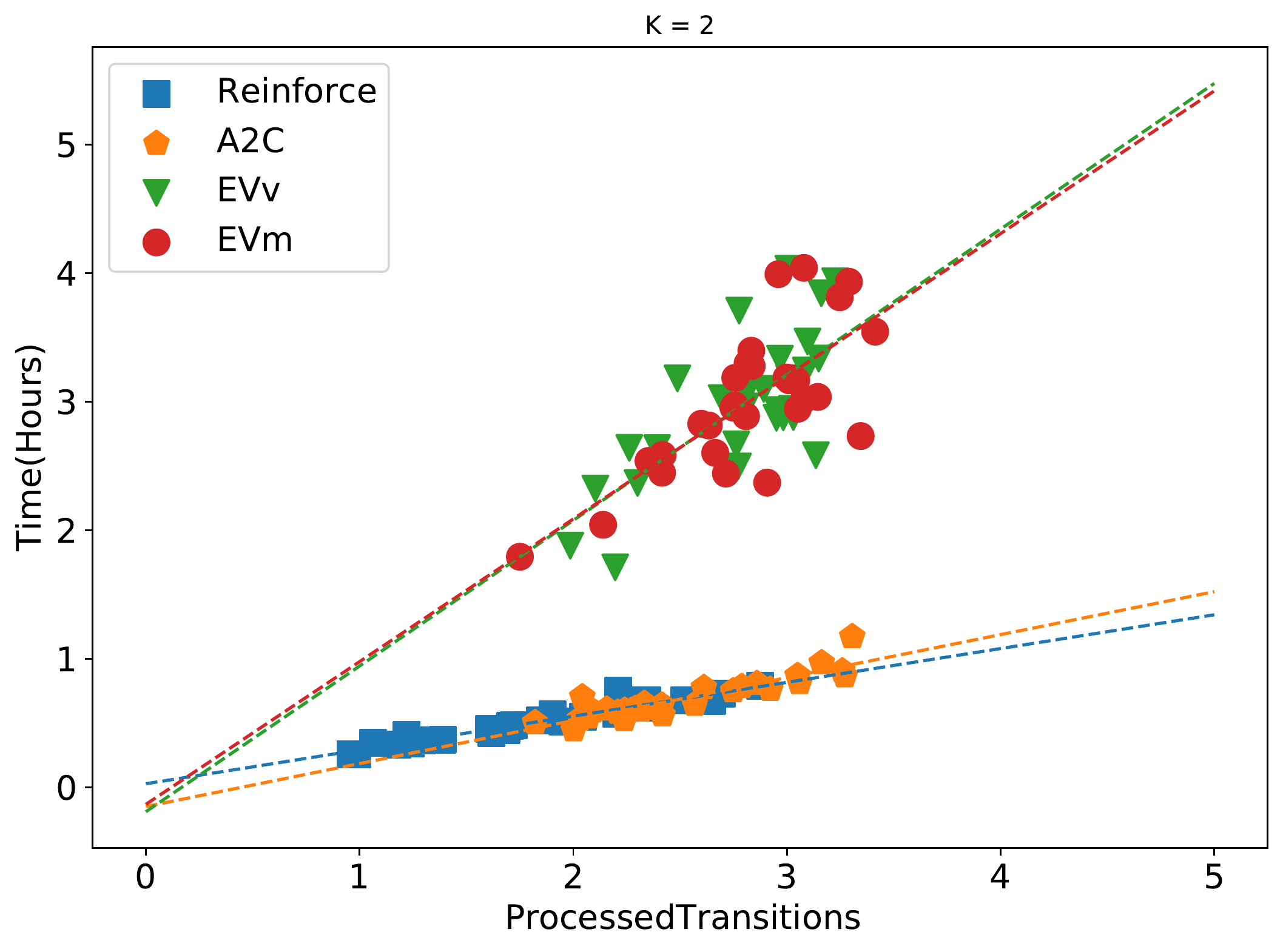} \\ 
    (c) \includegraphics[scale=.27]{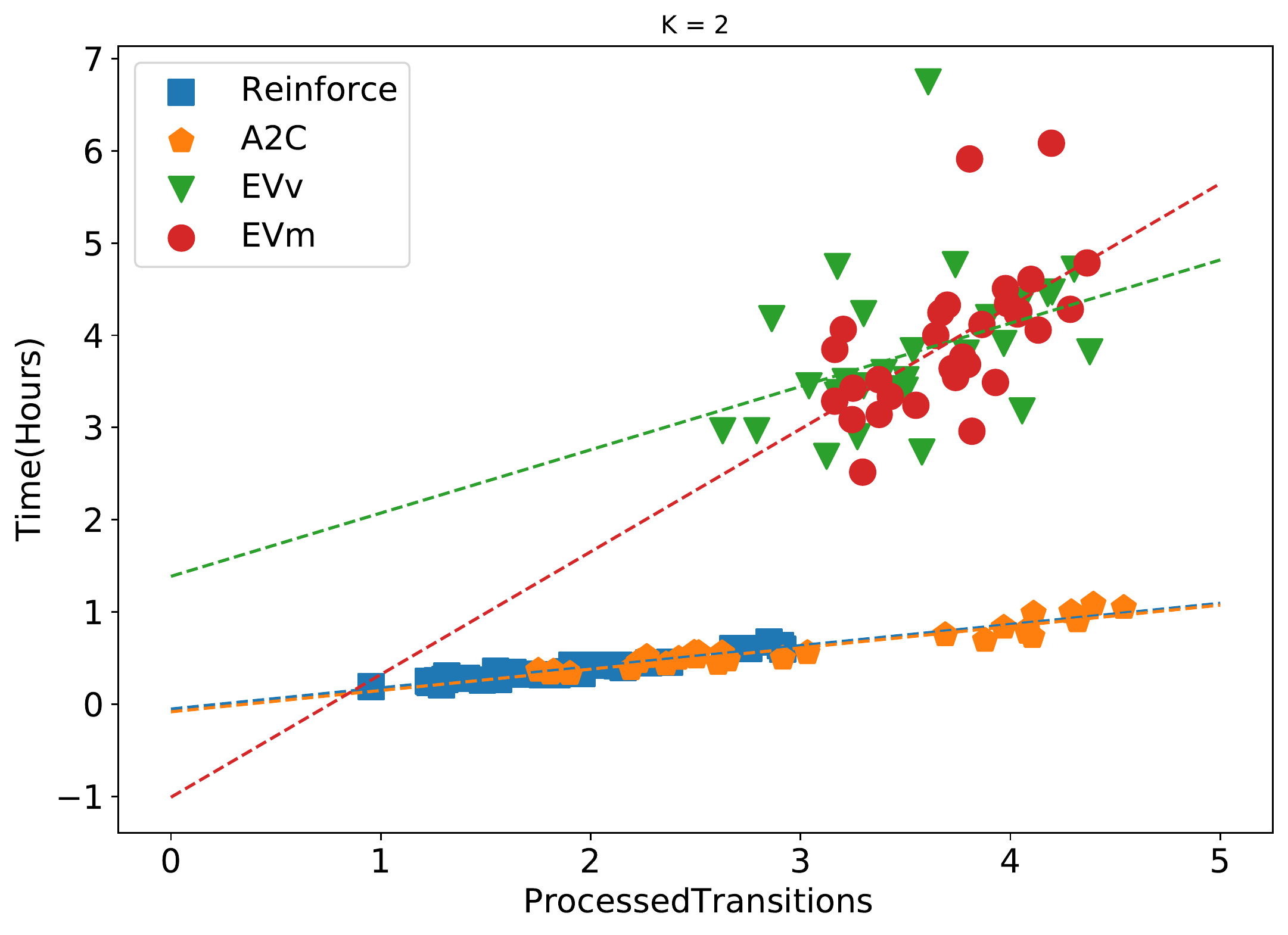}
    (d) \includegraphics[scale=.27]{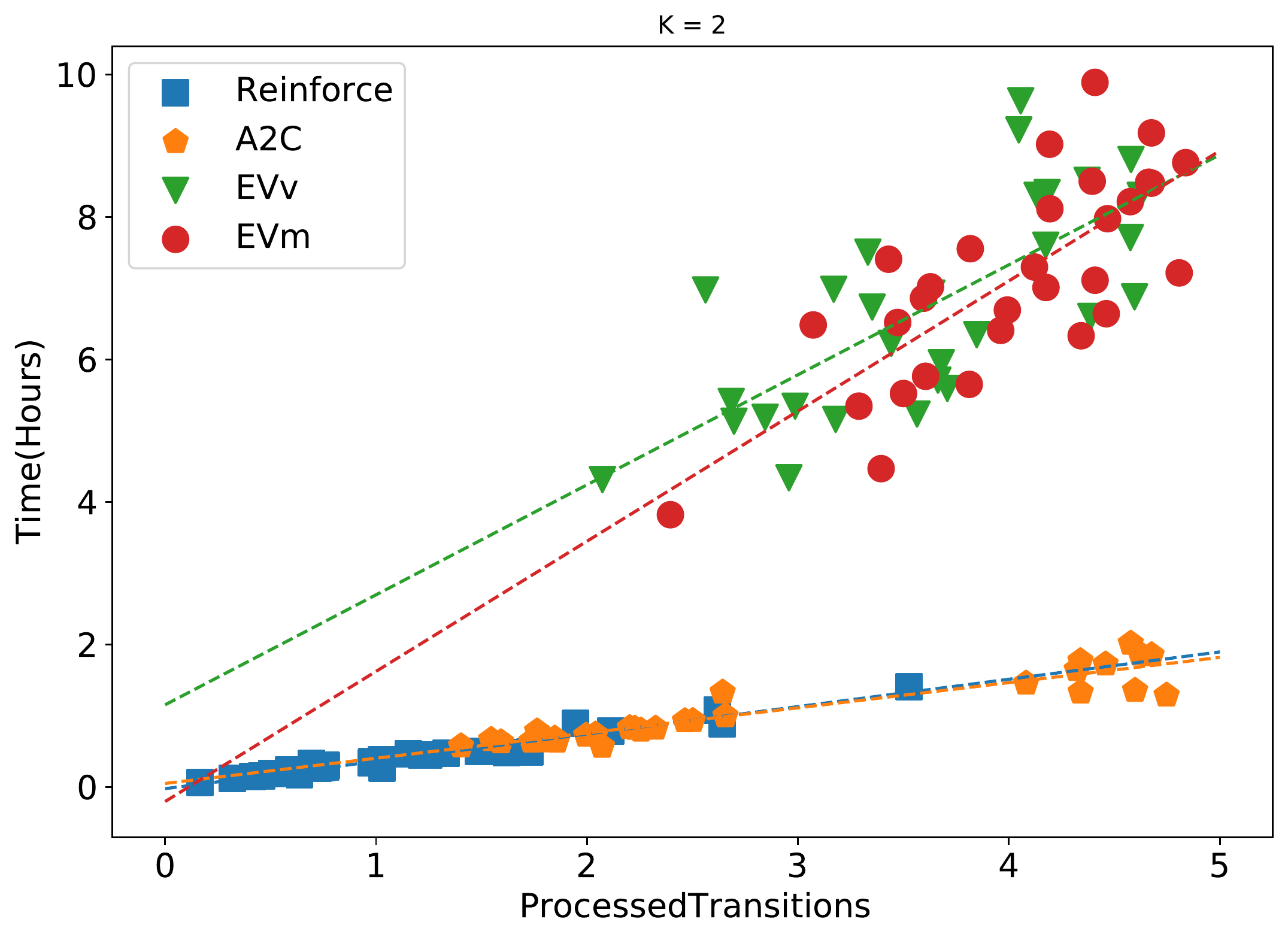} \\ 
    (e) \includegraphics[scale=.27]{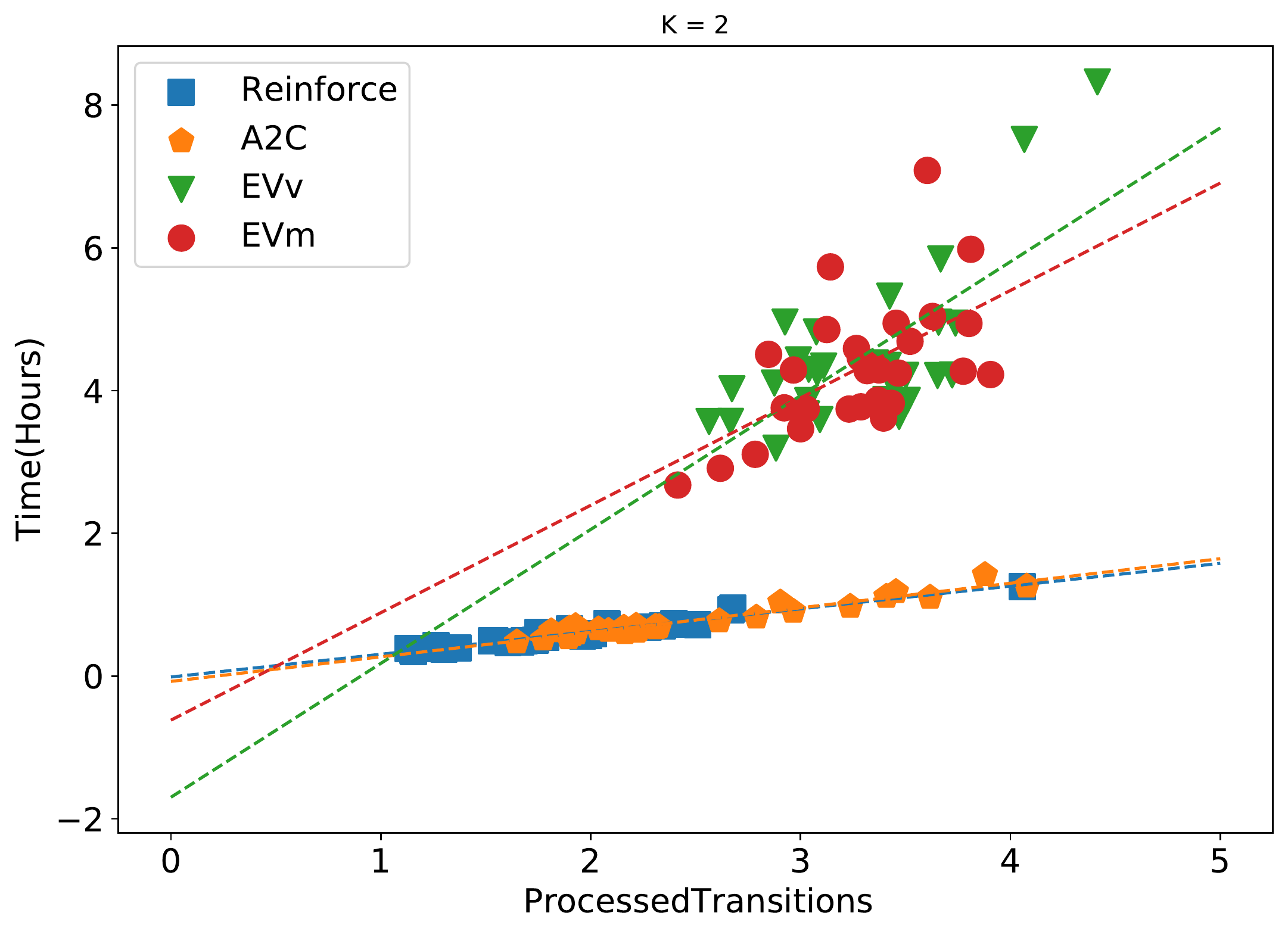}
    \end{tabular}
    \caption{The charts representing dependency of training time from number of the processed transitions(scale of millions) for CartPole environment: (a) config1, (b) config5, (c) config7, (d) config8, (e) config9}
    \label{fig:sup_CartPole_elapsedtime}
\end{figure}

\begin{figure}[h!]
    \centering
    \begin{tabular}{l}
    \includegraphics[scale=.27]{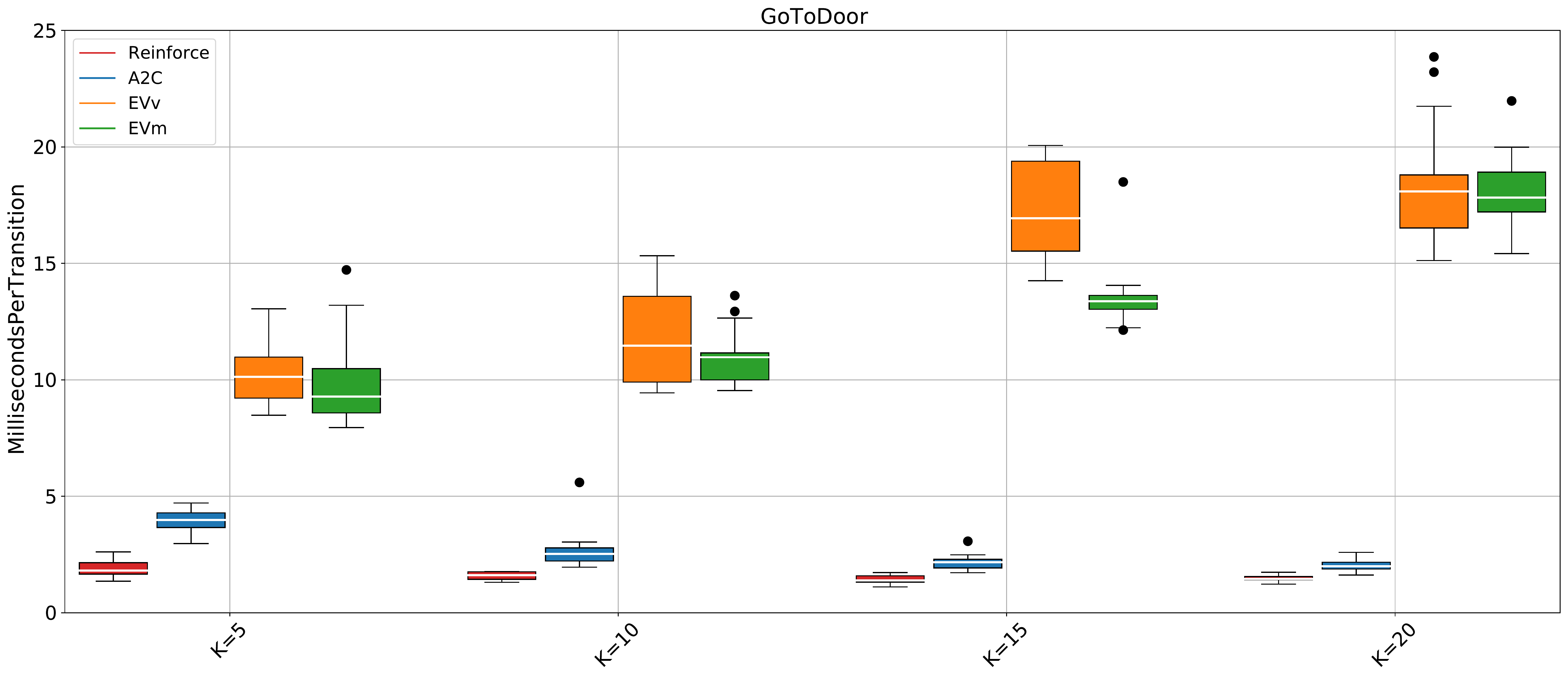}\\
    \end{tabular}
    \caption{The charts representing distribution of time per transition (scale of milliseconds) w.r.t. number of trajectories used for training in GoToDoor environment, $K=5,10,15,20$}
    \label{fig:sup_GoToDoor_time_fromK}
\end{figure}

\begin{figure}[h!]
    \centering
    \begin{tabular}{l}
    \includegraphics[scale=.27]{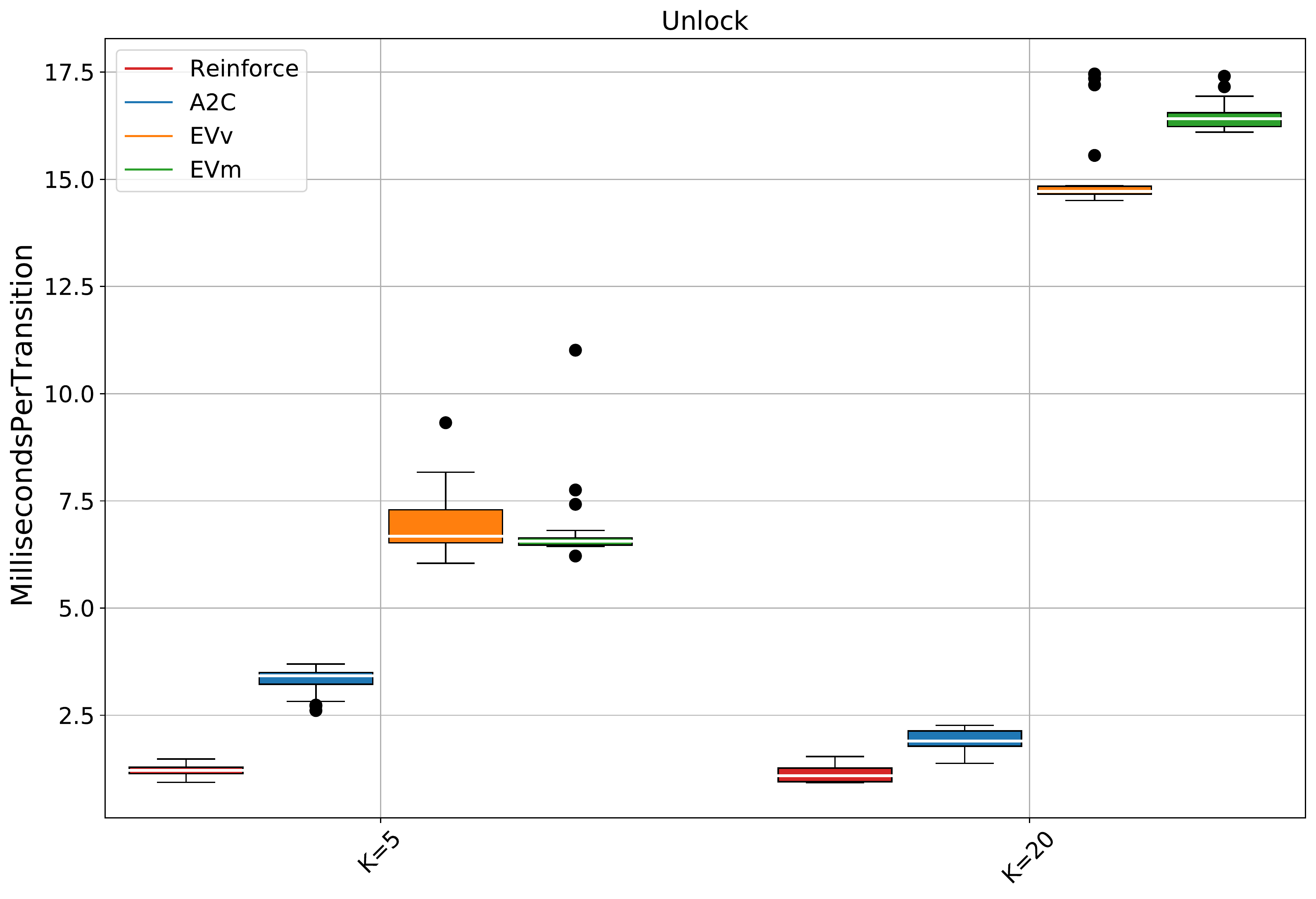}\\
    \end{tabular}
    \caption{The charts representing distribution of time per transition (scale of milliseconds) w.r.t. number of trajectories used for training in Unlock environment, $K=5,20$}
    \label{fig:sup_Unlock_time_fromK}
\end{figure}

\begin{figure}[h!]
    \centering
    \begin{tabular}{l}
    (a) \includegraphics[scale=.27]{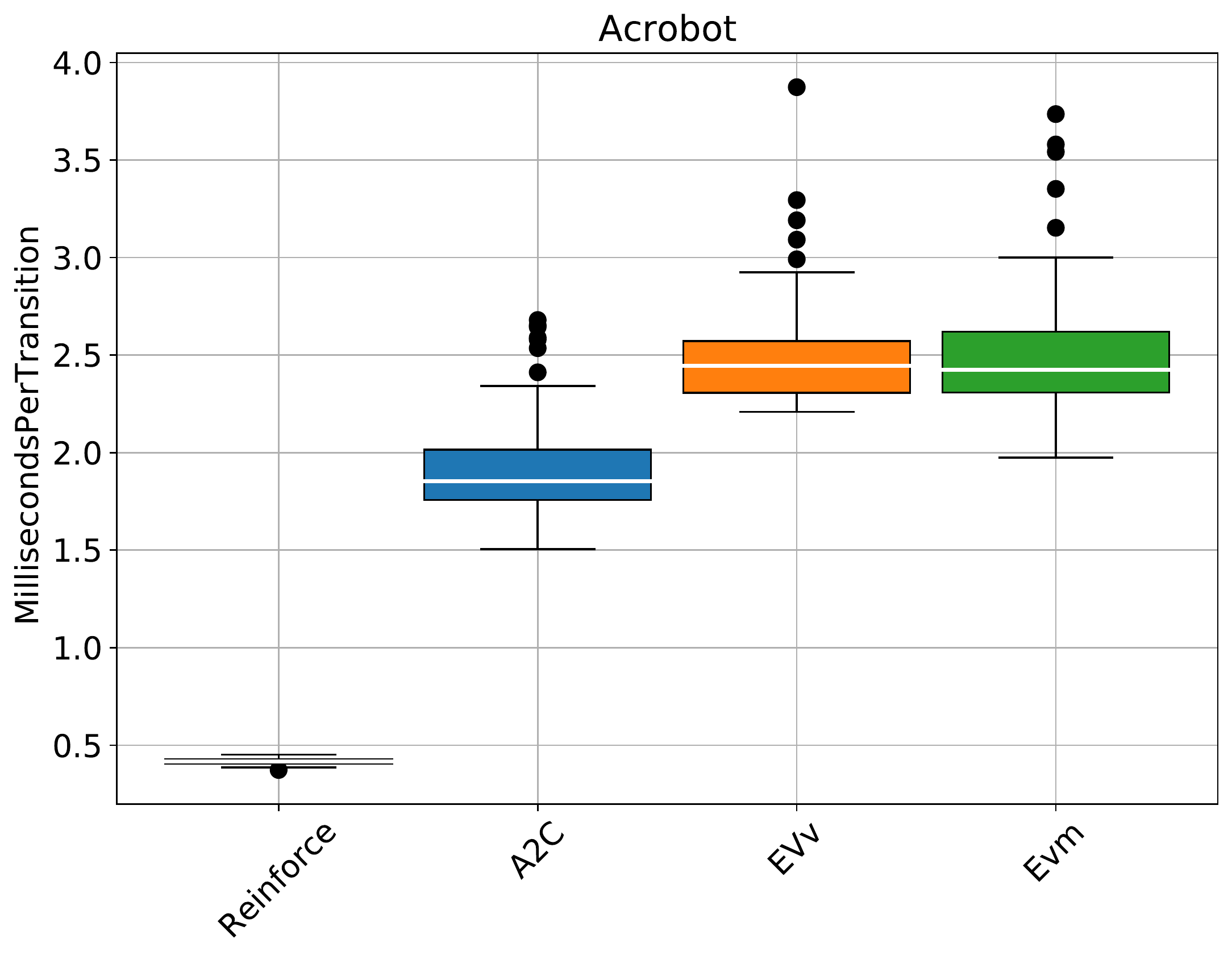}
    (b) \includegraphics[scale=.27]{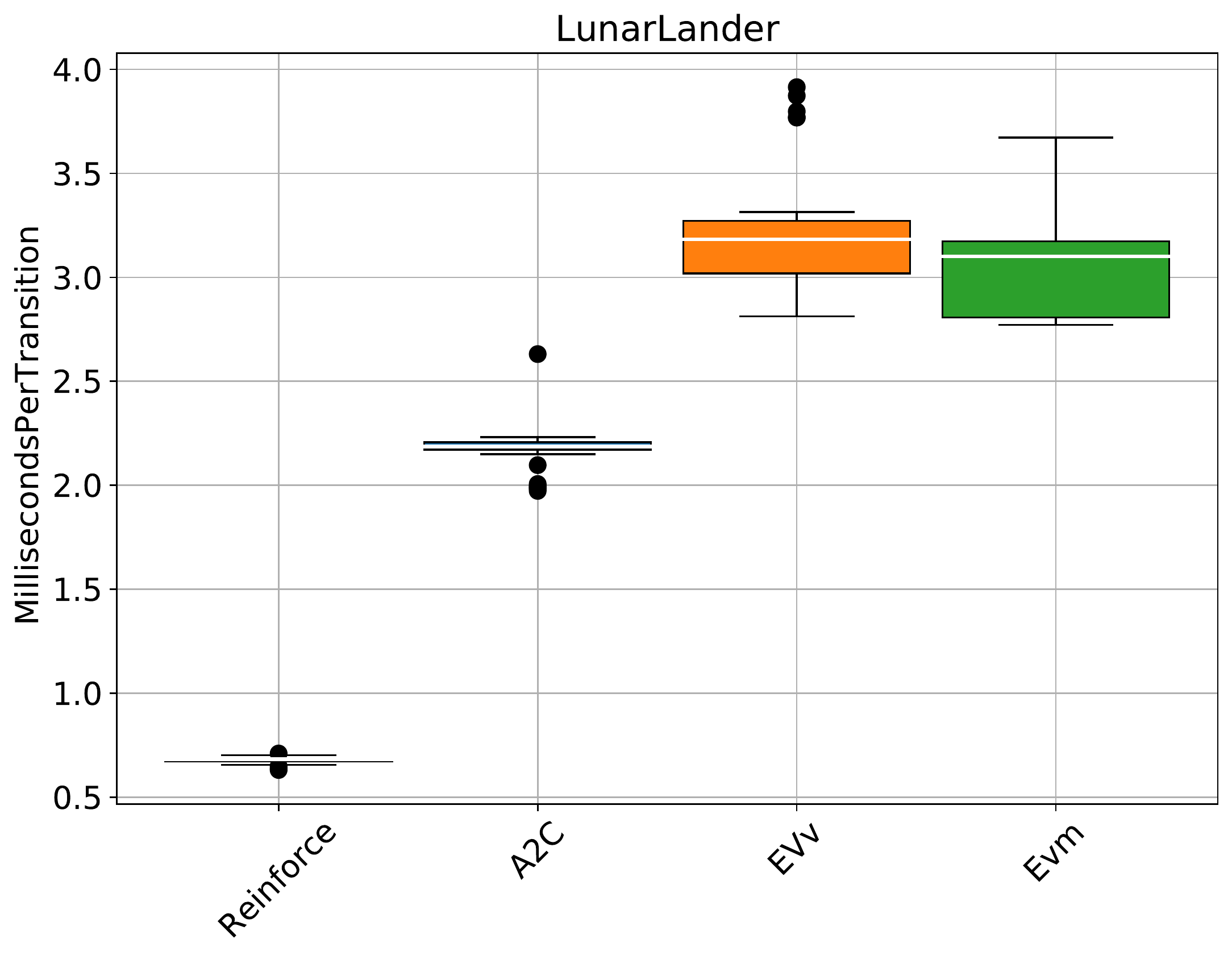}
    \end{tabular}
    \caption{The charts representing distribution of time per transition (scale of milliseconds) w.r.t. an algorithm for (a) Acrobot and for (b) LunarLander}
\end{figure}

\begin{figure}[h!]
    \centering
    \begin{tabular}{l}
    \includegraphics[scale=.27]{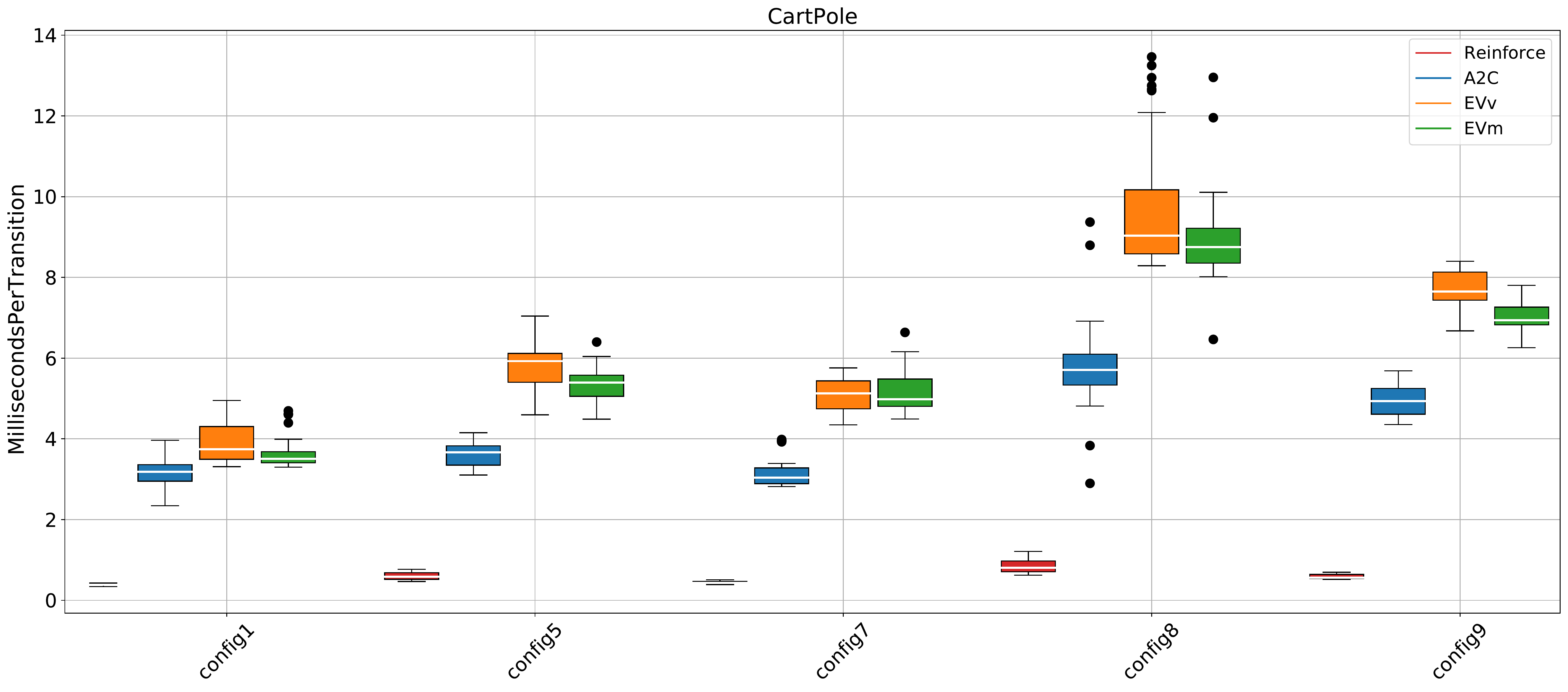}\\
    \end{tabular}
    \caption{The charts representing distribution of time per transition (scale of milliseconds) w.r.t. config number, CartPole environment.}
    \label{fig:sup_CartPole_time_fromConfigId}
\end{figure}